\documentclass{article}

\usepackage{arxiv}

\usepackage[utf8]{inputenc} 
\usepackage[T1]{fontenc}    
\usepackage{hyperref}       
\usepackage{url}            
\usepackage{booktabs}       
\usepackage{amsfonts}       
\usepackage{nicefrac}       
\usepackage{microtype}      
\usepackage{lipsum}
\usepackage{amsthm}
\usepackage{amssymb}
\usepackage{threeparttable}
\usepackage{amsmath}
\usepackage{verbatim}
\usepackage{xcolor}
\usepackage{tikz}
\usepackage{courier}
\usepackage{colortbl}
\usepackage{arydshln}
\usepackage{float}
\usepackage[bottom]{footmisc}

\definecolor{ciano}{rgb}{0.854,0.909,0.988}
\definecolor{cinza}{rgb}{0.7529,0.7529,0.7529}

\usepackage{enumitem,booktabs}
\usepackage[referable]{threeparttablex}
\renewlist{tablenotes}{enumerate}{1}
\makeatletter
\setlist[tablenotes]{label=\tnote{\alph*},ref=\alph*,itemsep=\z@,topsep=\z@skip,partopsep=\z@skip,parsep=\z@,itemindent=\z@,labelindent=\tabcolsep,labelsep=.2em,leftmargin=*,align=left,before={\footnotesize}}

\title{Attention, please! A survey of Neural Attention Models in Deep Learning}

\author{Alana~de~Santana~Correia,
        and~Esther~Luna~Colombini,
\thanks{Laboratory of Robotics and Cognitive Systems (LaRoCS)}\\
Institute of Computing, University of Campinas, Av. Albert Einstein, 1251 - Campinas, SP - Brazil\\ e-mail: \{alana.correia, esther\}@ic.unicamp.br.}

\begin{document}
\maketitle

\begin{abstract}
In humans, Attention is a core property of all perceptual and cognitive operations. Given our limited ability to process competing sources, attention mechanisms select, modulate, and focus on the information most relevant to behavior. For decades, concepts and functions of attention have been studied in philosophy, psychology, neuroscience, and computing. For the last six years, this property has been widely explored in deep neural networks. Currently, the state-of-the-art in Deep Learning is represented by neural attention models in several application domains. This survey provides a comprehensive overview and analysis of developments in neural attention models. We systematically reviewed hundreds of architectures in the area, identifying and discussing those in which attention has shown a significant impact. We also developed and made public an automated methodology to facilitate the development of reviews in the area. By critically analyzing 650 works, we describe the primary uses of attention in convolutional, recurrent networks and generative models, identifying common subgroups of uses and applications. Furthermore, we describe the impact of attention in different application domains and their impact on neural networks' interpretability. Finally, we list possible trends and opportunities for further research, hoping that this review will provide a succinct overview of the main attentional models in the area and guide researchers in developing future approaches that will drive further improvements.
\end{abstract}

\keywords{Survey \and Attention Mechanism \and Neural Networks \and Deep Learning \and Attention Models.}

\section{Introduction}
\label{sec:introduction}

Attention is a behavioral and cognitive process of focusing selectively on a discrete aspect of information, whether subjective or objective, while ignoring other perceptible information~\cite{colombini2014attentional}, playing an essential role in human cognition and the survival of living beings in general. In animals of lower levels in the evolutionary scale, it provides perceptual resource allocation allowing these beings to respond correctly to the environment's stimuli to escape predators and capture preys efficiently. In human beings, attention acts on practically all mental processes, from reactive responses to unexpected stimuli in the environment - guaranteeing our survival in the presence of danger - to complex mental processes, such as planning, reasoning, and emotions. Attention is necessary because, at any moment, the environment presents much more perceptual information than can be effectively processed, the memory contains more competing traits than can be remembered, and the choices, tasks, or motor responses available are much greater than can be dealt with~\cite{chun2011taxonomy}.

At early sensorial processing stages, data is separated between sight, hearing, touch, smell, and taste. At this level, Attention selects and modulates processing within each of the five modalities and directly impacts processing in the relevant cortical regions. For example, attention to visual stimuli increases discrimination and activates the relevant topographic areas in the retinotopic visual cortex~\cite{tootell1998retinotopy}, allowing observers to detect contrasting stimuli or make more precise discriminations. In hearing, attention allows listeners to detect weaker sounds or differences in extremely subtle tones but essential for recognizing emotions and feelings~\cite{woldorff1993modulation}. Similar effects of attention operate on the somatosensory cortex~\cite{johansen2000physiology}, olfactory cortex~\cite{zelano2005attentional}, and gustatory cortex~\cite{veldhuizen2007trying}. In addition to sensory perception, our cognitive control is intrinsically attentional. Our brain has severe cognitive limitations - the number of items that can be kept in working memory, the number of choices that can be selected, and the number of responses that can be generated at any time are limited. Hence, evolution has favored selective attention concepts as the brain has to prioritize.  

Long before contemporary psychologists entered the discussion on Attention, William James~\cite{James1890JAMPOP} offered us a precise definition that has been, at least, partially corroborated more than a century later by neurophysiological studies. According to James, ``Attention implies withdrawal from some things in order to deal effectively with others... Millions of items of the outward order are present to my senses which never properly enter into my experience. Why? Because they have no interest for me. My experience is what I agree to attend to. Only those items which I notice shape my mind — without selective interest, experience is an utter chaos.'' Indeed, the first scientific studies of Attention have been reported by Herman Von Helmholtz
(1821-1894) and William James (1890-1950) in the nineteenth century. They both conducted experiments to understand the role of Attention.


For the past decades, the concept of attention has permeated most aspects of research in perception and cognition, being considered as a property of multiple and different perceptual and cognitive operations~\cite{colombini2014attentional}. Thus, to the extent that these mechanisms are specialized and decentralized, attention reflects this organization. These mechanisms are in wide communication, and the executive control processes help set priorities for the system. Selection mechanisms operate throughout the brain and are involved in almost every stage, from sensory processing to decision making and awareness. Attention has become a broad term to define how the brain controls its information processing, and its effects can be measured through conscious introspection, electrophysiology, and brain imaging. Attention has been studied from different perspectives for a long time. 

\subsection{Pre-Deep Learning Models of Attention}
Computational attention systems based on psychophysical models, supported by neurobiological evidence, have existed for at least three decades~\cite{FrintropSurvey}. Treisman's Feature Integration Theory (FIT) ~\cite{treisman1980feature}, Wolfe's Guides Search ~\cite{wolfe1989guided}, Triadic architecture ~\cite{rensink2000dynamic}, Broadbent's Model ~\cite{broadbent2013perception}, Norman Attentional Model ~\cite{norman1968toward} ~\cite{kahneman1973attention}, Closed-loop Attention Model ~\cite{van2004clam}, SeLective Attention Model ~\cite{phaf1990slam}, among several other models, introduced the theoretical basis of computational attention systems.

Initially, attention was mainly studied with visual experiments where a subject looks at a scene that changes in time~\cite{frintrop2010computational}. In these models, the attentional system was restricted only to the selective attention component in visual search tasks, focusing on the extraction of multiple features through a sensor. Therefore, most of the attentional computational models occurred in computer vision to select important image regions. Koch and Ullman ~\cite{koch1987shifts} introduced the area's first visual attention architecture based on FIT ~\cite{treisman1980feature}. The idea behind it is that several features are computed in parallel, and their conspicuities are collected on a salience map. Winner-Take-All (WTA) determines the most prominent region on the map, which is finally routed to the central representation. From then on, only the region of interest proceeds to more specific processing. Neuromorphic Vision Toolkit (NVT), derived from the Koch-Ullman ~\cite{itti1998model} model, was the basis for developing research in computational visual attention for several years. Navalpakkam and Itti introduce a derivative of NVT which can deal with top-down cues ~\cite{navalpakkam2006integrated}. The idea is to learn the target's feature values from a training image in which a binary mask indicates the target. The attention system of Hamker ~\cite{hamker2005emergence} ~\cite{hamker2006modeling} calculates various features and contrast maps and turns them into perceptual maps. With target information influencing processing, they combine detection units to determine whether a region on the perceptual map is a candidate for eye movement. VOCUS ~\cite{frintrop2006vocus} introduced a way to combine bottom-up and top-down attention, overcoming the limitations of the time. Several other models have emerged in the literature, each with peculiarities according to the task. Many computational attention systems focus on the computation of mainly three features:
intensity, orientation, and color. These models employed neural networks or filter models that use classical linear filters to compute features.

Computational attention systems were used successfully before Deep Learning (DL) in object recognition ~\cite{salah2002selective}, image compression ~\cite{ouerhani2004visual}, image matching ~\cite{walther2006interactions}, image segmentation ~\cite{ouerhani2004visual}, object tracking ~\cite{walther2004detection}, active vision ~\cite{clark1988modal}, human-robot interaction ~\cite{breazeal1999context}, object manipulation in robotics ~\cite{rotenstein2007towards}, robotic navigation ~\cite{clark1992attentive}, and SLAM ~\cite{frintrop2008attentional}. In mid-1997, Scheier and Egner ~\cite{scheier1997visual} presented a mobile robot that uses attention for navigation. Still, in the 90s, Baluja and Pomerleau ~\cite{baluja1997expectation} used an attention system to navigate an autonomous car, which followed relevant regions of a projection map. Walther ~\cite{walther2006interactions} combined an attentional system with an object recognizer based on SIFT features and demonstrated that the attentional front-end enhanced the recognition results. Salah et al. ~\cite{salah2002selective} combined attention with neural networks in an Observable Markov model for handwritten digit recognition and face recognition. Ouerhani et al. ~\cite{ouerhani2004visual} proposed the focused image compression, which determines the number of bits to be allocated for encoding regions of an image according to their salience. High saliency regions have a high quality of reconstruction concerning the rest of the image.

\subsection{Deep Learning Models of Attention: the beginning}

\begin{figure}[htb]
    \centering
    \includegraphics[width=0.55\linewidth]{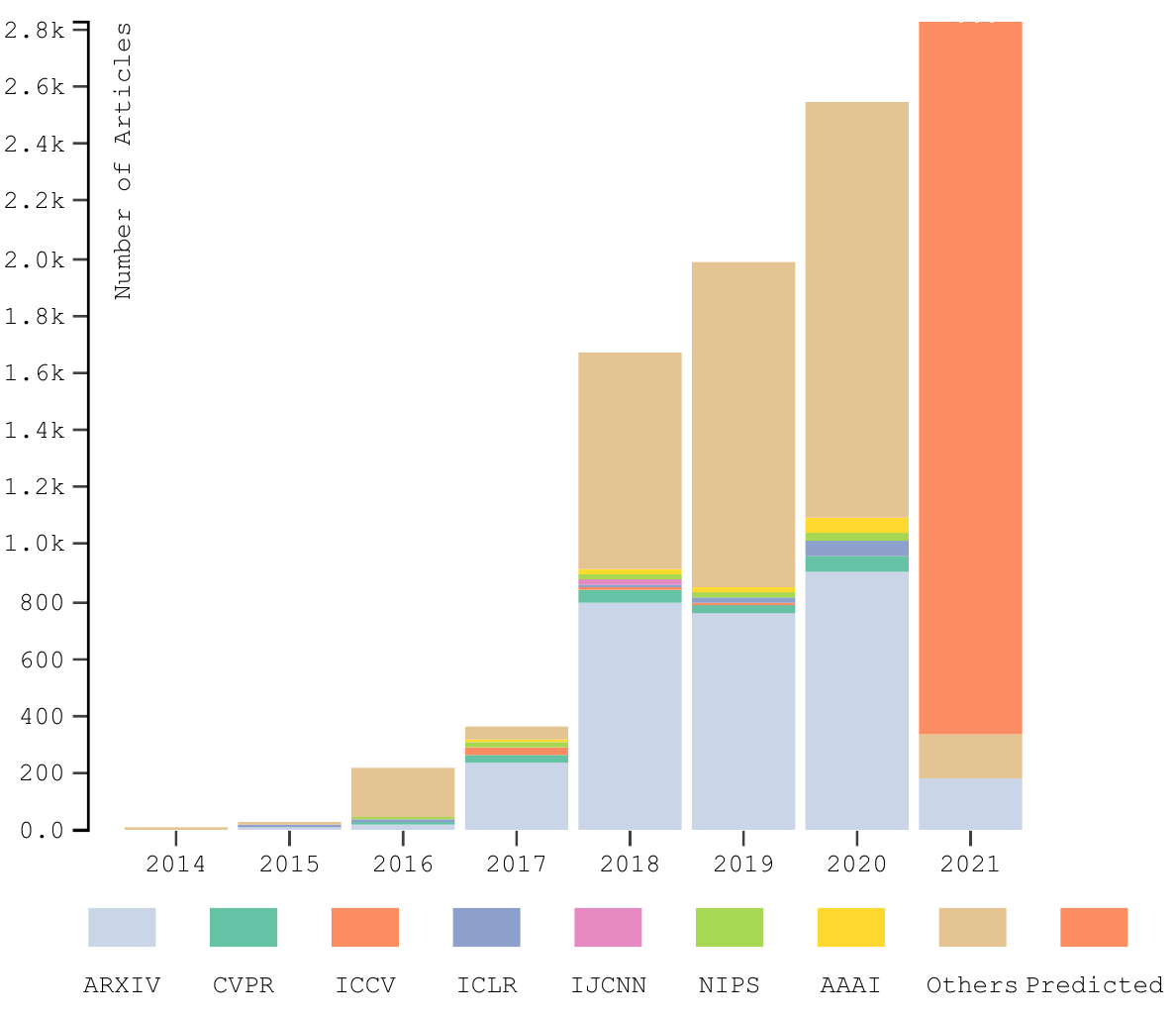}
    \caption{Works published by year between 01/01/2014 to 15/02/2021. The main sources collected are ArXiv, CVPR, ICCV, ICLR, IJCNN, NIPS, and AAAI. The other category refers mainly to the following publishing vehicles: ICML, ACL, ACM, EMNLP, ICRA, ICPR, ACCV, CORR, ECCV, ICASSP, ICLR, IEEE ACCESS, Neurocomputing, and several other magazines.}
    \label{fig:hist_works}
\end{figure}

By 2014, the DL community noticed attention as a fundamental concept for advancing deep neural networks. Currently, the state-of-the-art in the field uses neural attention models. As shown in figure~\ref{fig:hist_works}, the number of published works grows each year significantly in the leading repositories. In neural networks, attention mechanisms dynamically manage the flow of information, the features, and the resources available, improving learning. These mechanisms filter out irrelevant stimuli for the task and help the network to deal with long-time dependencies simply. Many neural attentional models are simple, scalable, flexible, and with promising results in several application domains~\cite{draw}~\cite{vaswani_attention_2017}~\cite{weston_2014_memory}. Given the current research extent, interesting questions related to neural attention models arise in the literature: \textbf{how these mechanisms help improve neural networks' performance}, \textbf{which classes of problems benefit from this approach}, and \textbf{how these benefits arise}.

To the best of our knowledge, most surveys available in the literature do not address all of these questions or are more specific to some domain.
Wang et al.~\cite{wang2016survey} propose a review on recurrent networks and applications in computer vision, Hu~\cite{hu2019introductory}, and Galassi et al.~\cite{galassi2020attention} offer surveys on attention in natural language processing (NLP). Lee et al.~\cite{lee_attention_2018} present a review on attention in graph neural networks, and Chaudhari et al.~\cite{chaudhari2019attentive} presented a more general, yet short, review. 

\subsection{Contributions}

To assess the breadth of attention applications in deep neural networks, we present a systemic review of the field in this survey. Throughout our review, we \textbf{critically analyzed 650 papers} while addressing \textbf{quantitatively 6,567}. 

As the main contributions of our work, we highlight:

\begin{enumerate}
    \item A replicable research methodology. We provide, in the Appendix, the detailed process conducted to collect our data and we make available the scripts to collect the papers and create the graphs we use;
    \item An in-depth overview of the field. We critically analyzed 650 papers and extracted different metrics from 6,567, employing various visualization techniques to highlight overall trends in the area;
    \item We describe the main attentional mechanisms; 
    \item We present the main neural architectures that employ attention mechanisms, describing how they have contributed to the NN field;
    \item We introduce how attentional modules or interfaces have been used in classic DL architectures extending the Neural Network Zoo diagrams;
    \item Finally, we present a broad description of application domains, trends, and research opportunities.
\end{enumerate}

This survey is structured as follows. In Section~\ref{sub:attention_overview} we present the field overview reporting the main events from 2014 to the present. Section~\ref{sub:attention_mechanisms} contains a description of attention main mechanisms. In Section~\ref{sub:attention_classic_architectures} we analyze how attentional modules are used in classic DL architectures. Section~\ref{sec:applications} explains the main classes of problems and applications of attention. Finally, in Section~\ref{sec:trends} we discuss limitations, open challenges, current trends, and future directions in the area, concluding our work in section~\ref{sec:conclusion} with directions for further improvements.
\section{Overview}
\label{sub:attention_overview}

\begin{figure*}[h]
    \centering
    \includegraphics[width=0.95\linewidth]{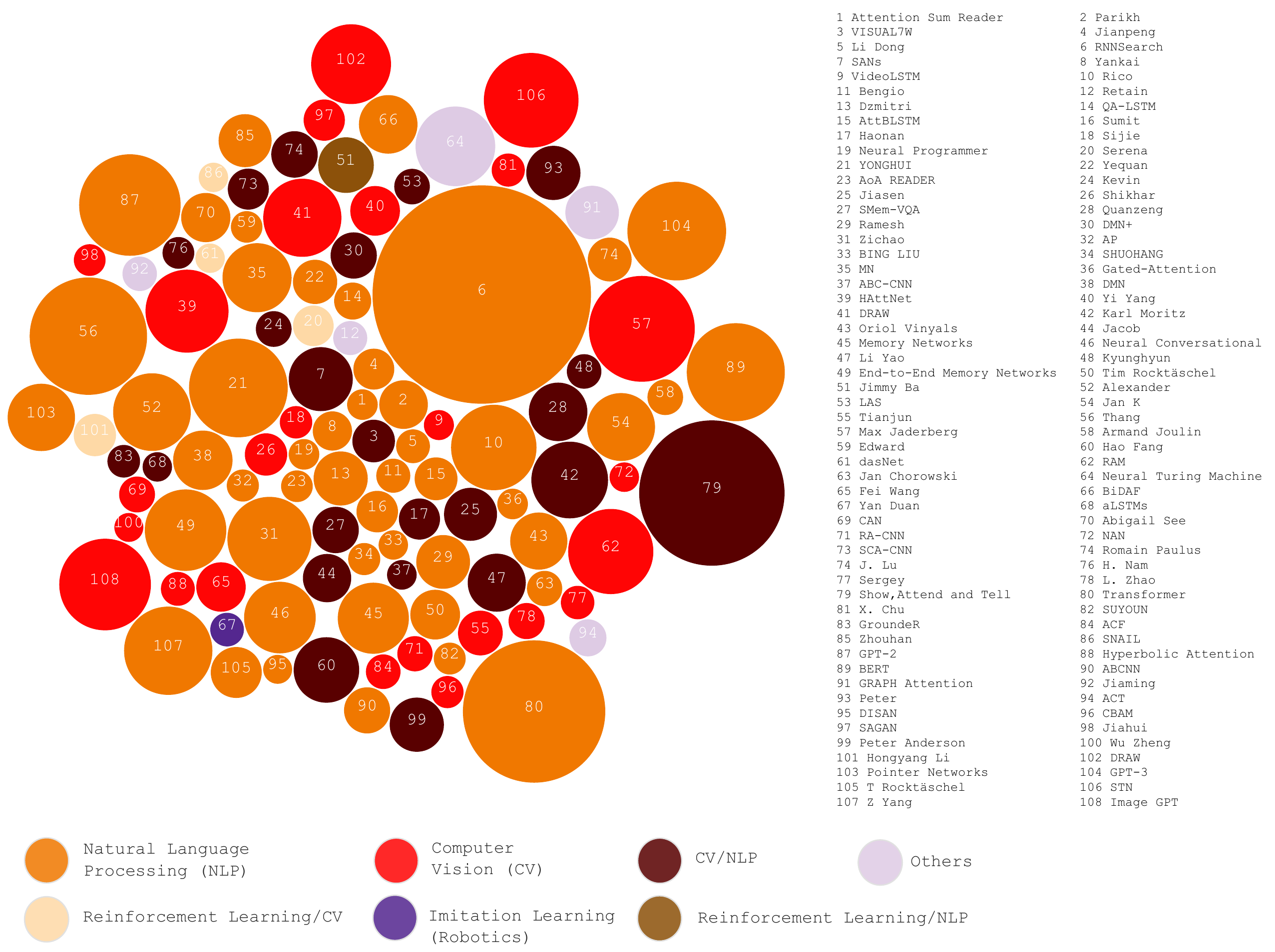}
    \caption{Main Neural Attention Networks (NAN). Each circle corresponds to an architecture. The radius of the circles is defined based on the impact of the NAN on the field. The impact was defined by the citation number and the architecture innovation level. The greater the radius of the circle, the more significant the impact of architecture, and vice versa. Architectures labels are color-coded as follows: orange - natural language processing, red - computer vision, dark brown - computer vision and natural language processing, dark yellow - reinforcement learning and computer vision, light yellow - reinforcement learning and natural language processing, blue - imitation learning and robotics, and purple - others.}
    \label{fig:main_architectures_field}
\end{figure*}

Historically, research in computational attention systems has existed since the 1980s. Only in mid-2014, the Neural Attentional Networks (NANs) emerged in Natural Language Processing (NLP), where attention provided significant advances, bringing promising results through scalable and straightforward networks. Attention allowed us to move towards the complex tasks of conversational machine comprehension, sentiment analysis, machine translation, question-answering, and transfer learning, previously challenging. Subsequently, NANs appeared in other fields equally important for artificial intelligence, such as computer vision, reinforcement learning, and robotics. There are currently numerous attentional architectures, but few of them have a significantly higher impact, as shown in Figure~\ref{fig:main_architectures_field}. In this image, we depict the most relevant group of works organized according to citation levels and innovations where RNNSearch~\cite{bahdanau_neural_2014}, Transformer~\cite{vaswani_attention_2017}, Memory Networks~\cite{weston_2014_memory}, ``show, attend and tell''~\cite{xu_show_2015}, and RAM~\cite{mnih_recurrent_2014} stand out as key developments.

The \textit{bottleneck problem} in the classic encoder-decoder framework worked as the initial motivation for attention research in Deep Learning. In this framework, the encoder encodes a source sentence into a fixed-length vector from which a decoder generates the translation. The main issue is that a neural network needs to compress all the necessary information from a source sentence into a fixed-length vector. Cho et al. ~\cite{cho2014properties} showed that the performance of the classic encoder-decoder deteriorates rapidly as the size of the input sentence increases. To minimize this bottleneck, Bahdanau et al.~\cite{bahdanau_neural_2014} proposed \textbf{RNNSearch}, an extension to the encoder-decoder model that learns to align and translate together. RNNSearch generates a translated word at each time-step, looking for a set of positions in the source sentence with the most relevant words. The model predicts a target word based on the context vectors associated with those source positions and all previously generated target words. The main advantage is that RNNSearch does not encode an entire input sentence into a single fixed-length vector. Instead, it encodes the input sentence into a sequence of vectors, choosing a subset of these vectors adaptively while generating the translation. The attention mechanism allows extra information to be propagated through the network, eliminating the fixed-size context vector's information bottleneck. This approach demonstrated that the attentive model outperforms classic encoder-decoder frameworks for long sentences for the first time.

RNNSearch was instrumental in introducing the first attention mechanism, \textbf{soft attention} (Section~\ref{sub:attention_mechanisms}). This mechanism has the main characteristic of smoothly selecting the network's most relevant elements. Based on RNNSearch, there have been numerous attempts to augment neural networks with new properties. Two research directions stand out as particularly interesting - \textbf{attentional interfaces} and \textbf{end-to-end attention}. Attentional interfaces treat attention as a module or set of elective modules, easily plugged into classic Deep Learning neural networks, just like RNNSearch. So far, this is the most explored research direction in the area, mainly for simplicity, general use, and the good results of generalization that the attentional interfaces bring. End-to-end attention is a younger research direction, where the attention block covers the entire neural network. High and low-level attentional layers act recursively or cascaded at all network abstraction levels to produce the desired output in these models. End-to-end attention models introduce a new class of neural networks in Deep Learning. End-to-end attention research makes sense since no isolated attention center exists in the human brain, and its mechanisms are used in different cognitive processes.

\subsection{Attentional interfaces}
\label{sub:attentional_interfaces}
RNNSearch is the basis for research on attentional interfaces. The attentional module of this architecture is widely used in several other applications. In voice recognition~\cite{chan2015listen}, allowing one RNN to process the audio while another examines it focusing on the relevant parts as it generates a description. In-text analysis~\cite{vinyals2015grammar}, it allows a model to look at the words as it generates an analysis tree. In conversational modeling~\cite{vinyals_neural_nodate}, it allows the model to focus on the last parts of the conversation as it generates its response. There are also important extensions to deal with other information bottlenecks in addition to the classic encoder-decoder problem. \textbf{BiDAF}~\cite{seo_bidirectional_2016} proposes a multi-stage hierarchical process to question-answering. It uses the bidirectional attention flow to build a multi-stage hierarchical network with context paragraph representations at different granularity levels. The attention layer does not summarize the context paragraph in a fixed-length vector. Instead, attention is calculated for each step, and the vector assisted at each step, along with representations of previous layers, can flow to the subsequent modeling layer. This reduces the loss of information caused by the early summary. At each stage of time, attention is only a function of the query and the paragraph of the context in the current stage and does not depend directly on the previous stage's attention. The hypothesis is that this simplification leads to a work division between the attention layer and the modeling layer, forcing the attention layer to focus on learning attention between the query and the context.

Yang et al.~\cite{yang2016hierarchical} proposed the \textbf{Hierarchical Attention Network (HAN)} to capture two essential insights about document structure. Documents have a hierarchical structure: words form sentences, sentences form a document. Humans, likewise, construct a document representation by first building representations of sentences and then aggregating them into a document representation. Different words and sentences in a document are differentially informative. Moreover, the importance of words and sentences is highly context-dependent, i.e., the same word or sentence may have different importance in different contexts. To include sensitivity to this fact, HAN consists of two levels of attention mechanisms - one at the word level and one at the sentence level - that let the model pay more or less attention to individual words and sentences when constructing the document's representation. Xiong et al.~\cite{xiong2016dynamic} created a coattentive encoder that captures the interactions between the question and the document with a dynamic pointing decoder that alternates between estimating the start and end of the answer span. To learn approximate solutions to computationally intractable problems, \textbf{Ptr-Net}~\cite{vinyals2015pointer} modifies the RNNSearch's attentional mechanism to represent variable-length dictionaries. It uses the attention mechanism as a pointer.

See et. al.~\cite{see_get_2017} used a hybrid  between classic sequence-to-sequence attentional models and a Ptr-Net~\cite{vinyals2015pointer} to abstractive text summarization. The hybrid pointer-generator~\cite{see_get_2017} copies words from the source text via pointing, which aids accurate reproduction of information while retaining the ability to produce novel words through the generator. Finally, it uses a mechanism to keep track of what has been summarized, which discourages repetition. FusionNet~\cite{huang_fusionnet:_2018} presents a novel concept of "history-of-word" to characterize attention information from the lowest word-embedding level up to the highest semantic-level representation. This concept considers that data input is gradually transformed into a more abstract representation, forming each word's history in human mental flow. FusionNet employs a fully-aware multi-level attention mechanism and an attention score-function that takes advantage of the history-of-word. Rocktäschel et al.~\cite{rocktaschel_reasoning_2015} introduce two-away attention for  recognizing textual entailment (RTE). The mechanism allows the model to attend over past output vectors, solving the LSTM's cell state bottleneck. The LSTM with attention does not need to capture the premise's whole semantics in the LSTM cell state. Instead, attention generates output vectors while reading the premise and accumulating a representation in the cell state that informs the second LSTM which of the premises' output vectors to attend to determine the RTE class. Luong, et al.~\cite{luong_effective_2015}, proposed global and local attention in machine translation. Global attention is similar to soft attention, while local is an improvement to make hard attention differentiable - the model first provides for a single position aligned to the current target word, and a window centered around the position is used to calculate a vector of context. 

Attentional interfaces have also emerged in architectures for computer vision tasks. Initially, they are based on human saccadic movements and robustness to change. The human visual attention mechanism can explore local differences in an image while highlighting the relevant parts. One person focuses attention on parts of the image simultaneously, glimpsing to quickly scan the entire image to find the main areas during the recognition process. In this process, the different regions' internal relationship guides the eyes' movement to find the next area to focus. Ignoring the irrelevant parts makes it easier to learn in the presence of disorder. Another advantage of glimpse and visual attention is its robustness. Our eyes can see an object in a real-world scene but ignore irrelevant parts. Convolutional neural networks (CNNs) are extremely different. CNNs are rigid, and the number of parameters grows linearly with the size of the image. Also, for the network to capture long-distance dependencies between pixels, the architecture needs to have many layers, compromising the model's convergence. Besides, the network treats all pixels in the same way. This process does not resemble the human visual system that contains visual attention mechanisms and a glimpse structure that provides unmatched performance in object recognition.

RAM~\cite{mnih_recurrent_2014} and STN are pioneering architectures with attentional interfaces based on human visual attention. \textbf{RAM}~\cite{mnih_recurrent_2014} can extract information from an image or video by adaptively selecting a sequence of regions, glimpses, only processing the selected areas at high resolution. The model is a Recurrent Neural Network that processes different parts of the images (or video frames) at each instant of time \textit{t}, building a dynamic internal representation of the scene via Reinforcement Learning training. The main model advantages are the reduced number of parameters and the architecture's independence to the input image size, which does not occur in convolutional neural networks. This approach is generic. It can use static images, videos, or a perceptual module of an agent that interacts with the environment. \textbf{STN (Spatial Transformer Network)}~\cite{jaderberg_spatial_2015} is a module robust to spatial transformation changes. In STN, if the input is transformed, the model must generate the correct classification label, even if it is distorted in unusual ways. STN works as an attentional module attachable -- with few modifications -- to any neural network to actively spatially transform feature maps. STN learns transformation during the training process. Unlike pooling layers, where receptive fields are fixed and local, a Spatial Transformer is a dynamic mechanism that can spatially transform an image, or feature map, producing the appropriate transformation for each input sample. The transformation is performed across the map and may include changes in scale, cut, rotations, and non-rigid body deformations. This approach allows the network to select the most relevant image regions (attention) and transform them into a desired canonical position by simplifying recognition in the following layers.

Following the RAM approach, the \textbf{Deep Recurrent Attentive Writer (DRAW)}~\cite{draw} represents a change to a more natural way of constructing the image in which parts of a scene are created independently of the others. This process is how human beings draw a scene by recreating a visual scene sequentially, refining all parts of the drawing for several iterations, and reevaluating their work after each modification. Although natural to humans, most approaches to automatic image generation aim to generate complete scenes at once. This means that all pixels are conditioned in a single latent distribution, making it challenging to scale large image approaches. DRAW belongs to the family of variational autoencoders. It has an encoder that compresses the images presented during training and a decoder that reconstructs the images. Unlike other generative models, DRAW iteratively constructs the scenes by accumulating modifications emitted by the decoder, each observed by the encoder. DRAW uses RAM attention mechanisms to attend to parts of the scene while ignoring others selectively. This mechanism's main challenge is to learn where to look, which is usually addressed by reinforcement learning techniques. However, at DRAW, the attention mechanism is differentiable, making it possible to use backpropagation.

\subsection{Multimodality}
\label{sub:multimodality}

The first attention interfaces' use in DL were limited to  NLP and computer vision domains to solve isolated tasks. Currently, attentional interfaces are studied in multimodal learning. Sensory multimodality in neural networks is a historical problem widely discussed by the scientific community~\cite{ramachandram2017deep}~\cite{gao2020survey}. Multimodal data improves the robustness of perception through complementarity and redundancy. The human brain continually deals with multimodal data and integrates it into a coherent representation of the world. However, employing different sensors present a series of challenges computationally, such as incomplete or spurious data, different properties (i.e. dimensionality or range of values), and the need for data alignment association. The integration of multiple sensors depends on a reasoning structure over the data to build a common representation, which does not exist in classical neural networks. Attentional interfaces adapted for multimodal perception are an efficient alternative for reasoning about misaligned data from different sensory sources.

\begin{figure}[htb]
    \centering
    \includegraphics[clip, trim=0.3cm 6cm 0.5cm 6cm, width=1.00\textwidth]{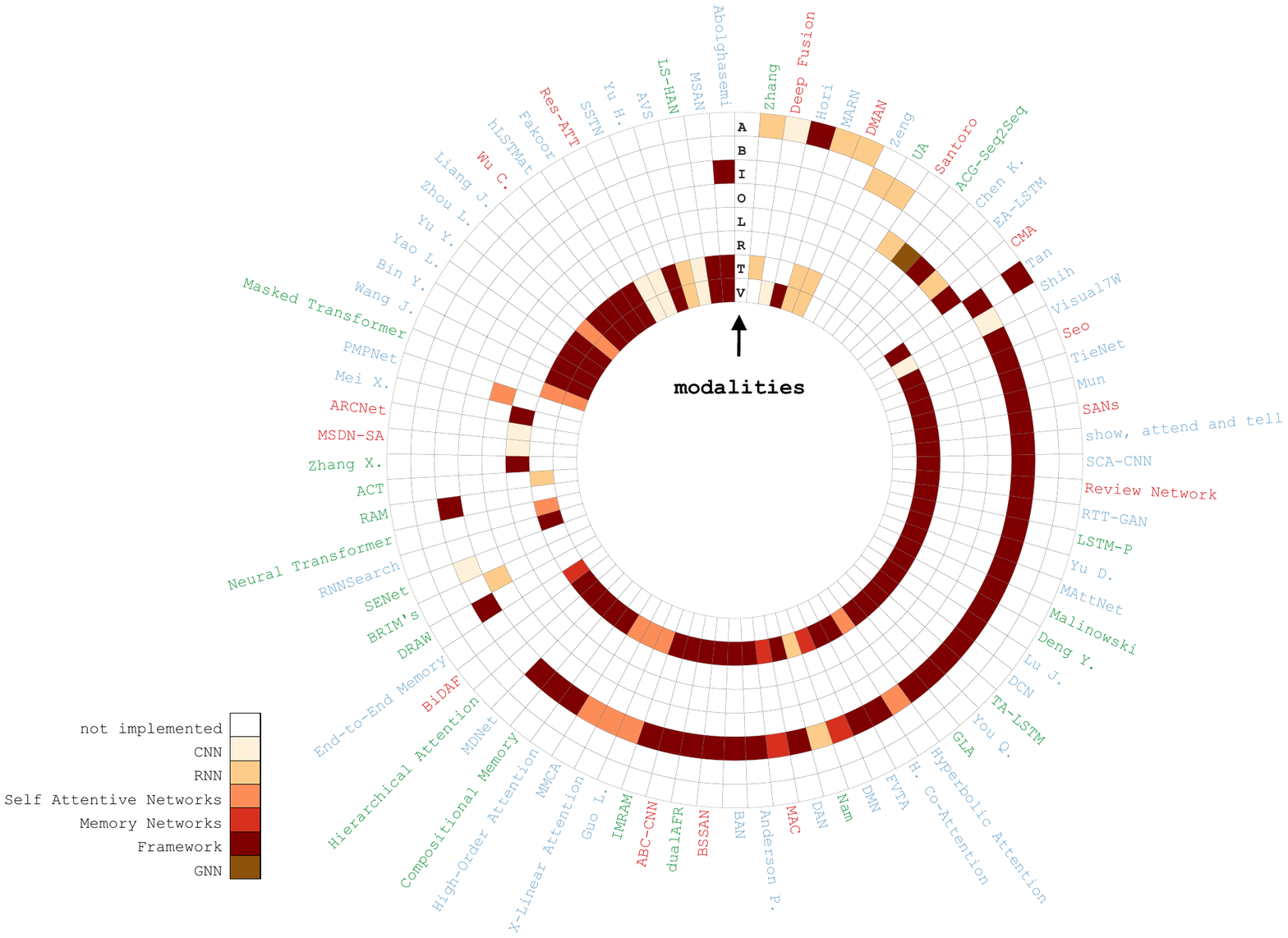}
    \caption{A diagram showing sensory modalities of neural attention models. Radial segments correspond to attention architectures, and each track corresponds to a modality. Modalities are: (A) audio, (B) biomedical signals, (I) image, (O) other sensors, (L) LiDAR, (R) remote sensing data, (T) text, and (V) video. The following coloring convention is used for the individual segments: white (the modality is not implemented), light yellow (CNN), light orange (RNN), orange (Self-attentive networks), red (Memory networks), dark red (framework), and brown (GNN). This diagram emphasizes multimodal architectures so that only the most representative single modality (i.e., text or image) architectures are shown. Most multimodal architectures use the image/text or video/text modalities.}
    \label{fig:modalities}
\end{figure}

The first widespread use of attention for multimodality occurs with the attentional interface between a convolutional neural network and an LSTM in image captioning~\cite{xu_show_2015}. In this model, a CNN processes the image, extracting high-level features, whereas the LSTM consumes the features to produce descriptive words, one by one. The attention mechanism guides the LSTM to relevant image information for each word's generation, equivalent to the human visual attention mechanism. The visualization of attention weights in multimodal tasks improved the understanding of how architecture works. This approach derived from countless other works with attentional interfaces that deal with video-text data~\cite{yao_describing_2015}~\cite{wu_hierarchical_2018}~\cite{fakoor2016memory}, image-text data~\cite{tian2018diagnostic}~\cite{pu_adaptive_2018}, monocular/RGB-D images~\cite{liu2017global}~\cite{zhang2018attention}~\cite{zhang2018adding}, RADAR~\cite{zhang2018attention}, remote sensing data~\cite{xiangrong_zhang;xin_wang;xu_tang;huiyu_zhou;chen_li_description_2019}~\cite{bei_fang;ying_li;haokui_zhang;jonathan_cheung-wai_chan_hyperspectral_2019}~\cite{qi_wang;shaoteng_liu;jocelyn_chanussot;xuelong_li_scene_2019}~\cite{xiaoguang_mei;erting_pan;yong_ma;xiaobing_dai;jun_huang;fan_fan;qinglei_du;hong_zheng;jiayi_ma_spectral-spatial_2019}, audio-video~\cite{hori_attention-based_2017}~\cite{zhang2019deep}, and diverse sensors~\cite{zadeh2018memory}~\cite{zadeh2018multi}~\cite{santoro2018relational}, as shown in Figure~\ref{fig:modalities}.

Zhang et al.~\cite{zheng_zhang;lizi_liao;minlie_huang;xiaoyan_zhu;tat-seng_chua_neural_2019} used an adaptive attention mechanism to learn to emphasize different visual and textual sources for dialogue systems for fashion retail. An adaptive attention scheme
automatically decided the evidence source for tracking dialogue states
based on visual and textual context. \textbf{Dual Attention Networks}~\cite{nam_dual_2017} presented attention mechanisms to capture the fine-grained interplay between images and textual information. The mechanism allows visual and textual attention to guide each other during collaborative inference. \textbf{HATT}~\cite{wu_hierarchical_2018} presented a new attention-based hierarchical fusion to explore the complementary features of multimodal features progressively, fusing temporal, motion, audio, and semantic label features for video representation. The model consists of three attention layers. First, the low-level attention layer deals with temporal, motion, and audio features inside each modality and across modalities. Second, high-level attention selectively focuses on semantic label features. Finally, the sequential attention layer incorporates hidden information generated by encoded low-level attention and high-level attention. Hori et. al.~\cite{hori_attention-based_2017} extended simple attention multimodal fusion. Unlike the simple multimodal fusion method, the feature-level attention weights can change according to the decoder state and the context vectors, enabling the decoder network to pay attention to a different set of features or modalities when predicting each subsequent word in the description. Memory Fusion Network~\cite{zadeh2018memory} presented the Delta-memory Attention module for multi-view sequential learning. First, an LSTM system, one for each of the modalities, encodes the modality-specific dynamics and interactions. Delta-memory attention discovers both cross-modality and temporal interactions in different memory dimensions of LSTMs. Finally, Multi-view Gated Memory (unifying memory) stores the cross-modality interactions over time.

Huang et al.~\cite{noauthor_bi-directional_nodate} investigated the problem of matching image-text by exploiting the bi-directional attention with fine-granularity correlations between visual regions and textual words. \textbf{Bi-directional attention} connects the word to regions and objects to words for learning mage-text matching. Li et. al.~\cite{yehao_li;ting_yao;yingwei_pan;hongyang_chao;tao_mei_pointing_2019} introduced \textbf{Long Short-Term Memory with Pointing} (LSTM-P) inspired by humans pointing behavior~\cite{matthews2012origins}, and Pointer Networks~\cite{vinyals2015pointer}. The pointing mechanism encapsulates dynamic contextual information (current input word and LSTM cell output) to deal with the image captioning scenario's novel objects. Liu et. al.~\cite{noauthor_improving_nodate} proposed a cross-modal attention-guided erasing approach for referring expressions. Previous attention models focus on only the most dominant features of both modalities and neglect textual-visual correspondences between images and referring expressions. To tackle this issue, cross-modal attention discards the most dominant information from either textual or visual domains to generate difficult training samples and drive the model to discover complementary textual-visual correspondences. Abolghasemi et al.~\cite{pay_attention} demonstrated an approach for augmenting a deep visuomotor policy trained through demonstrations with \textbf{Task Focused Visual Attention} (TFA). Attention receives as input a manipulation task specified in natural language text, an image with the environment, and returns as output the area with an object that the robot needs to manipulate. TFA allows the policy to be significantly more robust from the baseline policy, i.e., no visual attention. Pu et al.~\cite{pu_adaptive_2018} adaptively select features from the multiple CNN layers for video captioning. Previous models often use the output from a specific layer of a CNN as video features. However, this attention model adaptively and sequentially focuses on different layers of CNN features.

\subsection{Attention-augmented memory}
\label{sub:attention_memory}

Attentional interfaces also allow the neural network iteration with other cognitive elements (i.e., memories, working memory). Memory control and logic flow are essential for learning. However, they are elements that do not exist in classical architectures. The memory of classic RNNs, encoded by hidden states and weights, is usually minimal and is not sufficient to remember facts from the past accurately. Most Deep Learning models do not have a simple way to read and write data to an external memory component. The \textbf{Neural Turing Machine} (NTM)~\cite{graves_neural_2014} and Memory Networks (MemNN)~\cite{weston_2014_memory} - a new class of neural networks - introduced the possibility for a neural network dealing with addressable memory. NTM is a differentiable approach that can be trained with gradient descent algorithms, producing a practical learning program mechanism. NTM memory is a short-term storage space for information with its rules-based manipulation. Computationally, these rules are simple programs, where data are those programs' arguments. Therefore, an NTM resembles a working memory designed to solve tasks that require rules, where variables are quickly linked to memory slots. NTMs use an attentive process to read and write elements to memory selectively. This attentional mechanism makes the network learn to use working memory instead of implementing a fixed set of symbolic data rules.

\begin{figure}[htb]
  \centering
  \includegraphics[width=0.80\textwidth]{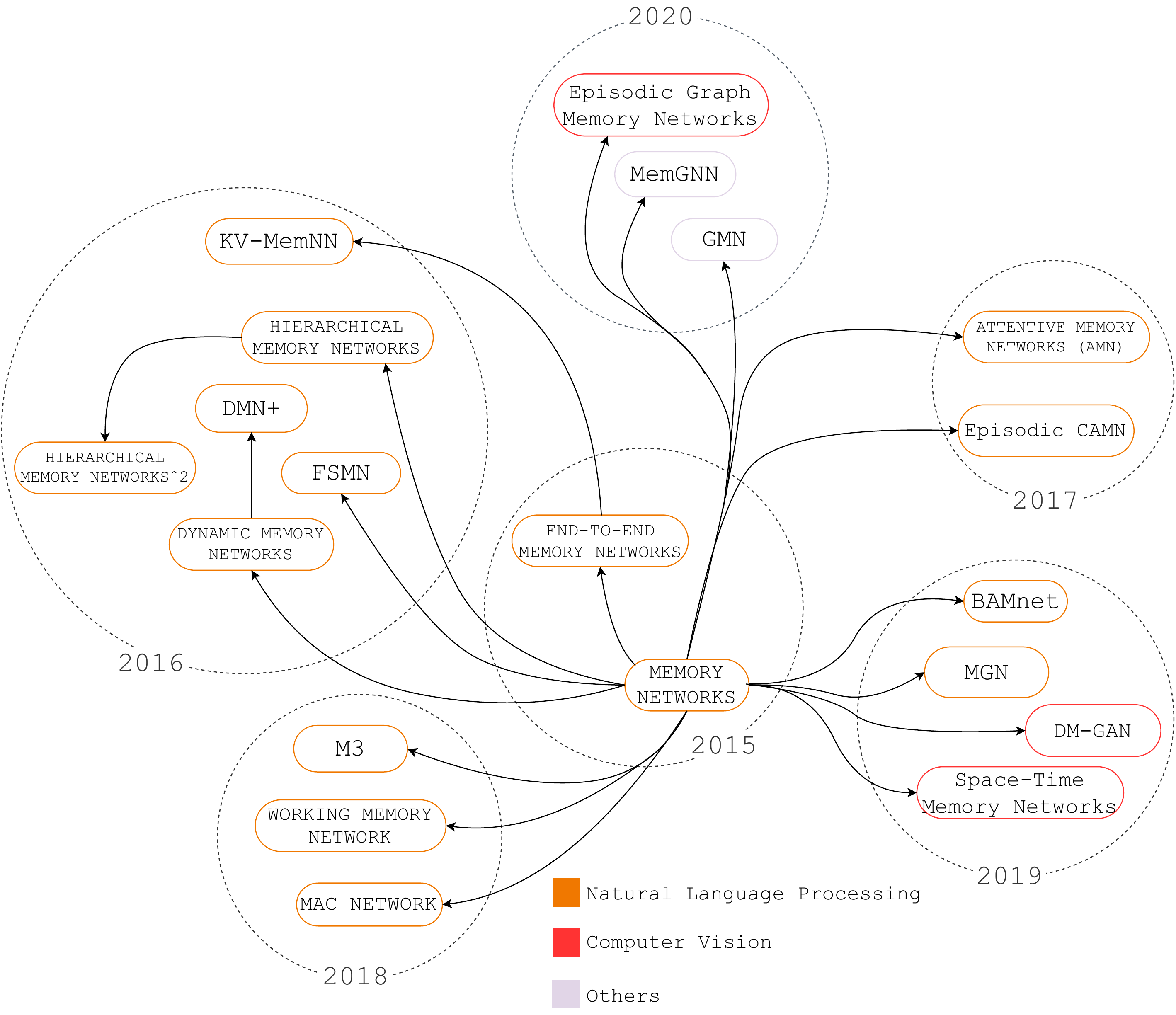}
  \caption{Memory-based neural networks (MemNN). Architectures labels are color-coded as follows: orange - natural language processing, red - computer vision, purple - others. The end-to-End Memory networks is the first end-to-end differentiable version of MemNN. GMN~\cite{khasahmadi2020memory} and MemGNN~\cite{khasahmadi2020memory} are the first graph networks with memory. DMN~\cite{xiong_dynamic_2016}, MemGNN~\cite{khasahmadi2020memory}, Episodic graph memory networks~\cite{lu2020video}, Episodic CAMN~\cite{abdulnabi2017episodic}, are the first instances of the episodic memory framework.}
  \label{fig:memory_networks_family}
\end{figure}

\textbf{Memory Networks}~\cite{weston_2014_memory} are a relatively new framework of models designed to alleviate the problem of learning long-term dependencies in sequential data by providing an explicit memory representation for each token in the sequence. Instead of forgetting the past, Memory Networks explicitly consider the input history, with a dedicated vector representation for each history element, effectively removing the chance to forget. The limit on memory size becomes a hyper-parameter to tune, rather than an intrinsic limitation of the model itself. This model was used in question-answering tasks where the long-term memory effectively acts as a (dynamic) knowledge base, and the output is a textual response. Large-scale question-answer tests were performed, and the reasoning power of memory networks that answer questions that require an in-depth analysis of verb intent was demonstrated. Mainly due to the success of MemNN, networks with external memory are a growing research direction in DL, with several branches under development as shown in figure~\ref{fig:memory_networks_family}.

\textbf{End-to-end Memory Networks}~\cite{sukhbaatar2015end} is the first version of MemNN applicable to realistic, trainable end-to-end scenarios, which requires low supervision during training. Aug Oh. et al.~\cite{oh2019video} extends Memory Networks to suit the task of semi-supervised segmentation of video objects. Frames with object masks are placed in memory, and a frame to be segmented acts as a query. The memory is updated with the new masks provided and faces challenges such as changes, occlusions, and accumulations of errors without online learning. The algorithm acts as an attentional space-time system calculating when and where to meet each query pixel to decide whether the pixel belongs to a foreground object or not. Kumar et al.~\cite{kumar_ask_2015} propose the first network with episodic memory - a type of memory extremely relevant to humans - to iterate over representations emitted by the input module updating its internal state through an attentional interface. In~\cite{lu2020video}, an episodic memory with a key-value retrieval mechanism chooses which parts of the input to focus on thorough attention. The module then produces a summary representation of the memory, taking into account the query and the stored memory. Finally, the latest research has invested in Graph Memory Networks (GMN), which are memories in GNNs~\cite{wu2020comprehensive}, to better handle unstructured data using key-value structured memories~\cite{miller2016key}~\cite{khasahmadi2020memory}~\cite{khasahmadi2020memory}~\cite{moon2019memory}.

\subsection{End-to-end attention models}
\label{sub:end_to_end_attention_models}
In mid-2017, research aiming at end-to-end attention models appeared in the area. The Neural Transformer (NT)~\cite{vaswani_attention_2017} and Graph Attention Networks~\cite{velickovic_graph_2018} - purely attentional architectures - demonstrated to the scientific community that attention is a key element for the future development in Deep Learning. The Transformer's goal is to use self-attention (Section~\ref{sub:attention_mechanisms}) to minimize traditional recurrent neural networks' difficulties. The \textbf{Neural Transformer} is the first neural architecture that uses only attentional modules and fully-connected neural networks to process sequential data successfully. It dispenses recurrences and convolutions, capturing the relationship between the sequence elements regardless of their distance. Attention allows the Transformer to be simple, parallelizable, and low training cost~\cite{vaswani_attention_2017}. \textbf{Graph Attention Networks} (GATs) are an end-to-end attention version of GNNs~\cite{wu2020comprehensive}. They have stacks of attentional layers that help the model focus on the unstructured data's most relevant parts to make decisions. The main purpose of attention is to avoid noisy parts of the graph by improving the signal-to-noise ratio (SNR) while also reducing the structure's complexity. Furthermore, they provide a more interpretable structure for solving the problem. For example, when analyzing the Attention of a model under different components in a graph, it is possible to identify the main factors contributing to achieving a particular response condition.

\begin{figure}[H]
  \centering
  \includegraphics[width=\textwidth]{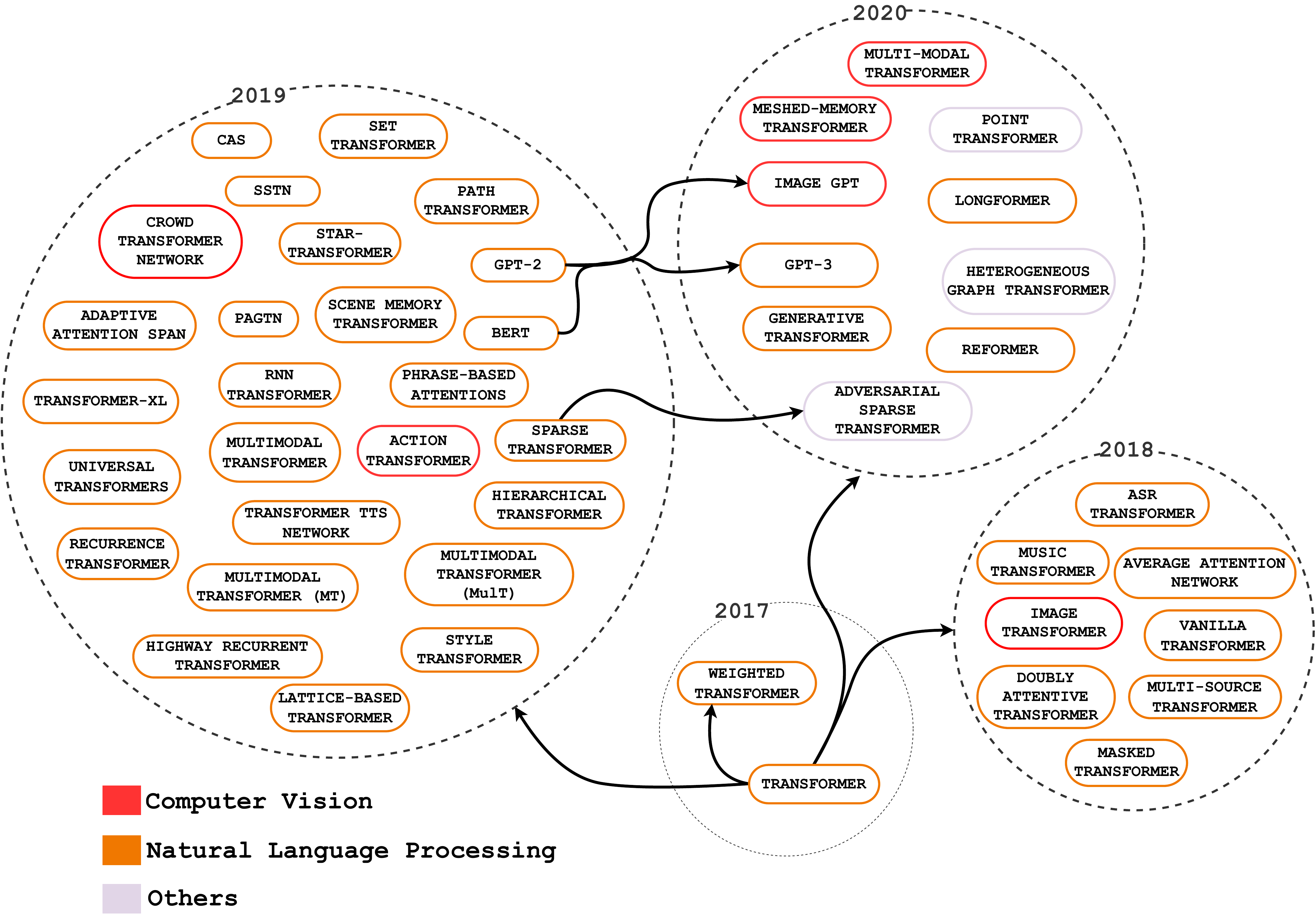}
  \caption{Transformer-based neural networks. Architectures labels are color-coded as follows: orange - natural language processing, red - computer vision, purple - others.}
  \label{fig:transformer_family}
\end{figure}

There is a growing interest in NT and GATs, and some extensions have been proposed~\cite{wang2019heterogeneous}~\cite{wang2019graph}~\cite{abu2018watch}~\cite{li2019relation}, with numerous Transformer-based architectures as shown figure~\ref{fig:transformer_family}. These architectures and all that use self-attention belong to a new category of neural networks, called Self-Attentive Neural Networks. They aim to explore self-attention in various tasks and improve the following drawbacks: 1) a Large number of parameters and training iterations to converge; 2) High memory cost per layer and quadratic growth of memory according to sequence length; 3) Auto-regressive model; 4) Low parallelization in the decoder layers. Specifically, Weighted Transformer~\cite{weighted_transformer} proposes modifications in the attention layers achieving a 40 \% faster convergence. The multi-head attention modules are replaced by modules called branched attention that the model learns to match during the training process. The Star-transformer~\cite{qipeng_guo;xipeng_qiu;pengfei_liu;yunfan_shao;xiangyang_xue;zheng_zhang_star-transformer_2019} proposes a lightweight alternative to reduce the model's complexity with a star-shaped topology. To reduce the cost of memory, Music Transformer~\cite{music_transformer}, and Sparse Transformer~\cite{rewon_child;scott_gray;alec_radford;ilya_sutskever_generating_2019} introduces relative self-attention and factored self-attention, respectively. Lee et al.~\cite{lee2018set} also features an attention mechanism that reduces self-attention from quadratic to linear, allowing scaling for high inputs and data sets.

Some approaches adapt the Transformer to new applications and areas. In natural language processing, several new architectures have emerged, mainly in multimodal learning. Doubly Attentive Transformer~\cite{doubly_attentive_transformer} proposes a multimodal machine-translation method, incorporating visual information. It modifies the attentional decoder, allowing textual features from a pre-trained CNN encoder and visual features. The Multi-source Transformer~\cite{multi_source_transformer} explores four different strategies for combining input into the multi-head attention decoder layer for multimodal translation. Style Transformer~\cite{style_transformer}, Hierarchical Transformer~\cite{hierarchical_transformer}, HighWay Recurrent Transformer~\cite {highway_transformer}, Lattice-Based Transformer~\cite {lattice_transformer}, Transformer TTS Network~\cite {li2019neural}, Phrase-Based Attention~\cite{phrase_attention} are some important architectures in style transfer, document summarization and machine translation. Transfer Learning in NLP is one of Transformer's major contribution areas. BERT~\cite{devlin_bert:_2018}, GPT-2~\cite{radford2019language}, and GPT-3~\cite{brown2020language} based NT architecture to solve the problem of Transfer Learning in NLP because current techniques restrict the power of pre-trained representations. In computer vision, the generation of images is one of the Transformer's great news. Image Transformer~\cite{parmar2018image}, SAGAN~\cite{zhang2018self}, and Image GPT~\cite{chen2020generative} uses self-attention mechanism to attend the local neighborhoods. The size of the images that the model can process in practice significantly increases, despite maintaining significantly larger receptive fields per layer than the typical convolutional neural networks. Recently, at the beginning of 2021, OpenAi introduced the scientific community to DALL·E~\cite{unpublished2021dalle}, the Newest language model based on Transformer and GPT-3, capable of generating images from texts extending the knowledge of GPT-3 for viewing with only 12 billions of parameters.

\subsection{Attention today}
\label{sub:attention_today}

\begin{figure*}[ht]
 \centering
 \includegraphics[width=\textwidth]{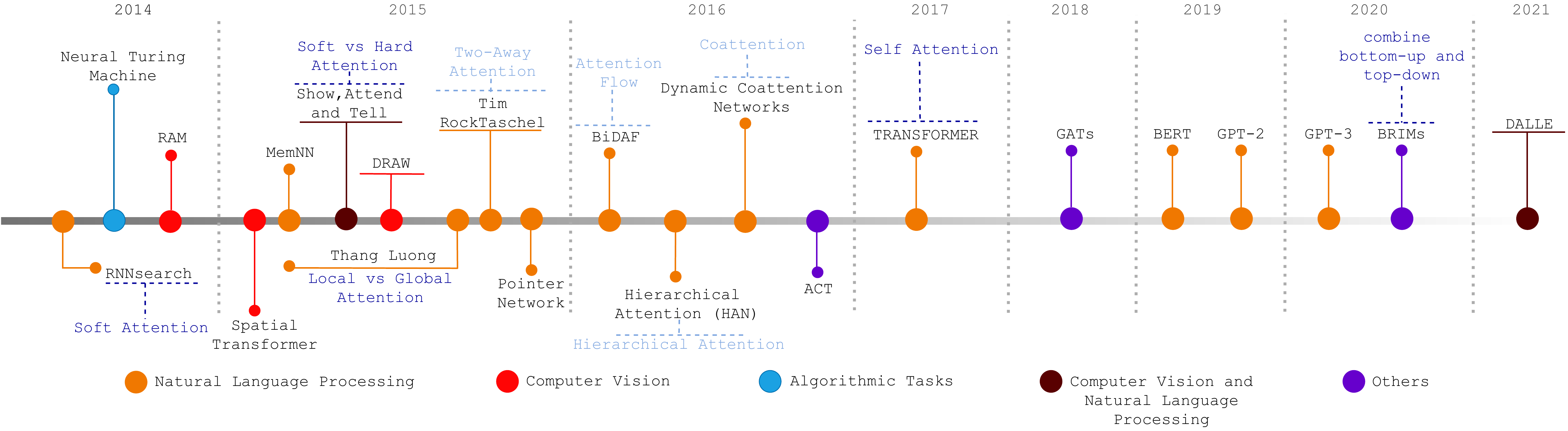}
 \caption{Key developments in Attention in DL Timeline. RNNSearch presented the first attention mechanism. Neural Turing machine and Memory networks introduced memory and dynamic flow control. RAM and DRAW learned to combine multi-glimpse, visual attention, and sequential processing. Spatial Transformer introduced a module to increase the robustness of CNNs to variations in spatial transformations. Show, attend and tell created attention for multimodality. The Pointer network used attention as a pointer. BiDAF, HAN, and DCN presented attentional techniques to align data with different hierarchical levels. ACT introduced the computation time topic. Transformer~\cite{vaswani_attention_2017} was the first self-attentive neural network with an end-to-end attention approach. GATs introduced attention in GNNs. BERT~\cite{devlin_bert:_2018}, GPT-2~\cite{radford2019language}, GPT-3~\cite{brown2020language}, and DALL·E~\cite{unpublished2021dalle} are the state-of-the-art in language models and text-to-image generation. Finally, BRIMs~\cite{mittal2020learning} learned to combine bottom-up and top-down signals.}
 \label{fig:timeline}
\end{figure*}

Currently, hybrid models that employ the main key developments in attention's use in Deep Learning (Figure~\ref{fig:timeline}) have aroused the scientific community's interest. Mainly, hybrid models based on Transformer, GATs, and Memory Networks have emerged for multimodal learning and several other application domains. Hyperbolic Attention Networks (HAN)~\cite{gulcehre_hyperbolic_2018}, Hyperbolic Graph Attention Networks (GHN)~\cite{zhang2019hyperbolic}, Temporal Graph Networks (TGN)~\cite{rossi2020temporal} and Memory-based Graph Networks (MGN)~\cite{khasahmadi2020memory} are some of the most promising developments. Hyperbolic networks are a new class of architecture that combine the benefits of self-attention, memory, graphs, and hyperbolic geometry in activating neural networks to reason with high capacity over embeddings produced by deep neural networks. Since 2019 these networks have stood out as a new research branch because they represent state-of-the-art generalization on neural machine
translation, learning on graphs, and visual question answering tasks while keeping the neural representations compact. Since 2019, GATs have also received much attention due to their ability to learn complex relationships or interactions in a wide spectrum of problems ranging from biology, particle physics, social networks to recommendation systems. To improve the representation of nodes and expand the capacity of GATs to deal with data of a dynamic nature (i.e. evolving features or connectivity over time), architectures that combine memory modules and the temporal dimension, like MGNs and TGNs, were proposed.
 
At the end of 2020, two research branches still little explored in the literature were strengthened: 1) explicit combination of bottom-up and top-down stimuli in bidirectional recurrent neural networks and 2) adaptive computation time. Classic recurrent neural networks perform recurring iteration within a particular level of representation instead of using a top-down iteration, in which higher levels act at lower levels. However, Mittal et al.~\cite{mittal2020learning} revisited the bidirectional recurrent layers with attentional mechanisms to explicitly route the flow of bottom-up and top-down information, promoting selection iteration between the two levels of stimuli. The approach separates the hidden state into several modules so that upward iterations between bottom-up and top-down signals can be appropriately focused. The layer structure has concurrent modules so that each hierarchical layer can send information both in the bottom-up and top-down directions.

The adaptive computation time is an interesting little-explored topic in the literature that began to expand only in 2020 despite initial studies emerging in 2017. ACT applies to different neural networks (e.g. RNNs, CNNs, LSTMs, Transformers). The general idea is that complex data might require more computation to produce a final result, while some unimportant or straightforward data might require less. The attention mechanism dynamically decides how long to process network training data. The seminal approach by Graves et al.~\cite{graves_adaptive_2016} made minor modifications to an RNN, allowing the network to perform a variable number of state transitions and a variable number of outputs at each stage of the input. The resulting output is a weighted sum of the intermediate outputs, i.e., soft attention. A halting unit decides when the network should stop or continue. To limit computation time, attention adds a time penalty to the cost function by preventing the network from processing data for unnecessary amounts of time. This approach has recently been updated and expanded to other architectures. Spatially Adaptive Computation Time (SACT)~\cite{figurnov2017spatially} adapts ACT to adjust the per-position amount of computation to each spatial position of the block in convolutional layers, learning to focus computing on the regions of interest and to stop when the features maps are "good enough". Finally, Differentiable Adaptive Computation Time (DACT)~\cite{eyzaguirre2020differentiable} introduced the first differentiable end-to-end approach to computation time on recurring networks.
\section{Attention Mechanisms}
\label{sub:attention_mechanisms}

Deep attention mechanisms can be categorized into soft attention (global attention), hard attention (local attention), and self-attention (intra-attention). 

\begin{figure}[ht]
  \centering
  \includegraphics[width=0.90\textwidth]{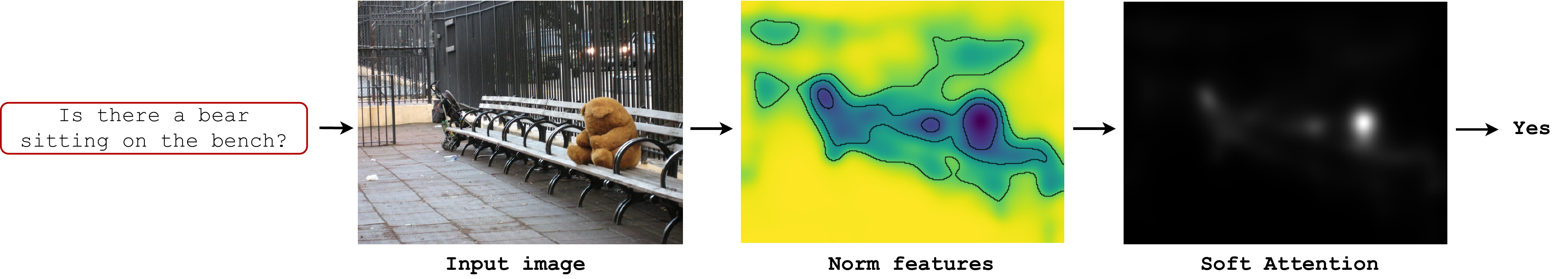}
  \caption{An intuitive example of Soft Attention. Visual QA architecture outputs an answer given an image and a textual question as input. It uses a soft attention mechanism that weighted visual features for the task for further processing. The premise is that the norm of the visual features correlates with their relevance. Besides, those feature vectors with high magnitudes correspond to image regions that contain relevant semantic content.}
  \label{fig:soft_attention_example}
\end{figure}

\textbf{Soft Attention}. Soft attention assigns a weight of 0 to 1 for each input element. It decides how much attention should be focused on each element, considering the interdependence between the input of the deep neural network's mechanism and target. It uses softmax functions in the attention layers to calculate weights so that the entire attentional model is deterministic and differentiable. Soft attention can act in the spatial and temporal context. The spatial context operates mainly to extract the features or the weighting of the most relevant features. For the temporal context, it works by adjusting the weights of all samples in sliding time windows, as samples at different times have different contributions. Despite being deterministic and differentiable, soft mechanisms have a high computational cost for large inputs. Figure~\ref{fig:soft_attention_example} shows an intuitive example of a soft attention mechanism.

\begin{figure}[ht]
  \centering
  \includegraphics[width=0.90\textwidth]{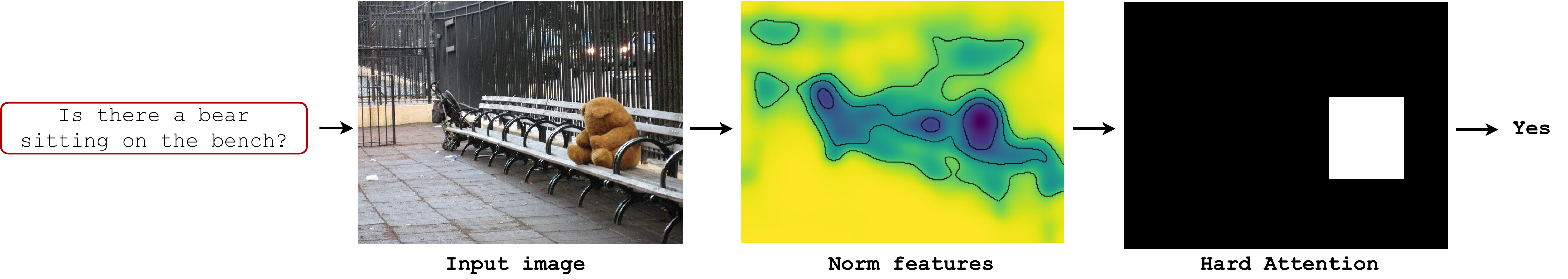}
  \caption{An intuitive example of Hard Attention. Given an image and a textual question as input, the Visual QA architecture outputs an answer. It uses a hard attention mechanism that \textbf{selects only} the important visual features for the task for further processing. }
  \label{fig:hard_attention_example}
\end{figure}

\textbf{Hard Attention}. Hard attention determines whether a part of the mechanism's input should be considered or not, reflecting the interdependence between the input of the mechanism and the target of the deep neural network. The weight assigned to an input part is either 0 or 1. Hence, as input elements are either seen, the objective is non-differentiable. The process involves making a sequence of selections on which part to attend. In the temporal context, for example, the model attends to a part of the input to obtain information, decidinng where to attend in the next step based on the known information. A neural network can make a selection based on this information. However, as there is no ground truth to indicate the correct selection policy, the hard-attention type mechanisms are represented by stochastic processes. As the model is not differentiable, reinforcement learning techniques are necessary to train models with hard attention. Inference time and  computational costs are reduced compared to soft mechanisms once the entire input is not being stored or processed. Figure~\ref{fig:hard_attention_example} shows an intuitive example of a hard attention mechanism.

\begin{figure}[ht]
  \centering
  \includegraphics[width=0.90\textwidth]{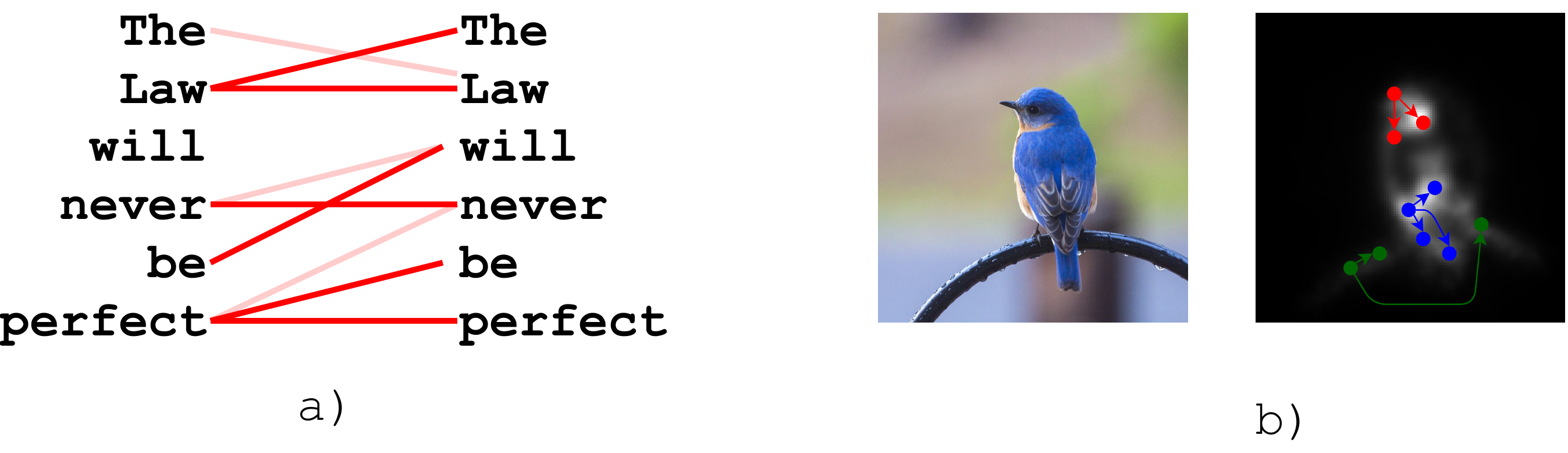}
  \caption{Self-Attention examples. a) Self-attention in sentences
  b) Self-attention in images. The first image shows five representative query locations with color-coded dots with the corresponding color-coded arrows summarizing the most-attended regions.}
  \label{fig:self_attention_example}
\end{figure}

\textbf{Self-Attention}. Self-attention quantifies the interdependence between the input elements of the mechanism. This mechanism allows the inputs to interact with each other "self" and determine what they should pay more attention to. The self-attention layer's main advantages compared to soft and hard mechanisms are parallel computing ability for a long input. This mechanism layer checks the attention with all the same input elements using simple and easily parallelizable matrix calculations. Figure~\ref{fig:self_attention_example} shows an intuitive example of a self-attention mechanism.

\section{Attention-based Classic Deep Learning Architectures}
\label{sub:attention_classic_architectures}

This section introduces details about attentional interfaces in classic DL architectures. Specifically, we present the uses of attention in convolutional, recurrent networks and generative models.


\subsection{Attention-based Convolutional Neural Networks (CNNs)}
\label{attention_cnns}
\begin{figure}[ht]
\begin{center}
\includegraphics[width=0.95\linewidth]{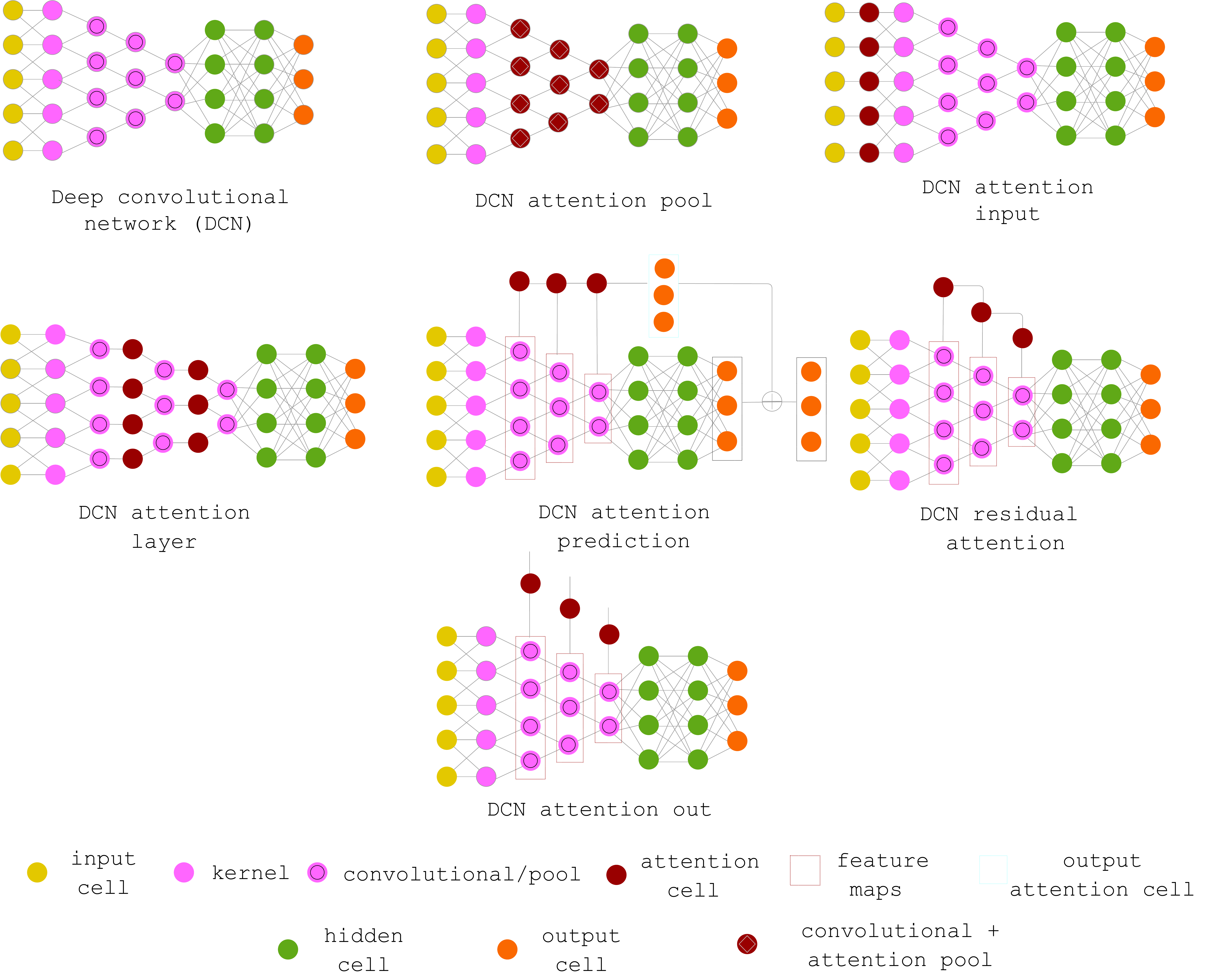}
\caption{Attention-based convolutional neural networks. DCN attention pool group uses an attention pool, instead of regular pooling, as a strategy to determine the importance of each individual in a given feature map window. The premise is that only a few of these windows are significant and must be selected concerning a particular objective. DCN attention input group uses structures similar to the human's visual attention. DCN attention layer group collects important stimuli of high (semantic level) and low level (saliency) for subsequent layers of architecture. DCN attention prediction group uses attention in the final stages of prediction, sometimes as an ensemble element. DCN residual attention group uses attention as a residual module between any convolutional layers to mitigate the vanishing problem, capturing only the relevant stimuli from each feature map. DCN attention out-group can represent the category of recurrent attention processes.}
\label{fig:dcn_atention}
\end{center}
\end{figure}

Attention emerges in CNNs to filter information and allocate resources to the neural network efficiently. There are numerous ways to use attention on CNNs, which makes it very difficult to summarize how this occurs and the impacts of each use. We divided the uses of attention into six distinct groups (Figure~\ref{fig:dcn_atention}): 1) DCN attention pool -- attention replaces the classic CNN pooling mechanism; 2) DCN attention input -- the attentional modules are filter masks for the input data. This mask assigns low weights to regions irrelevant to neural network processing and high weights to relevant areas; 3) DCN attention layer -- 
attention is between the convolutional layers; 4) DCN attention prediction -- attentional mechanisms assist the model directly in the prediction process; 5) DCN residual attention --  extracts information from the feature maps and presents a residual input connection to the next layer; 6) DCN attention out -- attention captures important stimuli of feature maps for other architectures, or other instances of the same architecture. To maintain consistency with the Deep Neural Network's area, we extend \textbf{The Neural Network Zoo} schematics~\footnotemark[1] to accommodate attention elements.

\footnotetext[1]{https://www.asimovinstitute.org/neural-network-zoo/}

 \textbf{DCN attention input} mainly uses attention to filter input data - a structure similar to the multi-glimpse mechanism and visual attention of human beings. Multi-glimpse refers to the ability to quickly scan the entire image and find the main areas relevant to the recognition process, while visual attention focuses on a critical area by extracting key features to understand the scene. When a person focuses on one part of the image, the different regions' internal relationship is captured, guiding eye movement to find the next relevant area—ignoring the irrelevant parts easy learning in the presence of disorder. For this reason, human vision has an incomparable performance in object recognition. The main contribution of attention at the CNNs' input is robustness. If our eyes see an object in a real-world scene, parts far from the object are ignored. Therefore, the distant background of the fixed object does not interfere in recognition. However, CNNs treat all parts of the image equally. The irrelevant regions confuse the classification and make it sensitive to visual disturbances, including background, changes in camera views, and lighting conditions. Attention in CNNs' input contributes to increasing robustness in several ways: 1) It makes architectures more scalable, in which the number of parameters does not vary linearly with the size of the input image; 2) Eliminates distractors; 3) Minimizes the effects of changing camera lighting, scale, and views. 4) It allows the extension of models for more complex tasks, i.e., fine-grained classification or segmentation. 5) Simplifies CNN encoding. 6) Facilitates learning by including relevant priorities for architecture.


Zhao et al.~\cite{zhao_deep_2017} used visual attention-based image processing to generate the focused image. Then, the focused image is input into CNN to be classified. According to the classification, the information entropy guides reinforcement learning agents to achieve a better image classification policy. Wang et al.~\cite{x._wang;_l._gao;_j._song;_h._shen_beyond_2017} used attention to create representations rich in motion information for action recognition. The attention extracts saliency maps using both motion and appearance information to calculate the objectness scores. For a video, attention process frame by frame to generate a saliency-aware map for each frame. The classic pipeline uses only CNN sequence features as input for LSTMs, failing to capture adjacent frames' motion information. The saliency-aware maps capture only regions with relevant movements making CNN encoding simple and representative for the task. Liu et al.~\cite{ning_liu;yongchao_long;changqing_zou;qun_niu;li_pan;hefeng_wu_adcrowdnet:_2019} used attention as input of a CNN to provide important priors in counting crowded tasks. An attention map generator first provides two types of priors for the system: candidate crowd regions and crowd regions' congestion degree. The priors guide subsequent CNNs to pay more attention to those regions with crowds and improving their capacity to be resistant to noise. Specifically, the congestion degree prior provides fine-grained density estimation for a system.

\begin{figure}[ht]
\begin{center}
    \includegraphics[width=1\linewidth]{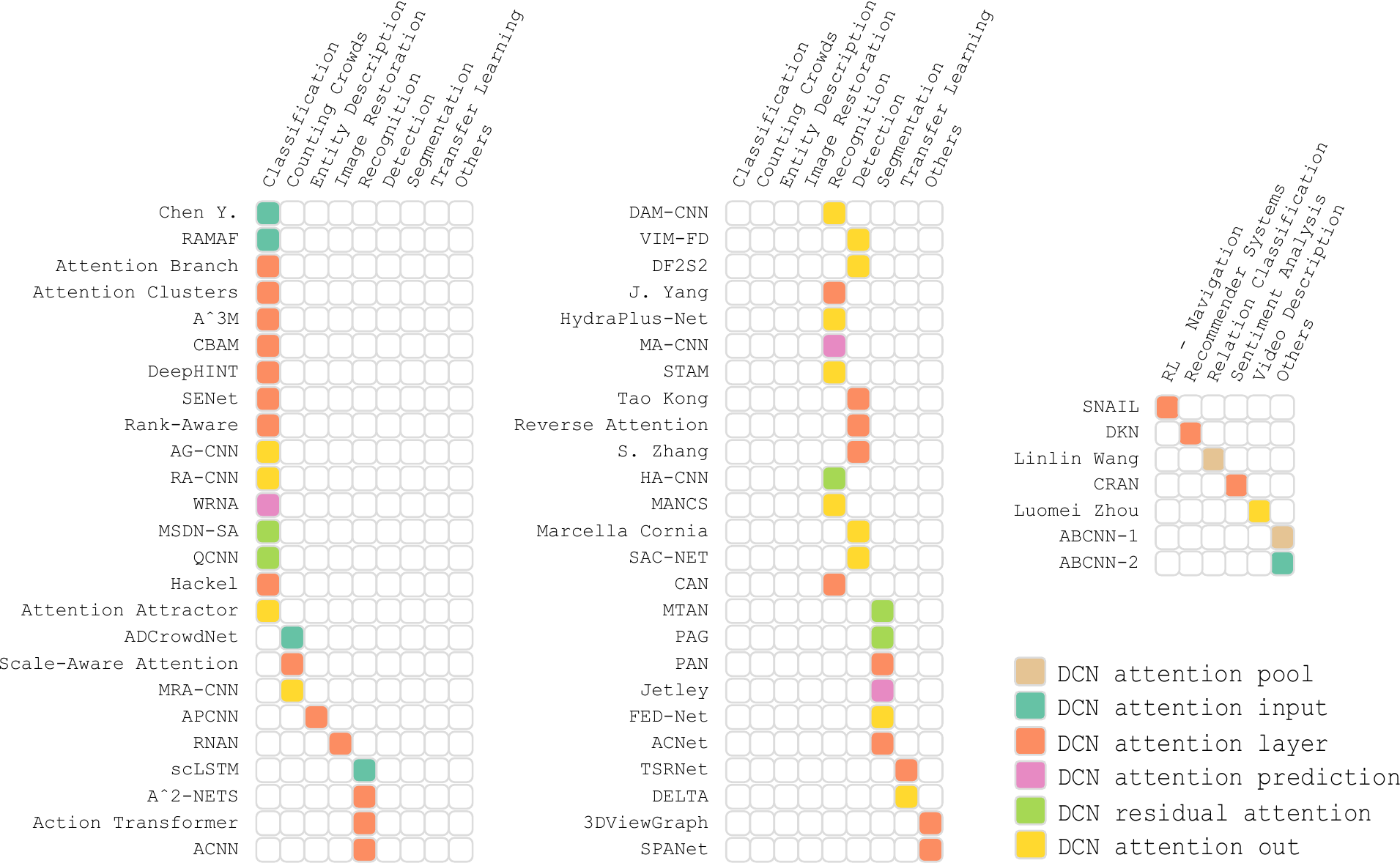}
\caption{Attention-based convolutional neural networks main architectures by task and attention use.}
\label{fig:dcns_architectures}
\end{center}
\end{figure}

In classic CNNs, the size of the receptive fields is relatively small. Most of them extract features locally with convolutional operations, which fail to capture long-range dependencies between pixels throughout the image. However, larger receptive fields allow for better use of training inputs, and much more context information is available at the expense of instability or even convergence in training. Also, traditional CNNs treat channel features equally. This naive treatment lacks the flexibility to deal with low and high-frequency information. Some frequencies may contain more relevant information for a task than others, but equal treatment by the network makes it difficult to converge the models. To mitigate such problems, most literature approaches use attention between convolutional layers (i.e., \textbf{DCN attention layer} and \textbf{DCN residual attention}), as shown in figure~\ref{fig:dcns_architectures}.  Between layers, attention acts mainly for feature recalibration, capturing long-term dependencies, internalizing, and correctly using past experiences.

The pioneering approach to adopting attention between convolutional layers is the Squeeze-and-Excitation Networks~\cite{hu_squeeze-and-excitation_2017} created in 2016 and winner of the ILSVRC in 2017. It is also the first architecture to model channel interdependencies to recalibrate filter responses in two steps, squeeze and excitation, i.e., SE blocks. To explore local dependencies, the squeeze module encodes spatial information into a channel descriptor. The output is a collection of local descriptors with expressive characteristics for the entire image. To make use of the information aggregated by the squeeze operation, excitation captures channel-wise dependencies by learning a non-linear and non-mutually exclusive relationship between channels, ensuring that multiple channels can be emphasized. In this sense, SE blocks intrinsically introduce attentional dynamics to boost feature discrimination between convolutional layers.

The inter-channel and intra-channel attention to capturing long-term dependencies and simultaneously taking advantage of high and low-level stimuli are widely explored in the literature. Zhang. et al.~\cite{zhang2019residual} proposed residual local and non-local attention
blocks consisting of trunk and mask branches. Their attention mechanism helps to learn local and non-local information from the hierarchical features, further preserving low-level features while maintaining a representational quality of high-level features. The Cbam~\cite{sanghyun_woo;jongchan_park;joon-young_lee;in_so_kweon_cbam:_2018} infers attentional maps in two separate dimensions, channel and spatial, for adaptive feature refinement. The double attention block in~\cite{chen_^2-nets:_2018} aggregates and propagates global informational features considering the entire spatio-temporal context of images and videos, allowing subsequent convolution layers to access resources from across space efficiently. In the first stage, attention gathers features from all space into a compact set employing groupings. In the second stage, it selects and adaptively distributes the resources for each architectural location. Following similar exploration proposals, several attentional modules can be easily plugged into classic CNNs~\cite{fukui2019attention}~\cite{han20193dviewgraph}~\cite{ji2017distant}~\cite{yang2017neural}.


Hackel et al.~\cite{hackel_inference_2018} explored attention to preserving sparsity in convolutional operations. Convolutions with kernels greater than $1 \times 1$ generate fill-in, reducing feature maps' sparse nature. Generally, the change in data sparsity has little influence on the network output, but memory consumption and execution time considerably increase when it occurs in many layers. To guarantee low memory consumption, attention acts as a $k-selection$ filter, which has two different versions of selection: 1) it acts on the output of the convolution, preferring the largest $k$ positive responses similar to a rectified linear unit; 2) it chooses the $k$ highest absolute values, expressing a preference for responses of great magnitude. The parameter $k$ controls the level of sparse data and, consequently, computational resources during training and inference. Results point out that training with attentional control of data sparsity can reduce in more than $200\%$ the forward pass runtime in one layer.

To previous aggregate information and dynamically point to past experiences, SNAIL~\cite{mishra_simple_2017} - a pioneering class of meta-learner based attention architectures - has proposed combining temporal convolutions with soft attention. This approach demonstrates that attention acts as a complement to the disadvantages of convolution. Attention allows precise access in an infinitely large context, while convolutions provide high-bandwidth access at the expense of a finite context. By merging convolutional layers with attentional layers, SNAIL can have unrestricted access to the number of previous experiences effectively, as well as the model can learn a more efficient representation of features. As additional benefits, SNAIL architectures become simpler to train than classic RNNs.

The \textbf{DCN attention out-group} uses attention to share relevant feature maps with other architectures or even with instances of the current architecture. Usually, the main objective is to facilitate the fusion of features, multimodality, and external knowledge. In some cases, attention regularly works by turning classic CNNs into recurrent convolutional neural networks - a new trend in Deep Learning to deal with challenging images' problems. RA-CNN~\cite{fu_look_2017} is a pioneering framework for recurrent convolutional networks. In their framework, attention proceeds along two dimensions, i.e., discriminative feature learning and sophisticated part localization. Given an input image, a classic CNN extracts feature maps, and the attention proposal network maps convolutional features to a feature vector that could be matched with the category entries. Then, attention estimates the focus region for the next CNN instance, i.e., the next finer scale. Once the focus region is located, the system cuts and enlarges the region to a finer scale with higher resolution to extract more refined features. Thus, each CNN in the stack generates a prediction so that the stack's deepest layers generate more accurate predictions.

For merging features, Cheng. et. al.~\cite{xueying_chen;rong_zhang;pingkun_yan_feature_2019} presented Feature-fusion Encoder-Decoder Network (FED-net) to image segmentation. Their model uses attention to fuse features of different levels of an encoder. At each encoder level, the attention module merges features from its current level with features from later levels. After the merger, the decoder performs convolutional upsampling with the information from each attention level, which contributes by modulating the most relevant stimuli for segmentation. Tian et al.~\cite{tian2018learning} used feature pyramid-based attention to combine meaningful semantic features with semantically weak but visually strong features in a face detection task. Their goal is to learn more discriminative hierarchical features with enriched semantics and details at all levels to detect hard-to-detect faces, like tiny or partially occluded faces. Their attention mechanism can fuse different feature maps from top to bottom recursively by combining transposed convolutions and element-wise multiplication maximizing mutual information between the lower and upper-level representations.

Delta~\cite{xingjian_li;haoyi_xiong;hanchao_wang;yuxuan_rao;liping_liu;jun_huan_delta:_2019} framework presented an efficient strategy for transfer learning. Their attention system acts as a behavior regulator between the source model and the target model. The attention identifies the source model's completely transferable channels, preserving their responses and identifying the non-transferable channels to dynamically modulate their signals, increasing the target model's generalization capacity. Specifically, the attentional system characterizes the distance between the source/target model through the feature maps' outputs and incorporates that distance to regularize the loss function. Optimization normally affects the weights of the neural network and assigns generalization capacity to the target model. Regularization modulated by attention on high and low semantic stimuli manages to take important steps in the semantic problem to plug in external knowledge.

The \textbf{DCN attention prediction} group uses attention directly in the prediction process. Various attentional systems capture features from different convolutional layers as input and generate a prediction as an output. Voting between different predictors generates the final prediction. Reusing activations of CNNs feature maps to find the most informative parts of the image at different depths makes prediction tasks more discriminative. Each attentional system learns to relate stimuli and part-based fine-grained features, which, although correlated, are not explored together in classical approaches. Zheng et al.~\cite{zheng2017learning} proposed a multi-attention mechanism to group channels, creating part classification sub-networks. The mechanism takes as input feature maps from convolutional layers and generates multiple single clusters spatially-correlated subtle patterns as a compact representation. The sub-network classifies an image by each individual part. The attention mechanism proposed in~\cite{rodriguez_painless_2018} uses a similar approach. However, instead of grouping features into clusters, the attentional system has the most relevant feature map regions selected by the attention heads. The output heads generate a hypothesis given the attended information, and the confidence gates generate a confidence score for each attention head.

Finally, the \textbf{DCN attention pool} group replaces classic pooling strategies with attention-based pooling. The objective is to create a non-linear encoding to select only stimuli relevant to the task, given that classical strategies select only the most contrasting stimuli. To modulate the resulting stimuli, attentional pooling layers generally capture different relationships between feature maps or between different layers. For example, Wang et al.~\cite{linlin_wang_zhu_cao_gerard_de_melo_zhiyuan_liu:_relation_nodate} created an attentional mechanism that captures pertinent relationships between convoluted context windows and the relation class embedding through a correlation matrix learned during training. The correlation matrix modulates the convolved windows, and finally, the mechanism selects only the most salient stimuli. A similar approach is also followed in~\cite{yin2016abcnn} for modeling sentence pairs.

\subsection{Attention-based Recurrent Neural Networks (RNNs)}
\label{subsec:attention_rnn}

\begin{figure}[ht]
\begin{center}
    \includegraphics[width=0.95\linewidth]{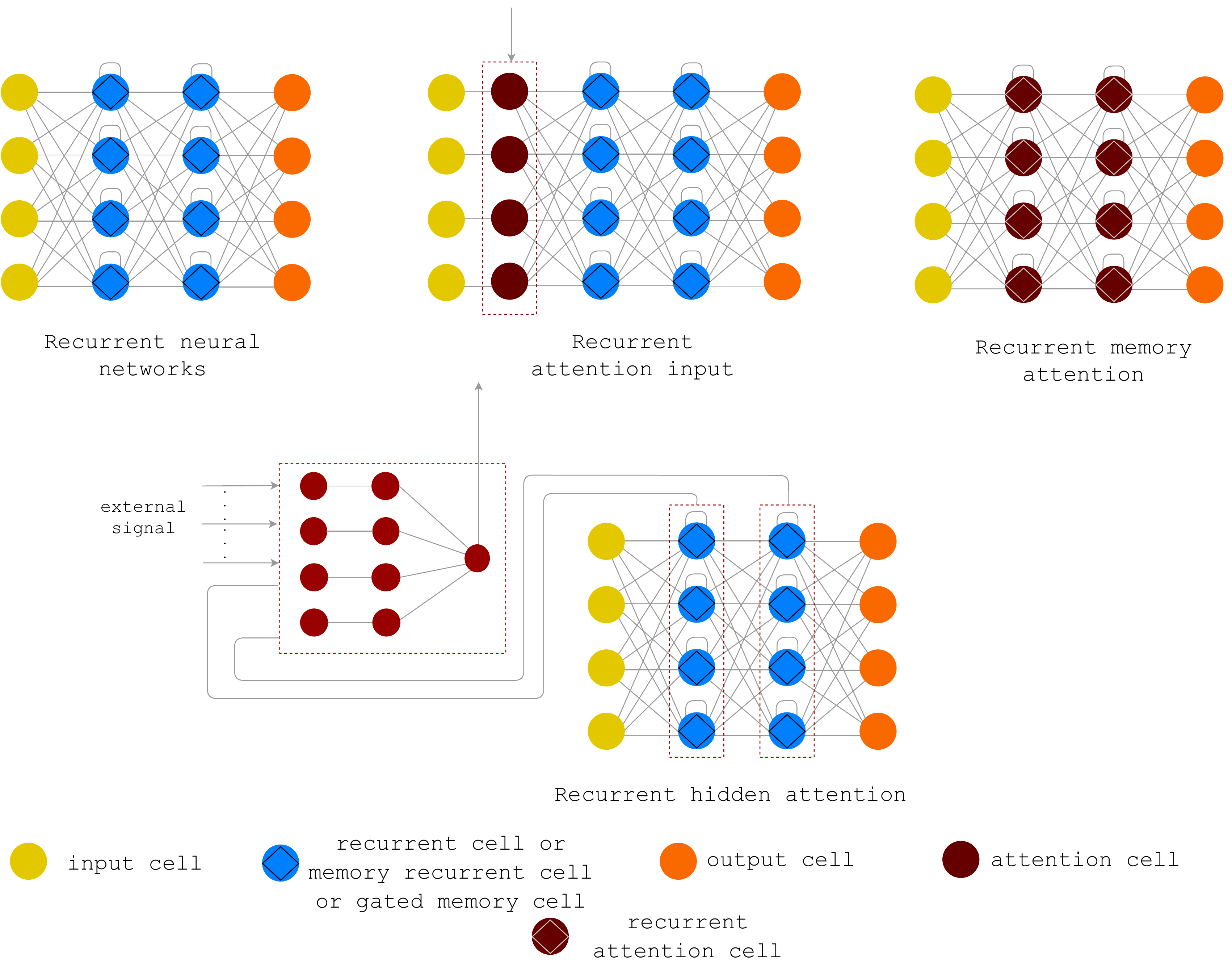}
\caption{Attention-based recurrent neural networks. The architecture is a classic recurrent network or a Bi-RNN when hidden layers are recurrent cells. When the hidden layer is a recurrent memory, the architecture is an LSTM or Bi-LSTM. Finally, the architecture is a GRU when the hidden layer is a gated memory cell. The recurrent attention input group uses attention to filter input data. Recurrent hidden attention groups automatically select relevant encoder hidden states across all time steps. Usually, this group implements attention in encoder-decoder frameworks. The recurrent memory attention group implements attention within the memory cell. There are not many architectures in this category as far as we know, but the main uses are related to filtering the input data and the weighting of different historical components for predicting the current time step.}
\label{fig:rnn_attention}
\end{center}
\end{figure}

Attention in RNNs is mainly responsible for capturing long-distance dependencies. Currently, there are not many ways to use attention on RNNs. RNNSearch's mechanism for encoder-decoder frameworks inspires most approaches~\cite{bahdanau_neural_2014}. We divided the uses of attention into three distinct groups (Figure~\ref{fig:rnn_attention}): 1) Recurrent attention input -- the first stage of attention to select elementary input stimulus, i.e., elementary features, 2) recurrent memory attention -- the first stage of attention to historical weight components, 3) Recurrent hidden attention -- the second stage of attention to select categorical information to the decode stage.

The \textbf{recurrent attention input} group main uses are item-wise hard, local-wise hard, item-wise soft, and local-wise soft selection. \textbf{Item-wise hard} selects discretely relevant input data for further processing, whereas \textbf{location-wise hard} discretely focuses only on the most relevant features for the task. \textbf{Item-wise soft} assigns a continuous weight to each input data given a sequence of items as input,  and  \textbf{location-wise soft} assigns a continuous weight between input features. Location-wise soft estimates high weights for features more correlated with the global context of the task. Hard selection for input elements are applied more frequently in computer vision approaches~\cite{mnih_recurrent_2014}~\cite{marcus_edel;joscha_lausch_capacity_2016}. On the other hand, soft mechanisms are often applied in other fields, mainly in Natural Language Processing. The soft selection normally weighs relevant parts of the series or input features, and the attention layer is a feed-forward network differentiable and with a low computational cost. Soft approaches are interesting to filter noise from time series and to dynamically learn the correlation between input features and output~\cite{qin2017dual}~\cite{geoman_2018}~\cite{du2017rpan}. Besides, this approach is useful for addressing graph-to-sequence learning problems that learn a mapping between graph-structured inputs to sequence outputs, which current Seq2Seq and Tree2Seq may be inadequate to handle~\cite{xu2018graph2seq}.

Hard mechanisms take inspiration from how humans perform visual sequence recognition tasks, such as reading by continually moving the fovea to the next relevant object or character, recognizing the individual entity, and adding the knowledge to our internal representation. A deep recurrent neural network, at each step, processes a multi-resolution crop of the input image, called a glimpse. The network uses information from the glimpse to update its internal representation and outputs the next glimpse location. The Glimpse network captures salient information about the input image at a specific position and region size. The internal state is formed by the hidden units $h_{t}$ of the recurrent neural network, which is updated over time by the core network. At each step, the location network estimates the next focus localization, and action networks depend on the task (e.g., for the classification task, the action network's outputs are a prediction for the class label.). Hard attention is not entirely differentiable and therefore uses reinforcement learning.

RAM~\cite{mnih_recurrent_2014} was the first architecture to use a recurrent network implementing hard selection for image classification tasks. While this model has learned successful strategies in various image data sets, it only uses several static glimpse sizes. CRAM~\cite{marcus_edel;joscha_lausch_capacity_2016} uses an additional sub-network to dynamically change the glimpse size, with the assumption to increase the performance, and in Jimmy et al.~\cite{ba_multiple_2014} explore modifications in RAM for real-world image tasks and multiple objects classification. CRAM is a similar RAM model except for two key differences: Firstly, a dynamically updated attention mechanism restrains the input region observed by the glimpse network and the next output region prediction from the emission network -- a network that incorporates the location and capacity information as well as past information.  In a more straightforward way, the sub-network decides at each time-step what the focus region's capacity should be. Secondly, the capacity sub-network outputs are successively added to the emission network's input that will ultimately generate the information for the next focus region—allowing the emission network to combine the information from the location and the capacity networks.

\begin{figure}[htb]
\begin{center}
    \includegraphics[width=1\linewidth]{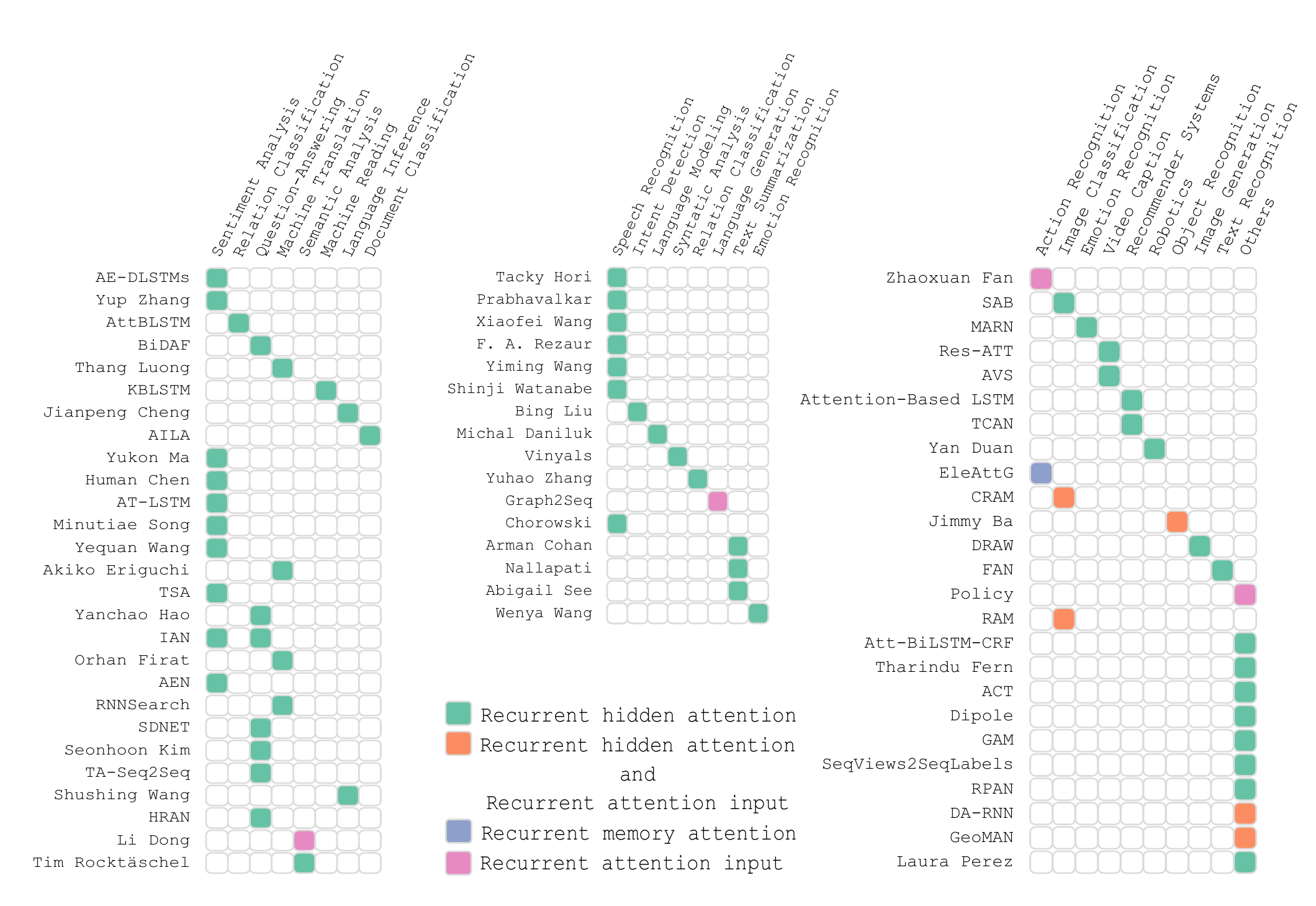}
\caption{Attention-based recurrent neural networks main architectures by task and attention use.}
\label{fig:recurrent_attention_table}
\end{center}
\end{figure}

Nearly all important works in the field belong to the \textbf{recurrent hidden attention} group, as shown in Figure~\ref{fig:recurrent_attention_table}. In this category, the attention mechanism selects elements that are in the RNN's hidden layers for inter-alignment, contextual embedding, multiple-input processing, memory management, and capturing long-term dependencies, a typical problem with recurrent neural networks. Inter-alignment involves the encoder-decoder framework, and the attention module between these two networks is the most common approach. This mechanism builds a context vector dynamically from all previous decoder hidden states and the current encoder hidden state. Attention in inter-alignment helps minimize the bottleneck problem, with RNNSearch~\cite{bahdanau_neural_2014} for machine translation tasks as its first representative. Further, several other architectures implemented the same approach in other tasks~\cite{cheng2016long}~\cite{yang2016hierarchical}~\cite{seo_bidirectional_2016}. For example, Zichao Yang et al.~\cite{yang2016hierarchical} extended the soft selection to the hierarchical attention structure, which allows the calculation of soft attention at the word level and the sentence level in the GRU networks encoder for document classification.

To create contextual embeddings and to manipulate multimodal inputs, co-attention is highly effective for text matching applications. Co-attention enables the learning of pairwise attention, i.e., learning to attend based on computing word-level affinity scores between two documents. Such a mechanism is designed for architectures comprised of queries and context, such as questions and answers and emotions analysis. Co-attention models can be fine-grained or coarse-grained. Fine-grained models consider each element of input concerning each element of the other input. Coarse-grained models calculate attention for each input, using an embedding of the other input as a query. Although efficient, co-attention suffers from information loss from the target and the context due to the anticipated summary. Attention flow emerges as an alternative to summary problems. Unlike co-attention, attention flow links and merges context and query information at each stage of time, allowing embeddings from previous layers to flow to subsequent modeling layers. The attention flow layer is not used to summarize the query and the context in vectors of unique features, reducing information loss. Attention is calculated in two directions, from the context to the query and from the query to the context. The output is the query-aware representations of context words. Attention flow allows a hierarchical process of multiple stages to represent the context at different granularity levels without an anticipated summary.

Hard attention mechanisms do not often occur on recurrent hidden attention networks. However, Nan Rosemary et al.~\cite{ke_sparse_2018} demonstrate that hard selection to retrieve past hidden states based on the current state mimics an effect similar to the brain's ability. Humans use a very sparse subset of past experiences and can access them directly and establish relevance with the present, unlike classic RNNs and self-attentive networks. Hard attention is an efficient mechanism for RNNs to recover sparse memories. It determines which memories will be selected on the forward pass, and therefore which will receive gradient updates. At time $t$, RNN receives a vector of hidden states $h^{t-1}$, a vector of cell states $c^{t-1}$, and an input $x^{t}$, and computes new cell states $c^{t}$ and a provisional hidden state vector $\widetilde{h}^{t}$ that also serves as a provisional output. First, the provisional hidden state vector $\widetilde{h}^{t}$ is concatenated to each memory vector $m_{i}$ in the memory $M$. MLP maps each vector to an attention weight $ a^{t}_{i}$, representing memory relevance $i$ in current moment $t$. With attention weights $ a^{t}_{i}$ sparse attention computes a hard decision. The attention mechanism is differentiable but implements a hard selection to forget memories with no prominence over others. This is quite different from typical approaches as the mechanism does not allow the gradient to flow directly to a previous step in the training process. Instead, it propagates to some local timesteps as a type of local credit given to a memory.

\begin{figure}[h]
\begin{center}
    \includegraphics[width=0.95\linewidth]{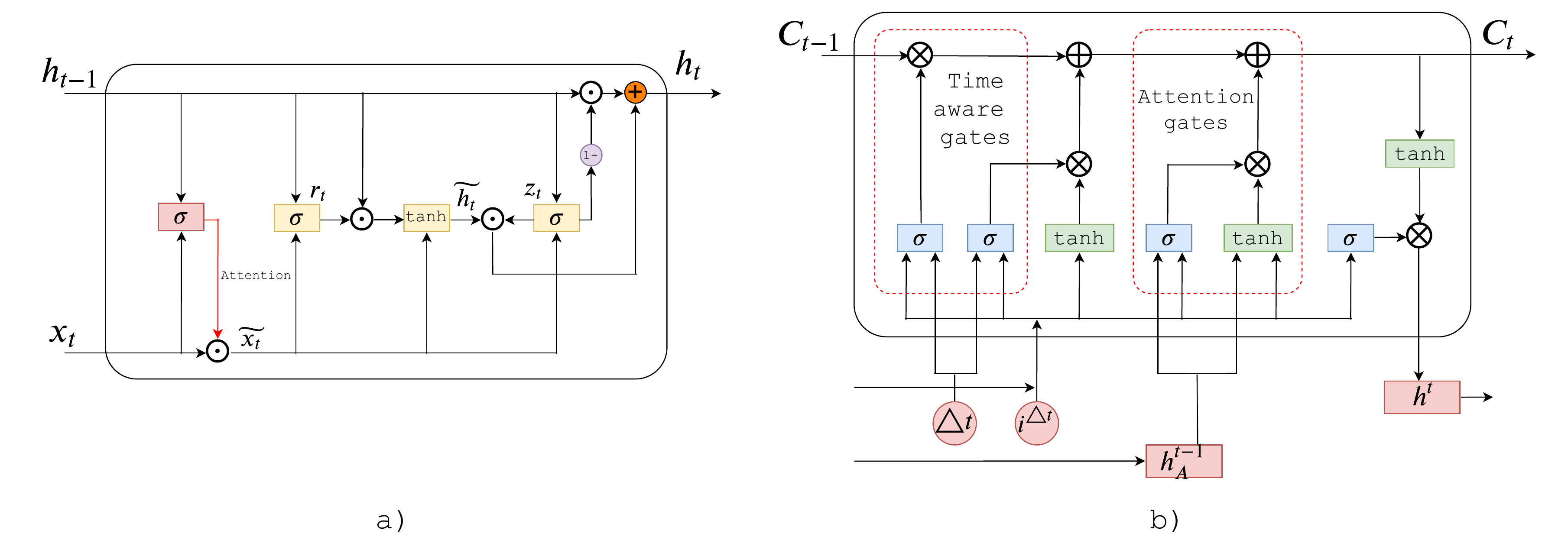}
\caption{recurrent memory attention approaches. a) Illustration of Element-wise-Attention Gate in GRU. Specifically, the input modulation is adaptable to the content and is performed in fine granularity, element-wise rather than input-wise. b) Gate attention in LSTM. The model calculates attention scores to weigh the relevance of different parts of history. There are two additional gates for updating the current cell using the previous state $h_{A}^{t-1}$. The first, input attention gate layer, analyzes the current input $i^{\triangle t}$ and $h_{A}^{t-1}$ to determine which values to be updated in current cell $C^{t}$. The second, the modulation attention gate, analyzes the current input $i^{\triangle t}$ and $h_{A}^{t-1}$. Then computes the set of candidate values that must be added when updating the current cell state $C^{t}$. The attention mechanisms in the memory cell help to more easily capture long-term dependencies and the problem of data scarcity.}
\label{fig:recurrent_memory_attention}
\end{center}
\end{figure}

Finally, \textbf{recurrent memory attention} groups implement attention within the memory cell. As far as our research goes, there are not many architectures in this category. Pengfei et al.~\cite{zhang2018adding} proposed an approach that modulates the input adaptively within the memory cell by assigning different levels of importance to each element/dimension of the input shown in Figure~\ref{fig:recurrent_memory_attention}a. Dilruk et al.~\cite{perera2020lstm} proposed mechanisms of attention within memory cell to improve the past encoding history in the cell's state vector since all parts of the data history are not equally relevant to the current prediction. As shown in Figure~\ref{fig:recurrent_memory_attention}b, the mechanism uses additional gates to update LSTM's current cell.

\subsection{Attention-based Generative Models}
\label{subsec:attention_generative_models}

\begin{figure}[ht]
\begin{center}
    \includegraphics[width=0.90\linewidth]{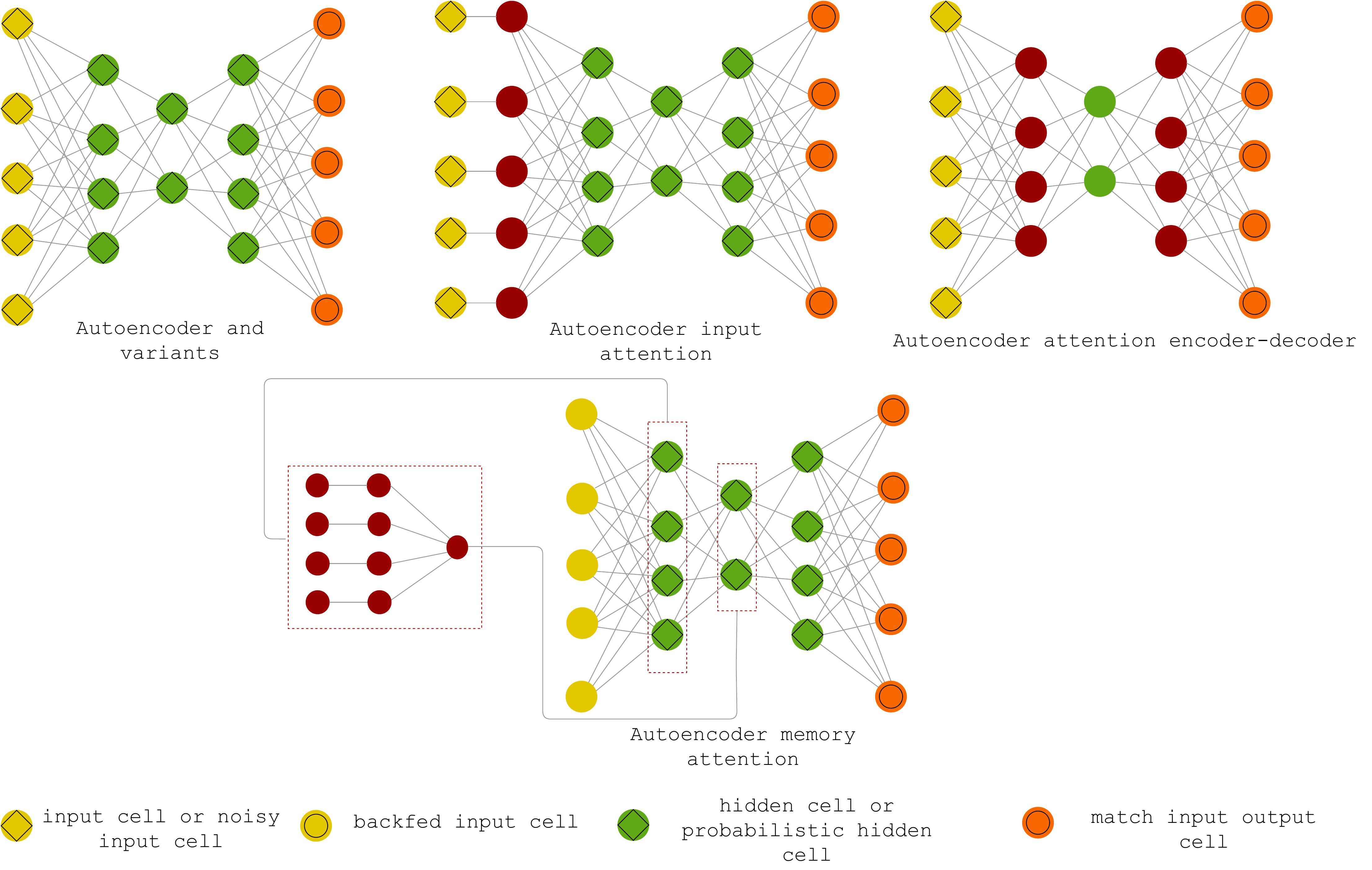}
\caption{Attention-based generative models. The Autoencoder input attention group uses attention to facilitate the decomposition of scenes in abstract building blocks. The input components extracted by the masks share significant properties and help to imagine new scenarios. This approach is very efficient for the network to learn to decompose challenging scenarios between semantically significant components. Autoencoder memory attention group handles a memory buffer that allows read/writes operations and is persistent over time. Such models generally handle input and output to the memory buffer using write/read operations guided by the attention system. The use of attention and memory history in autoencoders helps to increase the generalizability of architecture. Autoencoder attention encoder-decoder using a model (usually self-attentive model) to increase the ability to generalize.}
\label{fig:generative_models}
\end{center}
\end{figure}

Attention emerges in generative models essentially to augmented memory. Currently, there are not many ways to use attention on generative models. Since GANs are not a neural network architecture but a framework, we do not discuss the use of attention in GANs but autoencoders. We divided the uses of attention into three distinct groups (Figure~\ref{fig:generative_models}): 1) Autoencoder input attention -- attention provides spatial masks corresponding to all the parts for a given input, while a component autoencoder (e.g., AE, VAE, SAE) independently models each of the parts indicated by the masks. 2) Autoencoder memory attention -- attention module acts as a layer between the encoder-decoder to augmented memory., 3) Autoencoder attention encoder-decoder -- a fully attentive architecture acts on the encoder, decoder, or both.

MONet~\cite{burgess2019monet} is one of the few architectures to implement attention at the VAE input. A VAE is a neural network with an encoder parameterized by $\phi $ and a decoder parameterized by $\theta$. The encoder parameterizes a distribution over the component latent $z_{k}$, conditioned on both the input data \textit{x} and an attention mask $m_{k}$. The mask indicates which regions of the input the VAE should focus on representing via its latent posterior distribution, $q\phi(z_{k}|x, m_{k})$. During training, the VAE's
decoder likelihood term in the loss $p_{\theta}(x|z_{k})$ is weighted according to the mask, such that it is
unconstrained outside of the masked regions. In~\cite{li2016learning}, the authors use soft attention with learned memory contents to augment models to have more parameters in the autoencoder. In~\cite{bartunov2016fast}, Generative Matching Networks use attention to access the exemplar memory, with the address weights computed based on a learned similarity function between an observation at the address and a function of the latent state of the generative model. In~\cite{rezende2016one}, external memory and attention work as a way of implementing one-shot generalization by treating the exemplars conditioned on as memory entries accessed through a soft attention mechanism at each step of the incremental generative process similar to DRAW~\cite{draw}. Although most approaches use soft attention to address the memory, in~\cite{bornschein2017variational} the authors use a stochastic, hard attention approach, which allows using variational inference about it in a context of few-shot learning. 

In~\cite{escolano2018self}, self-attentive networks increase the autoencoder ability to generalize. The advantage of using this model instead of other alternatives, such as recurrent or convolutional encoders, is that this model is based only on self-attention and traditional attention over the whole representation created by the encoder. This approach allows us to easily employ the different components of the networks (encoder and decoder) as modules that, during inference, can be used with other parts of the network without the need for previous step information.
\section{Applications}
\label{sec:applications}
\begin{figure*}[htb]
  \centering
  \includegraphics[width=0.95\textwidth]{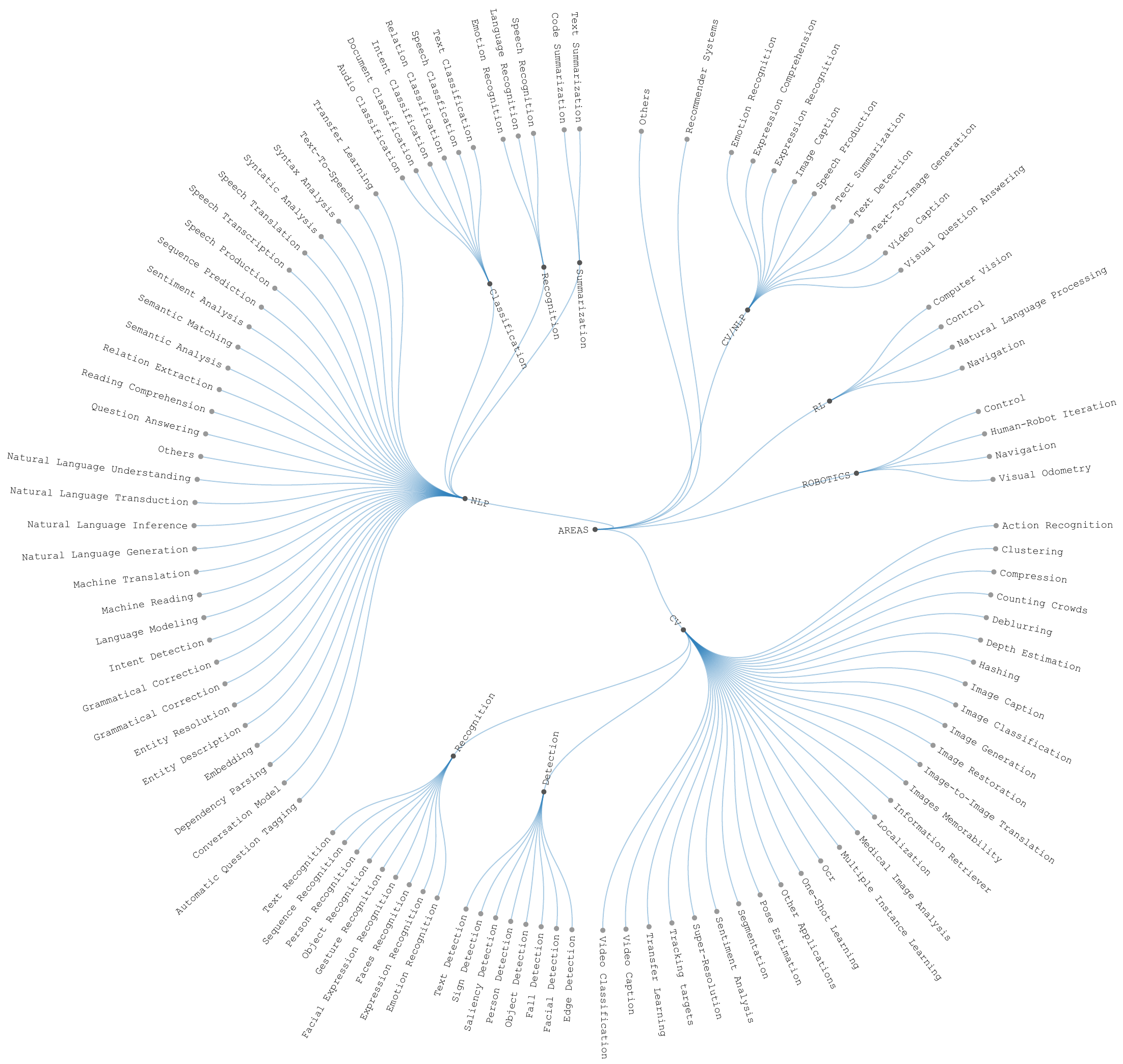}
  \caption{Diagram showing the main existing applications of neural attention networks. The main areas are Natural language processing (NLP), Computer Vision (CV), multimodal tasks (mainly with images and texts - CV/NLP), reinforcement learning (RL), robotics, recommendation systems, and others (e.i., graph embeddings, interpretability.).}
  \label{fig:applications}
\end{figure*}

In a few years, neural attention networks have been used in numerous domains due to versatility, interpretability, and significance of results. These networks have been explored mainly in computer vision, natural language processing, and multi-modal tasks, as shown in figure~\ref{fig:applications}. In some applications, these models transformed the area entirely (i.e., question-answering, machine translation, document representations/embeddings, graph embeddings), mainly due to significant performance impacts on the task in question. In others, they helped learn better representations and deal with temporal dependencies over long distances. This section explores a list of application domains and subareas, mainly discussing each domain's main models and how it benefits from attention. We also present the most representative instances within each area and list them with reference approaches in a wide range of applications.


\subsection{\textbf{Natural Language Processing (NLP)}}
\label{sec:application_nlp}


In the NLP domain, attention plays a vital role in many sub-areas, as shown in figure~\ref{fig:applications}. There are several state-of-the-art approaches, mainly in language modeling, machine translation, natural language inference, question answering, sentiment analysis, semantic analysis, speech recognition, and text summarization. Table~\ref{tab:main_nlp_app} groups works developed in each of these areas. Several applications have been facing an increasing expansion, with few representative works, such as emotion recognition, speech classification, sequence prediction, semantic matching, and grammatical correction, as shown in Table~\ref{tab:others_nlp_applications}.

\begin{table}[htb]
\small
\centering
\fontfamily{pag}\selectfont
\begin{tabular}{ll}
\hline
\textbf{Task}   & \textbf{References}   \\ \hline

\textbf{Language Modeling} & \begin{tabular}[c]{@{}l@{}}~\cite{dehghani_universal_2018}~\cite{joulin_inferring_2015}~\cite{ke_sparse_2018}~\cite{cheng2016long}~\cite{vinyals_matching_2016}~\cite{dai_transformer-xl:_2019}~\cite{sainbayar_sukhbaatar;edouard_grave;piotr_bojanowski;armand_joulin_adaptive_2019}~\cite{alexei_baevski;michael_auli_adaptive_2019}~\cite{michal_daniluk_tim_rocktaschel_johannes_welbl_sebastian_riedel:_frustratingly_nodate}\end{tabular} \\ \hline

\textbf{Machine Translation} &                                          \begin{tabular}[c]{@{}l@{}} ~\cite{kim_structured_2017}~\cite{vaswani_attention_2017}~\cite{gulcehre_hyperbolic_2018}~\cite{feng_neural_2018}~\cite{dehghani_universal_2018}~\cite{bahdanau_neural_2014}~\cite{luong_effective_2015}~\cite{shaw_self-attention_2018}~\cite{raffel_online_2017} \\ ~\cite{cho_learning_2014}~\cite{sennrich_neural_2016}~\cite{thomas_zenkel;joern_wuebker;john_denero_adding_2019}~\cite{baosong_yang;jian_li;derek_wong;lidia_s._chao;xing_wang;zhaopeng_tu_context-aware_2019}~\cite{baosong_yang;longyue_wang;derek_f._wong;lidia_s._chao;zhaopeng_tu_convolutional_2019}~\cite{jie_hao;xing_wang;baosong_yang;longyue_wang;jinfeng_zhang;zhaopeng_tu_modeling_2019}~\cite{felix_hieber;tobias_domhan;michael_denkowski;david_vilar;artem_sokolov;ann_clifton;matt_post_sockeye:_2018}~\cite{xiangwen_zhang;jinsong_su;yue_qin;yang_liu;rongrong_ji;hongji_wang_asynchronous_2018}~\cite{joost_bastings;ivan_titov;wilker_aziz;diego_marcheggiani;khalil_simaan_graph_2017}\\
~\cite{jie_zhou;ying_cao;xuguang_wang;peng_li;wei_xu_deep_2016}~\cite{yonghui_wu;mike_schuster;zhifeng_chen;quoc_v._le;mohammad_norouzi;wolfgang_macherey;maxim_krikun;yuan_cao;qin_gao;klaus_macherey;jeff_klingner;apurva_shah;melvin_johnson;xiaobing_liu;lukasz_kaiser;stephan_gouws;yoshikiyo_kato;taku_kudo;hideto_kazawa;keith_stevens;george_kurian;nishant_patil;wei_wang;cliff_young;jason_smith_googles_2016}~\cite{akiko_eriguchi_kazuma_hashimoto_yoshimasa_tsuruoka:_tree--sequence_nodate}~\cite{jian_li;baosong_yang;zi-yi_dou;xing_wang;michael_r._lyu;zhaopeng_tu_information_2019}~\cite{orhan_firat;kyunghyun_cho;yoshua_bengio_multi-way;_2016}~\cite{deng_latent_2018}~\cite{h._zhang;_j._li;_y._ji;_h._yue_understanding_2017} \end{tabular} \\ \hline

\textbf{Natural Language Inference}  &                                      \begin{tabular}[c]{@{}l@{}}~\cite{kim_structured_2017}~\cite{parikh_decomposable_2016}~\cite{shen_disan:_nodate}~\cite{cheng2016long}~\cite{tao_shen_tianyi_zhou_guodong_long_jing_jiang_chengqi_zhang:_bi-directional_nodate}~\cite{qipeng_guo;xipeng_qiu;pengfei_liu;yunfan_shao;xiangyang_xue;zheng_zhang_star-transformer_2019}~\cite{shuohang_wang;jing_jiang_learning_2016}~\cite{yang_liu;chengjie_sun;lei_lin;xiaolong_wang_learning_2016}\end{tabular} \\ \hline

\textbf{Question Answering}  &                                          
\begin{tabular}[c]{@{}l@{}}~\cite{kim_structured_2017}~\cite{neelakantan_neural_2015}~\cite{wu_hierarchical_2017}~\cite{dehghani_universal_2018}~\cite{weston_2014_memory}~\cite{kumar_ask_2015}~\cite{vinyals_neural_nodate}~\cite{yang_anmm:_2018} \\ ~\cite{tay_densely_2018}~\cite{seo_bidirectional_2016}~\cite{adams_wei_yu_david_dohan_minh-thang_luong_rui_zhao_kai_chen_mohammad_norouzi_quoc_v._le:_qanet:_nodate}~\cite{wei_wang_chen_wu_ming_yan:_multi-granularity_nodate}~\cite{xiangyang_zhou_lu_li_daxiang_dong_yi_liu_ying_chen_wayne_xin_zhao_dianhai_yu_hua_wu:_multi-turn_nodate}~\cite{cicero_dos_santos;ming_tan;bing_xiang;bowen_zhou_attentive_2016}~\cite{victor_zhong;caiming_xiong;nitish_shirish_keskar;richard_socher_coarse-grain_2019}~\cite{louis_shao;stephan_gouws;denny_britz;anna_goldie;brian_strope;ray_kurzweil_generating_2017}~\cite{chenguang_zhu;michael_zeng;xuedong_huang_sdnet:_2019}\\ ~\cite{ming_tan;cicero_dos_santos;bing_xiang;bowen_zhou_lstm-based_2016}~\cite{bhuwan_dhingra_hanxiao_liu_zhilin_yang_william_w._cohen_ruslan_salakhutdinov:_gated-attention_nodate}~\cite{yanchao_hao_yuanzhe_zhang_kang_liu_shizhu_he_zhanyi_liu_hua_wu_jun_zhao:_end--end_nodate}~\cite{rudolf_kadlec_martin_schmid_ondrej_bajgar_jan_kleindienst:_text_nodate}~\cite{sukhbaatar2015end}~\cite{tsendsuren_munkhdalai;hong_yu_neural_2017}~\cite{seonhoon_kim;inho_kang;nojun_kwak_semantic_2018}~\cite{lisa_bauer;yicheng_wang;mohit_bansal_commonsense_2019}~\cite{qiu_ran;peng_li;weiwei_hu;jie_zhou_option_2019} \\ ~\cite{hermann2015teaching}~\cite{huang_fusionnet:_2018}~\cite{chen_xing;wei_wu;yu_wu;jie_liu;yalou_huang;ming_zhou;wei-ying_ma_topic_2016}~\cite{sordoni_iterative_2017}~\cite{bingning_wang_kang_liu_jun_zhao:_inner_nodate}  \end{tabular} \\ \hline

\textbf{Sentiment Analysis}    &                  
\begin{tabular}[c]{@{}l@{}}~\cite{wang_attention-based_2017}~\cite{shen_disan:_nodate}~\cite{cheng2016long}~\cite{dehong_ma;sujian_li;xiaodong_zhang;houfeng_wang_interactive_2017}~\cite{kai_shuang_xintao_ren_qianqian_yang_rui_li_jonathan_loo:_aela-dlstms:_nodate}~\cite{christos_baziotis_nikos_pelekis_christos_doulkeridis:_datastories_nodate}~\cite{wei_xue;tao_li_aspect_2018}~\cite{shi_feng;yang_wang;liran_liu;daling_wang;ge_yu_attention_2019}~\cite{youwei_song;jiahai_wang;tao_jiang;zhiyue_liu;yanghui_rao_attentional_2019} \\ ~\cite{bonggun_shin;timothy_lee;jinho_d._choi_lexicon_2017}~\cite{minchae_song;hyunjung_park;kyung-shik_shin_attention-based_2019}~\cite{jiachen_du;lin_gui;yulan_he;ruifeng_xu;xuan_wang_convolution-based_2019}~\cite{peng_chen;zhongqian_sun;lidong_bing;wei_yang_recurrent_2017}~\cite{yukun_ma;haiyun_peng;erik_cambria_targeted_2018}~\cite{yue_zhang;jiangming_liu_attention_2017}~\cite{huimin_chen;maosong_sun;cunchao_tu;yankai_lin;zhiyuan_liu_neural_2016}~\cite{jiangfeng_zeng;xiao_ma;ke_zhou_enhancing_2019}  \end{tabular} \\ \hline

\textbf{Semantic Analysis} &                  \begin{tabular}[c]{@{}l@{}}~\cite{rocktaschel_reasoning_2015}~\cite{she_distant_2018}~\cite{zhang_learning_2018}~\cite{shen_reinforced_2018}~\cite{zhixing_tan;mingxuan_wang;jun_xie;yidong_chen;xiaodong_shi_deep_2017}~\cite{li_dong_mirella_lapata:_language_nodate}~\cite{wenya_wu;yufeng_chen;jinan_xu;yujie_zhang_attention-based_2018} \end{tabular} \\ \hline

\textbf{Speech Recognition} &                                          \begin{tabular}[c]{@{}l@{}} ~\cite{chorowski_attention-based_2015}~\cite{raffel_online_2017}~\cite{prabhavalkar_minimum_2018}~\cite{kazuki_irie;albert_zeyer;ralf_schluter;hermann_ney_language_2019}~\cite{christoph_luscher;eugen_beck;kazuki_irie;markus_kitza;wilfried_michel;albert_zeyer;ralf_schluter;hermann_ney_rwth_2019}~\cite{linhao_dong;feng_wang;bo_xu_self-attention_2019}~\cite{julian_salazar;katrin_kirchhoff;zhiheng_huang_self-attention_2019}~\cite{xiaofei_wang;ruizhi_li;sri_harish_mallid;takaaki_hori;shinji_watanabe;hynek_hermansky_stream_2019}~\cite{shiyu_zhou;linhao_dong;shuang_xu;bo_xu_syllable-based_2018} \\ ~\cite{shinji_watanabe;takaaki_hori;suyoun_kim;john_r._hershey;tomoki_hayashi_hybrid_2017}~\cite{dzmitry_bahdanau;jan_chorowski;dmitriy_serdyuk;philemon_brakel;yoshua_bengio_end--end_2016}~\cite{jan_chorowski;dzmitry_bahdanau;kyunghyun_cho;yoshua_bengio_end--end_2014}~\cite{f_a_rezaur_rahman_chowdhury;quan_wang;ignacio_lopez_moreno;li_wan_attention-based_2018}~\cite{yiming_wang;xing_fan;i-fan_chen;yuzong_liu;tongfei_chen;bjorn_hoffmeister_end--end_2019}~\cite{albert_zeyer;kazuki_irie;ralf_schluter;hermann_ney_improved_2018}~\cite{suyoun_kim;takaaki_hori;shinji_watanabe_joint_2017}~\cite{koji_okabe;takafumi_koshinaka;koichi_shinoda_attentive_2018}~\cite{takaaki_hori;shinji_watanabe;yu_zhang;william_chan_advances_2017} \end{tabular} \\ \hline

\textbf{Text Summarization} &       
\begin{tabular}[c]{@{}l@{}}~\cite{chopra_abstractive_2016}~\cite{paulus_deep_2017}~\cite{nallapati_abstractive_2016}~\cite{see_get_2017}~\cite{raffel_online_2017}~\cite{alexander_m._rush;sumit_chopra;jason_weston_neural_2015}~\cite{angela_fan;mike_lewis;yann_dauphin_hierarchical_2018}~\cite{b._rekabdar;_c._mousas;_b._gupta_generative_2019}~\cite{arman_cohan;franck_dernoncourt;doo_soon_kim;trung_bui;seokhwan_kim;walter_chang;nazli_goharian_discourse-aware_2018} \end{tabular} \\ \hline
\end{tabular}
\caption{Summary state-of-the-art approaches in several natural language processing sub-areas.}
\label{tab:main_nlp_app}
\end{table}

For \textbf{machine translation} (MT), \textbf{question answering} (QA), and \textbf{automatic speech recognition} (ASR), attention works mainly in the alignment input and output sequences capturing long-range dependencies. For example, in ASR tasks, attention aligns acoustic frames extracting information from anchor words to recognize the main speaker while ignoring background noise and interfering speech. Hence, only information on the desired speech is used for the decoder as it provides a straightforward way to align each output symbol with different input frames with selective noise decoding. In MT, automatic alignment translates long sentences more efficiently. It is a powerful tool for \textbf{multilingual machine translation} (NMT), efficiently capturing subjects, verbs, and nouns in sentences of different languages that differ significantly in their syntactic structure and semantics.

In QA, alignment usually occurs between a query and the content, looking for key terms to answer the question. The classic QA approaches do not support very long sequences and fail to correctly model the meaning of context-dependent words. Different words can have different meanings, which increases the difficulty of extracting the essential semantic logical flow of each sentence in different paragraphs of context. These models are unable to address uncertain situations that require additional information to answer a particular question. In contrast, attention networks allow rich dialogues through addressing mechanisms for explicit memories or alignment structures in the query-context and context-query directions.

Attention also contributes to \textbf{summarize or classify texts/documents}. It mainly helps build more effective embeddings that generally consider contextual, semantic, and hierarchical information between words, phrases, and paragraphs. Specifically, in summarization tasks, attention minimizes critical problems involving: 1) modeling of keywords; 2) summary of abstract sentences; 3) capture of the sentence's hierarchical structure; 4) repetitions of inconsistent phrases; and 5) generation of short sentences preserving their meaning.

\begin{figure}[htb]
  \centering
  \includegraphics[width=\linewidth]{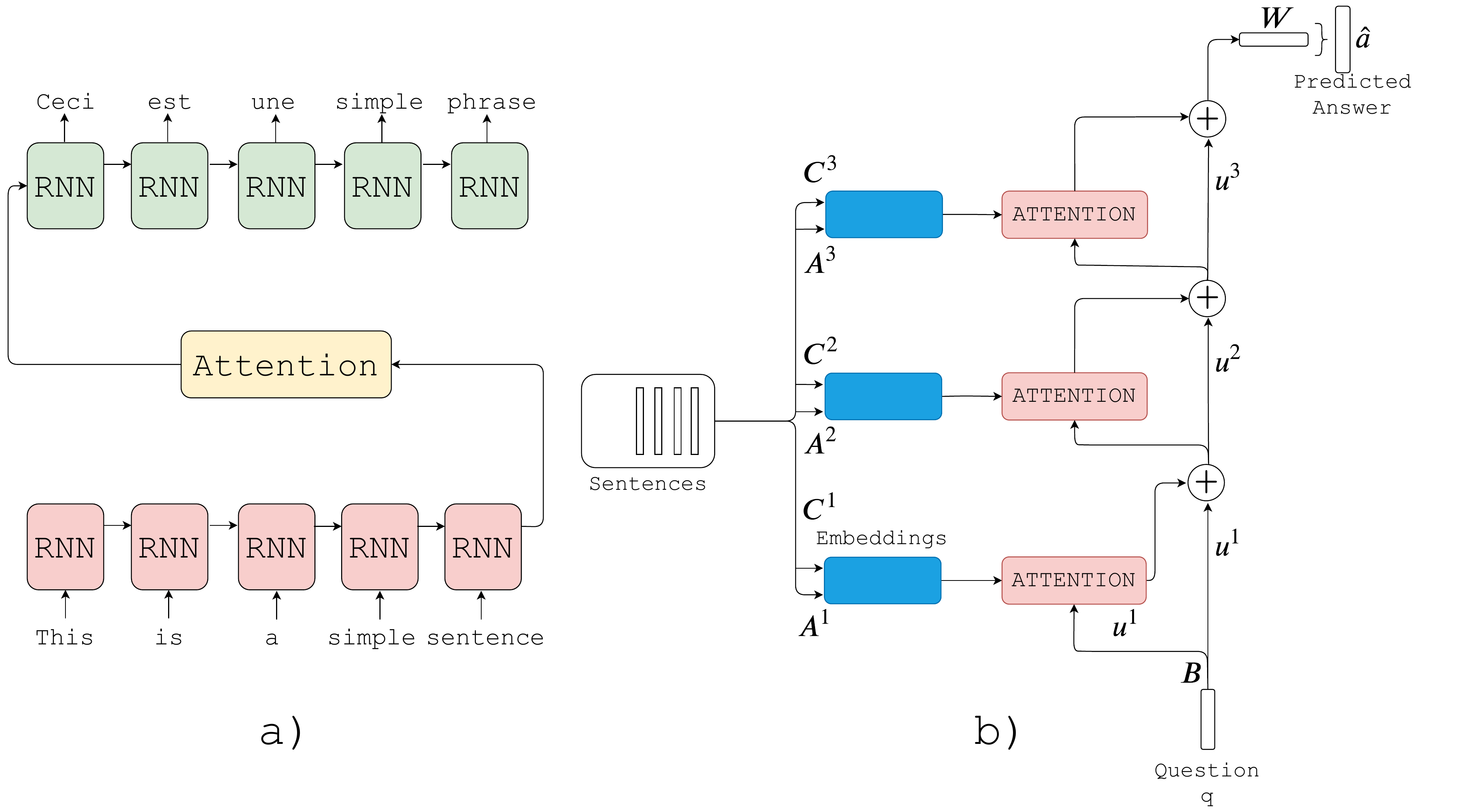}  
  \caption{Illustration of RNNSearch~\cite{bahdanau_neural_2014} for machine translation, and End-to-End Memory Networks~\cite{sukhbaatar2015end} for question answering. a) The RNNSearch architecture. The attention guided by the decoder's previous state dynamically searches for important source words for the next time step. b) The End-to-End Memory Networks. The architecture consists of external memory and several stacked attention modules. To generate a response, the model makes several hops in memory using only attentional layers.}
  \label{fig:nlp_apps_1}
\end{figure}

\begin{figure}[htb]
  \centering
  \includegraphics[width=\linewidth]{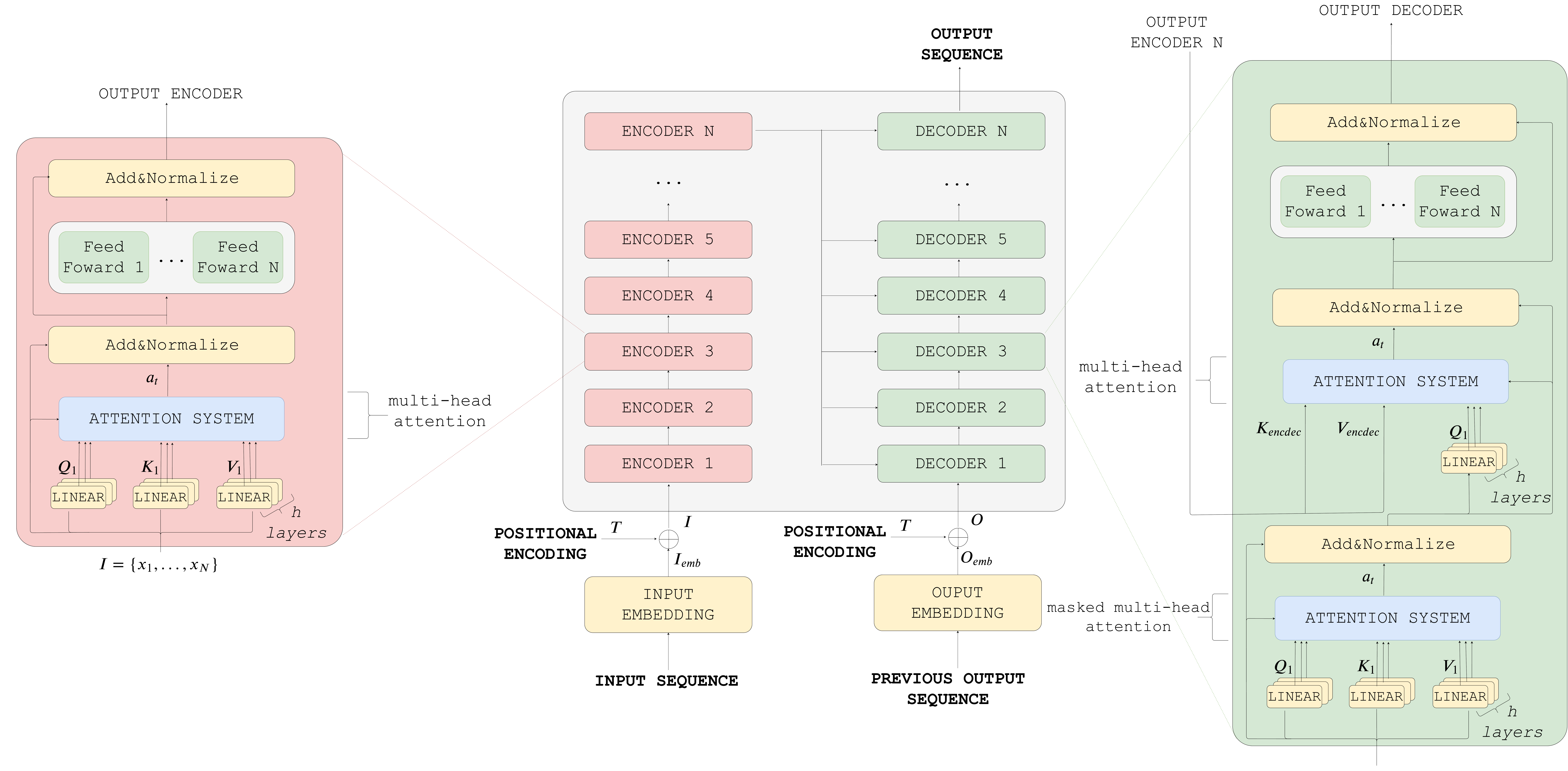}  
  \caption{Illustration of Neural Transformer~\cite{vaswani_attention_2017} for machine translation. The architecture consists of several stacked attentional encoders and decoders. The encoder process is massively parallel and eliminates recurrences, and the decoder generates the translated words sequentially. Each encoder uses multiple heads of the self-attention mechanism followed by fusion and normalization layers. Similarly, to generate the output, each decoder has multiple heads of self-attention and masked-self attention to mask words not yet generated.}
  \label{fig:nlp_apps_2}
\end{figure}

Figure~\ref{fig:nlp_apps_1} illustrates two models working in NLP tasks: RNNSearch~\cite{bahdanau_neural_2014}, in machine translation, and End-to-End Memory Networks~\cite{sukhbaatar2015end} in question answering. In \textbf{RNNSearch}, the attention guided by the decoder's previous state dynamically searches for important source words for the next time step. It consists of an encoder followed by a decoder. The encoder is a bidirectional RNN (BiRNN)~\cite{schuster1997bidirectional} that consists of forward and backward RNN's. The forward RNN reads the input sequence in order and calculates the forward hidden state sequence. The backward RNN reads the sequence in the reverse order, resulting in the backward hidden states sequence. The decoder has an RNN and an attention system that calculates a probability distribution for all possible output symbols from a context vector.

In \textbf{End-to-End Memory Networks}, attention looks for the memory elements most related to query $q$ using an alignment function that dispenses the RNNs' complex structure. It consists of a memory and a stack of identical attentional systems. Each layer $i$ takes as input set $\left \{ x_{1}, ..., x_{N} \right \}$ to store in the memory. The input set is converted in memory vectors $\left \{ m_{1}, ..., m_{N} \right \}$ and $\left \{ h_{1}, ..., h_{N} \right \}$, in the simplest case using the embedding matrix $A^{i} \in \mathbb{R}^{d \times V}$ to generate each $m_{i} \in \mathbb{R}^{d}$, and the matrix $C^{i} \in \mathbb{R}^{d \times V}$ to generate each $h_{i} \in \mathbb{R}^{d}$. In the first, layer the query $q$ is also embedded, via embedding matrix $B^{1}$ to obtain an internal state $u^{1}$. From the second layer, the internal state $u^{i+1}$ is the sum of the $i$ layer output and the internal state $u^{i}$. Finally, the last layer generates $\hat{a}$.

The \textbf{Neural Transformer}~\cite{vaswani_attention_2017}, illustrated in figure~\ref{fig:nlp_apps_2}, is the basis model for state-of-the-art results in NLP. The architecture consists of an arbitrary amount of stacked encoders and decoders. Each encoder has linear layers, an attention system, feed-forward neural networks, and normalization layers. The attention system has several parallel heads. Each head has $N$ attentional subsystems that perform the same task but have different contextual inputs. The encoder receives a word embedding matrix $I = \left \{ x_{1}, ..., x_{N} \right \}$, $I$ $\in \mathbb{R}^{N \times d_{emb}}$ as input. As the architecture does not use recurrences, the input tokens' position information is not explicit, but it is necessary. To represent the spatial position information, the Transformer adds a positional encoding to each embedding vector. Positional encoding is fixed and uses sinusoidal functions.

The input $I$ goes through linear layers and generates, for each word, a \textbf{query vector} ($q_{i}$), a \textbf{key vector} ($k_{i}$), and a \textbf{value vector} ($v_{i}$).The attentional system receives all $Q$, $K$, and $V$ arrays as input and uses several parallel attention heads. The motivation for using a multi-head structure is to explore multiple subspaces since each head gets a different projection of the data. Each head learns a different aspect of attention to the input, calculating different attentional distributions. Having multiple heads on the Transformer is similar to having multiple feature extraction filters on CNNs. The head outputs an attentional mask that relates all queries to a certain key. In a simplified way, the operation performed by a head is a matrix multiplication between a matrix of queries and keys.

Finally, the data is added to the residual output from the previous layer and normalized, representing the encoder output. This data is input to the next encoder. The last encoder's data are transformed into the attention matrices $K_{encdec}$ and $V_{encdec}$. They are input to all decoder layers. This data help the decoder to focus on the appropriate locations in the input sequence. The decoder has two layers of attention, Feed-Foward layers and normalization layers. The attentional layers are the masked multi-head attention and the decoder multi-head attention. 

The masked multi-head attention is very similar to the encoder multi-head attention, with the difference that the attention matrices $Q$, $K$, and $V$ are created only with the previous data words, masking future positions with $-\infty$ values before the softmax step. The decoder multi-head attention is equal to the encoder multi-head attention, except it creates the \textit{Q} matrix from the data of the previous layer and uses the $K_{encdec}$ and $V_{encdec}$ matrices of the encoder output. The $K_{encdec}$ and $V_{encdec}$ matrices are the memory structure of the network, storing context information of the input sequence, and given the previous words in the output decoder, the relevant information is selected in memory for the prediction of the next word. Finally, a linear layer followed by a softmax function projects the decoder vector by the last decoder into a probability vector in which each position defines the probability of the output word being a given vocabulary word. At each time step $t$, the position with the highest probability value is chosen, and the word associated with it is the output.

\begin{table}[htb]
\small
\centering
\fontfamily{pag}\selectfont
\begin{tabular}{lll}
\hline
\textbf{Task}                           & \textbf{References}                                     \\ \hline
\textbf{Code Summarization}          & ~\cite{allamanis2016convolutional} \\ \hline
\textbf{Language Recognition}     & ~\cite{huang2018video}               \\ \hline
\textbf{Emotion Recognition}     &                
\begin{tabular}[c]{@{}l@{}}~\cite{seyedmahdad_mirsamadi;emad_barsoum;cha_zhang_automatic_2017}~\cite{le_hoang_son;akshi_kumar;saurabh_raj_sangwan;anshika_arora;an;nayyar;mohamed_abdel-basset_sarcasm_2019}~\cite{kun-yi_huang;chung-hsien_wu;ming-hsiang_su_attention-based_2019}~\cite{navonil_majumder;soujanya_poria;devamanyu_hazarika;rada_mihalcea;alexander_gelbukh;erik_cambria_dialoguernn:_2019}~\cite{yuanyuan_zhang;jun_du;zirui_wang;jianshu_zhang_attention_2019}~\cite{michael_neumann;ngoc_thang_vu_attentive_2017}~\cite{wenya_wang;sinno_jialin_pan;daniel_dahlmeier;xiaokui_xiao_coupled_2017} \end{tabular} \\ \hline

\textbf{Text Classification}   & ~\cite{liu2019bidirectional}~\cite{qipeng_guo;xipeng_qiu;pengfei_liu;yunfan_shao;xiangyang_xue;zheng_zhang_star-transformer_2019} \\ \hline

\textbf{Speech Classification}  &                  ~\cite{norouzian2019exploring}~\cite{li2019multi} \\ \hline

\textbf{Relation Classification}        &                  
\begin{tabular}[c]{@{}l@{}}~\cite{patrick_verga;emma_strubell;andrew_mccallum_simultaneously_2018}~\cite{xiaoyu_guo;hui_zhang;haijun_yang;lianyuan_xu;zhiwen_ye_single_2019}~\cite{linlin_wang_zhu_cao_gerard_de_melo_zhiyuan_liu:_relation_nodate}~\cite{peng_zhou_wei_shi_jun_tian_zhenyu_qi_bingchen_li_hongwei_hao_bo_xu:_attention-based_nodate}~\cite{yankai_lin_shiqi_shen_zhiyuan_liu_huanbo_luan_maosong_sun:_neural_nodate}~\cite{yuhao_zhang;victor_zhong;danqi_chen;gabor_angeli;christopher_d._manning_position-aware_2017}  \end{tabular} \\ \hline

\textbf{Intent Classification}   &                  ~\cite{chen2019bert}                                       \\ \hline

\textbf{Document Classification}      &                  ~\cite{choi2019aila}~\cite{yang2016hierarchical}           \\ \hline

\textbf{Audio Classification}          &                  ~\cite{kong2018audio}                                       \\ \hline

\textbf{Transfer Learning}        &                  ~\cite{devlin2018bert}~\cite{alt2019improving}              \\ \hline

\textbf{Text-to-Speech}            &                  ~\cite{yasuda2019investigation}~\cite{zhang2019joint}~\cite{li2019neural}      \\ \hline

\textbf{Syntax Analysis}       &                  ~\cite{kuncoro2016recurrent}                                \\ \hline

\textbf{Speech Translation}           &                  ~\cite{sperber2019attention}                                \\ \hline

\textbf{Speech Transcription}          &                  ~\cite{chan2015listen}                                       \\ \hline

\textbf{Speech Production}             &                  ~\cite{tachibana2018efficiently}                             \\ \hline

\textbf{Sequence Prediction}           &                  ~\cite{mensch2018differentiable}                             \\ \hline

\textbf{Semantic Matching}             &                  ~\cite{zhang2018multiresolution}                              \\ \hline

\textbf{Relation Extraction}           &                ~\cite{zhang2019long}                                     \\ \hline

\textbf{Reading Comprehension}         &                  ~\cite{tao_shen_tianyi_zhou_guodong_long_jing_jiang_chengqi_zhang:_bi-directional_nodate}~\cite{wei_wang_chen_wu_ming_yan:_multi-granularity_nodate}~\cite{yiming_cui_zhipeng_chen_si_wei_shijin_wang_ting_liu_guoping_hu:_attention-over-attention_nodate}~\cite{s._liu;_s._zhang;_x._zhang;_h._wang_r-trans:_2019}~\cite{yiming_cui;ting_liu;zhipeng_chen;shijin_wang;guoping_hu_consensus_2018}                                       \\ \hline

\textbf{Natural Language Understanding}   &                  ~\cite{kim2018efficient}                                       \\ \hline

\textbf{Natural Language Transduction}    &                   ~\cite{grefenstette2015learning}                              \\ \hline

\textbf{Natural Language Generation}     &                  ~\cite{xu2018graph2seq}                                       \\ \hline

\textbf{Machine Reading}               &                  ~\cite{yang2019leveraging}                                    \\ \hline

\textbf{Intent Detection}              &                  ~\cite{liu2016attention}                                       \\ \hline

\textbf{Grammatical Correction}        &                  ~\cite{ahmadi2018attention}                                    \\ \hline

\textbf{Entity Resolution}             &                  ~\cite{das2016chains}~\cite{ganea2017deep}                     \\ \hline

\textbf{Entity Description}            &                  ~\cite{ji2017distant}                                          \\ \hline

\textbf{Embedding}                     &                  ~\cite{lin2017structured}~\cite{schick2019attentive}~\cite{zhu2018self}
\\ \hline

\textbf{Dependency Parsing}            &                  ~\cite{dozat2016deep}~\cite{strubell2018linguistically}        \\ \hline

\textbf{Conversation Model}          &                  ~\cite{zhou2018commonsense}~\cite{zhang2019sequence}           \\ \hline
\textbf{Automatic Question Tagging}    &                  ~\cite{sun2018automatic}                                       \\ \hline

\end{tabular}
\caption{Summary others applications of attention in natural language processing.}
\label{tab:others_nlp_applications}
\end{table}


\subsection{Computer Vision (CV)} 
\label{sec:application_cv}

Visual attention has become popular in many CV tasks. Action recognition, counting crowds, image classification, image generation, object detection, person recognition, segmentation, saliency detection, text recognition, and tracking targets are the most explored sub-areas, as shown in Table~\ref{tab:main_cv_app}. Applications in other sub-areas still have few representative works, such as clustering, compression, deblurring, depth estimation, image restoration, among others, as shown in Table~\ref{tab:other_cv_application}.


Visual attention in \textbf{image classification} tasks was first addressed by Graves et al.~\cite{mnih_recurrent_2014}. In this domain, there are sequential approaches inspired by human saccadic movements~\cite{mnih_recurrent_2014} and feedforward-augmented structures CNNs (Section~\ref{sub:attention_classic_architectures}). The general goal is usually to amplify fine-grained recognition, improve classification in the presence of occlusions, sudden variations in points of view, lighting, and rotation. Some approaches aim to learn to \textit{look} at the most relevant parts of the input image, while others try to discern between discriminating regions through feature recalibration and ensemble predictors via attention. To \textbf{fine-grained recognition}, important advances have been achieved through recurrent convolutional networks in the classification of bird subspecies~\cite{fu_look_2017} and architectures trained via RL to classify vehicle subtypes~\cite{zhao_deep_2017}.

\begin{table}[htb]
\small
\centering
\fontfamily{pag}\selectfont
\begin{tabular}{ll}
\hline
\textbf{Task}       & \textbf{References}       \\ \hline

\textbf{Action Recognition} & \begin{tabular}[c]{@{}l@{}} ~\cite{sharma_action_2015}~\cite{girdhar2019video}~\cite{girdhar_attentional_2017}~\cite{song_end--end_2016}~\cite{zhang2018adding}~\cite{chen_^2-nets:_2018}~\cite{liu_global_2017}~\cite{bazzani2016recurrent}~\cite{du2017rpan}~\cite{li2018videolstm}\\ ~\cite{ma_attend_2017}~\cite{wang_eidetic_2018}~\cite{dong_li;ting_yao;ling-yu_duan;tao_mei;yong_rui_unified_2019}~\cite{p._zhang;_c._lan;_j._xing;_w._zeng;_j._xue;_n._zheng_view_nodate}~\cite{wu_zheng;lin_li;zhaoxiang_zhang;yan_huang;liang_wang_relational_2019}~\cite{harshala_gammulle;simon_denman;sridha_sridharan;clinton_fookes_two_2017}~\cite{x._wang;_l._gao;_j._song;_h._shen_beyond_2017}~\cite{zhaoxuan_fan;xu_zhao;tianwei_lin;haisheng_su_attention-based_2019}~\cite{chenyang_si;wentao_chen;wei_wang;liang_wang;tieniu_tan_attention_2019}\\ ~\cite{swathikiran_sudhakaran;sergio_escalera;oswald_lanz_lsta:_2019}~\cite{noauthor_skeleton_nodate} \end{tabular} \\ \hline

\textbf{Counting Crowds}  & \begin{tabular}[c]{@{}l@{}} ~\cite{liu_decidenet:_2018}~\cite{lingbo_liu;hongjun_wang;guanbin_li;wanli_ouyang;liang_lin_crowd_2018}~\cite{mohammad_asiful_hossain;mehrdad_hosseinzadeh;omit_chanda;yang_wang_crowd_2019}~\cite{youmei_zhang_chunluan_zhou_faliang_chang_alex_c._kot:_multi-resolution_nodate}~\cite{ning_liu;yongchao_long;changqing_zou;qun_niu;li_pan;hefeng_wu_adcrowdnet:_2019} \end{tabular} \\ \hline

\textbf{Image Classification}    &                          \begin{tabular}[c]{@{}l@{}} ~\cite{fu_look_2017}~\cite{serra_overcoming_2018}~\cite{han_attribute-aware_2019}~\cite{sermanet_attention_2014}~\cite{hackel_inference_2018}~\cite{kang_deep_2018}~\cite{mnih_recurrent_2014}~\cite{chen_^2-nets:_2018}~\cite{ke_sparse_2018}~\cite{rodriguez_painless_2018}\\ ~\cite{wang_residual_2017}~\cite{jaderberg_spatial_2015}~\cite{seo_hierarchical_2016}~\cite{vinyals_matching_2016}~\cite{hu_squeeze-and-excitation_2017}~\cite{z._wang;_t._chen;_g._li;_r._xu;_l._lin_multi-label_nodate}~\cite{irwan_bello;barret_zoph;ashish_vaswani;jonathon_shlens;quoc_v._le_attention_2019}~\cite{lin_wu;yang_wang;xue_li;junbin_gao_deep_2019}~\cite{bo_zhao;xiao_wu;jiashi_feng;qiang_peng;shuicheng_yan_diversified_2017}~\cite{bo_zhao;xiao_wu;jiashi_feng;qiang_peng;shuicheng_yan_diversified_2017}\\ ~\cite{han-jia_ye;hexiang_hu;de-chuan_zhan;fei_sha_learning_2019}~\cite{emanuele_pesce;petros-pavlos_ypsilantis;samuel_withey;robert_bakewell;vicky_goh;giovanni_montana_learning_2019}~\cite{xiaoguang_mei;erting_pan;yong_ma;xiaobing_dai;jun_huang;fan_fan;qinglei_du;hong_zheng;jiayi_ma_spectral-spatial_2019}~\cite{bin_zhao;xuelong_li;xiaoqiang_lu;zhigang_wang_cnn-rnn_2019}~\cite{q._yin;_j._wang;_x._luo;_j._zhai;_s._k._jha;_y._shi_quaternion_2019}~\cite{tianjun_xiao;_yichong_xu;_kuiyuan_yang;_jiaxing_zhang;_yuxin_peng;_z._zhang_application_nodate}~\cite{hiroshi_fukui;tsubasa_hirakawa;takayoshi_yamashita;hironobu_fujiyoshi_attention_2019}~\cite{sanghyun_woo;jongchan_park;joon-young_lee;in_so_kweon_cbam:_2018}~\cite{hailin_hu;an_xiao;sai_zhang;yangyang_li;xuanling_shi;tao_jiang;linqi_zhang;lei_zhang;jianyang_zeng_deephint:_2019}~\cite{qingji_guan;yaping_huang;zhun_zhong;zhedong_zheng;liang_zheng;yi_yang_diagnose_2018}\\~\cite{bei_fang;ying_li;haokui_zhang;jonathan_cheung-wai_chan_hyperspectral_2019}~\cite{jianming_zhang;sarah_adel_bargal;zhe_lin;jonathan_br;t;xiaohui_shen;stan_sclaroff_top-down_2018}~\cite{w._wang;_w._wang;_y._xu;_j._shen;_s._zhu_attentive_2018}~\cite{wonsik_kim;bhavya_goyal;kunal_chawla;jungmin_lee;keunjoo_kwon_attention-based_2018}~\cite{mengye_ren;renjie_liao;ethan_fetaya;richard_s._zemel_incremental_2019}~\cite{yuxin_peng;xiangteng_he;junjie_zhao_object-part_2018}~\cite{qi_wang;shaoteng_liu;jocelyn_chanussot;xuelong_li_scene_2019}~\cite{hazel_doughty;walterio_mayol-cuevas;dima_damen_pros_2019}~\cite{noauthor_object-part_nodate}\end{tabular}   
\\ \hline

\textbf{Image Generation}  &   
\multicolumn{1}{l}{\begin{tabular}[l]{@{}l@{}}~\cite{draw}~\cite{zhang2018self}~\cite{kastaniotis_attention-aware_2018}~\cite{yu_generative_2018}~\cite{bornschein2017variational}~\cite{reed_few-shot_2017}~\cite{parmar2018image}~\cite{xinyuan_chen;chang_xu;xiaokang_yang;dacheng_tao_attention-gan_2018}~\cite{rewon_child;scott_gray;alec_radford;ilya_sutskever_generating_2019}~\cite{jianan_li;jimei_yang;aaron_hertzmann;jianming_zhang;tingfa_xu_layoutgan:_2019}\\
~\cite{hao_tang;dan_xu;nicu_sebe;yanzhi_wang;jason_j._corso;yan_yan_multi-channel_2019} \end{tabular}} \\ \hline

\textbf{Object Recognition}  &                          \multicolumn{1}{l}{\begin{tabular}[c]{@{}c@{}}~\cite{q._chu;_w._ouyang;_h._li;_x._wang;_b._liu;_n._yu_online_nodate}~\cite{zheng2017learning}~\cite{zhedong_zheng;liang_zheng;yi_yang_pedestrian_2017}~\cite{x._liu;_h._zhao;_m._tian;_l._sheng;_j._shao;_s._yi;_j._yan;_x._wang_hydraplus-net:_nodate}~\cite{he;_anfeng;luo;_chong;tian;_xinmei;zeng;_wenjun_twofold_2018}~\cite{xinlei_chen;li-jia_li;li_fei-fei;abhinav_gupta_iterative_2018}~\cite{jianlou_si;honggang_zhang;chun-guang_li;jason_kuen;xiangfei_kong;alex_c._kot;gang_wang_dual_2018}~\cite{chunfeng_song;yan_huang;wanli_ouyang;liang_wang_mask-guided_2018}~\cite{yi_zhou;ling_shao_viewpoint-aware_2018} \end{tabular}} \\ \hline

\textbf{Object Detection}                               &                          \multicolumn{1}{l}{\begin{tabular}[c]{@{}c@{}}~\cite{zhang_progressive_2018}~\cite{tao_kong;fuchun_sun;wenbing_huang;huaping_liu_deep_2018}~\cite{guanbin_li;yukang_gan;hejun_wu;nong_xiao;liang_lin_cross-modal_2019}~\cite{shuhan_chen;xiuli_tan;ben_wang;xuelong_hu_reverse_2019}~\cite{hao_chen;youfu_li_three-stream_2019}~\cite{xudong_wang;zhaowei_cai;dashan_gao;nuno_vasconcelos_towards_2019} \end{tabular}} \\ \hline

\textbf{Person Recognition}      &                          \multicolumn{1}{l}{\begin{tabular}[l]{@{}l@{}}~\cite{li_harmonious_2018}~\cite{hao_liu;jiashi_feng;meibin_qi;jianguo_jiang;shuicheng_yan_end--end_2017}~\cite{xingyu_liao;lingxiao_he;zhouwang_yang;chi_zhang_video-based_2019}~\cite{d._chen;_h._li;_t._xiao;_s._yi;_x._wang_video_2018}~\cite{l._zhao;_x._li;_y._zhuang;_j._wang_deeply-learned_nodate}~\cite{z._zhou;_y._huang;_w._wang;_l._wang;_t._tan_see_nodate}~\cite{jing_xu;rui_zhao;feng_zhu;huaming_wang;wanli_ouyang_attention-aware_2018}~\cite{shuang_li;slawomir_bak;peter_carr;xiaogang_wang_diversity_2018}~\cite{meng_zheng;srikrishna_karanam;ziyan_wu;richard_j._radke_re-identification_2019}~\cite{deqiang_ouyang;yonghui_zhang;jie_shao_video-based_2019} \\ ~\cite{cheng_wang;qian_zhang;chang_huang;wenyu_liu;xinggang_wang_mancs:_2018}~\cite{noauthor_jointly_nodate}~\cite{noauthor_scan_nodate} \end{tabular}} \\ \hline

\textbf{Segmentation}                                                                                                    &                          
\multicolumn{1}{l}{\begin{tabular}[l]{@{}l@{}}~\cite{fu_dual_2018}~\cite{li_tell_2018}~\cite{ren_end--end_2017}~\cite{zhang_deep_2019}~\cite{chen_attention_2016}~\cite{jetley_learn_2018}~\cite{shikun_liu;edward_johns;andrew_j._davison_end--end_2019}~\cite{yuhui_yuan;jingdong_wang_ocnet:_2019}~\cite{hanchao_li;pengfei_xiong;jie_an;lingxue_wang_pyramid_2018}~\cite{xiaowei_hu;chi-wing_fu;lei_zhu;jing_qin;pheng-ann_heng_direction-aware_2019} \\ ~\cite{z._zeng;_w._xie;_y._zhang;_y._lu_ric-unet:_2019}~\cite{b._shuai;_z._zuo;_b._wang;_g._wang_scene_2018}~\cite{xinxin_hu;kailun_yang;lei_fei;kaiwei_wang_acnet:_2019}~\cite{ozan_oktay;jo_schlemper;loic_le_folgoc;matthew_lee;mattias_heinrich;kazunari_misawa;kensaku_mori;steven_mcdonagh;nils_y_hammerla;bernhard_kainz;ben_glocker;daniel_rueckert_attention_2018}~\cite{xueying_chen;rong_zhang;pingkun_yan_feature_2019}~\cite{ruirui_li;mingming_li;jiacheng_li;yating_zhou_connection_2019}~\cite{shu_kong;charless_c._fowlkes_pixel-wise_2019}~\cite{xiaoxiao_li;chen_change_loy_video_2018}~\cite{hengshuang_zhao;yi_zhang;shu_liu;jianping_shi;chen_change_loy;dahua_lin;jiaya_jia_psanet:_2018} \end{tabular}} \\ \hline

\textbf{Saliency Detection}  &                   ~\cite{j._kuen;_z._wang;_g._wang_recurrent_nodate}~\cite{nian_liu;junwei_han;ming-hsuan_yang_picanet:_2018}~\cite{nian_liu;junwei_han;ming-hsuan_yang_picanet:_2018}~\cite{marcella_cornia;lorenzo_baraldi;giuseppe_serra;rita_cucchiara_predicting_2018}~\cite{xiaowei_hu;chi-wing_fu;lei_zhu;pheng-ann_heng_sac-net:_2019}                                      \\ \hline

\textbf{Text Recognition}   &    ~\cite{he_end--end_2018}~\cite{cheng_focusing_2017}~\cite{canjie_luo;lianwen_jin;zenghui_sun_moran:_2019}~\cite{hongtao_xie;shancheng_fang;zheng-jun_zha;yating_yang;yan_li;yongdong_zhang_convolutional_2019}~\cite{hui_li;peng_wang;chunhua_shen;guyu_zhang_show_2019}~\cite{noauthor_focusing_nodate}             \\ \hline

\textbf{Tracking Targets}   &                            
\multicolumn{1}{l}{\begin{tabular}[l]{@{}l@{}}~\cite{fu_dual_2018}~\cite{li_tell_2018}~\cite{ren_end--end_2017}~\cite{zhang_deep_2019}~\cite{chen_attention_2016}~\cite{jetley_learn_2018}~\cite{shikun_liu;edward_johns;andrew_j._davison_end--end_2019}~\cite{yuhui_yuan;jingdong_wang_ocnet:_2019}~\cite{hanchao_li;pengfei_xiong;jie_an;lingxue_wang_pyramid_2018}~\cite{xiaowei_hu;chi-wing_fu;lei_zhu;jing_qin;pheng-ann_heng_direction-aware_2019} \\ ~\cite{z._zeng;_w._xie;_y._zhang;_y._lu_ric-unet:_2019}~\cite{b._shuai;_z._zuo;_b._wang;_g._wang_scene_2018}~\cite{xinxin_hu;kailun_yang;lei_fei;kaiwei_wang_acnet:_2019}~\cite{ozan_oktay;jo_schlemper;loic_le_folgoc;matthew_lee;mattias_heinrich;kazunari_misawa;kensaku_mori;steven_mcdonagh;nils_y_hammerla;bernhard_kainz;ben_glocker;daniel_rueckert_attention_2018}~\cite{xueying_chen;rong_zhang;pingkun_yan_feature_2019} \end{tabular}} \\ \hline
\end{tabular}
\caption{Summary state-of-art approaches in computer vision sub-areas.}
\label{tab:main_cv_app}
\end{table}

Visual attention also provides significant benefits for \textbf{action recognition} tasks by capturing spatio-temporal relationships. The biggest challenge's classical approaches are capturing discriminative features of movement in the sequences of images or videos. The attention allows the network to focus the processing only on the relevant joints or on the movement features easily. Generally, the main approaches use the following strategies: 1) \textbf{saliency maps}: spatiotemporal attention models learn \textit{where to look} in video directly human fixation data. These models express the probability of saliency for each pixel. Deep 3D CNNs extract features only high saliency regions to represent spatial and short time relations at clip level, and LSTMs expand the temporal domain from few frames to seconds~\cite{bazzani2016recurrent}; 2) \textbf{self-attention}: modeling context-dependencies. The person being classified is the Query (Q), and the clip around the person is the memory, represented by keys (K) and values (V) vectors. The network process the query and memory to generate an updated query vector. Intuitively self-attention adds context to other people and objects in the clip to assist in subsequent classification~\cite{girdhar2019video}; 3) \textbf{recurrent attention mechanisms}: captures relevant positions of joints or movement features and, through a recurring structure, refines the attentional focus at each time step~\cite{liu2017global}~\cite{du2017rpan}; and 4) \textbf{temporal attention}: captures relevant spatial-temporal locations~\cite{li2018videolstm}~\cite{li2018videolstm}~\cite{song_end--end_2016}~\cite {xin2016recurrent}~\cite{zang2018attention}~\cite{pei2017temporal}.

\begin{figure}[htb]
  \centering
  \includegraphics[width=\linewidth]{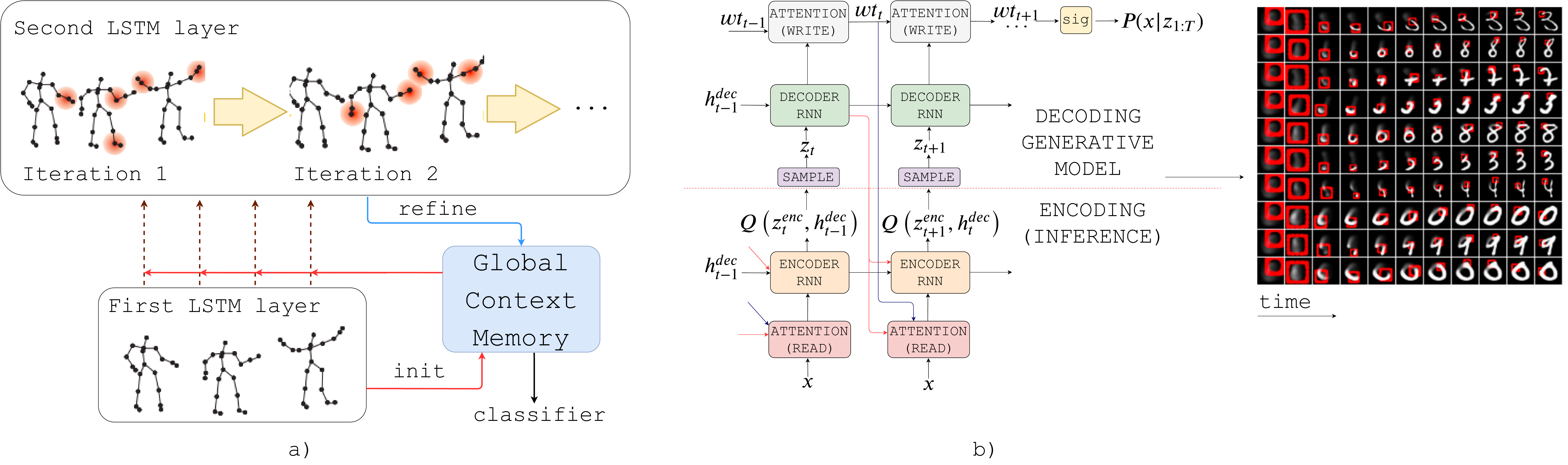}  
  \caption{Illustration of Global Context-Aware Attention~\cite{liu2017global} for action recognition, and DRAW~\cite{draw} for image generation. a) The Context-Aware Attention Network. The first LSTM layer encodes the skeleton sequence and generates an initial global context memory. The second layer performs attention over the inputs with global context memory assistance and generates a refined representation. The refine representation is then used back to the global context. Multiple attention iterations are carried out to refine the global context progressively. Finally, memory information is used for classification. b) The DRAW architecture. At each time step $t$, the input is read by attention and passed to the encoder RNN. The encoder's output is used to compute the approximate posterior over the latent variables. On the right, an illustration shows the iterative process of generating some images. Each row shows successive stages in the generation of a single digit. The network guided by attention draws and refine regions successively. The red rectangle delimits the area attended to by the network at each time-step.}
  \label{fig:cv_apps}
\end{figure}

Liu et al.~\cite{liu2017global} model is a recurrent attention approach to capturing the person's relevant positions. This model presented a pioneering approach using two layers of LSTMs and the context memory cell that recurrently interact with each other, as shown in figure~\ref{fig:cv_apps}a. First, a layer of LSTMs generates an encoding of a skeleton sequence, initializing the context memory cell. The memory representation is input to the second layer of LSTMs and helps the network selectively focus on each frame's informational articulations. Finally, attentional representation feeds back the context memory cell to refine the focus's orientation again by paying attention more reliably.
Similarly, Du et al.~\cite{du2017rpan} proposed RPAN - a recurrent attention approach between sequentially modeling by LSTMs and convolutional features extractors. First, CNNs extract features from the current frame, and the attentional mechanism guided by the LSTM's previous hidden state estimates a series of features related to human articulations related to the semantics of movements of interest. Then, these highly discriminative features feed LSTM time sequences.

In \textbf{image generation}, there were also notable benefits. DRAW~\cite{draw} introduced visual attention with an innovative approach - image patches are generated sequentially and gradually refined, in which to generate the entire image in a single pass (figure~\ref{fig:cv_apps}b). Subsequently, attentional mechanisms emerged in generative adversarial networks (GANs) to minimize the challenges in modeling images with structural constraints. Naturally, GANs efficiently synthesize elements differentiated by texture (i.e., oceans, sky, natural landscapes) but suffer to generate geometric patterns (i.e., faces, animals, people, fine details). The central problem is the convolutions that fail to model dependencies between distant regions. Besides, the statistical and computational efficiency of the model suffers from the stacking of many layers. The attentional mechanisms, especially self-attention, offered a computationally inexpensive alternative to model long-range dependencies easily. Self-attention as a complement to convolution contributes significantly to the advancement of the area with approaches capable of generating fine details~\cite{zhang2018self}, high-resolution images, and with intricate geometric patterns~\cite{chen2020generative}.

In \textbf{expression recognition}, attention optimizes the entire segmentation process by scanning input as a whole sequence, choosing the most relevant region to describe a segmented symbol or implicit space operator~\cite{zhang2017gru}. In \textbf{information retriever}, attention helps obtain appropriate semantic resources using individual class semantic resources to progressively orient visual aids to generate an attention map to ponder the importance of different local regions.~\cite{ji2018stacked}.

In \textbf{medical image analysis}, attention helps implicitly learn to suppress irrelevant areas in an input image while highlighting useful resources for a specific task. This allows us to eliminate the need to use explicit external tissue/organ localization modules using convolutional neural networks (CNNs). Besides, it allows generating both images and maps of attention in unsupervised learning useful for data annotation. For this, there are the ATA-GANS~\cite{kastaniotis_attention-aware_2018} and attention gate~\cite{schlemper_attention_2018} modules that work with unsupervised and supervised learning, respectively.

\begin{table}[htb]
\small
\centering
\fontfamily{pag}\selectfont

\begin{tabular}{lll}
\hline
\textbf{Task}           & \textbf{References}                                     \\ \hline
\textbf{Clustering}    & \begin{tabular}[c]{@{}l@{}} ~\cite{lee2018set} \end{tabular} \\ \hline

\textbf{Compression}    &  ~\cite{meng2018mganet}   \\\hline

\textbf{Deblurring}   &  ~\cite{park2019down}  \\ \hline

\textbf{Depth Estimation}         &  ~\cite{xu2018structured}~\cite{liu2019end}  \\ \hline

\textbf{Image Restoration}      &   ~\cite{zhang2019residual}~\cite{suganuma2019attention}~\cite{qian2018attentive}                                                      \\ \hline

\textbf{Image-to-Image Translation}      & ~\cite{mejjati2018unsupervised}~\cite{tang2019attention} \\ \hline

\textbf{Information Retriever}        & ~\cite{ji2018stacked}~\cite{jin2018deep}~\cite{yang2019deep} \\ \hline

\textbf{Medical Image Analysis}    &             ~\cite{kastaniotis_attention-aware_2018}~\cite{schlemper_attention_2018}~\cite{huiyan_jiang;tianyu_shi;zhiqi_bai;liangliang_huang_ahcnet:_2019}                                   \\ \hline

\textbf{Multiple Instance Learning}     &    ~\cite{ilse2018attention}    \\ \hline

\textbf{Ocr}      &                 ~\cite{lee2016recursive}                 \\ \hline

\textbf{One-Shot Learning}     &                 ~\cite{rezende2016one}                                    \\ \hline

\textbf{Pose Estimation}    &                 ~\cite{chu2017multi}                                    \\ \hline

\textbf{Super-Resolution}      &                   ~\cite{zhang2018image}                  \\ \hline

\textbf{Transfer Learning}     &                   ~\cite{zagoruyko_paying_2016}~\cite{xiao-yu_zhang;haichao_shi;changsheng_li;kai_zheng;xiaobin_zhu;lixin_duan_learning_2019}~\cite{xingjian_li;haoyi_xiong;hanchao_wang;yuxuan_rao;liping_liu;jun_huan_delta:_2019}                                      \\ \hline

\textbf{Video Classification}     &                   ~\cite{bielski2018pay}~\cite{long2018attention}        \\ \hline

\textbf{Facial Detection}      &                   ~\cite{tian2018learning}~\cite{yundong_zhang;xiang_xu;xiaotao_liu_robust_2019}~\cite{shengtao_xiao;jiashi_feng;junliang_xing;hanjiang_lai;shuicheng_yan;ashraf_a._kassim_robust_2016}                                      \\ \hline

\textbf{Fall Detection}   &                   ~\cite{lu2018deep}                                      \\ \hline

\textbf{Person Detection}      &                   ~\cite{zhang2019cross}~\cite{zhang2018occluded}         \\ \hline

\textbf{Sign Detection}    &                   ~\cite{yuan2019vssa}                                      \\ \hline

\textbf{Text Detection}   &                   ~\cite{he_end--end_2018}~\cite{wojna_attention-based_2017}~\cite{he_single_2017}~\cite{bhunia_script_2019}                                      \\ \hline

\textbf{Face Recognition}     &                   ~\cite{yang2017neural}                 \\ \hline

\textbf{Facial Expression Recognition}   &                   ~\cite{siyue_xie;haifeng_hu;yongbo_wu_deep_2019}~\cite{shervin_minaee;amirali_abdolrashidi_deep-emotion:_2019}~\cite{yong_li;jiabei_zeng;shiguang_shan;xilin_chen_occlusion_2019} \\ \hline

\textbf{Sequence Recognition}        &                   ~\cite{yang2019fully}                                    \\ \hline

\end{tabular}

\caption{Summary others applications of attention in computer vision.}
\label{tab:other_cv_application}
\end{table}


For \textbf{person recognition}, attention has become essential in in-person re-identification (re-id)~\cite{han_attribute-aware_2019}~\cite{ meng_zheng;srikrishna_karanam;ziyan_wu;richard_j._radke_re-identification_2019}~\cite{li_harmonious_2018}. Re-id aims to search for people seen from a surveillance camera implanted in different locations. In classical approaches, the bounding boxes of detected people were not optimized for re-identification suffering from misalignment problems, background disorder, occlusion, and absent body parts. Misalignment is one of the biggest challenges, as people are often captured in various poses, and the system needs to compare different images. In this sense, neural attention models started to lead the developments mainly with multiple attentional mechanisms of alignment between different bounding boxes.

There are still less popular applications, but for which attention plays an essential role. Self-attention models iterations between the input set for clustering tasks~\cite{lee2018set}. Attention refines and merges multi-scale feature maps in-depth estimation and edge detection~\cite{xu2017learning}~\cite{xu2018structured}. In video classification, attention helps capture global and local resources generating a comprehensive representation~\cite{xie2019semantic}. It also measures each time interval's relevance in a sequence~\cite{pei2017temporal}, promoting a more intuitive interpretation of the impact of content on the video's popularity, providing the regions that contribute the most to the prediction~\cite{bielski2018pay}. In face detection, attention dynamically selects the main reference points of the face~\cite{shengtao_xiao;jiashi_feng;junliang_xing;hanjiang_lai;shuicheng_yan;ashraf_a._kassim_robust_2016}. It improves deblurring in each convolutional layer in deblurring, preserving fine details~\cite{park2019down}. Finally, in emotion recognition, it captures complex relationships between audio and video data by obtaining regions where both signals relate to emotion~\cite{zhang2019deep}.

\subsection{\textbf{Multimodal Tasks (CV/NLP)}}
\label{sec:application_cv_nlp}

\begin{figure}[htb]
  \centering
  \includegraphics[width=\linewidth]{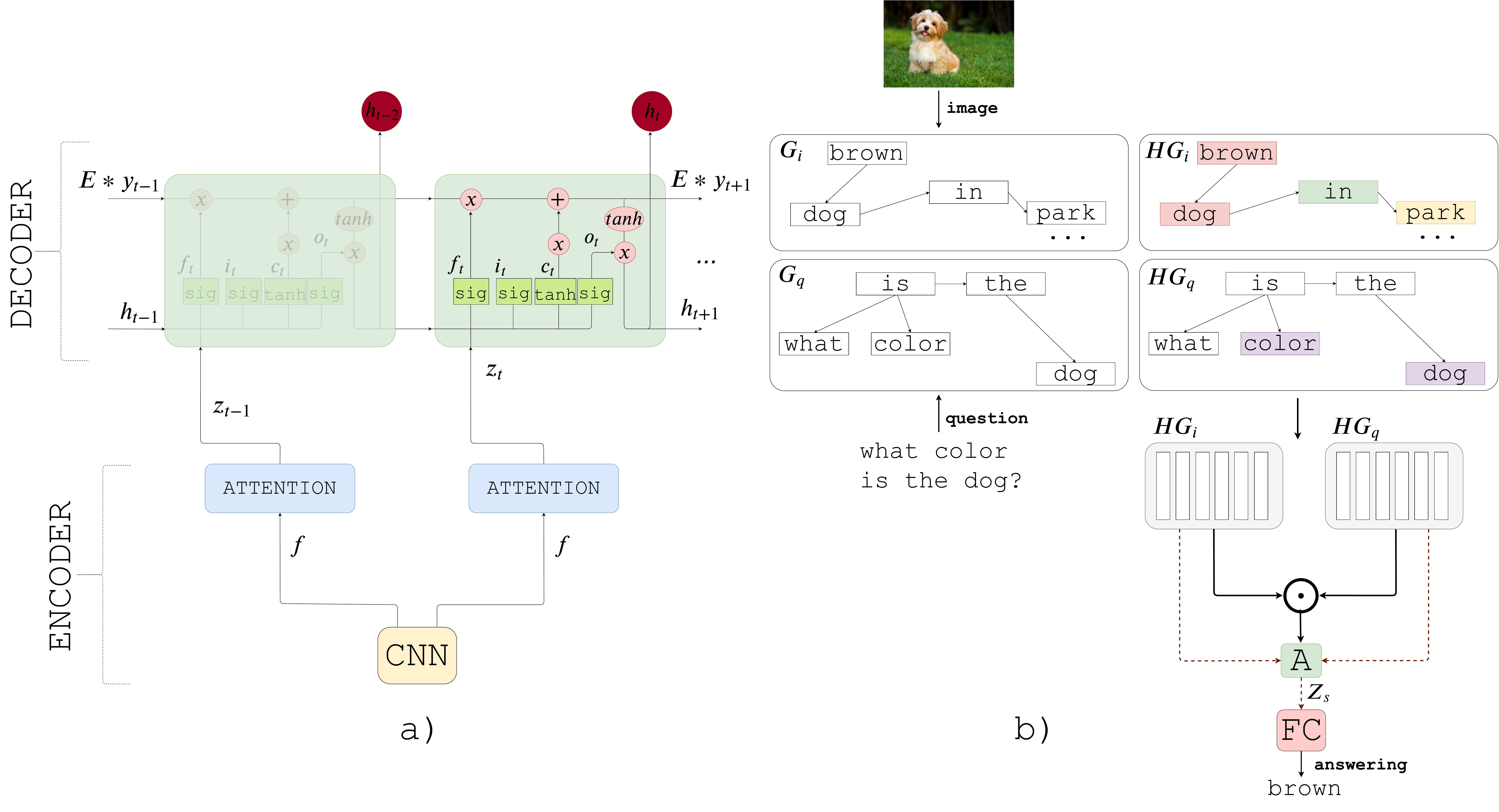}  
  \caption{Illustration of classic attentive encoder-decoder framework by Xu et al.~\cite{xu_show_2015} and Hypergraph Attention Network~\cite{kim2020hypergraph} for question answering tasks. a) The \textit{show, attend and tell} framework. At each time step $t$, attention takes as input visual feature maps and previous hidden state of the decoder and produces a dynamic context vector with essential visual features to predict the next word. b) The Hypergraph Attention Network. For a given pair of images and questions, two symbolic graphs $G_{i}$, and $G_{q}$ are constructed. After, two hypergraphs $HG_{i}$, and $HG_{q}$ with random-walk hyperedge are constructed and combined via co-attention map $A$. Finally, the final representation $Z_{s}$ is used to predict an answer for the given question.}
  \label{fig:multimodal_apps}
\end{figure}

Attention has been used extensively in multimodal learning, mainly for mapping complex relationships between different sensory modalities. In this domain, the importance of attention is quite intuitive, given that communication and human sensory processing are completely multimodal. The first approaches emerged from 2015 inspired by an attentive encoder-decoder framework entitled ``Show, attend and tell: Neural image caption generation with visual attention'' by Xu et al.~\cite{xu_show_2015}. In this framework, depicted in figure~\ref{fig:multimodal_apps}a at each time step $t$, attention generates a vector with a dynamic context of visual features based on the words previously generated - a principle very similar to that presented in RNNSearch~\cite{bahdanau_neural_2014}. Later, more elaborate methods using visual and textual sources were developed mainly in image captioning, video captioning, and visual question answering, as shown in the Table~\ref{table:cv_nlp_applications}.

\begin{table}[htb]
\small
\centering
\fontfamily{pag}\selectfont

\begin{tabular}{ll}
\hline
\textbf{Task}  & \textbf{References}  \\ \hline

\textbf{Emotion Recognition} & ~\cite{tan2019multimodal}~\cite{zadeh2018memory}~\cite{zadeh2018multi} \\ \hline

\textbf{Expression Comprehension} & ~\cite{yu2018mattnet}     \\ \hline

\textbf{Image Classification} & ~\cite{wang2018tienet}     \\ \hline

\textbf{Text-to-Image Generation} & ~\cite{xu2018attngan}     \\ \hline

\textbf{Image-to-Text Generation} & ~\cite{poulos2017character}     \\ \hline

\textbf{Image Captioning}  & \begin{tabular}[c]{@{}l@{}} ~\cite{noauthor_bottom-up_nodate}~\cite{zhu_image_2018}~\cite{xu_show_2015}~\cite{ma_da-gan:_2018-1}~\cite{lu_knowing_2017}~\cite{you_image_2016}~\cite{chen_sca-cnn:_2017}~\cite{he_vd-san:_2019}\\ ~\cite{noauthor_bi-directional_nodate}~\cite{fang_captions_2015}~\cite{yang_review_2016}~\cite{pedersoli_areas_2016}~\cite{ting_yao;yingwei_pan;yehao_li;tao_mei_exploring_2018}~\cite{yehao_li;ting_yao;yingwei_pan;hongyang_chao;tao_mei_pointing_2019}~\cite{x._liang;_z._hu;_h._zhang;_c._gan;_e._p._xing_recurrent_nodate}~\cite{xiangrong_zhang;xin_wang;xu_tang;huiyu_zhou;chen_li_description_2019}~\cite{anna_rohrbach;marcus_rohrbach;ronghang_hu;trevor_darrell;bernt_schiele_grounding_2017}\\ ~\cite{lukasz_kaiser;aidan_n._gomez;noam_shazeer;ashish_vaswani;niki_parmar;llion_jones;jakob_uszkoreit_one_2017}~\cite{kuang-huei_lee;xi_chen;gang_hua;houdong_hu;xiaodong_he_stacked_2018}~\cite{pan2020x}~\cite{noauthor_bottom-up_nodate} \end{tabular} 
\\ \hline

\textbf{Video Captioning}   & 
\begin{tabular}[c]{@{}l@{}}~\cite{cho_describing_2015}~\cite{wu_hierarchical_2018}~\cite{zhou2018end}~\cite{zhu2020actbert}~\cite{pu_adaptive_2018}~\cite{yao_describing_2015}~\cite{yu_video_2015}~\cite{hori_attention-based_2017}~\cite{krishna_dense-captioning_2017} \\~\cite{ma_attend_2017}~\cite{yi_bin;yang_yang;fumin_shen;ning_xie;heng_tao_shen;xuelong_li_describing_2019}~\cite{zhou2018end}~\cite{l._baraldi;_c._grana;_r._cucchiara_hierarchical_nodate}~\cite{silvio_olivastri;gurkirt_singh;fabio_cuzzolin_end--end_2019}~\cite{zhong_ji;kailin_xiong;yanwei_pang;xuelong_li_video_2018}~\cite{jingkuan_song;zhao_guo;lianli_gao;wu_liu;dongxiang_zhang;heng_tao_shen_hierarchical_2017}~\cite{xiangpeng_li;zhilong_zhou;lijiang_chen;lianli_gao_residual_2019}~\cite{lianli_gao;zhao_guo;hanwang_zhang;xing_xu;heng_tao_shen_video_2017}\\
~\cite{jingwen_wang;wenhao_jiang;lin_ma;wei_liu;yong_xu_bidirectional_2018} \end{tabular} \\ \hline
\textbf{Visual Question Answering}  & \begin{tabular}[c]{@{}l@{}} ~\cite{kim2020modality}~\cite{jiangfantastic}~\cite{liang_focal_2018}~\cite{hudson_compositional_2018}~\cite{gulcehre_hyperbolic_2018}~\cite{osman_dual_2018}~\cite{lu_hierarchical_2016}\\ ~\cite{nam_dual_2017}~\cite{deng_latent_2018}~\cite{kim_bilinear_2018}~\cite{nguyen_improved_2018}~\cite{andreas_deep_2015}~\cite{kim_multimodal_2016}~\cite{shih_where_2015}~\cite{xiong_dynamic_2016}~\cite{zhu_visual7w:_2015}\\ ~\cite{jiasen_lu;anitha_kannan;jianwei_yang;devi_parikh;dhruv_batra_best_2017}~\cite{zhou_yu;jun_yu;chenchao_xiang;jianping_fan;dacheng_tao_beyond_2019}~\cite{aishwarya_agrawal;dhruv_batra;devi_parikh;aniruddha_kembhavi_dont_2018}~\cite{yan_zhang;jonathon_hare;adam_prugel-bennett_learning_2018}~\cite{zhe_gan;yu_cheng;ahmed_el_kholy;linjie_li;jingjing_liu;jianfeng_gao_multi-step_2019}~\cite{kan_chen;jiang_wang;liang-chieh_chen;haoyuan_gao;wei_xu;ram_nevatia_abc-cnn:_2016}~\cite{huijuan_xu;kate_saenko_ask;_2016}~\cite{yundong_zhang;juan_carlos_niebles;alvaro_soto_interpretable_2019}~\cite{zheng_zhang;lizi_liao;minlie_huang;xiaoyan_zhu;tat-seng_chua_neural_2019}\\ ~\cite{noauthor_focal_nodate}~\cite{kim2020hypergraph}~\cite{yang_stacked_2016}~\cite{yu_multi-level_2017}~\cite{noauthor_bottom-up_nodate}~\cite{z._yu;_j._yu;_j._fan;_d._tao_multi-modal_nodate}\end{tabular} 
\\ \hline
\end{tabular}
\caption{Summary of state-of-art approaches in multimodal tasks (CV/NLP).}
\label{table:cv_nlp_applications}
\end{table}

For image captioning, Yan et al.~\cite{yang_review_2016} extended the seminal framework by Xu et al.~\cite{xu_show_2015} with review attention, a sequence of modules that capture global information in various stages of reviewing hidden states and generate more compact, abstract, and global context vectors.
Zhu et al.~\cite{zhu_image_2018} presented a triple attention model which enhances object information at the text generation stage. Two attention mechanisms capture semantic visual information in input, and a mechanism in the prediction stage integrates word and image information better. Lu et al.~\cite{lu_knowing_2017} presented an adaptive attention encoder-decoder framework that decides when to trust visual signals and when to trust only the language model. Specifically, their mechanism has two complementary elements: the \textit{visual sentinel} vector decides when to look at the image, and the \textit{sentinel gate} decides how much new information the decoder wants from the image. Recently, Pan et al.~\cite{pan2020x} created attentional mechanisms based on bilinear pooling capable of capturing high order interactions between multi-modal features, unlike the classic mechanisms that capture only first-order feature interactions.

Similarly, in visual question-answering tasks, methods seek to align salient textual features with visual features via feedforward or recurrent soft attention methods~\cite{noauthor_bottom-up_nodate}~\cite{osman_dual_2018}~\cite{lu_hierarchical_2016}~\cite{yang_stacked_2016}. More recent approaches aim to generate complex inter-modal representations. In this line, Kim et al.~\cite{kim2020hypergraph} proposed Hypergraph Attention Networks (HANs), a solution to minimize the disparity between different levels of abstraction from different sensory sources. So far, HAN is the first approach to define a common semantic space with symbolic graphs of each modality and extract an inter-modal representation based on co-attention maps in the constructed semantic space, as shown in figure~\ref{fig:multimodal_apps}b. Liang et al.~\cite{liang_focal_2018} used attention to capture hierarchical relationships between sequences of image-text pairs not directly related. The objective is to answer questions and justify what results in the system were based on answers.

For video captioning most approaches generally align textual features and spatio-temporal representations of visual features via simple soft attention mechanisms~\cite{cho_describing_2015}~\cite{yu_video_2015}~\cite{yao_describing_2015}~\cite{hori_attention-based_2017}. For example, Pu et al.~\cite{pu_adaptive_2018} design soft attention to adaptively emphasize different CNN layers while also imposing attention within local spatiotemporal regions of the feature maps at particular layers. These mechanisms define the importance of regions and layers to produce a word based on word-history information. Recently, self-attention mechanisms have also been used to capture more complex and explicit relationships between different modalities. Zhu et al.~\cite{zhu2020actbert} introduced ActBERT, a transformer-based approach trained via self-supervised learning to encode complex relations between global actions and local, regional objects and linguistic descriptions. Zhou et al.~\cite{zhou2018end} proposed a multimodal transformer via supervised learning, which employs a masking network to restrict its attention to the proposed event over the encoding feature.

Other applications also benefit from the attention. In the emotion recognition domain, the main approaches use memory fusion structures inspired by the human brain's communication understanding mechanisms. Biologically, different regions process and understand different modalities connected via neural links to integrate multimodal information over time. Similarly, in existing approaches, an attentional component models view-specific dynamics within each modality via recurrent neural networks, and a second component simultaneously finds multiple cross-view dynamics in each recurrence timestep by storing them in hybrid memories. Memory updates occur based on all the sequential data seen. Finally, to generate the output, the predictor integrates the two levels of information: view-specific and multiple cross-view memory information~\cite{zadeh2018memory}~\cite{zadeh2018multi}.

There are still few multimodal methods for classification. Whang et al.~\cite{wang2018tienet} presented a pioneering framework for classifying and describing image regions simultaneously from textual and visual sources. Their framework detects, classifies, and generates explanatory reports regarding abnormalities observed in chest X-ray images through multi-level attentional modules end-to-end in LSTMs and CNNs. In LSTMs, attention combines all hidden states and generates a dynamic context vector, then a spatial mechanism guided by a textual mechanism highlights the regions of the image with more meaningful information. Intuitively, the salient features of the image are extracted based on high-relevance textual regions.


\subsection{\textbf{Recommender Systems (RS)}}
\label{sec:application_recommender}

Attention has also been used in recommender systems for behavioral modeling of users. Capturing user interests is a challenging problem for neural networks, as some iterations are transient, some clicks are unintentional, and interests can change quickly in the same session. Classical approaches (i.e., Markov Chains and RNNs) have limited performance predicting the user's next actions, present different performances in sparse and dense datasets, and long-term memory problems. In this sense, attention has been used mainly to assign weights to a user's interacted items capturing long and short-term interests more effectively than traditional ones. Self-attention and memory approaches have been explored to improve the area's development. STAMP~\cite{liu2018stamp} model, based on attention and memory, manages users' general interests in long-term memories and current interests in short-term memories resulting in behavioral representations that are more coherent. The Collaborative Filtering~\cite{chen2017attentive} framework, and SASRec~\cite{kang_deep_2018} explored self-attention in capturing long-term semantics for finding the most relevant items in user's history.


\subsection{\textbf{Reinforcement Learning (RL)}}
\label{sec:application_rl}

Attention has been gradually introduced in reinforcement learning to deal with unstructured environments in which rewards and actions depend on past states and where it is challenging guaranteeing the Markov property. Specifically, the goals are to increase the agent's generalizability and minimize long-term memory problems. Currently, the main attentional reinforcement learning approaches are computer vision, graph reasoning, natural language processing, and virtual navigation, as shown in Table~\ref{tab:rl_applications}.

\begin{table}[htb]
\small
\centering
\fontfamily{pag}\selectfont

\begin{tabular}{lll}
\hline
\textbf{Task}   & \textbf{References}           \\ \hline

\textbf{Computer Vision} & 
\begin{tabular}[c]{@{}l@{}}~\cite{ba_multiple_2014}~\cite{li_action_2019}~\cite{rao_attention-aware_2017}~\cite{stollenga_deep_2014}~\cite{cao_attention-aware_2017} \\ ~\cite{zhao_deep_2017}~\cite{donghui_hu;shengnan_zhou;qiang_shen;shuli_zheng;zhongqiu_zhao;yuqi_fan_digital_2019}~\cite{marcus_edel;joscha_lausch_capacity_2016}~\cite{yeung_end--end_2016} \end{tabular}
 \\ \hline

\textbf{Graph reasoning}   & ~\cite{lee2018graph}                                                         \\ \hline
\textbf{Natural Language Processing}  &  ~\cite{santoro2018relational} \\ \hline

\textbf{Navigation}     &  ~\cite{mishra_simple_2017}~\cite{parisotto2017neural}~\cite{zambaldi2018relational}~\cite{santoro2018relational}~\cite{baker2019emergent}
\\ \hline

\end{tabular}
\caption{Summary of main state-of-art approaches in attentional reinforcement learning tasks.}
\label{tab:rl_applications}
\end{table}


To increase the ability to generalize in partially observable environments, some approaches use attention in the policy network. Mishra et al.~\cite{mishra_simple_2017} used attention to easily capture long-term temporal dependencies in convolutions in an agent's visual navigation task in random mazes. At each time step $t$, the model receives as input the current observation $o_ {t}$ and previous sequences of observations, rewards, and actions so that attention allows the policy to maintain a long memory of past episodes. Other approaches implement attention directly to the representation of the state. State representation is a classic and critical problem in RL, given that state space is one of the major bottlenecks for speed, efficiency, and generalization of training techniques. In this sense, the importance of attention on this topic is quite intuitive.

However, there are still few approaches exploring the representation of states. The neural map~\cite{parisotto2017neural} maintains an internal memory in the agent controlled via attention mechanisms. While the agent navigates the environment, an attentional mechanism alters the internal memory, dynamically constructing a history summary. At the same time, another generates a representation $o_{t}$, based on the contextual information of the memory and the state's current observation. Then, the policy network receives $o_{t}$ as input and generates the distribution of shares. Some more recent approaches affect the representation of the current $o_{t}$ observation of the state via self attention in iterative reasoning between entities in the scene~\cite{zambaldi2018relational}~\cite{baker2019emergent}, or between the current observation $o_{t}$ and memory units~\cite{santoro2018relational} to guide model-free policies.

\begin{figure}[!htb]
  \centering
  \includegraphics[width=\linewidth]{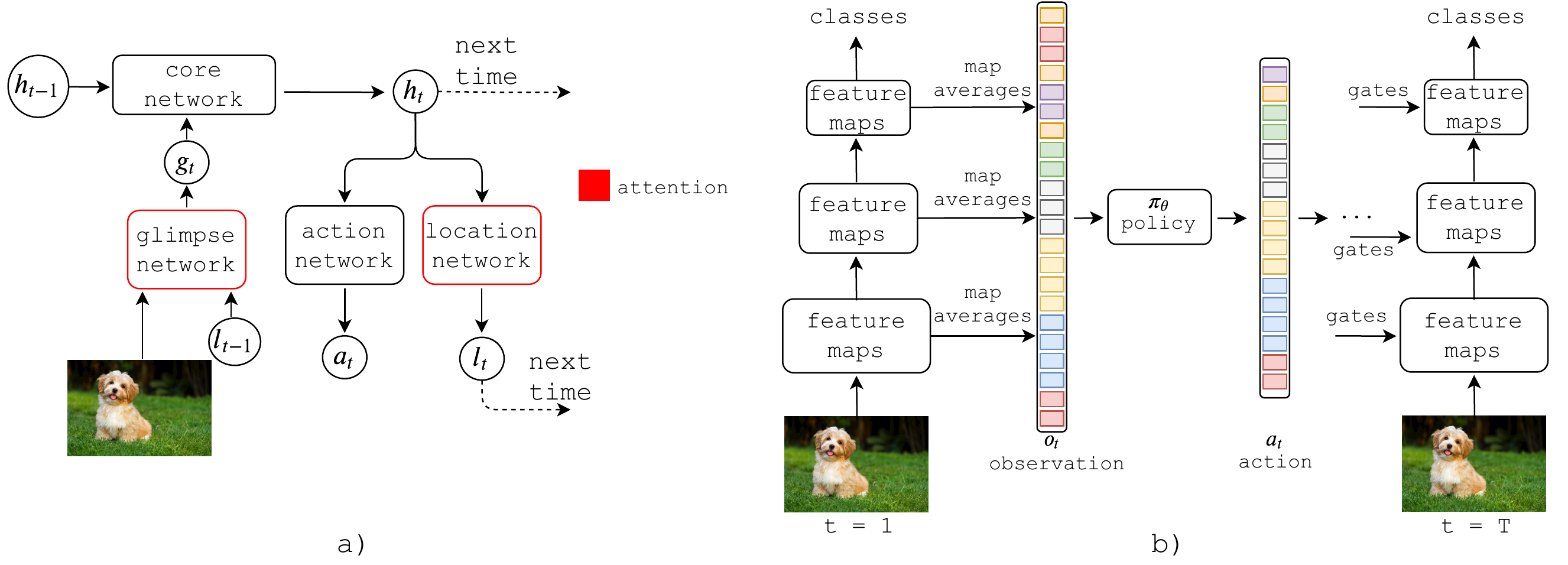}  
  \caption{Illustration of RAM~\cite{mnih_recurrent_2014} and dasNet~\cite{stollenga_deep_2014} for image classification tasks. a) The RAM framework. At each time step $t$, the glimpse networks extracts a retina-like representation based on input image and an focus location $l_{t-1}$. The core network takes the glimpse representation as input and combining it with the internal representation at the previous time step and produces the new internal state of the model $h_{t}$. The location network and action network use the $h_{t}$ to produce the next location $l_{t}$ to attend and the classification. b) The dasNet network. Each image is classified after $T$ passes through CNNs. After each forward propagation, the averages of feature maps are combined into an observation vector $o_{t}$ that is used by a deterministic policy to choose an action $a_{t}$ that changes the focus of all the feature maps for the next pass of the same image. Finally, after pass $T$ times, the output is used to classify the image.}
  \label{fig:rl_applications_ok}
\end{figure}

The most discussed topic is the use of the policy network to guide the attentional focus of the agent's glimpses sensors on the environment so that the representation of the state refers to only a small portion of the entire operating environment. This approach emerged initially by Graves et al.~\cite{mnih_recurrent_2014} using policy gradient methods (i.e., REINFORCE algorithm) in the hybrid training of recurrent networks in image classification tasks. Their model consists of a glimpse sensor that captures only a portion of the input image, a core network that maintains a summary of the history of patches seen by the agent, an action network that estimates the class of the image seen, and a location network trained via RL which estimates the focus of the glimpse on the next time step, as shown in figure~\ref{fig:rl_applications_ok}a. This structure considers the network as the agent, the image as the environment, and the reward is the number of correct network ratings in an episode. Stollenga et al.~\cite{stollenga_deep_2014} proposed a similar approach, however directly focused on CNNs, as shown in figure~\ref{fig:rl_applications_ok}b. The structure allows each layer to influence all the others through attentional bottom-up and top-down connections that modulate convolutional filters' activity. After supervised training, the attentional connections' weights implement a control policy via RL and SNES~\cite{schaul2011high}. The policy learns to suppress or enhance features at various levels by improving the classification of difficult cases not captured by the initial supervised training. Subsequently, variants similar to these approaches appeared in multiple image classification~\cite{ba_multiple_2014}~\cite{marcus_edel;joscha_lausch_capacity_2016}~\cite{zhao_deep_2017}, action recognition~\cite{yeung_end--end_2016}, and face hallucination~\cite{cao_attention-aware_2017}.

\subsection{\textbf{Robotics}} 
\label{sec:applications_robotics}

In robotics, there are still few applications with neural attentional models. A small portion of current work is focused on control, visual odometry, navigation, and human-robot interaction, as shown in Table~\ref{tab:robotics_applications}.

\begin{table}[htb]
\small
\centering
\fontfamily{pag}\selectfont
\begin{tabular}{lll}
\hline
\textbf{Task}    &  \textbf{References} \\ \hline
\textbf{Control}   &  ~\cite{one_shot}~\cite{pay_attention} \\ \hline
\textbf{Visual odometry}  & ~\cite{xue2020deep}~\cite{johnston2020self}~\cite{kuo2020dynamic}~\cite{damirchi2020exploring}~\cite{gao2020attentional}~\cite{li2021transformer}     \\ \hline
\textbf{Navigation}  &  ~\cite{sophie}~\cite{social_attention}~\cite{crowd}~\cite{scene_memory} \\ \hline
\textbf{Human-robot interaction}  &   ~\cite{translating_navigation}             \\ \hline
\end{tabular}
\caption{Summary state-of-art main approaches in robotics.}
\label{tab:robotics_applications}
\end{table}

\textbf{Navigation} and visual odometry are the most explored domains, although still with very few published works. For classic DL approaches, navigating tasks in real or complex environments are still very challenging. These approaches have limited performance in dynamic and unstructured environments and over long horizon tasks. In real environments, the robot must deal with dynamic and unexpected changes in humans and other obstacles around it. Also, decision-making depends on the information received in the past and the ability to infer the future state of the environment. Some seminal approaches in the literature have demonstrated the potential of attention to minimizing these problems without compromising the techniques' computational cost. Sadeghian et al.~\cite{sophie} proposed Sophie: an interpretable framework based on GANs for robotic agents in environments with human crowds. Their framework via attention extracts two levels of information: a physical extractor learns spatial and physical constraints generating a $C$ context vector that focuses on viable paths for each agent. In contrast, a social extractor learns the interactions between agents and their influence on each agent's future path. Finally, LSTMs based on GAN generate realistic samples capturing the nature of future paths, and the attentional mechanisms allow the framework to predict physically and socially feasible paths for agents, achieving cutting-edge performances on several different trajectories.

\begin{figure}[htb]
  \centering
  \includegraphics[width=\linewidth]{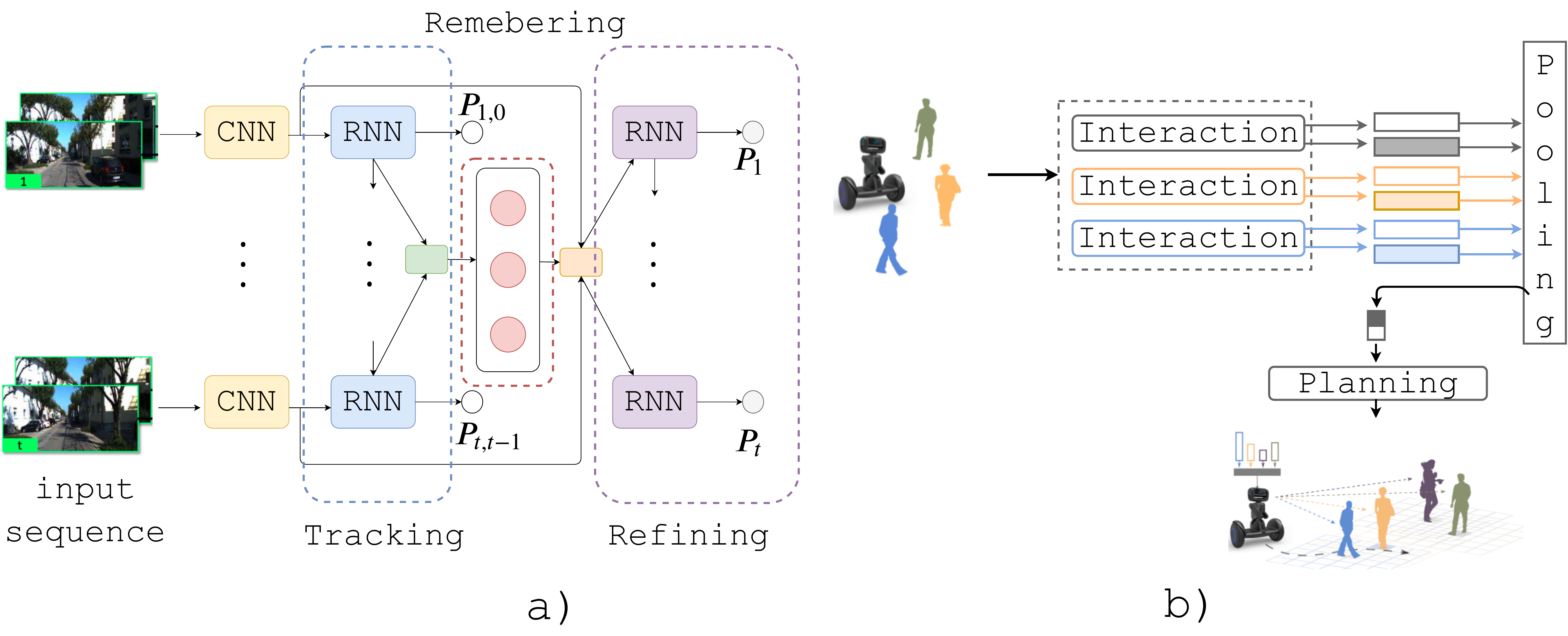}  
  \caption{Illustration of deep visual odometry with Adaptive Memory~\cite{xue2020deep} and Crowd-Robot Interaction~\cite{crowd} for navigation. a) The deep visual odometry with adaptive memory framework. This framework introduces two important components called \textit{remembering} and \textit{refining}, as opposed to classic frameworks that treat VO as just a tracking task. \textit{Remembering} preserves long-time information by adopting an adaptive context selection, and \textit{refining} improves previous outputs using spatial-temporal feature reorganization mechanism. b) Crowd-Robot Interaction network. The network consists of three modules: interaction, pooling, and planning. Interaction extracts robot-human interactions in crowds. Pooling aggregates all interaction information, and planning estimates the state's value based on robots and humans for navigation in crowds.}
  \label{fig:robotic_apps}
\end{figure}

Vemula et al.~\cite{social_attention} proposed a trajectory prediction model that captures each person's relative importance when navigating in the crowd, regardless of their proximity via spatio-temporal graphs. Chen et al.~\cite{crowd} proposed the crowd-aware robot navigation with attention-based deep reinforcement learning. Specifically, a self-attention mechanism models interactions between human-robot and human-human pairs, improving the robot's inference capacity of future environment states. It also captures how human-human interactions can indirectly affect decision-making, as shown in figure~\ref{fig:robotic_apps} in a). Fang et. al.~\cite{scene_memory} proposed the novel memory-based policy (i.e., scene memory transformer - SMT) for embodied agents in long-horizon tasks. The SMT policy consists of two modules: 1) scene memory which stores all past observations in an embedded form, and 2) an attention-based policy network that uses the updated scene memory to compute a distribution over actions. The SMT model is based on an encoder-decoder Transformer and showed strong performance as the agent moves in a large environment, and the number of observations grows rapidly.

In \textbf{visual odometry} (VO), the classic learning-based methods consider the VO task a problem of pure tracking through the recovery of camera poses from fragments of the image, leading to the accumulation of errors. Such approaches often disregard crucial global information to alleviate accumulated errors. However, it is challenging to preserve this information in end-to-end systems effectively. Attention represents an alternative that is still little explored in this area to alleviate such disadvantages. Xue et al.~\cite{xue2020deep} proposed an adaptive memory approach to avoid the network's catastrophic forgetfulness. Their framework consists mainly of a memory, a remembering, and a refining module, as shown in figure~~\ref{fig:robotic_apps}b). First, it remembers to select the main hidden states based on camera movement while preserving selected hidden states in the memory slot to build a global map. The memory stores the global information of the entire sequence, allowing refinements on previous results. Finally, the refining module estimates each view's absolute pose, allowing previously refined outputs to pass through recurrent units, thus improving the next estimate.

Another common problem in VO classical approaches is selecting the features to derive ego-motion between consecutive frames. In scenes, there are dynamic objects and non-textured surfaces that generate inconsistencies in the estimation of movement. Recently, self-attention mechanisms have been successfully employed in dynamic reweighting of features, and in the semantic selection of image regions to extract more refined egomotion~\cite{kuo2020dynamic}~\cite{damirchi2020exploring}~\cite{gao2020attentional}. Additionally, self-attentive neural networks have been used to replace traditional recurrent networks that consume training time and are inaccurate in the temporal integration of long sequences~\cite{li2021transformer}.

In \textbf{human-robot interaction}, Zang et al.~\cite{translating_navigation} proposed a framework that interprets navigation instructions in natural language and finds a mapping of commands in an executable navigation plan. The attentional mechanisms correlate navigation instructions very efficiently with the commands to be executed by the robot in only one trainable end-to-end model, unlike the classic approaches that use decoupled training and external interference during the system's operation. In \textbf{control}, existing applications mainly use manipulator robots in visual-motor tasks. Duan et al.~\cite{one_shot} used attention to improve the model's generalization capacity in imitation learning approaches with a complex manipulator arm. The objective is to build a one-shot learning system capable of successfully performing instances of tasks not seen in the training. Thus, it employs soft attention mechanisms to process a long sequence of (states, actions) demonstration pairs. Finally, Abolghasemi et al.~\cite{pay_attention} proposed a deep visual engine policy through task-focused visual attention to make the policy more robust and to prevent the robot from releasing manipulated objects even under physical attacks.

\subsection{\textbf{Interpretability}}
\label{sec:applications_interpretability}

A long-standing criticism of neural network models is
their lack of interpretability~\cite{li2016understanding}. Academia and industry have a great interest in the development of interpretable models mainly for the following aspects: 1) \textbf{critical decisions}: when critical decisions need to be made (e.i., medical analysis, stock market, autonomous cars), it is essential to provide explanations to increase the confidence of the specialist human results; 2) \textbf{failure analysis}: an interpretable model can retrospectively inspect where bad decisions were made and understand how to improve the system; 3) \textbf{verification}: there is no evidence of the models' robustness and convergence even with small errors in the test set. It is difficult to explain the influence of spurious correlations on performance and why the models are sometimes excellent in some test cases and flawed in others; and 4) \textbf{model improvements}: interpretability can guide improvements in the model's structure if the results are not acceptable.

Attention as an interpretability tool is still an open discussion. For some researchers, it allows to inspect the models' internal dynamics -- the hypothesis is that the attentional weights' magnitude is correlated with the data's relevance for predicting the output. Li et al.~\cite{li2016understanding} proposed a general methodology to analyze the effect of erasing particular representations of neural networks' input. When analyzing the effects of erasure, they found that attentional focuses are essential to understand networks' internal functioning. In~\cite{serrano2019attention} the results showed that higher attentional weights generally contribute with more significant impact to the model's decision, but multiple weights generally do not fully identify the most relevant representations for the final decision. In this investigation, the researchers concluded that attention is an ideal tool to identify which elements are responsible for the output but do not yet fully explain the model's decisions. 

Some studies have also shown that attention encodes linguistic notions relevant to understanding NLP models~\cite{vig2019analyzing}~\cite{tenney2019bert}~\cite{clark2019does}. However, Jain et al.~\cite{attention_is_not_explanation} showed that although attention improves NLP results, its ability to provide transparency or significant explanations for the model's predictions is questionable. Specifically, the researchers investigated the relationship between attentional weights and model results by answering the following questions: (i) to what extent do weights of attention correlate with metrics of the importance of features, specifically those resulting from gradient? Moreover, (ii) do different attentional maps produce different predictions? The results showed that the correlation between intuitive metrics about the features' importance (e.i., gradient approaches, erasure of features) and attentional weights is low in recurrent encoders. Besides, the selection of features other than the attentional distribution did not significantly impact the output as attentional weights exchanged at random also induced minimal output changes. The researchers also concluded that such results depend significantly on the type of architecture, given that feedforward encoders obtained more coherent relationships between attentional weights and output than other models.

Vashishth et al.~\cite{vashishth2019attention} systematically investigated explanations for the researchers' distinct views through experiments on NLP tasks with single sequence models, pair sequence models, and self-attentive neural networks. The experiments showed that attentional weights in single sequences tasks work like gates and do not reflect the reasoning behind the model's prediction, justifying the observations made by Jain et al.~\cite{attention_is_not_explanation}. However, for pair sequence tasks, attentional weights were essential to explaining the model's reasoning. Manual tests have also shown that attentional weights are highly similar to the manual assessment of human observers' attention. Recently, Wiegreffe et al.~\cite{wiegreffe2019attention} also investigated these issues in depth through an extensive protocol of experiments. The authors observed that attention as an explanation depends on the definition of explainability considered. If the focus is on plausible explainability, the authors concluded that attention could help interpret model insights. However, if the focus is a faithful and accurate interpretation of the link that the model establishes between inputs and outputs, results are not always positive. These authors confirmed that good alternatives distributions could be found in LSTMs and classification tasks, as hypothesized by Jain et al.~\cite{attention_is_not_explanation}. However, in some experiments, adversarial training's alternative distributions had poor performances concerning attention's traditional mechanisms. These results indicate that the attention mechanisms trained mainly in RNNs learn something significant about the relationship between tokens and prediction, which cannot be easily hacked. In the end, they showed that attention efficiency as an explanation depends on the data set and the model's properties.

\section{Trends and Opportunities}
\label{sec:trends}


Attention has been one of the most influential ideas in the Deep Learning community in recent years, with several profound advances, mainly in computer vision and natural language processing. However, there is much space to grow, and many contributions are still to appear. In this section, we highlight some gaps and opportunities in this scenario.

\subsection{End-To-End Attention models} 

Over the past eight years, most of the papers published in the literature have involved attentional mechanisms. Models that are state of the art in DL use attention. Specifically, we note that end-to-end attention networks, such as Transformers~\cite{vaswani_attention_2017} and Graph Attention Networks~\cite{velickovic_graph_2018}, have been expanding significantly and have been used successfully in tasks across multiple domains (Section~\ref{sub:attention_overview}). In particular, Transformer has introduced a new form of computing in which the neural network's core is fully attentional. Transformer-based language models like BERT~\cite{devlin2018bert}, GPT2~\cite{radford2019language}, and GPT3~\cite{brown2020language} are the most advanced language models in NLP. Image GPT~\cite{chen2020generative} has recently revolutionized the results of unsupervised learning in imaging. It is already a trend to propose Transfomer based models with sparse attentional mechanisms to reduce the Transformer's complexity from quadratic to linear and use attentional mechanisms to deal with multimodality in GATs. However, Transformer is still an autoregressive architecture in the decoder and does not use other cognitive mechanisms such as memory. As research in attention and DL is still at early stages, there is still plenty of space in the literature for new attentional mechanisms, and we believe that end-to-end attention architectures might be very influential in Deep Learning's future models.


\subsection{Learning Multimodality} 

Attention has played a crucial role in the growth of learning from multimodal data. Multimodality is extremely important for learning complex tasks. Human beings use different sensory signals all the time to interpret situations and decide which action to take. For example, while recognizing emotions, humans use visual data, gestures, and voice tones to analyze feelings. Attention allowed models to learn the synergistic relationship between the different sensory data, even if they are not synchronized, allowing the development of increasingly complex applications mainly in emotion recognition,~\cite{zhang2019deep}, feelings~\cite{tan2019multimodal}, and language-based image generation~\cite{unpublished2021dalle}. We note that multimodal applications are continually growing in recent years. However, most research efforts are still focused on relating a pair of sensory data, mostly visual and textual data. Architectures that can scale easily to handle more than one pair of sensors are not yet widely explored. Multimodal learning exploring voice data, RGBD images, images from monocular cameras, data from various sensors, such as accelerometers, gyroscopes, GPS, RADAR, biomedical sensors, are still scarce in the literature.

\subsection{Cognitive Elements} 

Attention proposed a new way of thinking about the architecture of neural networks. For many years, the scientific community neglected using other cognitive elements in neural network architectures, such as memory and logic flow control. Attention has made possible including in neural networks other elements that are widely important in human cognition. Memory Networks~\cite{weston_2014_memory}, and Neural Turing Machine~\cite{graves_neural_2014} are essential approaches in which attention makes updates and recoveries in external memory. However, research on this topic is at an early stage. The Neural Turing Machine has not yet been explored in several application domains, being used only in simple datasets for algorithmic tasks, with a slow and unstable convergence. We believe that there is plenty of room to explore the advantages of NTM in a wide range of problems and develop more stable and efficient models. Still, Memory Networks~\cite{weston_2014_memory} presents some developments (Section~\ref{sub:attention_overview}), but few studies explore the use of attention to managing complex and hierarchical structures of memory. Attention to managing different memory types simultaneously (i.e., working memory, declarative, non-declarative, semantic, and long and short term) is still absent in the literature. To the best of our knowledge, the most significant advances have been made in Dynamic Memory Networks~\cite{kumar_ask_2015} with the use of episodic memory. Another open challenge is how to use attention to plug external knowledge into memory and make training faster. Finally, undoubtedly one of the biggest challenges still lies in including other human cognition elements such as imagination, reasoning, creativity, and consciousness working in harmony with attentional structures.

\subsection{Computer Vision} 

Recurrent Attention Models (RAM)~\cite{mnih_recurrent_2014} introduced a new form of image computing using glimpses and hard attention. The architecture is simple, scalable, and flexible. Spatial Transformer (STN)~\cite{jaderberg_spatial_2015} presented a simple module for learning image transformations that can be easily plugged into different architectures. We note that RAM has a high potential for many tasks in which convolutional neural networks have difficulties, such as large, high-resolution images. However, currently, RAM has been explored with simple datasets. We believe that it is interesting to validate RAM in complex classification and regression tasks. Another proposal is to add new modules to the architecture, such as memory, multimodal glimpses, and scaling. It is interesting to explore STN in conjunction with RAM in classification tasks or use STN to predict transformations between sets of images. RAM aligned with STN can help address robostusnees to spatial transformation, learn the system dynamics in Visual Odometry tasks, enhance multiple-instance learning, addressing multiple view-points.


\subsection{Capsule Neural Network} 

Capsule networks (CapsNets), a new class of deep neural network architectures proposed recently by Hinton et al.~\cite{sabour2017dynamic}, have shown excellent performance in many fields, particularly in image recognition and natural language processing. However, few studies in the literature implement attention in capsule networks. AR CapsNet~\cite{choi2019attention} implements a dynamic routing algorithm where routing between capsules is made through
an attention module. The attention routing is a fast forward-pass while keeping spatial information. DA-CapsNet~\cite{huang2020capsnet} proposes a dual attention mechanism, the first layer is added after the convolution layer, and the second layer is added after the primary caps. SACN~\cite{hoogi2019self} is the first model that incorporates the self-attention mechanism as an integral layer. Recently, Tsai. et al.~\cite{tsai2020capsules} introduced a new attentional routing mechanism in which a daughter capsule is routed to a parent capsule-based between the father's state and the daughter's vote. We particularly believe that attention is essential to improve the relational and hierarchical nature that CapsNets propose. The development of works aiming at the dynamic attentional routing of the capsules and incorporating attentional capsules of self-attention, soft and hard attention can bring significant results to current models.


\subsection{Neural-Symbolic Learning and Reasoning} 

According to LeCun~\cite{lecun2015deep} one of the great challenges of artificial intelligence is to combine the robustness of connectionist systems (i.e., neural networks) with symbolic representation to perform complex reasoning tasks. While symbolic representation is highly recursive and declarative, neural networks encode knowledge implicitly by adjusting weights. For many decades exploring the fusion between connectionist and symbolic systems has been overlooked by the scientific community. Only over the past decade, research with hybrid approaches using the two families of AI methodologies has grown again. Approaches such as statistical relational learning (SRL)~\cite{khosravi2010survey} and neural-symbolic learning~\cite{besold2017neural} were proposed. Recently, attention mechanisms have been integrated into some neural-symbolic models, the development of which is still at an early stage. Memory Networks~\cite{weston_2014_memory} (Section~\ref{sub:attention_overview}) and Neural Turing Machine~\cite{graves_neural_2014} (Section~\ref{sub:attention_overview}) were the first initiatives to include reasoning in deep connectionist models.


In the context of neural logic programming, attention has been exploited to reason about knowledge graphs or memory structures to combine the learning of parameters and structures of logical rules. Neural Logic Programming~\cite{yang2017differentiable} uses attention on a neural controller that learns to select a subset of operations and memory content to execute first-order rules. Logic Attention Networks~\cite{wang2019logic} facilitates inductive KG embedding and uses attention to aggregate information coming from graph neighbors with rules and attention weights. A pGAT~\cite{harsha2020probabilistic} uses attention to knowledge base completion, which involves the prediction of missing relations between entities in a knowledge graph. While producing remarkable advances, recent approaches to reasoning with deep networks do not adequately address the task of symbolic reasoning. Current efforts are only about using attention to ensure efficient memory management. We believe that attention can be better explored to understand which pieces of knowledge are relevant to formulate a hypothesis to provide a correct answer, which are rarely present in current neural systems of reasoning.

\subsection{Incremental Learning} 

Incremental learning is one of the challenges for the DL community in the coming years. Machine learning classifiers are trained to recognize a fixed set of classes. However, it is desirable to have the flexibility to learn additional classes with limited data without re-training in the complete training set. Attention can significantly contribute to advances in the area and has been little explored. Ren et al.~\cite{mengye_ren;renjie_liao;ethan_fetaya;richard_s._zemel_incremental_2019} were the first to introduce seminal work in the area. They use Attention Attractor Networks to regularize the learning of new classes. In each episode, a set of new weights is trained to recognize new classes until they converge. Attention Attractor Networks helps recognize new classes while remembering the classes beforehand without revising the original training set.

\subsection{Credit Assignment Problem (CAP)} 

In Reinforcement Learning (RL), an action that leads to a higher final cumulative reward should have more value. Therefore, more "credit" should be assigned to it than an action that leads to a lower final reward. However, measuring the individual contribution of actions to future rewards is not simple and has been studied by the RL community for years. There are at least three variations of the CAP problem that have been explored. The temporal CAP refers to identifying which actions were useful or useless in obtaining the final feedback. The structural CAP seeks to find the set of sensory situations in which a given sequence of actions will produce the same result. Transfer CAP refers to learning how to generalize a sequence of actions in tasks. Few works in the literature explore attention to the CAP problem. We believe that attention will be fundamental to advance credit assignment research. Recently, Ferret et al.~\cite{ferret2019self} started the first research in the area by proposing a seminal work with attention to learn how to assign credit through a separate supervised problem and transfer credit assignment capabilities to new environments.

\subsection{Attention and Interpretability} 

There are investigations to verify attention as an interpretability tool. Some recent studies suggest that attention can be considered reliable for this purpose. However, other researchers criticize the use of attention weights as an analytical tool. Jain and Wallace~\cite{attention_is_not_explanation} proved that attention is not consistent with other explainability metrics and that it is easy to create distributions similar to those of the trained model but to produce a different result. Their conclusion is that changing attention weights does not significantly affect the model's prediction, contrary to research by Rudin~\cite{rudin2018please} and Riedl~\cite{riedl2019human} (Section~\ref{sec:applications_interpretability}). On the other hand, some studies have found how attention in neural models captures various notions of syntax and co-reference~\cite{vig2019analyzing}~\cite{clark2019does}~\cite{tenney2019bert}. Amid such confusion, Vashishth et al.~\cite{vashishth2019attention} investigated attention more systematically. They attempted to justify the two types of observation (that is, when attention is interpretable and not), employing various experiments on various NLP tasks. The conclusion was that attention weights are interpretable and are correlated with metrics of the importance of features. However, this is only valid for cases where weights are essential for predicting models and cannot simply be reduced to a gating unit. Despite the existing studies, there are numerous research opportunities to develop systematic methodologies to analyze attention as an interpretability tool. The current conclusions are based on experiments with few architectures in a specific set of applications in NLP.


\subsection{Unsupervised Learning} 

In the last decade, unsupervised learning has also been recognized as one of the most critical challenges of machine learning since, in fact, human learning is mainly unsupervised~\cite{lecun2015deep}. Some works have recently successfully explored attention within purely unsupervised models. In GANs, attention has been used to improve the global perception of a model (i.e., the model learns which part of the image gives more attention to the others). SAGAN~\cite{zhang2018self} was one of the pioneering efforts to incorporate self-attention in Convolutional Gans to improve the quality of the images generated. Image Transformer is an end-to-end attention network created to generate high-resolution images that significantly surpassed state-of-the-art in ImageNet in 2018. AttGan~\cite{he2019attgan} uses attention to easily take advantage of multimodality to improve the generation of images. Combining a region of the image with a corresponding part of the word-context vector helps to generate new features with more details in each stage.


Attention has still been little explored to make generative models simpler, scalable, and more stable. Perhaps the only approach in the literature to explore such aspects more deeply is DRAW~\cite{draw}, which presents a sequential and straightforward way to generate images, being possible to refine image patches while more information is captured sequentially. However, the architecture was tested only in simple datasets, leaving open spaces for new developments. There is not much exploration of attention using autoencoders. Using VAEs, Bornschein et al.~\cite{bornschein2017variational} increased the generative models with external memory and used an attentional system to address and retrieve the corresponding memory content.


In Natural Language Processing, attention is explored in unsupervised models mainly to extract aspects of sentiment analysis. It is also used within autoencoders to generate semantic representations of phrases~\cite{zhang2017battrae}\cite{tian2019attention}. However, most studies still use supervised learning attention, and few approaches still focus on computer vision and NLP. Therefore, we believe that there is still a great path for research and exploration of attention in the unsupervised context, particularly we note that the construction of purely bottom-up attentional systems is not explored in the literature and especially in the context of unsupervised learning, these systems can great value, accompanied by inhibition and return mechanisms.



\subsection{New Tasks and Robotics} 

Although attention has been used in several domains, there are still potential applications that can benefit from it. The prediction of time series, medical applications, and robotics applications are little-explored areas of the literature. Predicting time series becomes challenging as the size of the series increases. Attentional neural networks can contribute significantly to improving results. Specifically, we believe that exploring RAM~\cite{mnih_recurrent_2014} with multiple glimpses looking at different parts of the series or different frequency ranges can introduce a new way of computing time series. In medical applications, there are still few works that explore biomedical signals in attentional architectures. There are opportunities to apply attention to all applications, ranging from segmentation and image classification, support for disease diagnosis to support treatments such as Parkinson's, Alzheimer's, and other chronic diseases.

For robotics, there are countless opportunities. For years the robotics community has been striving for robots to perform tasks in a safe manner and with behaviors closer to humans. However, DL techniques need to cope well with multimodality, active learning, incremental learning, identify unknowns, uncertainty estimation, object and scene semantics, reasoning, awareness, and planning for this task. Architectures like RAM~\cite{mnih_recurrent_2014}, DRAW~\cite{draw} and Transformer~\cite{vaswani_attention_2017} can contribute a lot by being applied to visual odometry, SLAM and mapping tasks.

\section{Conclusions}
\label{sec:conclusion}

In this survey, we presented a systematic review of the literature on attention in Deep Learning to overview the area from its main approaches, historical landmarks, uses of attention, applications, and research opportunities. In total, we critically analyzed more than 600 relevant papers published from 2014 to the present. To the best of our knowledge, this is the broadest survey in the literature, given that most of the existing reviews cover only particular domains with a slightly smaller number of reviewed works. Throughout the paper, we have identified and discussed the relationship between attention mechanisms in established deep neural network models, emphasizing CNNs, RNNs, and generative models. We discussed how attention led to performance gains, improvements in computational efficiency, and a better understanding of networks' knowledge. We present an exhaustive list of application domains discussing the main benefits of attention, highlighting each domain's most representative instances. We also showed recent discussions about attention on the explanation and interpretability of models, a branch of research that is widely discussed today. Finally, we present what we consider trends and opportunities for new developments around attentive models. We hope that this survey will help the audience understand the different existing research directions and provide significant scientific community background in generating future research.

It is worth mentioning that our survey results from an extensive and exhaustive process of searching, filtering, and critical analysis of papers published between 01/01/2014 until 15/02/2021 in the central publication repositories for machine learning and related areas. In total, we collected more than 20,000 papers. After successive automatic and manual filtering, we selected approximately 650 papers for critical analysis and more than 6,000 for quantitative analyses, which correspond mainly to identifying the main application domains, places of publication, and main architectures. For automatic filtering, we use keywords from the area and set up different combinations of filters to eliminate noise from psychology and classic computational visual attention techniques (i.e., saliency maps). In manual filtering, we separate the papers by year and define the originality and number of citations of the work as the main selection criteria. In the appendix, we provide our complete methodology and links to our search codes to facilitate improving future revisions on any topic in the area.

We are currently complementing this survey with a theoretical analysis of the main neural attention models. This complementary survey will help to address an urgent need for an attentional framework supported by taxonomies based on theoretical aspects of attention, which predate the era of Deep Learning. The few existing taxonomies in the area do not yet use theoretical concepts and are challenging to extend to various architectures and application domains. Taxonomies inspired by classical concepts are essential to understand how attention has acted in deep neural networks and whether the roles played corroborate with theoretical foundations studied for more than 40 years in psychology and neuroscience. This study is already in the final stages of development by our team and will hopefully help researchers develop new attentional structures with functions still little explored in the literature. We hope to make it available to the scientific community as soon as possible.
\section*{Appendix}
\label{sec:systematic_review}

\begin{figure}[htb]
  \centering
  \includegraphics[width=0.50\linewidth]{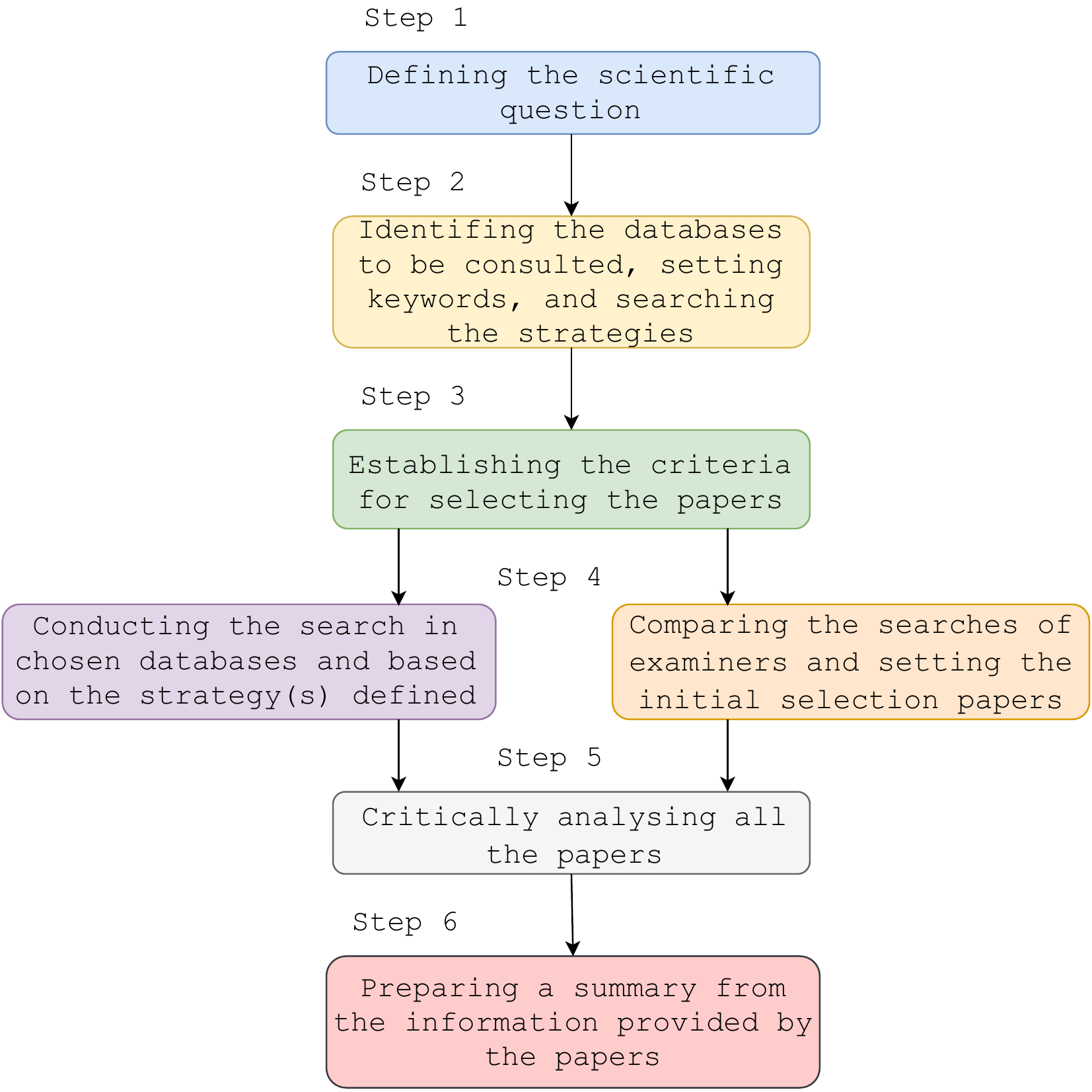}
  \caption{Steps of the systematic review used in this survey.}
  \label{fig:systematic_review_steps}
\end{figure}

This survey employs a systematic review (SR) approach aiming to collect, critically evaluate, and synthesize the results of multiple primary studies concerning Attention in Deep Learning. The selection and evaluation of the works should be meticulous and easily reproducible. Also, SR should be objective, systematic, transparent, and replicable. Although recent, the use of attention in Deep Learning is extensive. Therefore, we systematically reviewed the literature, collecting works from a variety of sources. SR consists of the following steps: defining the scientific questions, identifying the databases, establishing the criteria for selecting papers, searching the databases, performing a critical analysis to choose the most relevant works, and preparing a critical summary of the most relevant papers, as shown Figure~\ref{fig:systematic_review_steps}. 

This survey covers the following aspects: 1) The uses of attention in Deep Learning; 2) Attention mechanisms; 3) Uses of attention; 4) Attention applications; 5) Attention and interpretability; 6) Trends and challenges. These aspects provide the main topics regarding attention in Deep Learning, which can help understand the field's fundamentals. The second step identifies the main databases in the machine learning area, such as arXiv, DeepMind, Google AI, OpenAI, Facebook AI research, Microsoft research, Amazon research, Google Scholar, IEEE Xplore, DBLP, ACM, NIPS, ICML, ICLR, AAAI, CVPR, ICCV, CoRR, IJCNN, Neurocomputing, and Google general search (including blogs, distill, and Quora). Our searching period comprises 01/01/2014 to 06/30/2019 (first stage) and 07/01/2019 to 02/15/2021 (second stage), and the search was performed via a Phyton script~\footnotemark[2]. The papers' title, abstract, year, DOI, and source publication were downloaded and stored in a JSON file. The most appropriate set of keywords to perform the searches was defined by partially searching the field with expert knowledge from our research group. The final set of keywords wore: attention, attentional, and attentive.

\footnotetext[2]{https://github.com/larocs/attention\_dl}

However, these keywords are also relevant in psychology and visual attention. Hence, we performed a second selection to eliminate these papers and remmove duplicate papers unrelated to the DL field. After removing duplicates, 18,257 different papers remained. In the next selection step, we performed a sequential combination of three types of filters: 1) Filter I: Selecting the works with general terms of attention (i.e., attention, attentive, attentional, saliency, top-down, bottom-up, memory, focus, and mechanism); 2) Filter II: Selecting the works with terms related to DL (i.e. deep learning, neural network, ann, dnn deep neural, encoder, decoder, recurrent neural network, recurrent network, rnn, long short term memory, long short-term memory, lstm, gated recurrent unit, gru, autoencoder, ae, variational autoencoder, vae, denoising ae, dae, sparse ae, sae, markov chain, mc, hopfield network, boltzmann machine, em, restricted boltzmann machine, rbm, deep belief network, dbn, deep convolutional network, dcn, deconvolution network, dn, deep convolutional inverse graphics network, dcign, generative adversarial network, gan, liquid state machine, lsm, extreme learnng, machine, elm, echo state network, esn, deep residual network, drn, konohen network, kn, turing machine, ntm, convolutional network, cnn, and capsule network); 3) Filter III: Selecting the works with specific words of attention in Deep Learning (i.e., attention network, soft attention, hard attention, self-attention, self attention deep attention, hierarchical attention, transformer, local attention, global attention, coattention, co-attention, flow attention, attention-over-attention, way attention, intra-attention, self-attentive, and self attentive).

\begin{figure}[H]
  \centering
  \includegraphics[width=0.62\linewidth]{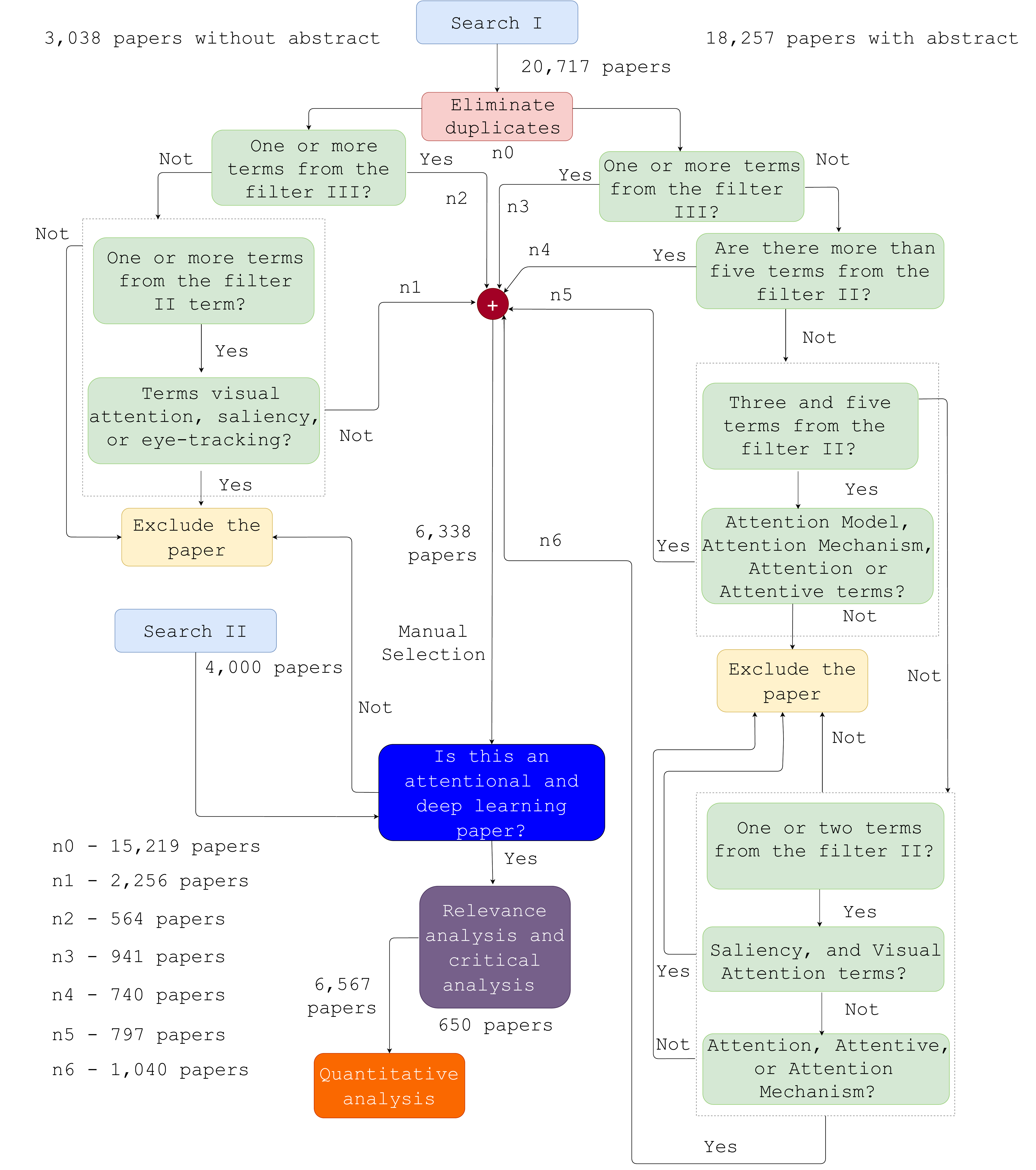}
  \caption{The filter strategies for selecting the relevant works. Search I corresponds to articles collected between 01/01/2014 to 06/30/2019 (first stage), and Search II corresponds to papers collected between 07/01/2019 to 02/15/2021 (second stage).}
  \label{fig:filter_papers_steps}
\end{figure}

The decision tree with the second selection is shown in Figure~\ref{fig:filter_papers_steps}. The third filtering selects works with at least one specific term of attention in deep learning. In the next filtering, we remove papers without abstract, the collection of filters verify if there is at least one specific term of Deep Learning and remove the works with the following keywords: visual attention, saliency, and eye-tracking. For the papers with abstract, the selection is more complex, requiring three cascade conditions: 1) First condition: Selecting the works that have more than five filter terms from filter II; 2) Second condition: selecting the works that have between three and five terms from filter II and where there is at least one of the following: attention model, attention mechanism, attention, or attentive; 3) Third condition: Selecting the works with one or two terms from filter II; without the terms: salience, visual attention, attentive, and attentional mechanism. A total of 6,338 works remained for manual selection. We manually excluded the papers without a title, abstract, or introduction related to the DL field. After manual selection, 3,567 works were stored in \textbf {Zotero}. Given the number of papers, we grouped them by year and chose those above a threshold (average citations in the group). Only works above average were read and classified as relevant or not for critical analysis. To find the number of citations, we automated the process with a Python script. 650 papers were considered relevant for this survey's critical analysis, and 6,567 were used to perform quantitative analyzes.

\bibliographystyle{unsrt}  
\bibliography{references}  

\begin{thebibliography}{100}

\bibitem{colombini2014attentional}
Esther~Luna Colombini, A~da~Silva~Simoes, and CHC Ribeiro.
\newblock {\em An attentional model for intelligent robotics agents.}
\newblock PhD thesis, Instituto Tecnol{\'o}gico de Aeron{\'a}utica, S{\~a}o
  Jos{\'e} dos Campos, Brazil, 2014.

\bibitem{chun2011taxonomy}
Marvin~M Chun, Julie~D Golomb, and Nicholas~B Turk-Browne.
\newblock A taxonomy of external and internal attention.
\newblock {\em Annual review of psychology}, 62:73--101, 2011.

\bibitem{tootell1998retinotopy}
Roger~BH Tootell, Nouchine Hadjikhani, E~Kevin Hall, Sean Marrett, Wim
  Vanduffel, J~Thomas Vaughan, and Anders~M Dale.
\newblock The retinotopy of visual spatial attention.
\newblock {\em Neuron}, 21(6):1409--1422, 1998.

\bibitem{woldorff1993modulation}
Marty~G Woldorff, Christopher~C Gallen, Scott~A Hampson, Steven~A Hillyard,
  Christo Pantev, David Sobel, and Floyd~E Bloom.
\newblock Modulation of early sensory processing in human auditory cortex
  during auditory selective attention.
\newblock {\em Proceedings of the National Academy of Sciences},
  90(18):8722--8726, 1993.

\bibitem{johansen2000physiology}
Heidi Johansen-Berg and Donna~M Lloyd.
\newblock The physiology and psychology of selective attention to touch.
\newblock {\em Front Biosci}, 5:D894--D904, 2000.

\bibitem{zelano2005attentional}
Christina Zelano, Moustafa Bensafi, Jess Porter, Joel Mainland, Brad Johnson,
  Elizabeth Bremner, Christina Telles, Rehan Khan, and Noam Sobel.
\newblock Attentional modulation in human primary olfactory cortex.
\newblock {\em Nature neuroscience}, 8(1):114--120, 2005.

\bibitem{veldhuizen2007trying}
Maria~G Veldhuizen, Genevieve Bender, R~Todd Constable, and Dana~M Small.
\newblock Trying to detect taste in a tasteless solution: modulation of early
  gustatory cortex by attention to taste.
\newblock {\em Chemical Senses}, 32(6):569--581, 2007.

\bibitem{James1890JAMPOP}
William James.
\newblock {\em The Principles of Psychology}.
\newblock Dover Publications, 1890.

\bibitem{FrintropSurvey}
Simone Frintrop, Erich Rome, and Henrik Christensen.
\newblock Computational visual attention systems and their cognitive
  foundations: A survey.
\newblock {\em ACM Transactions on Applied Perception}, 7, 01 2010.

\bibitem{treisman1980feature}
Anne~M Treisman and Garry Gelade.
\newblock A feature-integration theory of attention.
\newblock {\em Cognitive psychology}, 12(1):97--136, 1980.

\bibitem{wolfe1989guided}
Jeremy~M Wolfe, Kyle~R Cave, and Susan~L Franzel.
\newblock Guided search: an alternative to the feature integration model for
  visual search.
\newblock {\em Journal of Experimental Psychology: Human perception and
  performance}, 15(3):419, 1989.

\bibitem{rensink2000dynamic}
Ronald~A Rensink.
\newblock The dynamic representation of scenes.
\newblock {\em Visual cognition}, 7(1-3):17--42, 2000.

\bibitem{broadbent2013perception}
Donald~Eric Broadbent.
\newblock {\em Perception and communication}.
\newblock Elsevier, 2013.

\bibitem{norman1968toward}
Donald~A Norman.
\newblock Toward a theory of memory and attention.
\newblock {\em Psychological review}, 75(6):522, 1968.

\bibitem{kahneman1973attention}
Daniel Kahneman.
\newblock {\em Attention and effort}, volume 1063.
\newblock Citeseer, 1973.

\bibitem{van2004clam}
Frank Van~der Velde, Marc de~Kamps, et~al.
\newblock Clam: Closed-loop attention model for visual search.
\newblock {\em Neurocomputing}, 58:607--612, 2004.

\bibitem{phaf1990slam}
R~Hans Phaf, AHC Van~der Heijden, and Patrick~TW Hudson.
\newblock Slam: A connectionist model for attention in visual selection tasks.
\newblock {\em Cognitive psychology}, 22(3):273--341, 1990.

\bibitem{frintrop2010computational}
Simone Frintrop, Erich Rome, and Henrik~I Christensen.
\newblock Computational visual attention systems and their cognitive
  foundations: A survey.
\newblock {\em ACM Transactions on Applied Perception (TAP)}, 7(1):1--39, 2010.

\bibitem{koch1987shifts}
Christof Koch and Shimon Ullman.
\newblock Shifts in selective visual attention: towards the underlying neural
  circuitry.
\newblock In {\em Matters of intelligence}, pages 115--141. Springer, 1987.

\bibitem{itti1998model}
Laurent Itti, Christof Koch, and Ernst Niebur.
\newblock A model of saliency-based visual attention for rapid scene analysis.
\newblock {\em IEEE PAMI}, 20(11):1254--1259, 1998.

\bibitem{navalpakkam2006integrated}
Vidhya Navalpakkam and Laurent Itti.
\newblock An integrated model of top-down and bottom-up attention for
  optimizing detection speed.
\newblock In {\em 2006 IEEE CVPR)}, volume~2, pages 2049--2056. IEEE, 2006.

\bibitem{hamker2005emergence}
Fred~H Hamker.
\newblock The emergence of attention by population-based inference and its role
  in distributed processing and cognitive control of vision.
\newblock {\em Computer Vision and Image Understanding}, 100(1-2):64--106,
  2005.

\bibitem{hamker2006modeling}
Fred~H Hamker.
\newblock Modeling feature-based attention as an active top-down inference
  process.
\newblock {\em BioSystems}, 86(1-3):91--99, 2006.

\bibitem{frintrop2006vocus}
Simone Frintrop.
\newblock {\em VOCUS: A visual attention system for object detection and
  goal-directed search}, volume 3899.
\newblock Springer, 2006.

\bibitem{salah2002selective}
Albert~Ali Salah, Ethem Alpaydin, and Lale Akarun.
\newblock A selective attention-based method for visual pattern recognition
  with application to handwritten digit recognition and face recognition.
\newblock {\em IEEE PAMI}, 24(3):420--425, 2002.

\bibitem{ouerhani2004visual}
Nabil Ouerhani.
\newblock {\em Visual attention: from bio-inspired modeling to real-time
  implementation}.
\newblock PhD thesis, Universit{\'e} de Neuch{\^a}tel, 2003.

\bibitem{walther2006interactions}
Dirk Walther.
\newblock {\em Interactions of visual attention and object recognition:
  computational modeling, algorithms, and psychophysics}.
\newblock PhD thesis, California Institute of Technology, 2006.

\bibitem{walther2004detection}
Dirk Walther, Duane~R Edgington, and Christof Koch.
\newblock Detection and tracking of objects in underwater video.
\newblock In {\em Proc. of the IEEE CVPR}, volume~1, pages I--I. IEEE, 2004.

\bibitem{clark1988modal}
James~J Clark and Nicola~J Ferrier.
\newblock Modal control of an attentive vision system.
\newblock In {\em IEEE ICCV}, pages 514--523. IEEE, 1988.

\bibitem{breazeal1999context}
Cynthia Breazeal and Brian Scassellati.
\newblock A context-dependent attention system for a social robot.
\newblock {\em rn}, 255:3, 1999.

\bibitem{rotenstein2007towards}
A~Rotenstein, Alexander Andreopoulos, Ehzan Fazl, David Jacob, Matt Robinson,
  Ksenia Shubina, Yuliang Zhu, and J~Tsotsos.
\newblock Towards the dream of intelligent, visually-guided wheelchairs.
\newblock In {\em Proc. 2nd Int’l Conf. on Technology and Aging}, 2007.

\bibitem{clark1992attentive}
James~J Clark and Nicola~J Ferrier.
\newblock Attentive visual servoing.
\newblock In {\em Active vision}. Citeseer, 1992.

\bibitem{frintrop2008attentional}
Simone Frintrop and Patric Jensfelt.
\newblock Attentional landmarks and active gaze control for visual slam.
\newblock {\em IEEE Transactions on Robotics}, 24(5):1054--1065, 2008.

\bibitem{scheier1997visual}
Christian Scheier and Steffen Egner.
\newblock Visual attention in a mobile robot.
\newblock In {\em ISIE'97 Proceeding of the IEEE International Symposium on
  Industrial Electronics}, volume~1, pages SS48--SS52. IEEE, 1997.

\bibitem{baluja1997expectation}
Shumeet Baluja and Dean~A Pomerleau.
\newblock Expectation-based selective attention for visual monitoring and
  control of a robot vehicle.
\newblock {\em Robotics and autonomous systems}, 22(3-4):329--344, 1997.

\bibitem{draw}
Karol Gregor, Ivo Danihelka, Alex Graves, Danilo~Jimenez Rezende, and Daan
  Wierstra.
\newblock Draw: A recurrent neural network for image generation.
\newblock {\em arXiv preprint arXiv:1502.04623}, 2015.

\bibitem{vaswani_attention_2017}
Ashish Vaswani, Noam Shazeer, Niki Parmar, Jakob Uszkoreit, Llion Jones,
  Aidan~N. Gomez, Lukasz Kaiser, and Illia Polosukhin.
\newblock Attention is all you need.
\newblock {\em arXiv:1706.03762 [cs]}, June 2017.
\newblock arXiv: 1706.03762.

\bibitem{weston_2014_memory}
Jason Weston, Sumit Chopra, and Antoine Bordes.
\newblock Memory networks.
\newblock {\em arXiv preprint arXiv:1410.3916}, 2014.

\bibitem{wang2016survey}
Feng Wang and David~MJ Tax.
\newblock Survey on the attention based rnn model and its applications in
  computer vision.
\newblock {\em arXiv preprint arXiv:1601.06823}, 2016.

\bibitem{hu2019introductory}
Dichao Hu.
\newblock An introductory survey on attention mechanisms in nlp problems.
\newblock In {\em Proceedings of SAI Intelligent Systems Conference}, pages
  432--448. Springer, 2019.

\bibitem{galassi2020attention}
Andrea Galassi, Marco Lippi, and Paolo Torroni.
\newblock Attention in natural language processing.
\newblock {\em IEEE Transactions on Neural Networks and Learning Systems},
  2020.

\bibitem{lee_attention_2018}
John~Boaz Lee, Ryan~A. Rossi, Sungchul Kim, Nesreen~K. Ahmed, and Eunyee Koh.
\newblock Attention models in graphs: A survey.
\newblock {\em arXiv:1807.07984 [cs]}, July 2018.
\newblock arXiv: 1807.07984.

\bibitem{chaudhari2019attentive}
Sneha Chaudhari, Gungor Polatkan, Rohan Ramanath, and Varun Mithal.
\newblock An attentive survey of attention models.
\newblock {\em arXiv preprint arXiv:1904.02874}, 2019.

\bibitem{bahdanau_neural_2014}
Dzmitry Bahdanau, Kyunghyun Cho, and Yoshua Bengio.
\newblock Neural machine translation by jointly learning to align and
  translate.
\newblock {\em arXiv preprint arXiv:1409.0473}, 2014.

\bibitem{xu_show_2015}
Kelvin Xu, Jimmy Ba, Ryan Kiros, Kyunghyun Cho, Aaron Courville, Ruslan
  Salakhudinov, Rich Zemel, and Yoshua Bengio.
\newblock Show, attend and tell: Neural image caption generation with visual
  attention.
\newblock In {\em International {Conference} on {Machine} {Learning}}, pages
  2048--2057, June 2015.

\bibitem{mnih_recurrent_2014}
Volodymyr Mnih, Nicolas Heess, Alex Graves, and Koray Kavukcuoglu.
\newblock Recurrent models of visual attention.
\newblock {\em arXiv preprint arXiv:1406.6247}, 2014.

\bibitem{cho2014properties}
Kyunghyun Cho, Bart Van~Merri{\"e}nboer, Dzmitry Bahdanau, and Yoshua Bengio.
\newblock On the properties of neural machine translation: Encoder-decoder
  approaches.
\newblock {\em arXiv preprint arXiv:1409.1259}, 2014.

\bibitem{chan2015listen}
William Chan, Navdeep Jaitly, Quoc~V Le, and Oriol Vinyals.
\newblock Listen, attend and spell.
\newblock {\em arXiv preprint arXiv:1508.01211}, 2015.

\bibitem{vinyals2015grammar}
Oriol Vinyals, {\L}ukasz Kaiser, Terry Koo, Slav Petrov, Ilya Sutskever, and
  Geoffrey Hinton.
\newblock Grammar as a foreign language.
\newblock In {\em Advances in neural information processing systems}, pages
  2773--2781, 2015.

\bibitem{vinyals_neural_nodate}
Oriol Vinyals and Quoc Le.
\newblock A neural conversational model.
\newblock {\em arXiv preprint arXiv:1506.05869}, 2015.

\bibitem{seo_bidirectional_2016}
Minjoon Seo, Aniruddha Kembhavi, Ali Farhadi, and Hannaneh Hajishirzi.
\newblock Bidirectional attention flow for machine comprehension.
\newblock {\em arXiv:1611.01603 [cs]}, November 2016.
\newblock arXiv: 1611.01603.

\bibitem{yang2016hierarchical}
Zichao Yang, Diyi Yang, Chris Dyer, Xiaodong He, Alex Smola, and Eduard Hovy.
\newblock Hierarchical attention networks for document classification.
\newblock In {\em Proceedings of the 2016 conference of the North American
  chapter of the association for computational linguistics: human language
  technologies}, pages 1480--1489, 2016.

\bibitem{xiong2016dynamic}
Caiming Xiong, Victor Zhong, and Richard Socher.
\newblock Dynamic coattention networks for question answering.
\newblock {\em arXiv preprint arXiv:1611.01604}, 2016.

\bibitem{vinyals2015pointer}
Oriol Vinyals, Meire Fortunato, and Navdeep Jaitly.
\newblock Pointer networks.
\newblock In {\em Advances in neural information processing systems}, pages
  2692--2700, 2015.

\bibitem{see_get_2017}
Abigail See, Peter~J Liu, and Christopher~D Manning.
\newblock Get to the point: Summarization with pointer-generator networks.
\newblock {\em arXiv preprint arXiv:1704.04368}, 2017.

\bibitem{huang_fusionnet:_2018}
Hsin-Yuan Huang, Chenguang Zhu, Yelong Shen, and Weizhu Chen.
\newblock Fusionnet: Fusing via fully-aware attention with application to
  machine comprehension.
\newblock {\em arXiv preprint arXiv:1711.07341}, 2017.

\bibitem{rocktaschel_reasoning_2015}
Tim Rockt{\"a}schel, Edward Grefenstette, Karl~Moritz Hermann, Tom{\'a}{\v{s}}
  Ko{\v{c}}isk{\`y}, and Phil Blunsom.
\newblock Reasoning about entailment with neural attention.
\newblock {\em arXiv preprint arXiv:1509.06664}, 2015.

\bibitem{luong_effective_2015}
Thang Luong, Hieu Pham, and Christopher~D. Manning.
\newblock Effective approaches to attention-based neural machine translation.
\newblock In {\em Proceedings of the 2015 {Conference} on {Empirical} {Methods}
  in {Natural} {Language} {Processing}}, pages 1412--1421, Lisbon, Portugal,
  2015. Association for Computational Linguistics.

\bibitem{jaderberg_spatial_2015}
Max Jaderberg, Karen Simonyan, Andrew Zisserman, and koray kavukcuoglu.
\newblock Spatial transformer networks.
\newblock In C.~Cortes, N.~D. Lawrence, D.~D. Lee, M.~Sugiyama, and R.~Garnett,
  editors, {\em Advances in {Neural} {Information} {Processing} {Systems} 28},
  pages 2017--2025. Curran Associates, Inc., 2015.

\bibitem{ramachandram2017deep}
Dhanesh Ramachandram and Graham~W Taylor.
\newblock Deep multimodal learning: A survey on recent advances and trends.
\newblock {\em IEEE Signal Processing Magazine}, 34(6):96--108, 2017.

\bibitem{gao2020survey}
Jing Gao, Peng Li, Zhikui Chen, and Jianing Zhang.
\newblock A survey on deep learning for multimodal data fusion.
\newblock {\em Neural Computation}, 32(5):829--864, 2020.

\bibitem{yao_describing_2015}
Li~Yao, Atousa Torabi, Kyunghyun Cho, Nicolas Ballas, Christopher Pal, Hugo
  Larochelle, and Aaron Courville.
\newblock Describing videos by exploiting temporal structure.
\newblock In {\em Proceedings of the IEEE international conference on computer
  vision}, pages 4507--4515, 2015.

\bibitem{wu_hierarchical_2018}
Chunlei Wu, Yiwei Wei, Xiaoliang Chu, Sun Weichen, Fei Su, and Leiquan Wang.
\newblock Hierarchical attention-based multimodal fusion for video captioning.
\newblock {\em Neurocomputing}, 315:362--370, 2018.

\bibitem{fakoor2016memory}
Rasool Fakoor, Abdel-rahman Mohamed, Margaret Mitchell, Sing~Bing Kang, and
  Pushmeet Kohli.
\newblock Memory-augmented attention modelling for videos.
\newblock {\em arXiv preprint arXiv:1611.02261}, 2016.

\bibitem{tian2018diagnostic}
Jiang Tian, Cong Li, Zhongchao Shi, and Feiyu Xu.
\newblock A diagnostic report generator from ct volumes on liver tumor with
  semi-supervised attention mechanism.
\newblock In {\em International Conference on Medical Image Computing and
  Computer-Assisted Intervention}, pages 702--710. Springer, 2018.

\bibitem{pu_adaptive_2018}
Yunchen Pu, Martin~Renqiang Min, Zhe Gan, and Lawrence Carin.
\newblock Adaptive feature abstraction for translating video to text.
\newblock In {\em Thirty-{Second} {AAAI} {Conference} on {Artificial}
  {Intelligence}}, April 2018.

\bibitem{liu2017global}
Jun Liu, Gang Wang, Ping Hu, Ling-Yu Duan, and Alex~C Kot.
\newblock Global context-aware attention lstm networks for 3d action
  recognition.
\newblock In {\em Proceedings of the IEEE Conference on Computer Vision and
  Pattern Recognition}, pages 1647--1656, 2017.

\bibitem{zhang2018attention}
Liang Zhang, Guangming Zhu, Lin Mei, Peiyi Shen, Syed Afaq~Ali Shah, and
  Mohammed Bennamoun.
\newblock Attention in convolutional lstm for gesture recognition.
\newblock In {\em Advances in Neural Information Processing Systems}, pages
  1953--1962, 2018.

\bibitem{zhang2018adding}
Pengfei Zhang, Jianru Xue, Cuiling Lan, Wenjun Zeng, Zhanning Gao, and Nanning
  Zheng.
\newblock Adding attentiveness to the neurons in recurrent neural networks.
\newblock In {\em Proceedings of the European Conference on Computer Vision
  (ECCV)}, pages 135--151, 2018.

\bibitem{xiangrong_zhang;xin_wang;xu_tang;huiyu_zhou;chen_li_description_2019}
Xiangrong Zhang, Xin Wang, Xu~Tang, Huiyu Zhou, and Chen Li.
\newblock Description generation for remote sensing images using attribute
  attention mechanism.
\newblock {\em Remote Sensing}, 11(6):612, 2019.

\bibitem{bei_fang;ying_li;haokui_zhang;jonathan_cheung-wai_chan_hyperspectral_2019}
Bei Fang, Ying Li, Haokui Zhang, and Jonathan Cheung-Wai Chan.
\newblock Hyperspectral images classification based on dense convolutional
  networks with spectral-wise attention mechanism.
\newblock {\em Remote Sensing}, 11(2):159, 2019.

\bibitem{qi_wang;shaoteng_liu;jocelyn_chanussot;xuelong_li_scene_2019}
Qi~Wang, Shaoteng Liu, Jocelyn Chanussot, and Xuelong Li.
\newblock Scene classification with recurrent attention of vhr remote sensing
  images.
\newblock {\em IEEE Transactions on Geoscience and Remote Sensing},
  57(2):1155--1167, 2018.

\bibitem{xiaoguang_mei;erting_pan;yong_ma;xiaobing_dai;jun_huang;fan_fan;qinglei_du;hong_zheng;jiayi_ma_spectral-spatial_2019}
Xiaoguang Mei, Erting Pan, Yong Ma, Xiaobing Dai, Jun Huang, Fan Fan, Qinglei
  Du, Hong Zheng, and Jiayi Ma.
\newblock Spectral-spatial attention networks for hyperspectral image
  classification.
\newblock {\em Remote Sensing}, 11(8):963, 2019.

\bibitem{hori_attention-based_2017}
Chiori Hori, Takaaki Hori, Teng-Yok Lee, Kazuhiro Sumi, John~R. Hershey, and
  Tim~K. Marks.
\newblock Attention-based multimodal fusion for video description.
\newblock {\em arXiv:1701.03126 [cs]}, January 2017.
\newblock arXiv: 1701.03126.

\bibitem{zhang2019deep}
Yuanyuan Zhang, Zi-Rui Wang, and Jun Du.
\newblock Deep fusion: An attention guided factorized bilinear pooling for
  audio-video emotion recognition.
\newblock In {\em 2019 International Joint Conference on Neural Networks
  (IJCNN)}, pages 1--8. IEEE, 2019.

\bibitem{zadeh2018memory}
Amir Zadeh, Paul~Pu Liang, Navonil Mazumder, Soujanya Poria, Erik Cambria, and
  Louis-Philippe Morency.
\newblock Memory fusion network for multi-view sequential learning.
\newblock In {\em Thirty-Second AAAI Conference on Artificial Intelligence},
  2018.

\bibitem{zadeh2018multi}
Amir Zadeh, Paul~Pu Liang, Soujanya Poria, Prateek Vij, Erik Cambria, and
  Louis-Philippe Morency.
\newblock Multi-attention recurrent network for human communication
  comprehension.
\newblock In {\em Thirty-Second AAAI Conference on Artificial Intelligence},
  2018.

\bibitem{santoro2018relational}
Adam Santoro, Ryan Faulkner, David Raposo, Jack Rae, Mike Chrzanowski,
  Theophane Weber, Daan Wierstra, Oriol Vinyals, Razvan Pascanu, and Timothy
  Lillicrap.
\newblock Relational recurrent neural networks.
\newblock {\em arXiv preprint arXiv:1806.01822}, 2018.

\bibitem{zheng_zhang;lizi_liao;minlie_huang;xiaoyan_zhu;tat-seng_chua_neural_2019}
Zheng Zhang, Lizi Liao, Minlie Huang, Xiaoyan Zhu, and Tat-Seng Chua.
\newblock Neural multimodal belief tracker with adaptive attention for dialogue
  systems.
\newblock In {\em The World Wide Web Conference}, pages 2401--2412, 2019.

\bibitem{nam_dual_2017}
Hyeonseob Nam, Jung-Woo Ha, and Jeonghee Kim.
\newblock Dual attention networks for multimodal reasoning and matching.
\newblock In {\em Proceedings of the IEEE conference on computer vision and
  pattern recognition}, pages 299--307, 2017.

\bibitem{noauthor_bi-directional_nodate}
Feiran Huang, Xiaoming Zhang, Zhonghua Zhao, and Zhoujun Li.
\newblock Bi-directional spatial-semantic attention networks for image-text
  matching.
\newblock {\em IEEE Transactions on Image Processing}, 28(4):2008--2020, 2018.

\bibitem{yehao_li;ting_yao;yingwei_pan;hongyang_chao;tao_mei_pointing_2019}
Yehao Li, Ting Yao, Yingwei Pan, Hongyang Chao, and Tao Mei.
\newblock Pointing novel objects in image captioning.
\newblock In {\em Proceedings of the IEEE/CVF Conference on Computer Vision and
  Pattern Recognition}, pages 12497--12506, 2019.

\bibitem{matthews2012origins}
Danielle Matthews, Tanya Behne, Elena Lieven, and Michael Tomasello.
\newblock Origins of the human pointing gesture: a training study.
\newblock {\em Developmental science}, 15(6):817--829, 2012.

\bibitem{noauthor_improving_nodate}
Xihui Liu, Zihao Wang, Jing Shao, Xiaogang Wang, and Hongsheng Li.
\newblock Improving referring expression grounding with cross-modal
  attention-guided erasing.
\newblock In {\em Proceedings of the IEEE/CVF Conference on Computer Vision and
  Pattern Recognition}, pages 1950--1959, 2019.

\bibitem{pay_attention}
Pooya Abolghasemi, Amir Mazaheri, Mubarak Shah, and Ladislau Boloni.
\newblock Pay attention!-robustifying a deep visuomotor policy through
  task-focused visual attention.
\newblock In {\em Proceedings of the IEEE Conference on Computer Vision and
  Pattern Recognition}, pages 4254--4262, 2019.

\bibitem{graves_neural_2014}
Alex Graves, Greg Wayne, and Ivo Danihelka.
\newblock Neural turing machines.
\newblock {\em arXiv:1410.5401 [cs]}, October 2014.
\newblock arXiv: 1410.5401.

\bibitem{khasahmadi2020memory}
Amir~Hosein Khasahmadi, Kaveh Hassani, Parsa Moradi, Leo Lee, and Quaid Morris.
\newblock Memory-based graph networks.
\newblock {\em arXiv preprint arXiv:2002.09518}, 2020.

\bibitem{xiong_dynamic_2016}
Caiming Xiong, Stephen Merity, and Richard Socher.
\newblock Dynamic memory networks for visual and textual question answering.
\newblock {\em arXiv:1603.01417 [cs]}, March 2016.
\newblock arXiv: 1603.01417.

\bibitem{lu2020video}
Xinkai Lu, Wenguan Wang, Martin Danelljan, Tianfei Zhou, Jianbing Shen, and Luc
  Van~Gool.
\newblock Video object segmentation with episodic graph memory networks.
\newblock {\em arXiv preprint arXiv:2007.07020}, 2020.

\bibitem{abdulnabi2017episodic}
Abrar~H Abdulnabi, Bing Shuai, Stefan Winkler, and Gang Wang.
\newblock Episodic camn: Contextual attention-based memory networks with
  iterative feedback for scene labeling.
\newblock In {\em Proceedings of the IEEE Conference on Computer Vision and
  Pattern Recognition}, pages 5561--5570, 2017.

\bibitem{sukhbaatar2015end}
Sainbayar Sukhbaatar, Jason Weston, Rob Fergus, et~al.
\newblock End-to-end memory networks.
\newblock In {\em Advances in neural information processing systems}, pages
  2440--2448, 2015.

\bibitem{oh2019video}
Seoung~Wug Oh, Joon-Young Lee, Ning Xu, and Seon~Joo Kim.
\newblock Video object segmentation using space-time memory networks.
\newblock In {\em Proceedings of the IEEE International Conference on Computer
  Vision}, pages 9226--9235, 2019.

\bibitem{kumar_ask_2015}
Ankit Kumar, Ozan Irsoy, Peter Ondruska, Mohit Iyyer, James Bradbury, Ishaan
  Gulrajani, Victor Zhong, Romain Paulus, and Richard Socher.
\newblock Ask me anything: Dynamic memory networks for natural language
  processing.
\newblock In {\em International conference on machine learning}, pages
  1378--1387. PMLR, 2016.

\bibitem{wu2020comprehensive}
Zonghan Wu, Shirui Pan, Fengwen Chen, Guodong Long, Chengqi Zhang, and S~Yu
  Philip.
\newblock A comprehensive survey on graph neural networks.
\newblock {\em IEEE Transactions on Neural Networks and Learning Systems},
  2020.

\bibitem{miller2016key}
Alexander Miller, Adam Fisch, Jesse Dodge, Amir-Hossein Karimi, Antoine Bordes,
  and Jason Weston.
\newblock Key-value memory networks for directly reading documents.
\newblock {\em arXiv preprint arXiv:1606.03126}, 2016.

\bibitem{moon2019memory}
Seungwhan Moon, Pararth Shah, Anuj Kumar, and Rajen Subba.
\newblock Memory graph networks for explainable memory-grounded question
  answering.
\newblock In {\em Proceedings of the 23rd Conference on Computational Natural
  Language Learning (CoNLL)}, pages 728--736, 2019.

\bibitem{velickovic_graph_2018}
Petar Veli{\v{c}}kovi{\'c}, Guillem Cucurull, Arantxa Casanova, Adriana Romero,
  Pietro Lio, and Yoshua Bengio.
\newblock Graph attention networks.
\newblock {\em arXiv preprint arXiv:1710.10903}, 2017.

\bibitem{wang2019heterogeneous}
Xiao Wang, Houye Ji, Chuan Shi, Bai Wang, Yanfang Ye, Peng Cui, and Philip~S
  Yu.
\newblock Heterogeneous graph attention network.
\newblock In {\em The World Wide Web Conference}, pages 2022--2032, 2019.

\bibitem{wang2019graph}
Lei Wang, Yuchun Huang, Yaolin Hou, Shenman Zhang, and Jie Shan.
\newblock Graph attention convolution for point cloud semantic segmentation.
\newblock In {\em Proceedings of the IEEE Conference on Computer Vision and
  Pattern Recognition}, pages 10296--10305, 2019.

\bibitem{abu2018watch}
Sami Abu-El-Haija, Bryan Perozzi, Rami Al-Rfou, and Alexander~A Alemi.
\newblock Watch your step: Learning node embeddings via graph attention.
\newblock In {\em Advances in Neural Information Processing Systems}, pages
  9180--9190, 2018.

\bibitem{li2019relation}
Linjie Li, Zhe Gan, Yu~Cheng, and Jingjing Liu.
\newblock Relation-aware graph attention network for visual question answering.
\newblock In {\em Proceedings of the IEEE International Conference on Computer
  Vision}, pages 10313--10322, 2019.

\bibitem{weighted_transformer}
Karim Ahmed, Nitish~Shirish Keskar, and Richard Socher.
\newblock Weighted transformer network for machine translation.
\newblock {\em arXiv preprint arXiv:1711.02132}, 2017.

\bibitem{qipeng_guo;xipeng_qiu;pengfei_liu;yunfan_shao;xiangyang_xue;zheng_zhang_star-transformer_2019}
Qipeng Guo, Xipeng Qiu, Pengfei Liu, Yunfan Shao, Xiangyang Xue, and Zheng
  Zhang.
\newblock Star-transformer.
\newblock {\em arXiv preprint arXiv:1902.09113}, 2019.

\bibitem{music_transformer}
Cheng-Zhi~Anna Huang, Ashish Vaswani, Jakob Uszkoreit, Noam Shazeer, Ian Simon,
  Curtis Hawthorne, Andrew~M Dai, Matthew~D Hoffman, Monica Dinculescu, and
  Douglas Eck.
\newblock Music transformer.
\newblock {\em arXiv preprint arXiv:1809.04281}, 2018.

\bibitem{rewon_child;scott_gray;alec_radford;ilya_sutskever_generating_2019}
Rewon Child, Scott Gray, Alec Radford, and Ilya Sutskever.
\newblock Generating long sequences with sparse transformers.
\newblock {\em arXiv preprint arXiv:1904.10509}, 2019.

\bibitem{lee2018set}
Juho Lee, Yoonho Lee, Jungtaek Kim, Adam~R Kosiorek, Seungjin Choi, and
  Yee~Whye Teh.
\newblock Set transformer: A framework for attention-based
  permutation-invariant neural networks.
\newblock {\em arXiv preprint arXiv:1810.00825}, 2018.

\bibitem{doubly_attentive_transformer}
Hasan~Sait Arslan, Mark Fishel, and Gholamreza Anbarjafari.
\newblock Doubly attentive transformer machine translation.
\newblock {\em arXiv preprint arXiv:1807.11605}, 2018.

\bibitem{multi_source_transformer}
Jind{\v{r}}ich Libovick{\`y}, Jind{\v{r}}ich Helcl, and David Mare{\v{c}}ek.
\newblock Input combination strategies for multi-source transformer decoder.
\newblock {\em arXiv preprint arXiv:1811.04716}, 2018.

\bibitem{style_transformer}
Ning Dai, Jianze Liang, Xipeng Qiu, and Xuanjing Huang.
\newblock Style transformer: Unpaired text style transfer without disentangled
  latent representation.
\newblock {\em arXiv preprint arXiv:1905.05621}, 2019.

\bibitem{hierarchical_transformer}
Yang Liu and Mirella Lapata.
\newblock Hierarchical transformers for multi-document summarization.
\newblock {\em arXiv preprint arXiv:1905.13164}, 2019.

\bibitem{highway_transformer}
Ting-Rui Chiang, Chao-Wei Huang, Shang-Yu Su, and Yun-Nung Chen.
\newblock Learning multi-level information for dialogue response selection by
  highway recurrent transformer.
\newblock {\em arXiv preprint arXiv:1903.08953}, 2019.

\bibitem{lattice_transformer}
Fengshun Xiao, Jiangtong Li, Hai Zhao, Rui Wang, and Kehai Chen.
\newblock Lattice-based transformer encoder for neural machine translation.
\newblock {\em arXiv preprint arXiv:1906.01282}, 2019.

\bibitem{li2019neural}
Naihan Li, Shujie Liu, Yanqing Liu, Sheng Zhao, and Ming Liu.
\newblock Neural speech synthesis with transformer network.
\newblock In {\em Proceedings of the AAAI Conference on Artificial
  Intelligence}, pages 6706--6713, 2019.

\bibitem{phrase_attention}
Phi~Xuan Nguyen and Shafiq Joty.
\newblock Phrase-based attentions.
\newblock {\em arXiv preprint arXiv:1810.03444}, 2018.

\bibitem{devlin_bert:_2018}
Jacob Devlin, Ming-Wei Chang, Kenton Lee, and Kristina Toutanova.
\newblock Bert: Pre-training of deep bidirectional transformers for language
  understanding.
\newblock {\em arXiv:1810.04805 [cs]}, October 2018.
\newblock arXiv: 1810.04805.

\bibitem{radford2019language}
Alec Radford, Jeffrey Wu, Rewon Child, David Luan, Dario Amodei, and Ilya
  Sutskever.
\newblock Language models are unsupervised multitask learners.
\newblock {\em OpenAI blog}, 1(8):9, 2019.

\bibitem{brown2020language}
Tom~B Brown, Benjamin Mann, Nick Ryder, Melanie Subbiah, Jared Kaplan, Prafulla
  Dhariwal, Arvind Neelakantan, Pranav Shyam, Girish Sastry, Amanda Askell,
  et~al.
\newblock Language models are few-shot learners.
\newblock {\em arXiv preprint arXiv:2005.14165}, 2020.

\bibitem{parmar2018image}
Niki Parmar, Ashish Vaswani, Jakob Uszkoreit, Lukasz Kaiser, Noam Shazeer,
  Alexander Ku, and Dustin Tran.
\newblock Image transformer.
\newblock In {\em International Conference on Machine Learning}, pages
  4055--4064. PMLR, 2018.

\bibitem{zhang2018self}
Han Zhang, Ian Goodfellow, Dimitris Metaxas, and Augustus Odena.
\newblock Self-attention generative adversarial networks.
\newblock In {\em International conference on machine learning}, pages
  7354--7363. PMLR, 2019.

\bibitem{chen2020generative}
Mark Chen, Alec Radford, Rewon Child, Jeff Wu, Heewoo Jun, Prafulla Dhariwal,
  David Luan, and Ilya Sutskever.
\newblock Generative pretraining from pixels.
\newblock In {\em Proceedings of the 37th International Conference on Machine
  Learning}, volume~1, 2020.

\bibitem{unpublished2021dalle}
Aditya Ramesh, Mikhail Pavlov, Gabriel Goh, and Scott Gray.
\newblock Dall·e: Creating images from text, 2021.

\bibitem{gulcehre_hyperbolic_2018}
Caglar Gulcehre, Misha Denil, Mateusz Malinowski, Ali Razavi, Razvan Pascanu,
  Karl~Moritz Hermann, Peter Battaglia, Victor Bapst, David Raposo, Adam
  Santoro, et~al.
\newblock Hyperbolic attention networks.
\newblock {\em arXiv preprint arXiv:1805.09786}, 2018.

\bibitem{zhang2019hyperbolic}
Yiding Zhang, Xiao Wang, Xunqiang Jiang, Chuan Shi, and Yanfang Ye.
\newblock Hyperbolic graph attention network.
\newblock {\em arXiv preprint arXiv:1912.03046}, 2019.

\bibitem{rossi2020temporal}
Emanuele Rossi, Ben Chamberlain, Fabrizio Frasca, Davide Eynard, Federico
  Monti, and Michael Bronstein.
\newblock Temporal graph networks for deep learning on dynamic graphs.
\newblock {\em arXiv preprint arXiv:2006.10637}, 2020.

\bibitem{mittal2020learning}
Sarthak Mittal, Alex Lamb, Anirudh Goyal, Vikram Voleti, Murray Shanahan,
  Guillaume Lajoie, Michael Mozer, and Yoshua Bengio.
\newblock Learning to combine top-down and bottom-up signals in recurrent
  neural networks with attention over modules.
\newblock In {\em International Conference on Machine Learning}, pages
  6972--6986. PMLR, 2020.

\bibitem{graves_adaptive_2016}
Alex Graves.
\newblock Adaptive computation time for recurrent neural networks.
\newblock {\em arXiv:1603.08983 [cs]}, March 2016.
\newblock arXiv: 1603.08983.

\bibitem{figurnov2017spatially}
Michael Figurnov, Maxwell~D Collins, Yukun Zhu, Li~Zhang, Jonathan Huang,
  Dmitry Vetrov, and Ruslan Salakhutdinov.
\newblock Spatially adaptive computation time for residual networks.
\newblock In {\em Proceedings of the IEEE Conference on Computer Vision and
  Pattern Recognition}, pages 1039--1048, 2017.

\bibitem{eyzaguirre2020differentiable}
Cristobal Eyzaguirre and Alvaro Soto.
\newblock Differentiable adaptive computation time for visual reasoning.
\newblock In {\em Proceedings of the IEEE/CVF Conference on Computer Vision and
  Pattern Recognition}, pages 12817--12825, 2020.

\bibitem{zhao_deep_2017}
Dongbin Zhao, Yaran Chen, and Le~Lv.
\newblock Deep reinforcement learning with visual attention for vehicle
  classification.
\newblock {\em IEEE Transactions on Cognitive and Developmental Systems},
  9(4):356--367, 2016.

\bibitem{x._wang;_l._gao;_j._song;_h._shen_beyond_2017}
Xuanhan Wang, Lianli Gao, Jingkuan Song, and Hengtao Shen.
\newblock Beyond frame-level cnn: saliency-aware 3-d cnn with lstm for video
  action recognition.
\newblock {\em IEEE Signal Processing Letters}, 24(4):510--514, 2016.

\bibitem{ning_liu;yongchao_long;changqing_zou;qun_niu;li_pan;hefeng_wu_adcrowdnet:_2019}
Ning Liu, Yongchao Long, Changqing Zou, Qun Niu, Li~Pan, and Hefeng Wu.
\newblock Adcrowdnet: An attention-injective deformable convolutional network
  for crowd understanding.
\newblock In {\em Proceedings of the IEEE/CVF Conference on Computer Vision and
  Pattern Recognition}, pages 3225--3234, 2019.

\bibitem{hu_squeeze-and-excitation_2017}
Jie Hu, Li~Shen, Samuel Albanie, Gang Sun, and Enhua Wu.
\newblock Squeeze-and-excitation networks.
\newblock {\em arXiv:1709.01507 [cs]}, September 2017.
\newblock arXiv: 1709.01507.

\bibitem{zhang2019residual}
Yulun Zhang, Kunpeng Li, Kai Li, Bineng Zhong, and Yun Fu.
\newblock Residual non-local attention networks for image restoration.
\newblock {\em arXiv preprint arXiv:1903.10082}, 2019.

\bibitem{sanghyun_woo;jongchan_park;joon-young_lee;in_so_kweon_cbam:_2018}
Sanghyun Woo, Jongchan Park, Joon-Young Lee, and In~So Kweon.
\newblock Cbam: Convolutional block attention module.
\newblock In {\em Proceedings of the European conference on computer vision
  (ECCV)}, pages 3--19, 2018.

\bibitem{chen_^2-nets:_2018}
Yunpeng Chen, Yannis Kalantidis, Jianshu Li, Shuicheng Yan, and Jiashi Feng.
\newblock A\textasciicircum2-nets: Double attention networks.
\newblock In S.~Bengio, H.~Wallach, H.~Larochelle, K.~Grauman, N.~Cesa-Bianchi,
  and R.~Garnett, editors, {\em Advances in {Neural} {Information} {Processing}
  {Systems} 31}, pages 352--361. Curran Associates, Inc., 2018.

\bibitem{fukui2019attention}
Hiroshi Fukui, Tsubasa Hirakawa, Takayoshi Yamashita, and Hironobu Fujiyoshi.
\newblock Attention branch network: Learning of attention mechanism for visual
  explanation.
\newblock In {\em Proceedings of the IEEE/CVF Conference on Computer Vision and
  Pattern Recognition}, pages 10705--10714, 2019.

\bibitem{han20193dviewgraph}
Zhizhong Han, Xiyang Wang, Chi-Man Vong, Yu-Shen Liu, Matthias Zwicker, and
  CL~Chen.
\newblock 3dviewgraph: Learning global features for 3d shapes from a graph of
  unordered views with attention.
\newblock {\em arXiv preprint arXiv:1905.07503}, 2019.

\bibitem{ji2017distant}
Guoliang Ji, Kang Liu, Shizhu He, and Jun Zhao.
\newblock Distant supervision for relation extraction with sentence-level
  attention and entity descriptions.
\newblock In {\em Proceedings of the AAAI Conference on Artificial
  Intelligence}, 2017.

\bibitem{yang2017neural}
Jiaolong Yang, Peiran Ren, Dongqing Zhang, Dong Chen, Fang Wen, Hongdong Li,
  and Gang Hua.
\newblock Neural aggregation network for video face recognition.
\newblock In {\em Proceedings of the IEEE conference on computer vision and
  pattern recognition}, pages 4362--4371, 2017.

\bibitem{hackel_inference_2018}
Timo Hackel, Mikhail Usvyatsov, Silvano Galliani, Jan~D. Wegner, and Konrad
  Schindler.
\newblock Inference, learning and attention mechanisms that exploit and
  preserve sparsity in convolutional networks.
\newblock {\em arXiv:1801.10585 [cs]}, January 2018.
\newblock arXiv: 1801.10585.

\bibitem{mishra_simple_2017}
Nikhil Mishra, Mostafa Rohaninejad, Xi~Chen, and Pieter Abbeel.
\newblock A simple neural attentive meta-learner.
\newblock {\em arXiv:1707.03141 [cs, stat]}, July 2017.
\newblock arXiv: 1707.03141.

\bibitem{fu_look_2017}
Jianlong Fu, Heliang Zheng, and Tao Mei.
\newblock Look closer to see better: Recurrent attention convolutional neural
  network for fine-grained image recognition.
\newblock In {\em 2017 {IEEE} {Conference} on {Computer} {Vision} and {Pattern}
  {Recognition} ({CVPR})}, pages 4476--4484, Honolulu, HI, July 2017. IEEE.

\bibitem{xueying_chen;rong_zhang;pingkun_yan_feature_2019}
Xueying Chen, Rong Zhang, and Pingkun Yan.
\newblock Feature fusion encoder decoder network for automatic liver lesion
  segmentation.
\newblock In {\em 2019 IEEE 16th international symposium on biomedical imaging
  (ISBI 2019)}, pages 430--433. IEEE, 2019.

\bibitem{tian2018learning}
Wanxin Tian, Zixuan Wang, Haifeng Shen, Weihong Deng, Yiping Meng, Binghui
  Chen, Xiubao Zhang, Yuan Zhao, and Xiehe Huang.
\newblock Learning better features for face detection with feature fusion and
  segmentation supervision.
\newblock {\em arXiv preprint arXiv:1811.08557}, 2018.

\bibitem{xingjian_li;haoyi_xiong;hanchao_wang;yuxuan_rao;liping_liu;jun_huan_delta:_2019}
Xingjian Li, Haoyi Xiong, Hanchao Wang, Yuxuan Rao, Liping Liu, Zeyu Chen, and
  Jun Huan.
\newblock Delta: Deep learning transfer using feature map with attention for
  convolutional networks.
\newblock {\em arXiv preprint arXiv:1901.09229}, 2019.

\bibitem{zheng2017learning}
Heliang Zheng, Jianlong Fu, Tao Mei, and Jiebo Luo.
\newblock Learning multi-attention convolutional neural network for
  fine-grained image recognition.
\newblock In {\em Proceedings of the IEEE international conference on computer
  vision}, pages 5209--5217, 2017.

\bibitem{rodriguez_painless_2018}
Pau Rodr{\'\i}guez, Guillem Cucurull, Jordi Gonz{\`a}lez, Josep~M Gonfaus, and
  Xavier Roca.
\newblock A painless attention mechanism for convolutional neural networks.
\newblock {\em ICLR 2018}, 2018.

\bibitem{linlin_wang_zhu_cao_gerard_de_melo_zhiyuan_liu:_relation_nodate}
Linlin Wang, Zhu Cao, Gerard De~Melo, and Zhiyuan Liu.
\newblock Relation classification via multi-level attention cnns.
\newblock In {\em Proceedings of the 54th Annual Meeting of the Association for
  Computational Linguistics (Volume 1: Long Papers)}, pages 1298--1307, 2016.

\bibitem{yin2016abcnn}
Wenpeng Yin, Hinrich Sch{\"u}tze, Bing Xiang, and Bowen Zhou.
\newblock Abcnn: Attention-based convolutional neural network for modeling
  sentence pairs.
\newblock {\em Transactions of the Association for Computational Linguistics},
  4:259--272, 2016.

\bibitem{marcus_edel;joscha_lausch_capacity_2016}
Marcus Edel and Joscha Lausch.
\newblock Capacity visual attention networks.
\newblock In {\em GCAI}, pages 72--80, 2016.

\bibitem{qin2017dual}
Yao Qin, Dongjin Song, Haifeng Chen, Wei Cheng, Guofei Jiang, and Garrison
  Cottrell.
\newblock A dual-stage attention-based recurrent neural network for time series
  prediction.
\newblock {\em arXiv preprint arXiv:1704.02971}, 2017.

\bibitem{geoman_2018}
Yuxuan Liang, Songyu Ke, Junbo Zhang, Xiuwen Yi, and Yu~Zheng.
\newblock Geoman: Multi-level attention networks for geo-sensory time series
  prediction.
\newblock In {\em IJCAI}, pages 3428--3434, 2018.

\bibitem{du2017rpan}
Wenbin Du, Yali Wang, and Yu~Qiao.
\newblock Rpan: An end-to-end recurrent pose-attention network for action
  recognition in videos.
\newblock In {\em Proceedings of the IEEE International Conference on Computer
  Vision}, pages 3725--3734, 2017.

\bibitem{xu2018graph2seq}
Kun Xu, Lingfei Wu, Zhiguo Wang, Yansong Feng, Michael Witbrock, and Vadim
  Sheinin.
\newblock Graph2seq: Graph to sequence learning with attention-based neural
  networks.
\newblock {\em arXiv preprint arXiv:1804.00823}, 2018.

\bibitem{ba_multiple_2014}
Jimmy Ba, Volodymyr Mnih, and Koray Kavukcuoglu.
\newblock Multiple object recognition with visual attention.
\newblock {\em arXiv:1412.7755 [cs]}, December 2014.
\newblock arXiv: 1412.7755.

\bibitem{cheng2016long}
Jianpeng Cheng, Li~Dong, and Mirella Lapata.
\newblock Long short-term memory-networks for machine reading.
\newblock {\em arXiv preprint arXiv:1601.06733}, 2016.

\bibitem{ke_sparse_2018}
Nan~Rosemary Ke, Anirudh Goyal, Olexa Bilaniuk, Jonathan Binas, Michael~C
  Mozer, Chris Pal, and Yoshua Bengio.
\newblock Sparse attentive backtracking: Temporal creditassignment through
  reminding.
\newblock {\em arXiv preprint arXiv:1809.03702}, 2018.

\bibitem{perera2020lstm}
Dilruk Perera and Roger Zimmermann.
\newblock Lstm networks for online cross-network recommendations.
\newblock {\em arXiv preprint arXiv:2008.10849}, 2020.

\bibitem{burgess2019monet}
Christopher~P Burgess, Loic Matthey, Nicholas Watters, Rishabh Kabra, Irina
  Higgins, Matt Botvinick, and Alexander Lerchner.
\newblock Monet: Unsupervised scene decomposition and representation.
\newblock {\em arXiv preprint arXiv:1901.11390}, 2019.

\bibitem{li2016learning}
Chongxuan Li, Jun Zhu, and Bo~Zhang.
\newblock Learning to generate with memory.
\newblock In {\em International Conference on Machine Learning}, pages
  1177--1186, 2016.

\bibitem{bartunov2016fast}
Sergey Bartunov and Dmitry~P Vetrov.
\newblock Fast adaptation in generative models with generative matching
  networks.
\newblock {\em arXiv preprint arXiv:1612.02192}, 2016.

\bibitem{rezende2016one}
Danilo~Jimenez Rezende, Shakir Mohamed, Ivo Danihelka, Karol Gregor, and Daan
  Wierstra.
\newblock One-shot generalization in deep generative models.
\newblock {\em arXiv preprint arXiv:1603.05106}, 2016.

\bibitem{bornschein2017variational}
J{\"o}rg Bornschein, Andriy Mnih, Daniel Zoran, and Danilo~J Rezende.
\newblock Variational memory addressing in generative models.
\newblock {\em arXiv preprint arXiv:1709.07116}, 2017.

\bibitem{escolano2018self}
Carlos Escolano, Marta~R Costa-juss{\`a}, and Jos{\'e}~AR Fonollosa.
\newblock (self-attentive) autoencoder-based universal language representation
  for machine translation.
\newblock {\em arXiv preprint arXiv:1810.06351}, 2018.

\bibitem{dehghani_universal_2018}
Mostafa Dehghani, Stephan Gouws, Oriol Vinyals, Jakob Uszkoreit, and Łukasz
  Kaiser.
\newblock Universal transformers.
\newblock {\em arXiv:1807.03819 [cs, stat]}, July 2018.
\newblock arXiv: 1807.03819.

\bibitem{joulin_inferring_2015}
Armand Joulin and Tomas Mikolov.
\newblock Inferring algorithmic patterns with stack-augmented recurrent nets.
\newblock {\em arXiv preprint arXiv:1503.01007}, 2015.

\bibitem{vinyals_matching_2016}
Oriol Vinyals, Charles Blundell, Timothy Lillicrap, Koray Kavukcuoglu, and Daan
  Wierstra.
\newblock Matching networks for one shot learning.
\newblock {\em arXiv:1606.04080 [cs, stat]}, June 2016.
\newblock arXiv: 1606.04080.

\bibitem{dai_transformer-xl:_2019}
Zihang Dai, Zhilin Yang, Yiming Yang, Jaime Carbonell, Quoc~V. Le, and Ruslan
  Salakhutdinov.
\newblock Transformer-xl: Attentive language models beyond a fixed-length
  context.
\newblock {\em arXiv:1901.02860 [cs, stat]}, January 2019.
\newblock arXiv: 1901.02860.

\bibitem{sainbayar_sukhbaatar;edouard_grave;piotr_bojanowski;armand_joulin_adaptive_2019}
Sainbayar Sukhbaatar, Edouard Grave, Piotr Bojanowski, and Armand Joulin.
\newblock Adaptive attention span in transformers.
\newblock {\em arXiv preprint arXiv:1905.07799}, 2019.

\bibitem{alexei_baevski;michael_auli_adaptive_2019}
Alexei Baevski and Michael Auli.
\newblock Adaptive input representations for neural language modeling.
\newblock {\em arXiv preprint arXiv:1809.10853}, 2018.

\bibitem{michal_daniluk_tim_rocktaschel_johannes_welbl_sebastian_riedel:_frustratingly_nodate}
Micha{\l} Daniluk, Tim Rockt{\"a}schel, Johannes Welbl, and Sebastian Riedel.
\newblock Frustratingly short attention spans in neural language modeling.
\newblock {\em arXiv preprint arXiv:1702.04521}, 2017.

\bibitem{kim_structured_2017}
Yoon Kim, Carl Denton, Luong Hoang, and Alexander~M. Rush.
\newblock Structured attention networks.
\newblock {\em arXiv:1702.00887 [cs]}, February 2017.
\newblock arXiv: 1702.00887.

\bibitem{feng_neural_2018}
Jiangtao Feng, Lingpeng Kong, Po-Sen Huang, Chong Wang, Da~Huang, Jiayuan Mao,
  Kan Qiao, and Dengyong Zhou.
\newblock Neural phrase-to-phrase machine translation.
\newblock {\em arXiv:1811.02172 [cs, stat]}, November 2018.
\newblock arXiv: 1811.02172.

\bibitem{shaw_self-attention_2018}
Peter Shaw, Jakob Uszkoreit, and Ashish Vaswani.
\newblock Self-attention with relative position representations, 2018.

\bibitem{raffel_online_2017}
Colin Raffel, Douglas Eck, Peter Liu, Ron~J. Weiss, and Thang Luong.
\newblock Online and linear-time attention by enforcing monotonic alignments,
  2017.

\bibitem{cho_learning_2014}
Kyunghyun Cho, Bart van Merrienboer, Caglar Gulcehre, Dzmitry Bahdanau, Fethi
  Bougares, Holger Schwenk, and Yoshua Bengio.
\newblock Learning phrase representations using rnn encoder-decoder for
  statistical machine translation.
\newblock {\em arXiv:1406.1078 [cs, stat]}, June 2014.
\newblock arXiv: 1406.1078.

\bibitem{sennrich_neural_2016}
Rico Sennrich, Barry Haddow, and Alexandra Birch.
\newblock Neural machine translation of rare words with subword units.
\newblock In {\em Proceedings of the 54th {Annual} {Meeting} of the
  {Association} for {Computational} {Linguistics} ({Volume} 1: {Long}
  {Papers})}, pages 1715--1725, Berlin, Germany, August 2016. Association for
  Computational Linguistics.

\bibitem{thomas_zenkel;joern_wuebker;john_denero_adding_2019}
Thomas Zenkel, Joern Wuebker, and John DeNero.
\newblock Adding interpretable attention to neural translation models improves
  word alignment.
\newblock {\em arXiv preprint arXiv:1901.11359}, 2019.

\bibitem{baosong_yang;jian_li;derek_wong;lidia_s._chao;xing_wang;zhaopeng_tu_context-aware_2019}
Baosong Yang, Jian Li, Derek~F Wong, Lidia~S Chao, Xing Wang, and Zhaopeng Tu.
\newblock Context-aware self-attention networks.
\newblock In {\em Proceedings of the AAAI Conference on Artificial
  Intelligence}, pages 387--394, 2019.

\bibitem{baosong_yang;longyue_wang;derek_f._wong;lidia_s._chao;zhaopeng_tu_convolutional_2019}
Baosong Yang, Longyue Wang, Derek Wong, Lidia~S Chao, and Zhaopeng Tu.
\newblock Convolutional self-attention networks.
\newblock {\em arXiv preprint arXiv:1904.03107}, 2019.

\bibitem{jie_hao;xing_wang;baosong_yang;longyue_wang;jinfeng_zhang;zhaopeng_tu_modeling_2019}
Jie Hao, Xing Wang, Baosong Yang, Longyue Wang, Jinfeng Zhang, and Zhaopeng Tu.
\newblock Modeling recurrence for transformer.
\newblock {\em arXiv preprint arXiv:1904.03092}, 2019.

\bibitem{felix_hieber;tobias_domhan;michael_denkowski;david_vilar;artem_sokolov;ann_clifton;matt_post_sockeye:_2018}
Felix Hieber, Tobias Domhan, Michael Denkowski, David Vilar, Artem Sokolov, Ann
  Clifton, and Matt Post.
\newblock Sockeye: A toolkit for neural machine translation.
\newblock {\em arXiv preprint arXiv:1712.05690}, 2017.

\bibitem{xiangwen_zhang;jinsong_su;yue_qin;yang_liu;rongrong_ji;hongji_wang_asynchronous_2018}
Xiangwen Zhang, Jinsong Su, Yue Qin, Yang Liu, Rongrong Ji, and Hongji Wang.
\newblock Asynchronous bidirectional decoding for neural machine translation.
\newblock In {\em Proceedings of the AAAI Conference on Artificial
  Intelligence}, 2018.

\bibitem{joost_bastings;ivan_titov;wilker_aziz;diego_marcheggiani;khalil_simaan_graph_2017}
Joost Bastings, Ivan Titov, Wilker Aziz, Diego Marcheggiani, and Khalil
  Sima'an.
\newblock Graph convolutional encoders for syntax-aware neural machine
  translation.
\newblock {\em arXiv preprint arXiv:1704.04675}, 2017.

\bibitem{jie_zhou;ying_cao;xuguang_wang;peng_li;wei_xu_deep_2016}
Jie Zhou, Ying Cao, Xuguang Wang, Peng Li, and Wei Xu.
\newblock Deep recurrent models with fast-forward connections for neural
  machine translation.
\newblock {\em Transactions of the Association for Computational Linguistics},
  4:371--383, 2016.

\bibitem{yonghui_wu;mike_schuster;zhifeng_chen;quoc_v._le;mohammad_norouzi;wolfgang_macherey;maxim_krikun;yuan_cao;qin_gao;klaus_macherey;jeff_klingner;apurva_shah;melvin_johnson;xiaobing_liu;lukasz_kaiser;stephan_gouws;yoshikiyo_kato;taku_kudo;hideto_kazawa;keith_stevens;george_kurian;nishant_patil;wei_wang;cliff_young;jason_smith_googles_2016}
Yonghui Wu, Mike Schuster, Zhifeng Chen, Quoc~V Le, Mohammad Norouzi, Wolfgang
  Macherey, Maxim Krikun, Yuan Cao, Qin Gao, Klaus Macherey, et~al.
\newblock Google's neural machine translation system: Bridging the gap between
  human and machine translation.
\newblock {\em arXiv preprint arXiv:1609.08144}, 2016.

\bibitem{akiko_eriguchi_kazuma_hashimoto_yoshimasa_tsuruoka:_tree--sequence_nodate}
Akiko Eriguchi, Kazuma Hashimoto, and Yoshimasa Tsuruoka.
\newblock Tree-to-sequence attentional neural machine translation.
\newblock In {\em Proceedings of the 54th Annual Meeting of the Association for
  Computational Linguistics (Volume 1: Long Papers)}, pages 823--833, 2016.

\bibitem{jian_li;baosong_yang;zi-yi_dou;xing_wang;michael_r._lyu;zhaopeng_tu_information_2019}
Jian Li, Baosong Yang, Zi-Yi Dou, Xing Wang, Michael~R Lyu, and Zhaopeng Tu.
\newblock Information aggregation for multi-head attention with
  routing-by-agreement.
\newblock {\em arXiv preprint arXiv:1904.03100}, 2019.

\bibitem{orhan_firat;kyunghyun_cho;yoshua_bengio_multi-way;_2016}
Orhan Firat, Kyunghyun Cho, and Yoshua Bengio.
\newblock Multi-way, multilingual neural machine translation with a shared
  attention mechanism.
\newblock {\em arXiv preprint arXiv:1601.01073}, 2016.

\bibitem{deng_latent_2018}
Yuntian Deng, Yoon Kim, Justin Chiu, Demi Guo, and Alexander Rush.
\newblock Latent alignment and variational attention.
\newblock In S.~Bengio, H.~Wallach, H.~Larochelle, K.~Grauman, N.~Cesa-Bianchi,
  and R.~Garnett, editors, {\em Advances in {Neural} {Information} {Processing}
  {Systems} 31}, pages 9712--9724. Curran Associates, Inc., 2018.

\bibitem{h._zhang;_j._li;_y._ji;_h._yue_understanding_2017}
Haijun Zhang, Jingxuan Li, Yuzhu Ji, and Heng Yue.
\newblock Understanding subtitles by character-level sequence-to-sequence
  learning.
\newblock {\em IEEE Transactions on Industrial Informatics}, 13(2):616--624,
  2016.

\bibitem{parikh_decomposable_2016}
Ankur~P. Parikh, Oscar Täckström, Dipanjan Das, and Jakob Uszkoreit.
\newblock A decomposable attention model for natural language inference.
\newblock {\em arXiv:1606.01933 [cs]}, June 2016.
\newblock arXiv: 1606.01933.

\bibitem{shen_disan:_nodate}
Tao Shen, Tianyi Zhou, Guodong Long, Jing Jiang, Shirui Pan, and Chengqi Zhang.
\newblock Disan: Directional self-attention network for rnn/cnn-free language
  understanding.
\newblock In {\em Proceedings of the AAAI Conference on Artificial
  Intelligence}, 2018.

\bibitem{tao_shen_tianyi_zhou_guodong_long_jing_jiang_chengqi_zhang:_bi-directional_nodate}
Tao Shen, Tianyi Zhou, Guodong Long, Jing Jiang, and Chengqi Zhang.
\newblock Bi-directional block self-attention for fast and memory-efficient
  sequence modeling.
\newblock {\em arXiv preprint arXiv:1804.00857}, 2018.

\bibitem{shuohang_wang;jing_jiang_learning_2016}
Shuohang Wang and Jing Jiang.
\newblock Learning natural language inference with lstm.
\newblock {\em arXiv preprint arXiv:1512.08849}, 2015.

\bibitem{yang_liu;chengjie_sun;lei_lin;xiaolong_wang_learning_2016}
Yang Liu, Chengjie Sun, Lei Lin, and Xiaolong Wang.
\newblock Learning natural language inference using bidirectional lstm model
  and inner-attention.
\newblock {\em arXiv preprint arXiv:1605.09090}, 2016.

\bibitem{neelakantan_neural_2015}
Arvind Neelakantan, Quoc~V. Le, and Ilya Sutskever.
\newblock Neural programmer: Inducing latent programs with gradient descent.
\newblock {\em arXiv:1511.04834 [cs, stat]}, November 2015.
\newblock arXiv: 1511.04834.

\bibitem{wu_hierarchical_2017}
Chen Xing, Yu~Wu, Wei Wu, Yalou Huang, and Ming Zhou.
\newblock Hierarchical recurrent attention network for response generation.
\newblock In {\em Proceedings of the AAAI Conference on Artificial
  Intelligence}, 2018.

\bibitem{yang_anmm:_2018}
Liu Yang, Qingyao Ai, Jiafeng Guo, and W.~Bruce Croft.
\newblock anmm: Ranking short answer texts with attention-based neural matching
  model.
\newblock {\em arXiv:1801.01641 [cs]}, January 2018.
\newblock arXiv: 1801.01641.

\bibitem{tay_densely_2018}
Yi~Tay, Anh~Tuan Luu, Siu~Cheung Hui, and Jian Su.
\newblock Densely connected attention propagation for reading comprehension.
\newblock In S.~Bengio, H.~Wallach, H.~Larochelle, K.~Grauman, N.~Cesa-Bianchi,
  and R.~Garnett, editors, {\em Advances in {Neural} {Information} {Processing}
  {Systems} 31}, pages 4906--4917. Curran Associates, Inc., 2018.

\bibitem{adams_wei_yu_david_dohan_minh-thang_luong_rui_zhao_kai_chen_mohammad_norouzi_quoc_v._le:_qanet:_nodate}
Adams~Wei Yu, David Dohan, Minh-Thang Luong, Rui Zhao, Kai Chen, Mohammad
  Norouzi, and Quoc~V Le.
\newblock Qanet: Combining local convolution with global self-attention for
  reading comprehension.
\newblock {\em arXiv preprint arXiv:1804.09541}, 2018.

\bibitem{wei_wang_chen_wu_ming_yan:_multi-granularity_nodate}
Wei Wang, Ming Yan, and Chen Wu.
\newblock Multi-granularity hierarchical attention fusion networks for reading
  comprehension and question answering.
\newblock In {\em Proceedings of the 56th Annual Meeting of the Association for
  Computational Linguistics}, pages 1705--1714, 2018.

\bibitem{xiangyang_zhou_lu_li_daxiang_dong_yi_liu_ying_chen_wayne_xin_zhao_dianhai_yu_hua_wu:_multi-turn_nodate}
Xiangyang Zhou, Lu~Li, Daxiang Dong, Yi~Liu, Ying Chen, Wayne~Xin Zhao, Dianhai
  Yu, and Hua Wu.
\newblock Multi-turn response selection for chatbots with deep attention
  matching network.
\newblock In {\em Proceedings of the 56th Annual Meeting of the Association for
  Computational Linguistics (Volume 1: Long Papers)}, pages 1118--1127, 2018.

\bibitem{cicero_dos_santos;ming_tan;bing_xiang;bowen_zhou_attentive_2016}
Cicero~dos Santos, Ming Tan, Bing Xiang, and Bowen Zhou.
\newblock Attentive pooling networks.
\newblock {\em arXiv preprint arXiv:1602.03609}, 2016.

\bibitem{victor_zhong;caiming_xiong;nitish_shirish_keskar;richard_socher_coarse-grain_2019}
Victor Zhong, Caiming Xiong, Nitish~Shirish Keskar, and Richard Socher.
\newblock Coarse-grain fine-grain coattention network for multi-evidence
  question answering.
\newblock {\em arXiv preprint arXiv:1901.00603}, 2019.

\bibitem{louis_shao;stephan_gouws;denny_britz;anna_goldie;brian_strope;ray_kurzweil_generating_2017}
Louis Shao, Stephan Gouws, Denny Britz, Anna Goldie, Brian Strope, and Ray
  Kurzweil.
\newblock Generating high-quality and informative conversation responses with
  sequence-to-sequence models.
\newblock {\em arXiv preprint arXiv:1701.03185}, 2017.

\bibitem{chenguang_zhu;michael_zeng;xuedong_huang_sdnet:_2019}
Chenguang Zhu, Michael Zeng, and Xuedong Huang.
\newblock Sdnet: Contextualized attention-based deep network for conversational
  question answering.
\newblock {\em arXiv preprint arXiv:1812.03593}, 2018.

\bibitem{ming_tan;cicero_dos_santos;bing_xiang;bowen_zhou_lstm-based_2016}
Ming Tan, Cicero~dos Santos, Bing Xiang, and Bowen Zhou.
\newblock Lstm-based deep learning models for non-factoid answer selection.
\newblock {\em arXiv preprint arXiv:1511.04108}, 2015.

\bibitem{bhuwan_dhingra_hanxiao_liu_zhilin_yang_william_w._cohen_ruslan_salakhutdinov:_gated-attention_nodate}
Bhuwan Dhingra, Hanxiao Liu, Zhilin Yang, William~W Cohen, and Ruslan
  Salakhutdinov.
\newblock Gated-attention readers for text comprehension.
\newblock {\em arXiv preprint arXiv:1606.01549}, 2016.

\bibitem{yanchao_hao_yuanzhe_zhang_kang_liu_shizhu_he_zhanyi_liu_hua_wu_jun_zhao:_end--end_nodate}
Yanchao Hao, Yuanzhe Zhang, Kang Liu, Shizhu He, Zhanyi Liu, Hua Wu, and Jun
  Zhao.
\newblock An end-to-end model for question answering over knowledge base with
  cross-attention combining global knowledge.
\newblock In {\em Proceedings of the 55th Annual Meeting of the Association for
  Computational Linguistics (Volume 1: Long Papers)}, pages 221--231, 2017.

\bibitem{rudolf_kadlec_martin_schmid_ondrej_bajgar_jan_kleindienst:_text_nodate}
Rudolf Kadlec, Martin Schmid, Ond{\v{r}}ej Bajgar, and Jan Kleindienst.
\newblock Text understanding with the attention sum reader network.
\newblock In {\em Proceedings of the 54th Annual Meeting of the Association for
  Computational Linguistics (Volume 1: Long Papers)}, pages 908--918, 2016.

\bibitem{tsendsuren_munkhdalai;hong_yu_neural_2017}
Tsendsuren Munkhdalai and Hong Yu.
\newblock Neural tree indexers for text understanding.
\newblock In {\em Proceedings of the conference. Association for Computational
  Linguistics. Meeting}, volume~1, page~11. NIH Public Access, 2017.

\bibitem{seonhoon_kim;inho_kang;nojun_kwak_semantic_2018}
Seonhoon Kim, Inho Kang, and Nojun Kwak.
\newblock Semantic sentence matching with densely-connected recurrent and
  co-attentive information.
\newblock In {\em Proceedings of the AAAI conference on artificial
  intelligence}, pages 6586--6593, 2019.

\bibitem{lisa_bauer;yicheng_wang;mohit_bansal_commonsense_2019}
Lisa Bauer, Yicheng Wang, and Mohit Bansal.
\newblock Commonsense for generative multi-hop question answering tasks.
\newblock {\em arXiv preprint arXiv:1809.06309}, 2018.

\bibitem{qiu_ran;peng_li;weiwei_hu;jie_zhou_option_2019}
Qiu Ran, Peng Li, Weiwei Hu, and Jie Zhou.
\newblock Option comparison network for multiple-choice reading comprehension.
\newblock {\em arXiv preprint arXiv:1903.03033}, 2019.

\bibitem{hermann2015teaching}
Karl~Moritz Hermann, Tomas Kocisky, Edward Grefenstette, Lasse Espeholt, Will
  Kay, Mustafa Suleyman, and Phil Blunsom.
\newblock Teaching machines to read and comprehend.
\newblock In {\em Advances in neural information processing systems}, pages
  1693--1701, 2015.

\bibitem{chen_xing;wei_wu;yu_wu;jie_liu;yalou_huang;ming_zhou;wei-ying_ma_topic_2016}
Chen Xing, Wei Wu, Yu~Wu, Jie Liu, Yalou Huang, Ming Zhou, and Wei-Ying Ma.
\newblock Topic aware neural response generation.
\newblock In {\em Proceedings of the AAAI Conference on Artificial
  Intelligence}, 2017.

\bibitem{sordoni_iterative_2017}
Alessandro Sordoni, Philip Bachman, Adam Trischler, and Yoshua Bengio.
\newblock Iterative alternating neural attention for machine reading.
\newblock {\em arXiv preprint arXiv:1606.02245}, 2016.

\bibitem{bingning_wang_kang_liu_jun_zhao:_inner_nodate}
Bingning Wang, Kang Liu, and Jun Zhao.
\newblock Inner attention based recurrent neural networks for answer selection.
\newblock In {\em Proceedings of the 54th Annual Meeting of the Association for
  Computational Linguistics (Volume 1: Long Papers)}, pages 1288--1297, 2016.

\bibitem{wang_attention-based_2017}
Yequan Wang, Minlie Huang, Xiaoyan Zhu, and Li~Zhao.
\newblock Attention-based lstm for aspect-level sentiment classification.
\newblock In {\em Proceedings of the 2016 conference on empirical methods in
  natural language processing}, pages 606--615, 2016.

\bibitem{dehong_ma;sujian_li;xiaodong_zhang;houfeng_wang_interactive_2017}
Dehong Ma, Sujian Li, Xiaodong Zhang, and Houfeng Wang.
\newblock Interactive attention networks for aspect-level sentiment
  classification.
\newblock {\em arXiv preprint arXiv:1709.00893}, 2017.

\bibitem{kai_shuang_xintao_ren_qianqian_yang_rui_li_jonathan_loo:_aela-dlstms:_nodate}
Kai Shuang, Xintao Ren, Qianqian Yang, Rui Li, and Jonathan Loo.
\newblock Aela-dlstms: Attention-enabled and location-aware double lstms for
  aspect-level sentiment classification.
\newblock {\em Neurocomputing}, 334:25--34, 2019.

\bibitem{christos_baziotis_nikos_pelekis_christos_doulkeridis:_datastories_nodate}
Christos Baziotis, Nikos Pelekis, and Christos Doulkeridis.
\newblock Datastories at semeval-2017 task 4: Deep lstm with attention for
  message-level and topic-based sentiment analysis.
\newblock In {\em Proceedings of the 11th international workshop on semantic
  evaluation (SemEval-2017)}, pages 747--754, 2017.

\bibitem{wei_xue;tao_li_aspect_2018}
Wei Xue and Tao Li.
\newblock Aspect based sentiment analysis with gated convolutional networks.
\newblock {\em arXiv preprint arXiv:1805.07043}, 2018.

\bibitem{shi_feng;yang_wang;liran_liu;daling_wang;ge_yu_attention_2019}
Shi Feng, Yang Wang, Liran Liu, Daling Wang, and Ge~Yu.
\newblock Attention based hierarchical lstm network for context-aware microblog
  sentiment classification.
\newblock {\em World Wide Web}, 22(1):59--81, 2019.

\bibitem{youwei_song;jiahai_wang;tao_jiang;zhiyue_liu;yanghui_rao_attentional_2019}
Youwei Song, Jiahai Wang, Tao Jiang, Zhiyue Liu, and Yanghui Rao.
\newblock Attentional encoder network for targeted sentiment classification.
\newblock {\em arXiv preprint arXiv:1902.09314}, 2019.

\bibitem{bonggun_shin;timothy_lee;jinho_d._choi_lexicon_2017}
Bonggun Shin, Timothy Lee, and Jinho~D Choi.
\newblock Lexicon integrated cnn models with attention for sentiment analysis.
\newblock {\em arXiv preprint arXiv:1610.06272}, 2016.

\bibitem{minchae_song;hyunjung_park;kyung-shik_shin_attention-based_2019}
Minchae Song, Hyunjung Park, and Kyung-shik Shin.
\newblock Attention-based long short-term memory network using sentiment
  lexicon embedding for aspect-level sentiment analysis in korean.
\newblock {\em Information Processing \& Management}, 56(3):637--653, 2019.

\bibitem{jiachen_du;lin_gui;yulan_he;ruifeng_xu;xuan_wang_convolution-based_2019}
Jiachen Du, Lin Gui, Yulan He, Ruifeng Xu, and Xuan Wang.
\newblock Convolution-based neural attention with applications to sentiment
  classification.
\newblock {\em IEEE Access}, 7:27983--27992, 2019.

\bibitem{peng_chen;zhongqian_sun;lidong_bing;wei_yang_recurrent_2017}
Peng Chen, Zhongqian Sun, Lidong Bing, and Wei Yang.
\newblock Recurrent attention network on memory for aspect sentiment analysis.
\newblock In {\em Proceedings of the 2017 conference on empirical methods in
  natural language processing}, pages 452--461, 2017.

\bibitem{yukun_ma;haiyun_peng;erik_cambria_targeted_2018}
Yukun Ma, Haiyun Peng, and Erik Cambria.
\newblock Targeted aspect-based sentiment analysis via embedding commonsense
  knowledge into an attentive lstm.
\newblock In {\em Proceedings of the AAAI Conference on Artificial
  Intelligence}, 2018.

\bibitem{yue_zhang;jiangming_liu_attention_2017}
Jiangming Liu and Yue Zhang.
\newblock Attention modeling for targeted sentiment.
\newblock In {\em Proceedings of the 15th Conference of the European Chapter of
  the Association for Computational Linguistics: Volume 2, Short Papers}, pages
  572--577, 2017.

\bibitem{huimin_chen;maosong_sun;cunchao_tu;yankai_lin;zhiyuan_liu_neural_2016}
Huimin Chen, Maosong Sun, Cunchao Tu, Yankai Lin, and Zhiyuan Liu.
\newblock Neural sentiment classification with user and product attention.
\newblock In {\em Proceedings of the 2016 conference on empirical methods in
  natural language processing}, pages 1650--1659, 2016.

\bibitem{jiangfeng_zeng;xiao_ma;ke_zhou_enhancing_2019}
Jiangfeng Zeng, Xiao Ma, and Ke~Zhou.
\newblock Enhancing attention-based lstm with position context for aspect-level
  sentiment classification.
\newblock {\em IEEE Access}, 7:20462--20471, 2019.

\bibitem{she_distant_2018}
Heng She, Bin Wu, Bai Wang, and Renjun Chi.
\newblock Distant supervision for relation extraction with hierarchical
  attention and entity descriptions.
\newblock In {\em Proc. of IEEE IJCNN}, pages 1--8. IEEE, 2018.

\bibitem{zhang_learning_2018}
Biao Zhang, Deyi Xiong, Jinsong Su, and Min Zhang.
\newblock Learning better discourse representation for implicit discourse
  relation recognition via attention networks.
\newblock {\em Neurocomputing}, 275:1241--1249, January 2018.

\bibitem{shen_reinforced_2018}
Tao Shen, Tianyi Zhou, Guodong Long, Jing Jiang, Sen Wang, and Chengqi Zhang.
\newblock Reinforced self-attention network: a hybrid of hard and soft
  attention for sequence modeling.
\newblock {\em arXiv:1801.10296 [cs]}, January 2018.
\newblock arXiv: 1801.10296.

\bibitem{zhixing_tan;mingxuan_wang;jun_xie;yidong_chen;xiaodong_shi_deep_2017}
Zhixing Tan, Mingxuan Wang, Jun Xie, Yidong Chen, and Xiaodong Shi.
\newblock Deep semantic role labeling with self-attention.
\newblock In {\em Proceedings of the AAAI Conference on Artificial
  Intelligence}, 2018.

\bibitem{li_dong_mirella_lapata:_language_nodate}
Li~Dong and Mirella Lapata.
\newblock Language to logical form with neural attention.
\newblock In {\em Proceedings of the 54th Annual Meeting of the Association for
  Computational Linguistics (Volume 1: Long Papers)}, pages 33--43, 2016.

\bibitem{wenya_wu;yufeng_chen;jinan_xu;yujie_zhang_attention-based_2018}
Wenya Wu, Yufeng Chen, Jinan Xu, and Yujie Zhang.
\newblock Attention-based convolutional neural networks for chinese relation
  extraction.
\newblock In {\em Chinese Computational Linguistics and Natural Language
  Processing Based on Naturally Annotated Big Data}, pages 147--158. Springer,
  2018.

\bibitem{chorowski_attention-based_2015}
Jan Chorowski, Dzmitry Bahdanau, Dmitriy Serdyuk, Kyunghyun Cho, and Yoshua
  Bengio.
\newblock Attention-based models for speech recognition.
\newblock In {\em Proceedings of the 28th International Conference on Neural
  Information Processing Systems-Volume 1}, pages 577--585, 2015.

\bibitem{prabhavalkar_minimum_2018}
Rohit Prabhavalkar, Tara Sainath, Yonghui Wu, Patrick Nguyen, Zhifeng Chen,
  Chung-Cheng Chiu, and Anjuli Kannan.
\newblock Minimum word error rate training for attention-based
  sequence-to-sequence models, 2018.

\bibitem{kazuki_irie;albert_zeyer;ralf_schluter;hermann_ney_language_2019}
Kazuki Irie, Albert Zeyer, Ralf Schl{\"u}ter, and Hermann Ney.
\newblock Language modeling with deep transformers.
\newblock {\em arXiv preprint arXiv:1905.04226}, 2019.

\bibitem{christoph_luscher;eugen_beck;kazuki_irie;markus_kitza;wilfried_michel;albert_zeyer;ralf_schluter;hermann_ney_rwth_2019}
Christoph L{\"u}scher, Eugen Beck, Kazuki Irie, Markus Kitza, Wilfried Michel,
  Albert Zeyer, Ralf Schl{\"u}ter, and Hermann Ney.
\newblock Rwth asr systems for librispeech: Hybrid vs attention--w/o data
  augmentation.
\newblock {\em arXiv preprint arXiv:1905.03072}, 2019.

\bibitem{linhao_dong;feng_wang;bo_xu_self-attention_2019}
Linhao Dong, Feng Wang, and Bo~Xu.
\newblock Self-attention aligner: A latency-control end-to-end model for asr
  using self-attention network and chunk-hopping.
\newblock In {\em ICASSP 2019-2019 IEEE International Conference on Acoustics,
  Speech and Signal Processing (ICASSP)}, pages 5656--5660. IEEE, 2019.

\bibitem{julian_salazar;katrin_kirchhoff;zhiheng_huang_self-attention_2019}
Julian Salazar, Katrin Kirchhoff, and Zhiheng Huang.
\newblock Self-attention networks for connectionist temporal classification in
  speech recognition.
\newblock In {\em ICASSP 2019-2019 IEEE International Conference on Acoustics,
  Speech and Signal Processing (ICASSP)}, pages 7115--7119. IEEE, 2019.

\bibitem{xiaofei_wang;ruizhi_li;sri_harish_mallid;takaaki_hori;shinji_watanabe;hynek_hermansky_stream_2019}
Xiaofei Wang, Ruizhi Li, Sri~Harish Mallidi, Takaaki Hori, Shinji Watanabe, and
  Hynek Hermansky.
\newblock Stream attention-based multi-array end-to-end speech recognition.
\newblock In {\em ICASSP 2019-2019 IEEE International Conference on Acoustics,
  Speech and Signal Processing (ICASSP)}, pages 7105--7109. IEEE, 2019.

\bibitem{shiyu_zhou;linhao_dong;shuang_xu;bo_xu_syllable-based_2018}
Shiyu Zhou, Linhao Dong, Shuang Xu, and Bo~Xu.
\newblock Syllable-based sequence-to-sequence speech recognition with the
  transformer in mandarin chinese.
\newblock {\em arXiv preprint arXiv:1804.10752}, 2018.

\bibitem{shinji_watanabe;takaaki_hori;suyoun_kim;john_r._hershey;tomoki_hayashi_hybrid_2017}
Shinji Watanabe, Takaaki Hori, Suyoun Kim, John~R Hershey, and Tomoki Hayashi.
\newblock Hybrid ctc/attention architecture for end-to-end speech recognition.
\newblock {\em IEEE Journal of Selected Topics in Signal Processing},
  11(8):1240--1253, 2017.

\bibitem{dzmitry_bahdanau;jan_chorowski;dmitriy_serdyuk;philemon_brakel;yoshua_bengio_end--end_2016}
Dzmitry Bahdanau, Jan Chorowski, Dmitriy Serdyuk, Philemon Brakel, and Yoshua
  Bengio.
\newblock End-to-end attention-based large vocabulary speech recognition.
\newblock In {\em 2016 IEEE international conference on acoustics, speech and
  signal processing (ICASSP)}, pages 4945--4949. IEEE, 2016.

\bibitem{jan_chorowski;dzmitry_bahdanau;kyunghyun_cho;yoshua_bengio_end--end_2014}
Jan Chorowski, Dzmitry Bahdanau, Kyunghyun Cho, and Yoshua Bengio.
\newblock End-to-end continuous speech recognition using attention-based
  recurrent nn: First results.
\newblock {\em arXiv preprint arXiv:1412.1602}, 2014.

\bibitem{f_a_rezaur_rahman_chowdhury;quan_wang;ignacio_lopez_moreno;li_wan_attention-based_2018}
FA~Rezaur rahman Chowdhury, Quan Wang, Ignacio~Lopez Moreno, and Li~Wan.
\newblock Attention-based models for text-dependent speaker verification.
\newblock In {\em 2018 IEEE International Conference on Acoustics, Speech and
  Signal Processing (ICASSP)}, pages 5359--5363. IEEE, 2018.

\bibitem{yiming_wang;xing_fan;i-fan_chen;yuzong_liu;tongfei_chen;bjorn_hoffmeister_end--end_2019}
Yiming Wang, Xing Fan, I-Fan Chen, Yuzong Liu, Tongfei Chen, and Bj{\"o}rn
  Hoffmeister.
\newblock End-to-end anchored speech recognition.
\newblock In {\em ICASSP 2019-2019 IEEE International Conference on Acoustics,
  Speech and Signal Processing (ICASSP)}, pages 7090--7094. IEEE, 2019.

\bibitem{albert_zeyer;kazuki_irie;ralf_schluter;hermann_ney_improved_2018}
Albert Zeyer, Kazuki Irie, Ralf Schl{\"u}ter, and Hermann Ney.
\newblock Improved training of end-to-end attention models for speech
  recognition.
\newblock {\em arXiv preprint arXiv:1805.03294}, 2018.

\bibitem{suyoun_kim;takaaki_hori;shinji_watanabe_joint_2017}
Suyoun Kim, Takaaki Hori, and Shinji Watanabe.
\newblock Joint ctc-attention based end-to-end speech recognition using
  multi-task learning.
\newblock In {\em 2017 IEEE international conference on acoustics, speech and
  signal processing (ICASSP)}, pages 4835--4839. IEEE, 2017.

\bibitem{koji_okabe;takafumi_koshinaka;koichi_shinoda_attentive_2018}
Koji Okabe, Takafumi Koshinaka, and Koichi Shinoda.
\newblock Attentive statistics pooling for deep speaker embedding.
\newblock {\em arXiv preprint arXiv:1803.10963}, 2018.

\bibitem{takaaki_hori;shinji_watanabe;yu_zhang;william_chan_advances_2017}
Takaaki Hori, Shinji Watanabe, Yu~Zhang, and William Chan.
\newblock Advances in joint ctc-attention based end-to-end speech recognition
  with a deep cnn encoder and rnn-lm.
\newblock {\em arXiv preprint arXiv:1706.02737}, 2017.

\bibitem{chopra_abstractive_2016}
Sumit Chopra, Michael Auli, and Alexander~M. Rush.
\newblock Abstractive sentence summarization with attentive recurrent neural
  networks.
\newblock In {\em Proceedings of the 2016 {Conference} of the {North}
  {American} {Chapter} of the {Association} for {Computational} {Linguistics}:
  {Human} {Language} {Technologies}}, pages 93--98, San Diego, California,
  2016. Association for Computational Linguistics.

\bibitem{paulus_deep_2017}
Romain Paulus, Caiming Xiong, and Richard Socher.
\newblock A deep reinforced model for abstractive summarization.
\newblock {\em arXiv preprint arXiv:1705.04304}, 2017.

\bibitem{nallapati_abstractive_2016}
Ramesh Nallapati, Bowen Zhou, Caglar Gulcehre, Bing Xiang, et~al.
\newblock Abstractive text summarization using sequence-to-sequence rnns and
  beyond.
\newblock {\em arXiv preprint arXiv:1602.06023}, 2016.

\bibitem{alexander_m._rush;sumit_chopra;jason_weston_neural_2015}
Alexander~M Rush, Sumit Chopra, and Jason Weston.
\newblock A neural attention model for abstractive sentence summarization.
\newblock {\em arXiv preprint arXiv:1509.00685}, 2015.

\bibitem{angela_fan;mike_lewis;yann_dauphin_hierarchical_2018}
Angela Fan, Mike Lewis, and Yann Dauphin.
\newblock Hierarchical neural story generation.
\newblock {\em arXiv preprint arXiv:1805.04833}, 2018.

\bibitem{b._rekabdar;_c._mousas;_b._gupta_generative_2019}
Banafsheh Rekabdar, Christos Mousas, and Bidyut Gupta.
\newblock Generative adversarial network with policy gradient for text
  summarization.
\newblock In {\em 2019 IEEE 13th international conference on semantic computing
  (ICSC)}, pages 204--207. IEEE, 2019.

\bibitem{arman_cohan;franck_dernoncourt;doo_soon_kim;trung_bui;seokhwan_kim;walter_chang;nazli_goharian_discourse-aware_2018}
Arman Cohan, Franck Dernoncourt, Doo~Soon Kim, Trung Bui, Seokhwan Kim, Walter
  Chang, and Nazli Goharian.
\newblock A discourse-aware attention model for abstractive summarization of
  long documents.
\newblock {\em arXiv preprint arXiv:1804.05685}, 2018.

\bibitem{schuster1997bidirectional}
Mike Schuster and Kuldip~K Paliwal.
\newblock Bidirectional recurrent neural networks.
\newblock {\em IEEE transactions on Signal Processing}, 45(11):2673--2681,
  1997.

\bibitem{allamanis2016convolutional}
Miltiadis Allamanis, Hao Peng, and Charles Sutton.
\newblock A convolutional attention network for extreme summarization of source
  code.
\newblock In {\em International conference on machine learning}, pages
  2091--2100, 2016.

\bibitem{huang2018video}
Jie Huang, Wengang Zhou, Qilin Zhang, Houqiang Li, and Weiping Li.
\newblock Video-based sign language recognition without temporal segmentation.
\newblock In {\em Thirty-Second AAAI Conference on Artificial Intelligence},
  2018.

\bibitem{seyedmahdad_mirsamadi;emad_barsoum;cha_zhang_automatic_2017}
Seyedmahdad Mirsamadi, Emad Barsoum, and Cha Zhang.
\newblock Automatic speech emotion recognition using recurrent neural networks
  with local attention.
\newblock In {\em 2017 IEEE International Conference on Acoustics, Speech and
  Signal Processing (ICASSP)}, pages 2227--2231. IEEE, 2017.

\bibitem{le_hoang_son;akshi_kumar;saurabh_raj_sangwan;anshika_arora;an;nayyar;mohamed_abdel-basset_sarcasm_2019}
Akshi Kumar, Saurabh~Raj Sangwan, Anshika Arora, Anand Nayyar, Mohamed
  Abdel-Basset, et~al.
\newblock Sarcasm detection using soft attention-based bidirectional long
  short-term memory model with convolution network.
\newblock {\em IEEE access}, 7:23319--23328, 2019.

\bibitem{kun-yi_huang;chung-hsien_wu;ming-hsiang_su_attention-based_2019}
Kun-Yi Huang, Chung-Hsien Wu, and Ming-Hsiang Su.
\newblock Attention-based convolutional neural network and long short-term
  memory for short-term detection of mood disorders based on elicited speech
  responses.
\newblock {\em Pattern Recognition}, 88:668--678, 2019.

\bibitem{navonil_majumder;soujanya_poria;devamanyu_hazarika;rada_mihalcea;alexander_gelbukh;erik_cambria_dialoguernn:_2019}
Navonil Majumder, Soujanya Poria, Devamanyu Hazarika, Rada Mihalcea, Alexander
  Gelbukh, and Erik Cambria.
\newblock Dialoguernn: An attentive rnn for emotion detection in conversations.
\newblock In {\em Proceedings of the AAAI Conference on Artificial
  Intelligence}, pages 6818--6825, 2019.

\bibitem{yuanyuan_zhang;jun_du;zirui_wang;jianshu_zhang_attention_2019}
Yuanyuan Zhang, Jun Du, Zirui Wang, Jianshu Zhang, and Yanhui Tu.
\newblock Attention based fully convolutional network for speech emotion
  recognition.
\newblock In {\em 2018 Asia-Pacific Signal and Information Processing
  Association Annual Summit and Conference (APSIPA ASC)}, pages 1771--1775.
  IEEE, 2018.

\bibitem{michael_neumann;ngoc_thang_vu_attentive_2017}
Michael Neumann and Ngoc~Thang Vu.
\newblock Attentive convolutional neural network based speech emotion
  recognition: A study on the impact of input features, signal length, and
  acted speech.
\newblock {\em arXiv preprint arXiv:1706.00612}, 2017.

\bibitem{wenya_wang;sinno_jialin_pan;daniel_dahlmeier;xiaokui_xiao_coupled_2017}
Wenya Wang, Sinno~Jialin Pan, Daniel Dahlmeier, and Xiaokui Xiao.
\newblock Coupled multi-layer attentions for co-extraction of aspect and
  opinion terms.
\newblock In {\em Proceedings of the AAAI Conference on Artificial
  Intelligence}, 2017.

\bibitem{liu2019bidirectional}
Gang Liu and Jiabao Guo.
\newblock Bidirectional lstm with attention mechanism and convolutional layer
  for text classification.
\newblock {\em Neurocomputing}, 337:325--338, 2019.

\bibitem{norouzian2019exploring}
Atta Norouzian, Bogdan Mazoure, Dermot Connolly, and Daniel Willett.
\newblock Exploring attention mechanism for acoustic-based classification of
  speech utterances into system-directed and non-system-directed.
\newblock In {\em ICASSP 2019-2019 IEEE International Conference on Acoustics,
  Speech and Signal Processing (ICASSP)}, pages 7310--7314. IEEE, 2019.

\bibitem{li2019multi}
Xinyu Li, Venkata Chebiyyam, and Katrin Kirchhoff.
\newblock Multi-stream network with temporal attention for environmental sound
  classification.
\newblock {\em arXiv preprint arXiv:1901.08608}, 2019.

\bibitem{patrick_verga;emma_strubell;andrew_mccallum_simultaneously_2018}
Patrick Verga, Emma Strubell, and Andrew McCallum.
\newblock Simultaneously self-attending to all mentions for full-abstract
  biological relation extraction.
\newblock {\em arXiv preprint arXiv:1802.10569}, 2018.

\bibitem{xiaoyu_guo;hui_zhang;haijun_yang;lianyuan_xu;zhiwen_ye_single_2019}
Xiaoyu Guo, Hui Zhang, Haijun Yang, Lianyuan Xu, and Zhiwen Ye.
\newblock A single attention-based combination of cnn and rnn for relation
  classification.
\newblock {\em IEEE Access}, 7:12467--12475, 2019.

\bibitem{peng_zhou_wei_shi_jun_tian_zhenyu_qi_bingchen_li_hongwei_hao_bo_xu:_attention-based_nodate}
Peng Zhou, Wei Shi, Jun Tian, Zhenyu Qi, Bingchen Li, Hongwei Hao, and Bo~Xu.
\newblock Attention-based bidirectional long short-term memory networks for
  relation classification.
\newblock In {\em Proceedings of the 54th annual meeting of the association for
  computational linguistics (volume 2: Short papers)}, pages 207--212, 2016.

\bibitem{yankai_lin_shiqi_shen_zhiyuan_liu_huanbo_luan_maosong_sun:_neural_nodate}
Yankai Lin, Shiqi Shen, Zhiyuan Liu, Huanbo Luan, and Maosong Sun.
\newblock Neural relation extraction with selective attention over instances.
\newblock In {\em Proceedings of the 54th Annual Meeting of the Association for
  Computational Linguistics (Volume 1: Long Papers)}, pages 2124--2133, 2016.

\bibitem{yuhao_zhang;victor_zhong;danqi_chen;gabor_angeli;christopher_d._manning_position-aware_2017}
Yuhao Zhang, Victor Zhong, Danqi Chen, Gabor Angeli, and Christopher~D Manning.
\newblock Position-aware attention and supervised data improve slot filling.
\newblock In {\em Proceedings of the 2017 Conference on Empirical Methods in
  Natural Language Processing}, pages 35--45, 2017.

\bibitem{chen2019bert}
Qian Chen, Zhu Zhuo, and Wen Wang.
\newblock Bert for joint intent classification and slot filling.
\newblock {\em arXiv preprint arXiv:1902.10909}, 2019.

\bibitem{choi2019aila}
Minsuk Choi, Cheonbok Park, Soyoung Yang, Yonggyu Kim, Jaegul Choo, and
  Sungsoo~Ray Hong.
\newblock Aila: Attentive interactive labeling assistant for document
  classification through attention-based deep neural networks.
\newblock In {\em Proceedings of the 2019 CHI Conference on Human Factors in
  Computing Systems}, pages 1--12, 2019.

\bibitem{kong2018audio}
Qiuqiang Kong, Yong Xu, Wenwu Wang, and Mark~D Plumbley.
\newblock Audio set classification with attention model: A probabilistic
  perspective.
\newblock In {\em 2018 IEEE International Conference on Acoustics, Speech and
  Signal Processing (ICASSP)}, pages 316--320. IEEE, 2018.

\bibitem{devlin2018bert}
Jacob Devlin, Ming-Wei Chang, Kenton Lee, and Kristina Toutanova.
\newblock Bert: Pre-training of deep bidirectional transformers for language
  understanding.
\newblock {\em arXiv preprint arXiv:1810.04805}, 2018.

\bibitem{alt2019improving}
Christoph Alt, Marc H{\"u}bner, and Leonhard Hennig.
\newblock Improving relation extraction by pre-trained language
  representations.
\newblock {\em arXiv preprint arXiv:1906.03088}, 2019.

\bibitem{yasuda2019investigation}
Yusuke Yasuda, Xin Wang, Shinji Takaki, and Junichi Yamagishi.
\newblock Investigation of enhanced tacotron text-to-speech synthesis systems
  with self-attention for pitch accent language.
\newblock In {\em ICASSP 2019-2019 IEEE International Conference on Acoustics,
  Speech and Signal Processing (ICASSP)}, pages 6905--6909. IEEE, 2019.

\bibitem{zhang2019joint}
Mingyang Zhang, Xin Wang, Fuming Fang, Haizhou Li, and Junichi Yamagishi.
\newblock Joint training framework for text-to-speech and voice conversion
  using multi-source tacotron and wavenet.
\newblock {\em arXiv preprint arXiv:1903.12389}, 2019.

\bibitem{kuncoro2016recurrent}
Adhiguna Kuncoro, Miguel Ballesteros, Lingpeng Kong, Chris Dyer, Graham Neubig,
  and Noah~A Smith.
\newblock What do recurrent neural network grammars learn about syntax?
\newblock {\em arXiv preprint arXiv:1611.05774}, 2016.

\bibitem{sperber2019attention}
Matthias Sperber, Graham Neubig, Jan Niehues, and Alex Waibel.
\newblock Attention-passing models for robust and data-efficient end-to-end
  speech translation.
\newblock {\em Transactions of the Association for Computational Linguistics},
  7:313--325, 2019.

\bibitem{tachibana2018efficiently}
Hideyuki Tachibana, Katsuya Uenoyama, and Shunsuke Aihara.
\newblock Efficiently trainable text-to-speech system based on deep
  convolutional networks with guided attention.
\newblock In {\em 2018 IEEE International Conference on Acoustics, Speech and
  Signal Processing (ICASSP)}, pages 4784--4788. IEEE, 2018.

\bibitem{mensch2018differentiable}
Arthur Mensch and Mathieu Blondel.
\newblock Differentiable dynamic programming for structured prediction and
  attention.
\newblock {\em arXiv preprint arXiv:1802.03676}, 2018.

\bibitem{zhang2018multiresolution}
Ting Zhang, Bang Liu, Di~Niu, Kunfeng Lai, and Yu~Xu.
\newblock Multiresolution graph attention networks for relevance matching.
\newblock In {\em Proceedings of the 27th ACM International Conference on
  Information and Knowledge Management}, pages 933--942, 2018.

\bibitem{zhang2019long}
Ningyu Zhang, Shumin Deng, Zhanlin Sun, Guanying Wang, Xi~Chen, Wei Zhang, and
  Huajun Chen.
\newblock Long-tail relation extraction via knowledge graph embeddings and
  graph convolution networks.
\newblock {\em arXiv preprint arXiv:1903.01306}, 2019.

\bibitem{yiming_cui_zhipeng_chen_si_wei_shijin_wang_ting_liu_guoping_hu:_attention-over-attention_nodate}
Yiming Cui, Zhipeng Chen, Si~Wei, Shijin Wang, Ting Liu, and Guoping Hu.
\newblock Attention-over-attention neural networks for reading comprehension.
\newblock In {\em Proceedings of the 55th Annual Meeting of the Association for
  Computational Linguistics (Volume 1: Long Papers)}, pages 593--602, 2017.

\bibitem{s._liu;_s._zhang;_x._zhang;_h._wang_r-trans:_2019}
Shanshan Liu, Sheng Zhang, Xin Zhang, and Hui Wang.
\newblock R-trans: Rnn transformer network for chinese machine reading
  comprehension.
\newblock {\em IEEE Access}, 7:27736--27745, 2019.

\bibitem{yiming_cui;ting_liu;zhipeng_chen;shijin_wang;guoping_hu_consensus_2018}
Yiming Cui, Ting Liu, Zhipeng Chen, Shijin Wang, and Guoping Hu.
\newblock Consensus attention-based neural networks for chinese reading
  comprehension.
\newblock {\em arXiv preprint arXiv:1607.02250}, 2016.

\bibitem{kim2018efficient}
Young-Bum Kim, Dongchan Kim, Anjishnu Kumar, and Ruhi Sarikaya.
\newblock Efficient large-scale domain classification with personalized
  attention.
\newblock {\em arXiv preprint arXiv:1804.08065}, 2018.

\bibitem{grefenstette2015learning}
Edward Grefenstette, Karl~Moritz Hermann, Mustafa Suleyman, and Phil Blunsom.
\newblock Learning to transduce with unbounded memory.
\newblock In {\em Advances in neural information processing systems}, pages
  1828--1836, 2015.

\bibitem{yang2019leveraging}
Bishan Yang and Tom Mitchell.
\newblock Leveraging knowledge bases in lstms for improving machine reading.
\newblock {\em arXiv preprint arXiv:1902.09091}, 2019.

\bibitem{liu2016attention}
Bing Liu and Ian Lane.
\newblock Attention-based recurrent neural network models for joint intent
  detection and slot filling.
\newblock {\em arXiv preprint arXiv:1609.01454}, 2016.

\bibitem{ahmadi2018attention}
Sina Ahmadi.
\newblock Attention-based encoder-decoder networks for spelling and grammatical
  error correction.
\newblock {\em arXiv preprint arXiv:1810.00660}, 2018.

\bibitem{das2016chains}
Rajarshi Das, Arvind Neelakantan, David Belanger, and Andrew McCallum.
\newblock Chains of reasoning over entities, relations, and text using
  recurrent neural networks.
\newblock {\em arXiv preprint arXiv:1607.01426}, 2016.

\bibitem{ganea2017deep}
Octavian-Eugen Ganea and Thomas Hofmann.
\newblock Deep joint entity disambiguation with local neural attention.
\newblock {\em arXiv preprint arXiv:1704.04920}, 2017.

\bibitem{lin2017structured}
Zhouhan Lin, Minwei Feng, Cicero Nogueira~dos Santos, Mo~Yu, Bing Xiang, Bowen
  Zhou, and Yoshua Bengio.
\newblock A structured self-attentive sentence embedding.
\newblock {\em arXiv preprint arXiv:1703.03130}, 2017.

\bibitem{schick2019attentive}
Timo Schick and Hinrich Sch{\"u}tze.
\newblock Attentive mimicking: Better word embeddings by attending to
  informative contexts.
\newblock {\em arXiv preprint arXiv:1904.01617}, 2019.

\bibitem{zhu2018self}
Yingke Zhu, Tom Ko, David Snyder, Brian Mak, and Daniel Povey.
\newblock Self-attentive speaker embeddings for text-independent speaker
  verification.
\newblock In {\em Interspeech}, pages 3573--3577, 2018.

\bibitem{dozat2016deep}
Timothy Dozat and Christopher~D Manning.
\newblock Deep biaffine attention for neural dependency parsing.
\newblock {\em arXiv preprint arXiv:1611.01734}, 2016.

\bibitem{strubell2018linguistically}
Emma Strubell, Patrick Verga, Daniel Andor, David Weiss, and Andrew McCallum.
\newblock Linguistically-informed self-attention for semantic role labeling.
\newblock {\em arXiv preprint arXiv:1804.08199}, 2018.

\bibitem{zhou2018commonsense}
Hao Zhou, Tom Young, Minlie Huang, Haizhou Zhao, Jingfang Xu, and Xiaoyan Zhu.
\newblock Commonsense knowledge aware conversation generation with graph
  attention.
\newblock In {\em IJCAI}, pages 4623--4629, 2018.

\bibitem{zhang2019sequence}
Jing-Xuan Zhang, Zhen-Hua Ling, Li-Juan Liu, Yuan Jiang, and Li-Rong Dai.
\newblock Sequence-to-sequence acoustic modeling for voice conversion.
\newblock {\em IEEE/ACM Transactions on Audio, Speech, and Language
  Processing}, 27(3):631--644, 2019.

\bibitem{sun2018automatic}
Bo~Sun, Yunzong Zhu, Yongkang Xiao, Rong Xiao, and Yungang Wei.
\newblock Automatic question tagging with deep neural networks.
\newblock {\em IEEE Transactions on Learning Technologies}, 12(1):29--43, 2018.

\bibitem{sharma_action_2015}
Shikhar Sharma, Ryan Kiros, and Ruslan Salakhutdinov.
\newblock Action recognition using visual attention.
\newblock {\em arXiv preprint arXiv:1511.04119}, 2015.

\bibitem{girdhar2019video}
Rohit Girdhar, Joao Carreira, Carl Doersch, and Andrew Zisserman.
\newblock Video action transformer network.
\newblock In {\em Proceedings of the IEEE/CVF Conference on Computer Vision and
  Pattern Recognition}, pages 244--253, 2019.

\bibitem{girdhar_attentional_2017}
Rohit Girdhar and Deva Ramanan.
\newblock Attentional pooling for action recognition.
\newblock {\em arXiv preprint arXiv:1711.01467}, 2017.

\bibitem{song_end--end_2016}
Sijie Song, Cuiling Lan, Junliang Xing, Wenjun Zeng, and Jiaying Liu.
\newblock An end-to-end spatio-temporal attention model for human action
  recognition from skeleton data.
\newblock In {\em Proceedings of the AAAI conference on artificial
  intelligence}, 2017.

\bibitem{liu_global_2017}
Jun Liu, Gang Wang, Ping Hu, Ling-Yu Duan, and Alex~C. Kot.
\newblock Global context-aware attention lstm networks for 3d action
  recognition.
\newblock In {\em The {IEEE} {Conference} on {Computer} {Vision} and {Pattern}
  {Recognition} ({CVPR})}, July 2017.

\bibitem{bazzani2016recurrent}
Loris Bazzani, Hugo Larochelle, and Lorenzo Torresani.
\newblock Recurrent mixture density network for spatiotemporal visual
  attention.
\newblock {\em arXiv preprint arXiv:1603.08199}, 2016.

\bibitem{li2018videolstm}
Zhenyang Li, Kirill Gavrilyuk, Efstratios Gavves, Mihir Jain, and Cees~GM
  Snoek.
\newblock Videolstm convolves, attends and flows for action recognition.
\newblock {\em Computer Vision and Image Understanding}, 166:41--50, 2018.

\bibitem{ma_attend_2017}
Chih-Yao Ma, Asim Kadav, Iain Melvin, Zsolt Kira, Ghassan AlRegib, and
  Hans~Peter Graf.
\newblock Attend and interact: Higher-order object interactions for video
  understanding.
\newblock {\em arXiv:1711.06330 [cs]}, November 2017.
\newblock arXiv: 1711.06330.

\bibitem{wang_eidetic_2018}
Yunbo Wang, Lu~Jiang, Ming-Hsuan Yang, Li-Jia Li, Mingsheng Long, and
  Li~Fei-Fei.
\newblock Eidetic 3d lstm: A model for video prediction and beyond.
\newblock In {\em International conference on learning representations}, 2018.

\bibitem{dong_li;ting_yao;ling-yu_duan;tao_mei;yong_rui_unified_2019}
Dong Li, Ting Yao, Ling-Yu Duan, Tao Mei, and Yong Rui.
\newblock Unified spatio-temporal attention networks for action recognition in
  videos.
\newblock {\em IEEE Transactions on Multimedia}, 21(2):416--428, 2018.

\bibitem{p._zhang;_c._lan;_j._xing;_w._zeng;_j._xue;_n._zheng_view_nodate}
Pengfei Zhang, Cuiling Lan, Junliang Xing, Wenjun Zeng, Jianru Xue, and Nanning
  Zheng.
\newblock View adaptive recurrent neural networks for high performance human
  action recognition from skeleton data.
\newblock In {\em Proceedings of the IEEE International Conference on Computer
  Vision}, pages 2117--2126, 2017.

\bibitem{wu_zheng;lin_li;zhaoxiang_zhang;yan_huang;liang_wang_relational_2019}
Wu~Zheng, Lin Li, Zhaoxiang Zhang, Yan Huang, and Liang Wang.
\newblock Relational network for skeleton-based action recognition.
\newblock In {\em 2019 IEEE International Conference on Multimedia and Expo
  (ICME)}, pages 826--831. IEEE, 2019.

\bibitem{harshala_gammulle;simon_denman;sridha_sridharan;clinton_fookes_two_2017}
Harshala Gammulle, Simon Denman, Sridha Sridharan, and Clinton Fookes.
\newblock Two stream lstm: A deep fusion framework for human action
  recognition.
\newblock In {\em 2017 IEEE Winter Conference on Applications of Computer
  Vision (WACV)}, pages 177--186. IEEE, 2017.

\bibitem{zhaoxuan_fan;xu_zhao;tianwei_lin;haisheng_su_attention-based_2019}
Zhaoxuan Fan, Xu~Zhao, Tianwei Lin, and Haisheng Su.
\newblock Attention-based multiview re-observation fusion network for skeletal
  action recognition.
\newblock {\em IEEE Transactions on Multimedia}, 21(2):363--374, 2018.

\bibitem{chenyang_si;wentao_chen;wei_wang;liang_wang;tieniu_tan_attention_2019}
Chenyang Si, Wentao Chen, Wei Wang, Liang Wang, and Tieniu Tan.
\newblock An attention enhanced graph convolutional lstm network for
  skeleton-based action recognition.
\newblock In {\em Proceedings of the IEEE/CVF Conference on Computer Vision and
  Pattern Recognition}, pages 1227--1236, 2019.

\bibitem{swathikiran_sudhakaran;sergio_escalera;oswald_lanz_lsta:_2019}
Swathikiran Sudhakaran, Sergio Escalera, and Oswald Lanz.
\newblock Lsta: Long short-term attention for egocentric action recognition.
\newblock In {\em Proc. of the IEEE/CVF CVPR}, pages 9954--9963, 2019.

\bibitem{noauthor_skeleton_nodate}
Jun Liu, Gang Wang, Ling-Yu Duan, Kamila Abdiyeva, and Alex~C Kot.
\newblock Skeleton-based human action recognition with global context-aware
  attention lstm networks.
\newblock {\em IEEE Transactions on Image Processing}, 27(4):1586--1599, 2017.

\bibitem{liu_decidenet:_2018}
Jiang Liu, Chenqiang Gao, Deyu Meng, and Alexander~G Hauptmann.
\newblock Decidenet: Counting varying density crowds through attention guided
  detection and density estimation.
\newblock In {\em Proceedings of the IEEE Conference on Computer Vision and
  Pattern Recognition}, pages 5197--5206, 2018.

\bibitem{lingbo_liu;hongjun_wang;guanbin_li;wanli_ouyang;liang_lin_crowd_2018}
Lingbo Liu, Hongjun Wang, Guanbin Li, Wanli Ouyang, and Liang Lin.
\newblock Crowd counting using deep recurrent spatial-aware network.
\newblock {\em arXiv preprint arXiv:1807.00601}, 2018.

\bibitem{mohammad_asiful_hossain;mehrdad_hosseinzadeh;omit_chanda;yang_wang_crowd_2019}
Mohammad Hossain, Mehrdad Hosseinzadeh, Omit Chanda, and Yang Wang.
\newblock Crowd counting using scale-aware attention networks.
\newblock In {\em 2019 IEEE Winter Conference on Applications of Computer
  Vision (WACV)}, pages 1280--1288. IEEE, 2019.

\bibitem{youmei_zhang_chunluan_zhou_faliang_chang_alex_c._kot:_multi-resolution_nodate}
Youmei Zhang, Chunluan Zhou, Faliang Chang, and Alex~C Kot.
\newblock Multi-resolution attention convolutional neural network for crowd
  counting.
\newblock {\em Neurocomputing}, 329:144--152, 2019.

\bibitem{serra_overcoming_2018}
Joan Serra, Didac Suris, Marius Miron, and Alexandros Karatzoglou.
\newblock Overcoming catastrophic forgetting with hard attention to the task.
\newblock In {\em International Conference on Machine Learning}, pages
  4548--4557. PMLR, 2018.

\bibitem{han_attribute-aware_2019}
Kai Han, Jianyuan Guo, Chao Zhang, and Mingjian Zhu.
\newblock Attribute-aware attention model for fine-grained representation
  learning.
\newblock {\em arXiv:1901.00392 [cs]}, January 2019.
\newblock arXiv: 1901.00392.

\bibitem{sermanet_attention_2014}
Pierre Sermanet, Andrea Frome, and Esteban Real.
\newblock Attention for fine-grained categorization.
\newblock {\em arXiv preprint arXiv:1412.7054}, 2014.

\bibitem{kang_deep_2018}
Guoliang Kang, Liang Zheng, Yan Yan, and Yi~Yang.
\newblock Deep adversarial attention alignment for unsupervised domain
  adaptation: the benefit of target expectation maximization.
\newblock {\em arXiv:1801.10068 [cs]}, January 2018.
\newblock arXiv: 1801.10068.

\bibitem{wang_residual_2017}
Fei Wang, Mengqing Jiang, Chen Qian, Shuo Yang, Cheng Li, Honggang Zhang,
  Xiaogang Wang, and Xiaoou Tang.
\newblock Residual attention network for image classification.
\newblock In {\em Proceedings of the IEEE conference on computer vision and
  pattern recognition}, pages 3156--3164, 2017.

\bibitem{seo_hierarchical_2016}
Paul~Hongsuck Seo, Zhe Lin, Scott Cohen, Xiaohui Shen, and Bohyung Han.
\newblock Hierarchical attention networks.
\newblock {\em CoRR, abs/1606.02393}, 2016.

\bibitem{z._wang;_t._chen;_g._li;_r._xu;_l._lin_multi-label_nodate}
Zhouxia Wang, Tianshui Chen, Guanbin Li, Ruijia Xu, and Liang Lin.
\newblock Multi-label image recognition by recurrently discovering attentional
  regions.
\newblock In {\em Proc. of the IEEE ICCV}, pages 464--472, 2017.

\bibitem{irwan_bello;barret_zoph;ashish_vaswani;jonathon_shlens;quoc_v._le_attention_2019}
Irwan Bello, Barret Zoph, Ashish Vaswani, Jonathon Shlens, and Quoc~V Le.
\newblock Attention augmented convolutional networks.
\newblock In {\em Proceedings of the IEEE/CVF International Conference on
  Computer Vision}, pages 3286--3295, 2019.

\bibitem{lin_wu;yang_wang;xue_li;junbin_gao_deep_2019}
Lin Wu, Yang Wang, Xue Li, and Junbin Gao.
\newblock Deep attention-based spatially recursive networks for fine-grained
  visual recognition.
\newblock {\em IEEE transactions on cybernetics}, 49(5):1791--1802, 2018.

\bibitem{bo_zhao;xiao_wu;jiashi_feng;qiang_peng;shuicheng_yan_diversified_2017}
Bo~Zhao, Xiao Wu, Jiashi Feng, Qiang Peng, and Shuicheng Yan.
\newblock Diversified visual attention networks for fine-grained object
  classification.
\newblock {\em IEEE Transactions on Multimedia}, 19(6):1245--1256, 2017.

\bibitem{han-jia_ye;hexiang_hu;de-chuan_zhan;fei_sha_learning_2019}
Han-Jia Ye, Hexiang Hu, De-Chuan Zhan, and Fei Sha.
\newblock Learning embedding adaptation for few-shot learning.
\newblock {\em arXiv preprint arXiv:1812.03664}, 7, 2018.

\bibitem{emanuele_pesce;petros-pavlos_ypsilantis;samuel_withey;robert_bakewell;vicky_goh;giovanni_montana_learning_2019}
E~Pesce, P~Ypsilantis, S~Withey, R~Bakewell, V~Goh, and G~Montana.
\newblock Learning to detect chest radiographs containing lung nodules using
  visual attention networks. corr.
\newblock {\em arXiv preprint arXiv:1712.00996}, 2017.

\bibitem{bin_zhao;xuelong_li;xiaoqiang_lu;zhigang_wang_cnn-rnn_2019}
Bin Zhao, Xuelong Li, Xiaoqiang Lu, and Zhigang Wang.
\newblock A cnn--rnn architecture for multi-label weather recognition.
\newblock {\em Neurocomputing}, 322:47--57, 2018.

\bibitem{q._yin;_j._wang;_x._luo;_j._zhai;_s._k._jha;_y._shi_quaternion_2019}
Qilin Yin, Jinwei Wang, Xiangyang Luo, Jiangtao Zhai, Sunil~Kr Jha, and
  Yun-Qing Shi.
\newblock Quaternion convolutional neural network for color image
  classification and forensics.
\newblock {\em IEEE Access}, 7:20293--20301, 2019.

\bibitem{tianjun_xiao;_yichong_xu;_kuiyuan_yang;_jiaxing_zhang;_yuxin_peng;_z._zhang_application_nodate}
Tianjun Xiao, Yichong Xu, Kuiyuan Yang, Jiaxing Zhang, Yuxin Peng, and Zheng
  Zhang.
\newblock The application of two-level attention models in deep convolutional
  neural network for fine-grained image classification.
\newblock In {\em Proc. of the IEEE CVPR}, pages 842--850, 2015.

\bibitem{hiroshi_fukui;tsubasa_hirakawa;takayoshi_yamashita;hironobu_fujiyoshi_attention_2019}
Hiroshi Fukui, Tsubasa Hirakawa, Takayoshi Yamashita, and Hironobu Fujiyoshi.
\newblock Attention branch network: Learning of attention mechanism for visual
  explanation.
\newblock In {\em Proceedings of the IEEE/CVF Conference on Computer Vision and
  Pattern Recognition}, pages 10705--10714, 2019.

\bibitem{hailin_hu;an_xiao;sai_zhang;yangyang_li;xuanling_shi;tao_jiang;linqi_zhang;lei_zhang;jianyang_zeng_deephint:_2019}
Hailin Hu, An~Xiao, Sai Zhang, Yangyang Li, Xuanling Shi, Tao Jiang, Linqi
  Zhang, Lei Zhang, and Jianyang Zeng.
\newblock Deephint: understanding hiv-1 integration via deep learning with
  attention.
\newblock {\em Bioinformatics}, 35(10):1660--1667, 2019.

\bibitem{qingji_guan;yaping_huang;zhun_zhong;zhedong_zheng;liang_zheng;yi_yang_diagnose_2018}
Qingji Guan, Yaping Huang, Zhun Zhong, Zhedong Zheng, Liang Zheng, and Yi~Yang.
\newblock Diagnose like a radiologist: Attention guided convolutional neural
  network for thorax disease classification.
\newblock {\em arXiv preprint arXiv:1801.09927}, 2018.

\bibitem{jianming_zhang;sarah_adel_bargal;zhe_lin;jonathan_br;t;xiaohui_shen;stan_sclaroff_top-down_2018}
Jianming Zhang, Sarah~Adel Bargal, Zhe Lin, Jonathan Brandt, Xiaohui Shen, and
  Stan Sclaroff.
\newblock Top-down neural attention by excitation backprop.
\newblock {\em International Journal of Computer Vision}, 126(10):1084--1102,
  2018.

\bibitem{w._wang;_w._wang;_y._xu;_j._shen;_s._zhu_attentive_2018}
Wenguan Wang, Yuanlu Xu, Jianbing Shen, and Song-Chun Zhu.
\newblock Attentive fashion grammar network for fashion landmark detection and
  clothing category classification.
\newblock In {\em Proceedings of the IEEE Conference on Computer Vision and
  Pattern Recognition}, pages 4271--4280, 2018.

\bibitem{wonsik_kim;bhavya_goyal;kunal_chawla;jungmin_lee;keunjoo_kwon_attention-based_2018}
Wonsik Kim, Bhavya Goyal, Kunal Chawla, Jungmin Lee, and Keunjoo Kwon.
\newblock Attention-based ensemble for deep metric learning.
\newblock In {\em Proceedings of the European Conference on Computer Vision
  (ECCV)}, pages 736--751, 2018.

\bibitem{mengye_ren;renjie_liao;ethan_fetaya;richard_s._zemel_incremental_2019}
Mengye Ren, Renjie Liao, Ethan Fetaya, and Richard Zemel.
\newblock Incremental few-shot learning with attention attractor networks.
\newblock In {\em Advances in Neural Information Processing Systems}, pages
  5275--5285, 2019.

\bibitem{yuxin_peng;xiangteng_he;junjie_zhao_object-part_2018}
Yuxin Peng, Xiangteng He, and Junjie Zhao.
\newblock Object-part attention model for fine-grained image classification.
\newblock {\em IEEE Transactions on Image Processing}, 27(3):1487--1500, 2017.

\bibitem{hazel_doughty;walterio_mayol-cuevas;dima_damen_pros_2019}
Hazel Doughty, Walterio Mayol-Cuevas, and Dima Damen.
\newblock The pros and cons: Rank-aware temporal attention for skill
  determination in long videos.
\newblock In {\em Proceedings of the IEEE/CVF Conference on Computer Vision and
  Pattern Recognition}, pages 7862--7871, 2019.

\bibitem{noauthor_object-part_nodate}
Yuxin Peng, Xiangteng He, and Junjie Zhao.
\newblock Object-part attention driven discriminative localization for
  fine-grained image classification.
\newblock {\em CoRR}, abs/1704.01740, 2017.

\bibitem{kastaniotis_attention-aware_2018}
Dimitris Kastaniotis, Ioanna Ntinou, Dimitrios Tsourounis, George Economou, and
  Spiros Fotopoulos.
\newblock Attention-aware generative adversarial networks (ata-gans).
\newblock In {\em 2018 IEEE 13th Image, Video, and Multidimensional Signal
  Processing Workshop (IVMSP)}, pages 1--5. IEEE, 2018.

\bibitem{yu_generative_2018}
Jiahui Yu, Zhe Lin, Jimei Yang, Xiaohui Shen, Xin Lu, and Thomas~S. Huang.
\newblock Generative image inpainting with contextual attention.
\newblock {\em arXiv:1801.07892 [cs]}, January 2018.
\newblock arXiv: 1801.07892.

\bibitem{reed_few-shot_2017}
Scott Reed, Yutian Chen, Thomas Paine, Aäron van~den Oord, S.~M.~Ali Eslami,
  Danilo Rezende, Oriol Vinyals, and Nando de~Freitas.
\newblock Few-shot autoregressive density estimation: Towards learning to learn
  distributions.
\newblock {\em arXiv:1710.10304 [cs]}, October 2017.
\newblock arXiv: 1710.10304.

\bibitem{xinyuan_chen;chang_xu;xiaokang_yang;dacheng_tao_attention-gan_2018}
Xinyuan Chen, Chang Xu, Xiaokang Yang, and Dacheng Tao.
\newblock Attention-gan for object transfiguration in wild images.
\newblock In {\em Proceedings of the European Conference on Computer Vision
  (ECCV)}, pages 164--180, 2018.

\bibitem{jianan_li;jimei_yang;aaron_hertzmann;jianming_zhang;tingfa_xu_layoutgan:_2019}
Jianan Li, Jimei Yang, Aaron Hertzmann, Jianming Zhang, and Tingfa Xu.
\newblock Layoutgan: Generating graphic layouts with wireframe discriminators.
\newblock {\em arXiv preprint arXiv:1901.06767}, 2019.

\bibitem{hao_tang;dan_xu;nicu_sebe;yanzhi_wang;jason_j._corso;yan_yan_multi-channel_2019}
Hao Tang, Dan Xu, Nicu Sebe, Yanzhi Wang, Jason~J Corso, and Yan Yan.
\newblock Multi-channel attention selection gan with cascaded semantic guidance
  for cross-view image translation.
\newblock In {\em Proceedings of the IEEE/CVF Conference on Computer Vision and
  Pattern Recognition}, pages 2417--2426, 2019.

\bibitem{q._chu;_w._ouyang;_h._li;_x._wang;_b._liu;_n._yu_online_nodate}
Qi~Chu, Wanli Ouyang, Hongsheng Li, Xiaogang Wang, Bin Liu, and Nenghai Yu.
\newblock Online multi-object tracking using cnn-based single object tracker
  with spatial-temporal attention mechanism.
\newblock In {\em Proceedings of the IEEE International Conference on Computer
  Vision}, pages 4836--4845, 2017.

\bibitem{zhedong_zheng;liang_zheng;yi_yang_pedestrian_2017}
Zhedong Zheng, Liang Zheng, and Yi~Yang.
\newblock Pedestrian alignment network for large-scale person
  re-identification.
\newblock {\em IEEE Transactions on Circuits and Systems for Video Technology},
  29(10):3037--3045, 2018.

\bibitem{x._liu;_h._zhao;_m._tian;_l._sheng;_j._shao;_s._yi;_j._yan;_x._wang_hydraplus-net:_nodate}
Xihui Liu, Haiyu Zhao, Maoqing Tian, Lu~Sheng, Jing Shao, Shuai Yi, Junjie Yan,
  and Xiaogang Wang.
\newblock Hydraplus-net: Attentive deep features for pedestrian analysis.
\newblock In {\em Proceedings of the IEEE international conference on computer
  vision}, pages 350--359, 2017.

\bibitem{he;_anfeng;luo;_chong;tian;_xinmei;zeng;_wenjun_twofold_2018}
Anfeng He, Chong Luo, Xinmei Tian, and Wenjun Zeng.
\newblock A twofold siamese network for real-time object tracking.
\newblock In {\em Proceedings of the IEEE Conference on Computer Vision and
  Pattern Recognition}, pages 4834--4843, 2018.

\bibitem{xinlei_chen;li-jia_li;li_fei-fei;abhinav_gupta_iterative_2018}
Xinlei Chen, Li-Jia Li, Li~Fei-Fei, and Abhinav Gupta.
\newblock Iterative visual reasoning beyond convolutions.
\newblock In {\em Proceedings of the IEEE Conference on Computer Vision and
  Pattern Recognition}, pages 7239--7248, 2018.

\bibitem{jianlou_si;honggang_zhang;chun-guang_li;jason_kuen;xiangfei_kong;alex_c._kot;gang_wang_dual_2018}
Jianlou Si, Honggang Zhang, Chun-Guang Li, Jason Kuen, Xiangfei Kong, Alex~C
  Kot, and Gang Wang.
\newblock Dual attention matching network for context-aware feature sequence
  based person re-identification.
\newblock In {\em Proceedings of the IEEE CVPR}, pages 5363--5372, 2018.

\bibitem{chunfeng_song;yan_huang;wanli_ouyang;liang_wang_mask-guided_2018}
Chunfeng Song, Yan Huang, Wanli Ouyang, and Liang Wang.
\newblock Mask-guided contrastive attention model for person re-identification.
\newblock In {\em Proc. of the IEEE CVPR}, pages 1179--1188, 2018.

\bibitem{yi_zhou;ling_shao_viewpoint-aware_2018}
Yi~Zhou and Ling Shao.
\newblock Aware attentive multi-view inference for vehicle re-identification.
\newblock In {\em Proc. of the IEEE CVPR}, pages 6489--6498, 2018.

\bibitem{zhang_progressive_2018}
Xiaoning Zhang, Tiantian Wang, Jinqing Qi, Huchuan Lu, and Gang Wang.
\newblock Progressive attention guided recurrent network for salient object
  detection.
\newblock In {\em {IEEE} CVPR)}, June 2018.

\bibitem{tao_kong;fuchun_sun;wenbing_huang;huaping_liu_deep_2018}
Tao Kong, Fuchun Sun, Chuanqi Tan, Huaping Liu, and Wenbing Huang.
\newblock Deep feature pyramid reconfiguration for object detection.
\newblock In {\em Proceedings of the European conference on computer vision
  (ECCV)}, pages 169--185, 2018.

\bibitem{guanbin_li;yukang_gan;hejun_wu;nong_xiao;liang_lin_cross-modal_2019}
Guanbin Li, Yukang Gan, Hejun Wu, Nong Xiao, and Liang Lin.
\newblock Cross-modal attentional context learning for rgb-d object detection.
\newblock {\em IEEE Transactions on Image Processing}, 28(4):1591--1601, 2018.

\bibitem{shuhan_chen;xiuli_tan;ben_wang;xuelong_hu_reverse_2019}
Shuhan Chen, Xiuli Tan, Ben Wang, and Xuelong Hu.
\newblock Reverse attention for salient object detection.
\newblock In {\em Proceedings of the European Conference on Computer Vision
  (ECCV)}, pages 234--250, 2018.

\bibitem{hao_chen;youfu_li_three-stream_2019}
Hao Chen and Youfu Li.
\newblock Three-stream attention-aware network for rgb-d salient object
  detection.
\newblock {\em IEEE Transactions on Image Processing}, 28(6):2825--2835, 2019.

\bibitem{xudong_wang;zhaowei_cai;dashan_gao;nuno_vasconcelos_towards_2019}
Xudong Wang, Zhaowei Cai, Dashan Gao, and Nuno Vasconcelos.
\newblock Towards universal object detection by domain attention.
\newblock In {\em Proceedings of the IEEE/CVF Conference on Computer Vision and
  Pattern Recognition}, pages 7289--7298, 2019.

\bibitem{li_harmonious_2018}
Wei Li, Xiatian Zhu, and Shaogang Gong.
\newblock Harmonious attention network for person re-identification.
\newblock In {\em The {IEEE} {Conference} on {Computer} {Vision} and {Pattern}
  {Recognition} ({CVPR})}, June 2018.

\bibitem{hao_liu;jiashi_feng;meibin_qi;jianguo_jiang;shuicheng_yan_end--end_2017}
Hao Liu, Jiashi Feng, Meibin Qi, Jianguo Jiang, and Shuicheng Yan.
\newblock End-to-end comparative attention networks for person
  re-identification.
\newblock {\em IEEE Transactions on Image Processing}, 26(7):3492--3506, 2017.

\bibitem{xingyu_liao;lingxiao_he;zhouwang_yang;chi_zhang_video-based_2019}
Xingyu Liao, Lingxiao He, Zhouwang Yang, and Chi Zhang.
\newblock Video-based person re-identification via 3d convolutional networks
  and non-local attention.
\newblock In {\em Asian Conference on Computer Vision}, pages 620--634.
  Springer, 2018.

\bibitem{d._chen;_h._li;_t._xiao;_s._yi;_x._wang_video_2018}
Dapeng Chen, Hongsheng Li, Tong Xiao, Shuai Yi, and Xiaogang Wang.
\newblock Video person re-identification with competitive snippet-similarity
  aggregation and co-attentive snippet embedding.
\newblock In {\em Proceedings of the IEEE Conference on Computer Vision and
  Pattern Recognition}, pages 1169--1178, 2018.

\bibitem{l._zhao;_x._li;_y._zhuang;_j._wang_deeply-learned_nodate}
Liming Zhao, Xi~Li, Yueting Zhuang, and Jingdong Wang.
\newblock Deeply-learned part-aligned representations for person
  re-identification.
\newblock In {\em Proceedings of the IEEE international conference on computer
  vision}, pages 3219--3228, 2017.

\bibitem{z._zhou;_y._huang;_w._wang;_l._wang;_t._tan_see_nodate}
Zhen Zhou, Yan Huang, Wei Wang, Liang Wang, and Tieniu Tan.
\newblock See the forest for the trees: Joint spatial and temporal recurrent
  neural networks for video-based person re-identification.
\newblock In {\em Proceedings of the IEEE Conference on Computer Vision and
  Pattern Recognition}, pages 4747--4756, 2017.

\bibitem{jing_xu;rui_zhao;feng_zhu;huaming_wang;wanli_ouyang_attention-aware_2018}
Jing Xu, Rui Zhao, Feng Zhu, Huaming Wang, and Wanli Ouyang.
\newblock Attention-aware compositional network for person re-identification.
\newblock In {\em Proceedings of the IEEE CVPR}, pages 2119--2128, 2018.

\bibitem{shuang_li;slawomir_bak;peter_carr;xiaogang_wang_diversity_2018}
Shuang Li, Slawomir Bak, Peter Carr, and Xiaogang Wang.
\newblock Diversity regularized spatiotemporal attention for video-based person
  re-identification.
\newblock In {\em Proceedings of the IEEE Conference on Computer Vision and
  Pattern Recognition}, pages 369--378, 2018.

\bibitem{meng_zheng;srikrishna_karanam;ziyan_wu;richard_j._radke_re-identification_2019}
Meng Zheng, Srikrishna Karanam, Ziyan Wu, and Richard~J Radke.
\newblock Re-identification with consistent attentive siamese networks.
\newblock In {\em Proceedings of the IEEE/CVF Conference on Computer Vision and
  Pattern Recognition}, pages 5735--5744, 2019.

\bibitem{deqiang_ouyang;yonghui_zhang;jie_shao_video-based_2019}
Deqiang Ouyang, Yonghui Zhang, and Jie Shao.
\newblock Video-based person re-identification via spatio-temporal attentional
  and two-stream fusion convolutional networks.
\newblock {\em Pattern Recognition Letters}, 117:153--160, 2019.

\bibitem{cheng_wang;qian_zhang;chang_huang;wenyu_liu;xinggang_wang_mancs:_2018}
Cheng Wang, Qian Zhang, Chang Huang, Wenyu Liu, and Xinggang Wang.
\newblock Mancs: A multi-task attentional network with curriculum sampling for
  person re-identification.
\newblock In {\em Proceedings of the European Conference on Computer Vision
  (ECCV)}, pages 365--381, 2018.

\bibitem{noauthor_jointly_nodate}
Shuangjie Xu, Yu~Cheng, Kang Gu, Yang Yang, Shiyu Chang, and Pan Zhou.
\newblock Jointly attentive spatial-temporal pooling networks for video-based
  person re-identification.
\newblock In {\em Proceedings of the IEEE international conference on computer
  vision}, pages 4733--4742, 2017.

\bibitem{noauthor_scan_nodate}
Ruimao Zhang, Jingyu Li, Hongbin Sun, Yuying Ge, Ping Luo, Xiaogang Wang, and
  Liang Lin.
\newblock Scan: Self-and-collaborative attention network for video person
  re-identification.
\newblock {\em IEEE Transactions on Image Processing}, 28(10):4870--4882, 2019.

\bibitem{fu_dual_2018}
Jun Fu, Jing Liu, Haijie Tian, Zhiwei Fang, and Hanqing Lu.
\newblock Dual attention network for scene segmentation.
\newblock {\em arXiv:1809.02983 [cs]}, September 2018.
\newblock arXiv: 1809.02983.

\bibitem{li_tell_2018}
Kunpeng Li, Ziyan Wu, Kuan-Chuan Peng, Jan Ernst, and Yun Fu.
\newblock Tell me where to look: Guided attention inference network.
\newblock {\em arXiv:1802.10171 [cs]}, February 2018.
\newblock arXiv: 1802.10171.

\bibitem{ren_end--end_2017}
Mengye Ren and Richard~S. Zemel.
\newblock End-to-end instance segmentation with recurrent attention.
\newblock In {\em The {IEEE} {Conference} on {Computer} {Vision} and {Pattern}
  {Recognition} ({CVPR})}, July 2017.

\bibitem{zhang_deep_2019}
Pingping Zhang, Wei Liu, Hongyu Wang, Yinjie Lei, and Huchuan Lu.
\newblock Deep gated attention networks for large-scale street-level scene
  segmentation.
\newblock {\em Pattern Recognition}, 88:702--714, April 2019.

\bibitem{chen_attention_2016}
Liang-Chieh Chen, Yi~Yang, Jiang Wang, Wei Xu, and Alan~L Yuille.
\newblock Attention to scale: Scale-aware semantic image segmentation.
\newblock In {\em Proceedings of the IEEE conference on computer vision and
  pattern recognition}, pages 3640--3649, 2016.

\bibitem{jetley_learn_2018}
Saumya Jetley, Nicholas~A. Lord, Namhoon Lee, and Philip H.~S. Torr.
\newblock Learn to pay attention.
\newblock {\em arXiv:1804.02391 [cs]}, April 2018.
\newblock arXiv: 1804.02391.

\bibitem{shikun_liu;edward_johns;andrew_j._davison_end--end_2019}
Shikun Liu, Edward Johns, and Andrew~J Davison.
\newblock End-to-end multi-task learning with attention.
\newblock In {\em Proceedings of the IEEE/CVF Conference on Computer Vision and
  Pattern Recognition}, pages 1871--1880, 2019.

\bibitem{yuhui_yuan;jingdong_wang_ocnet:_2019}
Yuhui Yuan and Jingdong Wang.
\newblock Ocnet: Object context network for scene parsing.
\newblock {\em arXiv preprint arXiv:1809.00916}, 2018.

\bibitem{hanchao_li;pengfei_xiong;jie_an;lingxue_wang_pyramid_2018}
Hanchao Li, Pengfei Xiong, Jie An, and Lingxue Wang.
\newblock Pyramid attention network for semantic segmentation.
\newblock {\em arXiv preprint arXiv:1805.10180}, 2018.

\bibitem{xiaowei_hu;chi-wing_fu;lei_zhu;jing_qin;pheng-ann_heng_direction-aware_2019}
Xiaowei Hu, Lei Zhu, Chi-Wing Fu, Jing Qin, and Pheng-Ann Heng.
\newblock Direction-aware spatial context features for shadow detection.
\newblock In {\em Proceedings of the IEEE Conference on Computer Vision and
  Pattern Recognition}, pages 7454--7462, 2018.

\bibitem{z._zeng;_w._xie;_y._zhang;_y._lu_ric-unet:_2019}
Zitao Zeng, Weihao Xie, Yunzhe Zhang, and Yao Lu.
\newblock Ric-unet: An improved neural network based on unet for nuclei
  segmentation in histology images.
\newblock {\em Ieee Access}, 7:21420--21428, 2019.

\bibitem{b._shuai;_z._zuo;_b._wang;_g._wang_scene_2018}
Bing Shuai, Zhen Zuo, Bing Wang, and Gang Wang.
\newblock Scene segmentation with dag-recurrent neural networks.
\newblock {\em IEEE PAMI}, 40(6):1480--1493, 2017.

\bibitem{xinxin_hu;kailun_yang;lei_fei;kaiwei_wang_acnet:_2019}
Xinxin Hu, Kailun Yang, Lei Fei, and Kaiwei Wang.
\newblock Acnet: Attention based network to exploit complementary features for
  rgbd semantic segmentation.
\newblock In {\em 2019 IEEE International Conference on Image Processing
  (ICIP)}, pages 1440--1444. IEEE, 2019.

\bibitem{ozan_oktay;jo_schlemper;loic_le_folgoc;matthew_lee;mattias_heinrich;kazunari_misawa;kensaku_mori;steven_mcdonagh;nils_y_hammerla;bernhard_kainz;ben_glocker;daniel_rueckert_attention_2018}
Ozan Oktay, Jo~Schlemper, Loic~Le Folgoc, Matthew Lee, Mattias Heinrich,
  Kazunari Misawa, Kensaku Mori, Steven McDonagh, Nils~Y Hammerla, Bernhard
  Kainz, et~al.
\newblock Attention u-net: Learning where to look for the pancreas.
\newblock {\em arXiv preprint arXiv:1804.03999}, 2018.

\bibitem{ruirui_li;mingming_li;jiacheng_li;yating_zhou_connection_2019}
Ruirui Li, Mingming Li, Jiacheng Li, and Yating Zhou.
\newblock Connection sensitive attention u-net for accurate retinal vessel
  segmentation.
\newblock {\em arXiv preprint arXiv:1903.05558}, 2019.

\bibitem{shu_kong;charless_c._fowlkes_pixel-wise_2019}
Shu Kong and Charless Fowlkes.
\newblock Pixel-wise attentional gating for scene parsing.
\newblock In {\em 2019 IEEE Winter Conference on Applications of Computer
  Vision (WACV)}, pages 1024--1033. IEEE, 2019.

\bibitem{xiaoxiao_li;chen_change_loy_video_2018}
Xiaoxiao Li and Chen~Change Loy.
\newblock Video object segmentation with joint re-identification and
  attention-aware mask propagation.
\newblock In {\em Proceedings of the European Conference on Computer Vision
  (ECCV)}, pages 90--105, 2018.

\bibitem{hengshuang_zhao;yi_zhang;shu_liu;jianping_shi;chen_change_loy;dahua_lin;jiaya_jia_psanet:_2018}
Hengshuang Zhao, Yi~Zhang, Shu Liu, Jianping Shi, Chen~Change Loy, Dahua Lin,
  and Jiaya Jia.
\newblock Psanet: Point-wise spatial attention network for scene parsing.
\newblock In {\em Proceedings of the European Conference on Computer Vision
  (ECCV)}, pages 267--283, 2018.

\bibitem{j._kuen;_z._wang;_g._wang_recurrent_nodate}
Jason Kuen, Zhenhua Wang, and Gang Wang.
\newblock Recurrent attentional networks for saliency detection.
\newblock In {\em Proceedings of the IEEE Conference on computer Vision and
  Pattern Recognition}, pages 3668--3677, 2016.

\bibitem{nian_liu;junwei_han;ming-hsuan_yang_picanet:_2018}
Nian Liu, Junwei Han, and Ming-Hsuan Yang.
\newblock Picanet: Learning pixel-wise contextual attention for saliency
  detection.
\newblock In {\em Proceedings of the IEEE Conference on Computer Vision and
  Pattern Recognition}, pages 3089--3098, 2018.

\bibitem{marcella_cornia;lorenzo_baraldi;giuseppe_serra;rita_cucchiara_predicting_2018}
Marcella Cornia, Lorenzo Baraldi, Giuseppe Serra, and Rita Cucchiara.
\newblock Predicting human eye fixations via an lstm-based saliency attentive
  model.
\newblock {\em IEEE Transactions on Image Processing}, 27(10):5142--5154, 2018.

\bibitem{xiaowei_hu;chi-wing_fu;lei_zhu;pheng-ann_heng_sac-net:_2019}
Xiaowei Hu, Chi-Wing Fu, Lei Zhu, Tianyu Wang, and Pheng-Ann Heng.
\newblock Sac-net: Spatial attenuation context for salient object detection.
\newblock {\em IEEE Transactions on Circuits and Systems for Video Technology},
  2020.

\bibitem{he_end--end_2018}
Tong He, Zhi Tian, Weilin Huang, Chunhua Shen, Yu~Qiao, and Changming Sun.
\newblock An end-to-end textspotter with explicit alignment and attention.
\newblock {\em arXiv:1803.03474 [cs]}, March 2018.
\newblock arXiv: 1803.03474.

\bibitem{cheng_focusing_2017}
Z.~Cheng, F.~Bai, Y.~Xu, G.~Zheng, S.~Pu, and S.~Zhou.
\newblock Focusing attention: Towards accurate text recognition in natural
  images.
\newblock In {\em 2017 {IEEE} {International} {Conference} on {Computer}
  {Vision} ({ICCV})}, pages 5086--5094, October 2017.

\bibitem{canjie_luo;lianwen_jin;zenghui_sun_moran:_2019}
Canjie Luo, Lianwen Jin, and Zenghui Sun.
\newblock Moran: A multi-object rectified attention network for scene text
  recognition.
\newblock {\em Pattern Recognition}, 90:109--118, 2019.

\bibitem{hongtao_xie;shancheng_fang;zheng-jun_zha;yating_yang;yan_li;yongdong_zhang_convolutional_2019}
Hongtao Xie, Shancheng Fang, Zheng-Jun Zha, Yating Yang, Yan Li, and Yongdong
  Zhang.
\newblock Convolutional attention networks for scene text recognition.
\newblock {\em ACM Transactions on Multimedia Computing, Communications, and
  Applications (TOMM)}, 15(1s):1--17, 2019.

\bibitem{hui_li;peng_wang;chunhua_shen;guyu_zhang_show_2019}
Hui Li, Peng Wang, Chunhua Shen, and Guyu Zhang.
\newblock Show, attend and read: A simple and strong baseline for irregular
  text recognition.
\newblock In {\em Proceedings of the AAAI Conference on Artificial
  Intelligence}, pages 8610--8617, 2019.

\bibitem{noauthor_focusing_nodate}
Zhanzhan Cheng, Fan Bai, Yunlu Xu, Gang Zheng, Shiliang Pu, and Shuigeng Zhou.
\newblock Focusing attention: Towards accurate text recognition in natural
  images.
\newblock In {\em Proc. of the IEEE ICCV}, pages 5076--5084, 2017.

\bibitem{xin2016recurrent}
Miao Xin, Hong Zhang, Mingui Sun, and Ding Yuan.
\newblock Recurrent temporal sparse autoencoder for attention-based action
  recognition.
\newblock In {\em 2016 International joint conference on neural networks
  (IJCNN)}, pages 456--463. IEEE, 2016.

\bibitem{zang2018attention}
Jinliang Zang, Le~Wang, Ziyi Liu, Qilin Zhang, Gang Hua, and Nanning Zheng.
\newblock Attention-based temporal weighted convolutional neural network for
  action recognition.
\newblock In {\em IFIP International Conference on Artificial Intelligence
  Applications and Innovations}, pages 97--108. Springer, 2018.

\bibitem{pei2017temporal}
Wenjie Pei, Tadas Baltrusaitis, David~Mj Tax, and Louis-Philippe Morency.
\newblock Temporal attention-gated model for robust sequence classification.
\newblock In {\em Proceedings of the IEEE Conference on Computer Vision and
  Pattern Recognition}, pages 6730--6739, 2017.

\bibitem{zhang2017gru}
Jianshu Zhang, Jun Du, and Lirong Dai.
\newblock A gru-based encoder-decoder approach with attention for online
  handwritten mathematical expression recognition.
\newblock In {\em 2017 14th IAPR International Conference on Document Analysis
  and Recognition (ICDAR)}, volume~1, pages 902--907. IEEE, 2017.

\bibitem{ji2018stacked}
Zhong Ji, Yanwei Fu, Jichang Guo, Yanwei Pang, Zhongfei~Mark Zhang, et~al.
\newblock Stacked semantics-guided attention model for fine-grained zero-shot
  learning.
\newblock In {\em Advances in Neural Information Processing Systems}, pages
  5995--6004, 2018.

\bibitem{schlemper_attention_2018}
Jo~Schlemper, Ozan Oktay, Michiel Schaap, Mattias Heinrich, Bernhard Kainz, Ben
  Glocker, and Daniel Rueckert.
\newblock Attention gated networks: Learning to leverage salient regions in
  medical images.
\newblock {\em Medical image analysis}, 53:197--207, 2019.

\bibitem{meng2018mganet}
Xiandong Meng, Xuan Deng, Shuyuan Zhu, Shuaicheng Liu, Chuan Wang, Chen Chen,
  and Bing Zeng.
\newblock Mganet: A robust model for quality enhancement of compressed video.
\newblock {\em arXiv preprint arXiv:1811.09150}, 2018.

\bibitem{park2019down}
Dongwon Park, Jisoo Kim, and Se~Young Chun.
\newblock Down-scaling with learned kernels in multi-scale deep neural networks
  for non-uniform single image deblurring.
\newblock {\em arXiv preprint arXiv:1903.10157}, 2019.

\bibitem{xu2018structured}
Dan Xu, Wei Wang, Hao Tang, Hong Liu, Nicu Sebe, and Elisa Ricci.
\newblock Structured attention guided convolutional neural fields for monocular
  depth estimation.
\newblock In {\em Proceedings of the IEEE Conference on Computer Vision and
  Pattern Recognition}, pages 3917--3925, 2018.

\bibitem{liu2019end}
Shikun Liu, Edward Johns, and Andrew~J Davison.
\newblock End-to-end multi-task learning with attention.
\newblock In {\em Proceedings of the IEEE/CVF Conference on Computer Vision and
  Pattern Recognition}, pages 1871--1880, 2019.

\bibitem{suganuma2019attention}
Masanori Suganuma, Xing Liu, and Takayuki Okatani.
\newblock Attention-based adaptive selection of operations for image
  restoration in the presence of unknown combined distortions.
\newblock In {\em Proceedings of the IEEE Conference on Computer Vision and
  Pattern Recognition}, pages 9039--9048, 2019.

\bibitem{qian2018attentive}
Rui Qian, Robby~T Tan, Wenhan Yang, Jiajun Su, and Jiaying Liu.
\newblock Attentive generative adversarial network for raindrop removal from a
  single image.
\newblock In {\em Proceedings of the IEEE conference on computer vision and
  pattern recognition}, pages 2482--2491, 2018.

\bibitem{mejjati2018unsupervised}
Youssef~Alami Mejjati, Christian Richardt, James Tompkin, Darren Cosker, and
  Kwang~In Kim.
\newblock Unsupervised attention-guided image-to-image translation.
\newblock In {\em Advances in Neural Information Processing Systems}, pages
  3693--3703, 2018.

\bibitem{tang2019attention}
Hao Tang, Dan Xu, Nicu Sebe, and Yan Yan.
\newblock Attention-guided generative adversarial networks for unsupervised
  image-to-image translation.
\newblock In {\em 2019 International Joint Conference on Neural Networks
  (IJCNN)}, pages 1--8. IEEE, 2019.

\bibitem{jin2018deep}
Sheng Jin, Hongxun Yao, Xiaoshuai Sun, Shangchen Zhou, Lei Zhang, and Xiansheng
  Hua.
\newblock Deep saliency hashing.
\newblock {\em arXiv preprint arXiv:1807.01459}, 2018.

\bibitem{yang2019deep}
Zhan Yang, Osolo~Ian Raymond, Wuqing Sun, and Jun Long.
\newblock Deep attention-guided hashing.
\newblock {\em IEEE Access}, 7:11209--11221, 2019.

\bibitem{huiyan_jiang;tianyu_shi;zhiqi_bai;liangliang_huang_ahcnet:_2019}
Huiyan Jiang, Tianyu Shi, Zhiqi Bai, and Liangliang Huang.
\newblock Ahcnet: An application of attention mechanism and hybrid connection
  for liver tumor segmentation in ct volumes.
\newblock {\em IEEE Access}, 7:24898--24909, 2019.

\bibitem{ilse2018attention}
Maximilian Ilse, Jakub~M Tomczak, and Max Welling.
\newblock Attention-based deep multiple instance learning.
\newblock {\em arXiv preprint arXiv:1802.04712}, 2018.

\bibitem{lee2016recursive}
Chen-Yu Lee and Simon Osindero.
\newblock Recursive recurrent nets with attention modeling for ocr in the wild.
\newblock In {\em Proceedings of the IEEE Conference on Computer Vision and
  Pattern Recognition}, pages 2231--2239, 2016.

\bibitem{chu2017multi}
Xiao Chu, Wei Yang, Wanli Ouyang, Cheng Ma, Alan~L Yuille, and Xiaogang Wang.
\newblock Multi-context attention for human pose estimation.
\newblock In {\em Proceedings of the IEEE Conference on Computer Vision and
  Pattern Recognition}, pages 1831--1840, 2017.

\bibitem{zhang2018image}
Yulun Zhang, Kunpeng Li, Kai Li, Lichen Wang, Bineng Zhong, and Yun Fu.
\newblock Image super-resolution using very deep residual channel attention
  networks.
\newblock In {\em Proceedings of the European Conference on Computer Vision
  (ECCV)}, pages 286--301, 2018.

\bibitem{zagoruyko_paying_2016}
Sergey Zagoruyko and Nikos Komodakis.
\newblock Paying more attention to attention: Improving the performance of
  convolutional neural networks via attention transfer.
\newblock {\em arXiv:1612.03928 [cs]}, December 2016.
\newblock arXiv: 1612.03928.

\bibitem{xiao-yu_zhang;haichao_shi;changsheng_li;kai_zheng;xiaobin_zhu;lixin_duan_learning_2019}
Xiao-Yu Zhang, Haichao Shi, Changsheng Li, Kai Zheng, Xiaobin Zhu, and Lixin
  Duan.
\newblock Learning transferable self-attentive representations for action
  recognition in untrimmed videos with weak supervision.
\newblock In {\em Proceedings of the AAAI Conference on Artificial
  Intelligence}, pages 9227--9234, 2019.

\bibitem{bielski2018pay}
Adam Bielski and Tomasz Trzcinski.
\newblock Pay attention to virality: understanding popularity of social media
  videos with the attention mechanism.
\newblock In {\em Proceedings of the IEEE CVPR Workshops}, pages 2335--2337,
  2018.

\bibitem{long2018attention}
Xiang Long, Chuang Gan, Gerard De~Melo, Jiajun Wu, Xiao Liu, and Shilei Wen.
\newblock Attention clusters: Purely attention based local feature integration
  for video classification.
\newblock In {\em Proceedings of the IEEE Conference on Computer Vision and
  Pattern Recognition}, pages 7834--7843, 2018.

\bibitem{yundong_zhang;xiang_xu;xiaotao_liu_robust_2019}
Yundong Zhang, Xiang Xu, and Xiaotao Liu.
\newblock Robust and high performance face detector.
\newblock {\em arXiv preprint arXiv:1901.02350}, 2019.

\bibitem{shengtao_xiao;jiashi_feng;junliang_xing;hanjiang_lai;shuicheng_yan;ashraf_a._kassim_robust_2016}
Shengtao Xiao, Jiashi Feng, Junliang Xing, Hanjiang Lai, Shuicheng Yan, and
  Ashraf Kassim.
\newblock Robust facial landmark detection via recurrent attentive-refinement
  networks.
\newblock In {\em European conference on computer vision}, pages 57--72.
  Springer, 2016.

\bibitem{lu2018deep}
Na~Lu, Yidan Wu, Li~Feng, and Jinbo Song.
\newblock Deep learning for fall detection: Three-dimensional cnn combined with
  lstm on video kinematic data.
\newblock {\em IEEE journal of biomedical and health informatics},
  23(1):314--323, 2018.

\bibitem{zhang2019cross}
Lu~Zhang, Zhiyong Liu, Shifeng Zhang, Xu~Yang, Hong Qiao, Kaizhu Huang, and
  Amir Hussain.
\newblock Cross-modality interactive attention network for multispectral
  pedestrian detection.
\newblock {\em Information Fusion}, 50:20--29, 2019.

\bibitem{zhang2018occluded}
Shanshan Zhang, Jian Yang, and Bernt Schiele.
\newblock Occluded pedestrian detection through guided attention in cnns.
\newblock In {\em Proceedings of the IEEE Conference on Computer Vision and
  Pattern Recognition}, pages 6995--7003, 2018.

\bibitem{yuan2019vssa}
Yuan Yuan, Zhitong Xiong, and Qi~Wang.
\newblock Vssa-net: vertical spatial sequence attention network for traffic
  sign detection.
\newblock {\em IEEE transactions on image processing}, 28(7):3423--3434, 2019.

\bibitem{wojna_attention-based_2017}
Zbigniew Wojna, Alexander~N Gorban, Dar-Shyang Lee, Kevin Murphy, Qian Yu,
  Yeqing Li, and Julian Ibarz.
\newblock Attention-based extraction of structured information from street view
  imagery.
\newblock In {\em 2017 14th IAPR International Conference on Document Analysis
  and Recognition (ICDAR)}, volume~1, pages 844--850. IEEE, 2017.

\bibitem{he_single_2017}
Pan He, Weilin Huang, Tong He, Qile Zhu, Yu~Qiao, and Xiaolin Li.
\newblock Single shot text detector with regional attention.
\newblock In {\em Proceedings of the IEEE international conference on computer
  vision}, pages 3047--3055, 2017.

\bibitem{bhunia_script_2019}
Ankan~Kumar Bhunia, Aishik Konwer, Ayan~Kumar Bhunia, Abir Bhowmick, Partha~P.
  Roy, and Umapada Pal.
\newblock Script identification in natural scene image and video frames using
  an attention based convolutional-lstm network.
\newblock {\em Pattern Recognition}, 85:172--184, January 2019.

\bibitem{siyue_xie;haifeng_hu;yongbo_wu_deep_2019}
Siyue Xie, Haifeng Hu, and Yongbo Wu.
\newblock Deep multi-path convolutional neural network joint with salient
  region attention for facial expression recognition.
\newblock {\em Pattern Recognition}, 92:177--191, 2019.

\bibitem{shervin_minaee;amirali_abdolrashidi_deep-emotion:_2019}
Shervin Minaee and Amirali Abdolrashidi.
\newblock Deep-emotion: Facial expression recognition using attentional
  convolutional network.
\newblock {\em arXiv preprint arXiv:1902.01019}, 2019.

\bibitem{yong_li;jiabei_zeng;shiguang_shan;xilin_chen_occlusion_2019}
Yong Li, Jiabei Zeng, Shiguang Shan, and Xilin Chen.
\newblock Occlusion aware facial expression recognition using cnn with
  attention mechanism.
\newblock {\em IEEE Transactions on Image Processing}, 28(5):2439--2450, 2018.

\bibitem{yang2019fully}
Fan Yang, Lianwen Jin, Songxuan Lai, Xue Gao, and Zhaohai Li.
\newblock Fully convolutional sequence recognition network for water meter
  number reading.
\newblock {\em IEEE Access}, 7:11679--11687, 2019.

\bibitem{xu2017learning}
Dan Xu, Wanli Ouyang, Xavier Alameda-Pineda, Elisa Ricci, Xiaogang Wang, and
  Nicu Sebe.
\newblock Learning deep structured multi-scale features using attention-gated
  crfs for contour prediction.
\newblock In {\em Advances in Neural Information Processing Systems}, pages
  3961--3970, 2017.

\bibitem{xie2019semantic}
De~Xie, Cheng Deng, Hao Wang, Chao Li, and Dapeng Tao.
\newblock Semantic adversarial network with multi-scale pyramid attention for
  video classification.
\newblock In {\em Proceedings of the AAAI Conference on Artificial
  Intelligence}, volume~33, pages 9030--9037, 2019.

\bibitem{tan2019multimodal}
Zhi-Xuan Tan, Arushi Goel, Thanh-Son Nguyen, and Desmond~C Ong.
\newblock A multimodal lstm for predicting listener empathic responses over
  time.
\newblock In {\em 2019 14th IEEE International Conference on Automatic Face \&
  Gesture Recognition (FG 2019)}, pages 1--4. IEEE, 2019.

\bibitem{yu2018mattnet}
Licheng Yu, Zhe Lin, Xiaohui Shen, Jimei Yang, Xin Lu, Mohit Bansal, and
  Tamara~L Berg.
\newblock Mattnet: Modular attention network for referring expression
  comprehension.
\newblock In {\em Proceedings of the IEEE Conference on Computer Vision and
  Pattern Recognition}, pages 1307--1315, 2018.

\bibitem{wang2018tienet}
Xiaosong Wang, Yifan Peng, Le~Lu, Zhiyong Lu, and Ronald~M Summers.
\newblock Tienet: Text-image embedding network for common thorax disease
  classification and reporting in chest x-rays.
\newblock In {\em Proceedings of the IEEE conference on computer vision and
  pattern recognition}, pages 9049--9058, 2018.

\bibitem{xu2018attngan}
Tao Xu, Pengchuan Zhang, Qiuyuan Huang, Han Zhang, Zhe Gan, Xiaolei Huang, and
  Xiaodong He.
\newblock Attngan: Fine-grained text to image generation with attentional
  generative adversarial networks.
\newblock In {\em Proceedings of the IEEE conference on computer vision and
  pattern recognition}, pages 1316--1324, 2018.

\bibitem{poulos2017character}
Jason Poulos and Rafael Valle.
\newblock Character-based handwritten text transcription with attention
  networks.
\newblock {\em arXiv preprint arXiv:1712.04046}, 2017.

\bibitem{noauthor_bottom-up_nodate}
Peter Anderson, Xiaodong He, Chris Buehler, Damien Teney, Mark Johnson, Stephen
  Gould, and Lei Zhang.
\newblock Bottom-up and top-down attention for image captioning and visual
  question answering.
\newblock In {\em Proceedings of the IEEE conference on computer vision and
  pattern recognition}, pages 6077--6086, 2018.

\bibitem{zhu_image_2018}
Xinxin Zhu, Lixiang Li, Jing Liu, Ziyi Li, Haipeng Peng, and Xinxin Niu.
\newblock Image captioning with triple-attention and stack parallel lstm.
\newblock {\em Neurocomputing}, 319:55--65, November 2018.

\bibitem{ma_da-gan:_2018-1}
Shuang Ma, Jianlong Fu, Chang~Wen Chen, and Tao Mei.
\newblock Da-gan: Instance-level image translation by deep attention generative
  adversarial networks.
\newblock In {\em Proceedings of the IEEE Conference on Computer Vision and
  Pattern Recognition}, pages 5657--5666, 2018.

\bibitem{lu_knowing_2017}
Jiasen Lu, Caiming Xiong, Devi Parikh, and Richard Socher.
\newblock Knowing when to look: Adaptive attention via a visual sentinel for
  image captioning.
\newblock In {\em Proceedings of the IEEE conference on computer vision and
  pattern recognition}, pages 375--383, 2017.

\bibitem{you_image_2016}
Quanzeng You, Hailin Jin, Zhaowen Wang, Chen Fang, and Jiebo Luo.
\newblock Image captioning with semantic attention.
\newblock In {\em The {IEEE} {Conference} on {Computer} {Vision} and {Pattern}
  {Recognition} ({CVPR})}, June 2016.

\bibitem{chen_sca-cnn:_2017}
Long Chen, Hanwang Zhang, Jun Xiao, Liqiang Nie, Jian Shao, Wei Liu, and
  Tat-Seng Chua.
\newblock Sca-cnn: Spatial and channel-wise attention in convolutional networks
  for image captioning.
\newblock In {\em The {IEEE} {Conference} on {Computer} {Vision} and {Pattern}
  {Recognition} ({CVPR})}, July 2017.

\bibitem{he_vd-san:_2019}
Xinwei He, Yang Yang, Baoguang Shi, and Xiang Bai.
\newblock Vd-san: Visual-densely semantic attention network for image caption
  generation.
\newblock {\em Neurocomputing}, 328:48--55, February 2019.

\bibitem{fang_captions_2015}
Hao Fang, Saurabh Gupta, Forrest Iandola, Rupesh~K Srivastava, Li~Deng, Piotr
  Doll{\'a}r, Jianfeng Gao, Xiaodong He, Margaret Mitchell, John~C Platt,
  et~al.
\newblock From captions to visual concepts and back.
\newblock In {\em Proceedings of the IEEE conference on computer vision and
  pattern recognition}, pages 1473--1482, 2015.

\bibitem{yang_review_2016}
Zhilin Yang, Ye~Yuan, Yuexin Wu, Ruslan Salakhutdinov, and William~W. Cohen.
\newblock Review networks for caption generation.
\newblock {\em arXiv:1605.07912 [cs]}, May 2016.
\newblock arXiv: 1605.07912.

\bibitem{pedersoli_areas_2016}
Marco Pedersoli, Thomas Lucas, Cordelia Schmid, and Jakob Verbeek.
\newblock Areas of attention for image captioning.
\newblock {\em arXiv:1612.01033 [cs]}, December 2016.
\newblock arXiv: 1612.01033.

\bibitem{ting_yao;yingwei_pan;yehao_li;tao_mei_exploring_2018}
Ting Yao, Yingwei Pan, Yehao Li, and Tao Mei.
\newblock Exploring visual relationship for image captioning.
\newblock In {\em Proceedings of the European conference on computer vision
  (ECCV)}, pages 684--699, 2018.

\bibitem{x._liang;_z._hu;_h._zhang;_c._gan;_e._p._xing_recurrent_nodate}
Xiaodan Liang, Zhiting Hu, Hao Zhang, Chuang Gan, and Eric~P Xing.
\newblock Recurrent topic-transition gan for visual paragraph generation.
\newblock In {\em Proc. of the IEEE ICCV}, pages 3362--3371, 2017.

\bibitem{anna_rohrbach;marcus_rohrbach;ronghang_hu;trevor_darrell;bernt_schiele_grounding_2017}
Anna Rohrbach, Marcus Rohrbach, Ronghang Hu, Trevor Darrell, and Bernt Schiele.
\newblock Grounding of textual phrases in images by reconstruction.
\newblock In {\em European Conference on Computer Vision}, pages 817--834.
  Springer, 2016.

\bibitem{lukasz_kaiser;aidan_n._gomez;noam_shazeer;ashish_vaswani;niki_parmar;llion_jones;jakob_uszkoreit_one_2017}
Lukasz Kaiser, Aidan~N Gomez, Noam Shazeer, Ashish Vaswani, Niki Parmar, Llion
  Jones, and Jakob Uszkoreit.
\newblock One model to learn them all.
\newblock {\em arXiv preprint arXiv:1706.05137}, 2017.

\bibitem{kuang-huei_lee;xi_chen;gang_hua;houdong_hu;xiaodong_he_stacked_2018}
Kuang-Huei Lee, Xi~Chen, Gang Hua, Houdong Hu, and Xiaodong He.
\newblock Stacked cross attention for image-text matching.
\newblock In {\em Proceedings of the European Conference on Computer Vision
  (ECCV)}, pages 201--216, 2018.

\bibitem{pan2020x}
Yingwei Pan, Ting Yao, Yehao Li, and Tao Mei.
\newblock X-linear attention networks for image captioning.
\newblock In {\em Proceedings of the IEEE/CVF Conference on Computer Vision and
  Pattern Recognition}, pages 10971--10980, 2020.

\bibitem{cho_describing_2015}
Kyunghyun Cho, Aaron Courville, and Yoshua Bengio.
\newblock Describing multimedia content using attention-based encoder--decoder
  networks.
\newblock {\em IEEE Transactions on Multimedia}, 17(11):1875--1886, November
  2015.
\newblock arXiv: 1507.01053.

\bibitem{zhou2018end}
Luowei Zhou, Yingbo Zhou, Jason~J Corso, Richard Socher, and Caiming Xiong.
\newblock End-to-end dense video captioning with masked transformer.
\newblock In {\em Proceedings of the IEEE Conference on Computer Vision and
  Pattern Recognition}, pages 8739--8748, 2018.

\bibitem{zhu2020actbert}
Linchao Zhu and Yi~Yang.
\newblock Actbert: Learning global-local video-text representations.
\newblock In {\em Proceedings of the IEEE/CVF Conference on Computer Vision and
  Pattern Recognition}, pages 8746--8755, 2020.

\bibitem{yu_video_2015}
Haonan Yu, Jiang Wang, Zhiheng Huang, Yi~Yang, and Wei Xu.
\newblock Video paragraph captioning using hierarchical recurrent neural
  networks.
\newblock {\em arXiv:1510.07712 [cs]}, October 2015.
\newblock arXiv: 1510.07712.

\bibitem{krishna_dense-captioning_2017}
Ranjay Krishna, Kenji Hata, Frederic Ren, Li~Fei-Fei, and Juan Carlos~Niebles.
\newblock Dense-captioning events in videos.
\newblock In {\em Proceedings of the IEEE international conference on computer
  vision}, pages 706--715, 2017.

\bibitem{yi_bin;yang_yang;fumin_shen;ning_xie;heng_tao_shen;xuelong_li_describing_2019}
Yi~Bin, Yang Yang, Fumin Shen, Ning Xie, Heng~Tao Shen, and Xuelong Li.
\newblock Describing video with attention-based bidirectional lstm.
\newblock {\em IEEE transactions on cybernetics}, 49(7):2631--2641, 2018.

\bibitem{l._baraldi;_c._grana;_r._cucchiara_hierarchical_nodate}
Lorenzo Baraldi, Costantino Grana, and Rita Cucchiara.
\newblock Hierarchical boundary-aware neural encoder for video captioning.
\newblock In {\em Proceedings of the IEEE Conference on Computer Vision and
  Pattern Recognition}, pages 1657--1666, 2017.

\bibitem{silvio_olivastri;gurkirt_singh;fabio_cuzzolin_end--end_2019}
Silvio Olivastri, Gurkirt Singh, and Fabio Cuzzolin.
\newblock An end-to-end baseline for video captioning.
\newblock {\em arXiv preprint arXiv:1904.02628}, 2019.

\bibitem{zhong_ji;kailin_xiong;yanwei_pang;xuelong_li_video_2018}
Zhong Ji, Kailin Xiong, Yanwei Pang, and Xuelong Li.
\newblock Video summarization with attention-based encoder--decoder networks.
\newblock {\em IEEE Transactions on Circuits and Systems for Video Technology},
  30(6):1709--1717, 2019.

\bibitem{jingkuan_song;zhao_guo;lianli_gao;wu_liu;dongxiang_zhang;heng_tao_shen_hierarchical_2017}
Jingkuan Song, Zhao Guo, Lianli Gao, Wu~Liu, Dongxiang Zhang, and Heng~Tao
  Shen.
\newblock Hierarchical lstm with adjusted temporal attention for video
  captioning.
\newblock {\em arXiv preprint arXiv:1706.01231}, 2017.

\bibitem{xiangpeng_li;zhilong_zhou;lijiang_chen;lianli_gao_residual_2019}
Xiangpeng Li, Zhilong Zhou, Lijiang Chen, and Lianli Gao.
\newblock Residual attention-based lstm for video captioning.
\newblock {\em World Wide Web}, 22(2):621--636, 2019.

\bibitem{lianli_gao;zhao_guo;hanwang_zhang;xing_xu;heng_tao_shen_video_2017}
Lianli Gao, Zhao Guo, Hanwang Zhang, Xing Xu, and Heng~Tao Shen.
\newblock Video captioning with attention-based lstm and semantic consistency.
\newblock {\em IEEE Transactions on Multimedia}, 19(9):2045--2055, 2017.

\bibitem{jingwen_wang;wenhao_jiang;lin_ma;wei_liu;yong_xu_bidirectional_2018}
Jingwen Wang, Wenhao Jiang, Lin Ma, Wei Liu, and Yong Xu.
\newblock Bidirectional attentive fusion with context gating for dense video
  captioning.
\newblock In {\em Proceedings of the IEEE Conference on Computer Vision and
  Pattern Recognition}, pages 7190--7198, 2018.

\bibitem{kim2020modality}
Junyeong Kim, Minuk Ma, Trung Pham, Kyungsu Kim, and Chang~D Yoo.
\newblock Modality shifting attention network for multi-modal video question
  answering.
\newblock In {\em Proceedings of the IEEE/CVF Conference on Computer Vision and
  Pattern Recognition}, pages 10106--10115, 2020.

\bibitem{jiangfantastic}
Ming Jiang, Shi Chen, Jinhui Yang, and Qi~Zhao.
\newblock Fantastic answers and where to find them: Immersive question-directed
  visual attention.
\newblock In {\em Proceedings of the IEEE/CVF Conference on Computer Vision and
  Pattern Recognition}, pages 2980--2989, 2020.

\bibitem{liang_focal_2018}
Junwei Liang, Lu~Jiang, Liangliang Cao, Jia Li, and Alexander Haupmann.
\newblock Focal visual-text attention for visual question answering.
\newblock In {\em {CVPR}}, 2018.

\bibitem{hudson_compositional_2018}
Drew~A Hudson and Christopher~D Manning.
\newblock Compositional attention networks for machine reasoning.
\newblock {\em arXiv preprint arXiv:1803.03067}, 2018.

\bibitem{osman_dual_2018}
Ahmed Osman and Wojciech Samek.
\newblock Dual recurrent attention units for visual question answering.
\newblock {\em arXiv:1802.00209 [cs, stat]}, February 2018.
\newblock arXiv: 1802.00209.

\bibitem{lu_hierarchical_2016}
Jiasen Lu, Jianwei Yang, Dhruv Batra, and Devi Parikh.
\newblock Hierarchical question-image co-attention for visual question
  answering.
\newblock In D.~D. Lee, M.~Sugiyama, U.~V. Luxburg, I.~Guyon, and R.~Garnett,
  editors, {\em Advances in {Neural} {Information} {Processing} {Systems} 29},
  pages 289--297. Curran Associates, Inc., 2016.

\bibitem{kim_bilinear_2018}
Jin-Hwa Kim, Jaehyun Jun, and Byoung-Tak Zhang.
\newblock Bilinear attention networks.
\newblock In S.~Bengio, H.~Wallach, H.~Larochelle, K.~Grauman, N.~Cesa-Bianchi,
  and R.~Garnett, editors, {\em Advances in {Neural} {Information} {Processing}
  {Systems} 31}, pages 1564--1574. Curran Associates, Inc., 2018.

\bibitem{nguyen_improved_2018}
Duy-Kien Nguyen and Takayuki Okatani.
\newblock Improved fusion of visual and language representations by dense
  symmetric co-attention for visual question answering.
\newblock In {\em The {IEEE} {Conference} on {Computer} {Vision} and {Pattern}
  {Recognition} ({CVPR})}, June 2018.

\bibitem{andreas_deep_2015}
Jacob Andreas, Marcus Rohrbach, Trevor Darrell, and Dan Klein.
\newblock Deep compositional question answering with neural module networks.
\newblock {\em CoRR}, abs/1511.02799, 2015.

\bibitem{kim_multimodal_2016}
Jin-Hwa Kim, Sang-Woo Lee, Donghyun Kwak, Min-Oh Heo, Jeonghee Kim, Jung-Woo
  Ha, and Byoung-Tak Zhang.
\newblock Multimodal residual learning for visual qa.
\newblock In D.~D. Lee, M.~Sugiyama, U.~V. Luxburg, I.~Guyon, and R.~Garnett,
  editors, {\em Advances in {Neural} {Information} {Processing} {Systems} 29},
  pages 361--369. Curran Associates, Inc., 2016.

\bibitem{shih_where_2015}
Kevin~J Shih, Saurabh Singh, and Derek Hoiem.
\newblock Where to look: Focus regions for visual question answering.
\newblock In {\em Proceedings of the IEEE conference on computer vision and
  pattern recognition}, pages 4613--4621, 2016.

\bibitem{zhu_visual7w:_2015}
Yuke Zhu, Oliver Groth, Michael Bernstein, and Li~Fei-Fei.
\newblock Visual7w: Grounded question answering in images.
\newblock {\em arXiv:1511.03416 [cs]}, November 2015.
\newblock arXiv: 1511.03416.

\bibitem{jiasen_lu;anitha_kannan;jianwei_yang;devi_parikh;dhruv_batra_best_2017}
Jiasen Lu, Anitha Kannan, Jianwei Yang, Devi Parikh, and Dhruv Batra.
\newblock Best of both worlds: Transferring knowledge from discriminative
  learning to a generative visual dialog model.
\newblock {\em arXiv preprint arXiv:1706.01554}, 2017.

\bibitem{zhou_yu;jun_yu;chenchao_xiang;jianping_fan;dacheng_tao_beyond_2019}
Zhou Yu, Jun Yu, Chenchao Xiang, Jianping Fan, and Dacheng Tao.
\newblock Beyond bilinear: Generalized multimodal factorized high-order pooling
  for visual question answering.
\newblock {\em IEEE Trans. Neural Netw. Learn. Syst.}, 29(12):5947--5959, 2018.

\bibitem{aishwarya_agrawal;dhruv_batra;devi_parikh;aniruddha_kembhavi_dont_2018}
Aishwarya Agrawal, Dhruv Batra, Devi Parikh, and Aniruddha Kembhavi.
\newblock Don't just assume; look and answer: Overcoming priors for visual
  question answering.
\newblock In {\em Proc. of the IEEE CVPR}, pages 4971--4980, 2018.

\bibitem{yan_zhang;jonathon_hare;adam_prugel-bennett_learning_2018}
Yan Zhang, Jonathon Hare, and Adam Pr{\"u}gel-Bennett.
\newblock Learning to count objects in natural images for visual question
  answering.
\newblock {\em arXiv preprint arXiv:1802.05766}, 2018.

\bibitem{zhe_gan;yu_cheng;ahmed_el_kholy;linjie_li;jingjing_liu;jianfeng_gao_multi-step_2019}
Zhe Gan, Yu~Cheng, Ahmed~El Kholy, Linjie Li, Jingjing Liu, and Jianfeng Gao.
\newblock Multi-step reasoning via recurrent dual attention for visual dialog.
\newblock {\em arXiv preprint arXiv:1902.00579}, 2019.

\bibitem{kan_chen;jiang_wang;liang-chieh_chen;haoyuan_gao;wei_xu;ram_nevatia_abc-cnn:_2016}
Kan Chen, Jiang Wang, Liang-Chieh Chen, Haoyuan Gao, Wei Xu, and Ram Nevatia.
\newblock Abc-cnn: An attention based convolutional neural network for visual
  question answering.
\newblock {\em arXiv preprint arXiv:1511.05960}, 2015.

\bibitem{huijuan_xu;kate_saenko_ask;_2016}
Huijuan Xu and Kate Saenko.
\newblock Ask, attend and answer: Exploring question-guided spatial attention
  for visual question answering.
\newblock In {\em European Conference on Computer Vision}, pages 451--466.
  Springer, 2016.

\bibitem{yundong_zhang;juan_carlos_niebles;alvaro_soto_interpretable_2019}
Yundong Zhang, Juan~Carlos Niebles, and Alvaro Soto.
\newblock Interpretable visual question answering by visual grounding from
  attention supervision mining.
\newblock In {\em 2019 IEEE Winter Conference on Applications of Computer
  Vision (WACV)}, pages 349--357. IEEE, 2019.

\bibitem{noauthor_focal_nodate}
Junwei Liang, Lu~Jiang, Liangliang Cao, Yannis Kalantidis, Li-Jia Li, and
  Alexander~G Hauptmann.
\newblock Focal visual-text attention for memex question answering.
\newblock {\em IEEE PAMI}, 41(8):1893--1908, 2019.

\bibitem{kim2020hypergraph}
Eun-Sol Kim, Woo~Young Kang, Kyoung-Woon On, Yu-Jung Heo, and Byoung-Tak Zhang.
\newblock Hypergraph attention networks for multimodal learning.
\newblock In {\em Proceedings of the IEEE/CVF Conference on Computer Vision and
  Pattern Recognition}, pages 14581--14590, 2020.

\bibitem{yang_stacked_2016}
Zichao Yang, Xiaodong He, Jianfeng Gao, Li~Deng, and Alex Smola.
\newblock Stacked attention networks for image question answering.
\newblock In {\em Proceedings of the IEEE conference on computer vision and
  pattern recognition}, pages 21--29, 2016.

\bibitem{yu_multi-level_2017}
Dongfei Yu, Jianlong Fu, Tao Mei, and Yong Rui.
\newblock Multi-level attention networks for visual question answering.
\newblock In {\em Proceedings of the IEEE Conference on Computer Vision and
  Pattern Recognition}, pages 4709--4717, 2017.

\bibitem{z._yu;_j._yu;_j._fan;_d._tao_multi-modal_nodate}
Zhou Yu, Jun Yu, Jianping Fan, and Dacheng Tao.
\newblock Multi-modal factorized bilinear pooling with co-attention learning
  for visual question answering.
\newblock In {\em Proceedings of the IEEE international conference on computer
  vision}, pages 1821--1830, 2017.

\bibitem{liu2018stamp}
Qiao Liu, Yifu Zeng, Refuoe Mokhosi, and Haibin Zhang.
\newblock Stamp: short-term attention/memory priority model for session-based
  recommendation.
\newblock In {\em Proc. of the 24th ACM SIGKDD Int. Conf. on Knowledge
  Discovery \& Data Mining}, pages 1831--1839, 2018.

\bibitem{chen2017attentive}
Jingyuan Chen, Hanwang Zhang, Xiangnan He, Liqiang Nie, Wei Liu, and Tat-Seng
  Chua.
\newblock Attentive collaborative filtering: Multimedia recommendation with
  item-and component-level attention.
\newblock In {\em Proceedings of the 40th International ACM SIGIR conference on
  Research and Development in Information Retrieval}, pages 335--344, 2017.

\bibitem{li_action_2019}
Hongyang Li, Jun Chen, Ruimin Hu, Mei Yu, Huafeng Chen, and Zengmin Xu.
\newblock Action recognition using visual attention with reinforcement
  learning.
\newblock In Ioannis Kompatsiaris, Benoit Huet, Vasileios Mezaris, Cathal
  Gurrin, Wen-Huang Cheng, and Stefanos Vrochidis, editors, {\em {MultiMedia}
  {Modeling}}, Lecture {Notes} in {Computer} {Science}, pages 365--376.
  Springer International Publishing, 2019.

\bibitem{rao_attention-aware_2017}
Yongming Rao, Jiwen Lu, and Jie Zhou.
\newblock Attention-aware deep reinforcement learning for video face
  recognition.
\newblock In {\em Proceedings of the IEEE international conference on computer
  vision}, pages 3931--3940, 2017.

\bibitem{stollenga_deep_2014}
Marijn Stollenga, Jonathan Masci, Faustino Gomez, and J{\"u}rgen Schmidhuber.
\newblock Deep networks with internal selective attention through feedback
  connections.
\newblock {\em arXiv preprint arXiv:1407.3068}, 2014.

\bibitem{cao_attention-aware_2017}
Qingxing Cao, Liang Lin, Yukai Shi, Xiaodan Liang, and Guanbin Li.
\newblock Attention-aware face hallucination via deep reinforcement learning.
\newblock In {\em Proceedings of the IEEE Conference on Computer Vision and
  Pattern Recognition}, pages 690--698, 2017.

\bibitem{donghui_hu;shengnan_zhou;qiang_shen;shuli_zheng;zhongqiu_zhao;yuqi_fan_digital_2019}
Donghui Hu, Shengnan Zhou, Qiang Shen, Shuli Zheng, Zhongqiu Zhao, and Yuqi
  Fan.
\newblock Digital image steganalysis based on visual attention and deep
  reinforcement learning.
\newblock {\em IEEE Access}, 7:25924--25935, 2019.

\bibitem{yeung_end--end_2016}
Serena Yeung, Olga Russakovsky, Greg Mori, and Li~Fei-Fei.
\newblock End-to-end learning of action detection from frame glimpses in
  videos.
\newblock In {\em 2016 {IEEE} {Conference} on {Computer} {Vision} and {Pattern}
  {Recognition} ({CVPR})}, pages 2678--2687, Las Vegas, NV, USA, June 2016.
  IEEE.

\bibitem{lee2018graph}
John~Boaz Lee, Ryan Rossi, and Xiangnan Kong.
\newblock Graph classification using structural attention.
\newblock In {\em Proc. of the 24th ACM SIGKDD International Conference on
  Knowledge Discovery \& Data Mining}, pages 1666--1674, 2018.

\bibitem{parisotto2017neural}
Emilio Parisotto and Ruslan Salakhutdinov.
\newblock Neural map: Structured memory for deep reinforcement learning.
\newblock {\em arXiv preprint arXiv:1702.08360}, 2017.

\bibitem{zambaldi2018relational}
Vinicius Zambaldi, David Raposo, Adam Santoro, Victor Bapst, Yujia Li, Igor
  Babuschkin, Karl Tuyls, David Reichert, Timothy Lillicrap, Edward Lockhart,
  et~al.
\newblock Relational deep reinforcement learning.
\newblock {\em arXiv preprint arXiv:1806.01830}, 2018.

\bibitem{baker2019emergent}
Bowen Baker, Ingmar Kanitscheider, Todor Markov, Yi~Wu, Glenn Powell, Bob
  McGrew, and Igor Mordatch.
\newblock Emergent tool use from multi-agent autocurricula.
\newblock {\em arXiv preprint arXiv:1909.07528}, 2019.

\bibitem{schaul2011high}
Tom Schaul, Tobias Glasmachers, and J{\"u}rgen Schmidhuber.
\newblock High dimensions and heavy tails for natural evolution strategies.
\newblock In {\em Proceedings of the 13th annual conference on Genetic and
  evolutionary computation}, pages 845--852, 2011.

\bibitem{one_shot}
Yan Duan, Marcin Andrychowicz, Bradly Stadie, OpenAI~Jonathan Ho, Jonas
  Schneider, Ilya Sutskever, Pieter Abbeel, and Wojciech Zaremba.
\newblock One-shot imitation learning.
\newblock In {\em Advances in neural information processing systems}, pages
  1087--1098, 2017.

\bibitem{xue2020deep}
Fei Xue, Xin Wang, Junqiu Wang, and Hongbin Zha.
\newblock Deep visual odometry with adaptive memory.
\newblock {\em arXiv preprint arXiv:2008.01655}, 2020.

\bibitem{johnston2020self}
Adrian Johnston and Gustavo Carneiro.
\newblock Self-supervised monocular trained depth estimation using
  self-attention and discrete disparity volume.
\newblock In {\em Proceedings of the IEEE/CVF Conference on Computer Vision and
  Pattern Recognition}, pages 4756--4765, 2020.

\bibitem{kuo2020dynamic}
Xin-Yu Kuo, Chien Liu, Kai-Chen Lin, and Chun-Yi Lee.
\newblock Dynamic attention-based visual odometry.
\newblock In {\em Proceedings of the IEEE/CVF CVPR Workshops}, pages 36--37,
  2020.

\bibitem{damirchi2020exploring}
Hamed Damirchi, Rooholla Khorrambakht, and Hamid~D Taghirad.
\newblock Exploring self-attention for visual odometry.
\newblock {\em arXiv preprint arXiv:2011.08634}, 2020.

\bibitem{gao2020attentional}
Feng Gao, Jincheng Yu, Hao Shen, Yu~Wang, and Huazhong Yang.
\newblock Attentional separation-and-aggregation network for self-supervised
  depth-pose learning in dynamic scenes.
\newblock {\em arXiv preprint arXiv:2011.09369}, 2020.

\bibitem{li2021transformer}
Xiangyu Li, Yonghong Hou, Pichao Wang, Zhimin Gao, Mingliang Xu, and Wanqing
  Li.
\newblock Transformer guided geometry model for flow-based unsupervised visual
  odometry.
\newblock {\em Neural Computing and Applications}, pages 1--12, 2021.

\bibitem{sophie}
Amir Sadeghian, Vineet Kosaraju, Ali Sadeghian, Noriaki Hirose, Hamid
  Rezatofighi, and Silvio Savarese.
\newblock Sophie: An attentive gan for predicting paths compliant to social and
  physical constraints.
\newblock In {\em Proceedings of the IEEE Conference on Computer Vision and
  Pattern Recognition}, pages 1349--1358, 2019.

\bibitem{social_attention}
Anirudh Vemula, Katharina Muelling, and Jean Oh.
\newblock Social attention: Modeling attention in human crowds.
\newblock In {\em IEEE ICRA}, pages 1--7. IEEE, 2018.

\bibitem{crowd}
Changan Chen, Yuejiang Liu, Sven Kreiss, and Alexandre Alahi.
\newblock Crowd-robot interaction: Crowd-aware robot navigation with
  attention-based deep reinforcement learning.
\newblock In {\em 2019 IEEE ICRA}, pages 6015--6022. IEEE, 2019.

\bibitem{scene_memory}
Kuan Fang, Alexander Toshev, Li~Fei-Fei, and Silvio Savarese.
\newblock Scene memory transformer for embodied agents in long-horizon tasks.
\newblock In {\em Proceedings of the IEEE CVPR}, pages 538--547, 2019.

\bibitem{translating_navigation}
Xiaoxue Zang, Ashwini Pokle, Marynel V{\'a}zquez, Kevin Chen, Juan~Carlos
  Niebles, Alvaro Soto, and Silvio Savarese.
\newblock Translating navigation instructions in natural language to a
  high-level plan for behavioral robot navigation.
\newblock {\em arXiv preprint arXiv:1810.00663}, 2018.

\bibitem{li2016understanding}
Jiwei Li, Will Monroe, and Dan Jurafsky.
\newblock Understanding neural networks through representation erasure.
\newblock {\em arXiv preprint arXiv:1612.08220}, 2016.

\bibitem{serrano2019attention}
Sofia Serrano and Noah~A Smith.
\newblock Is attention interpretable?
\newblock {\em arXiv preprint arXiv:1906.03731}, 2019.

\bibitem{vig2019analyzing}
Jesse Vig and Yonatan Belinkov.
\newblock Analyzing the structure of attention in a transformer language model.
\newblock {\em arXiv preprint arXiv:1906.04284}, 2019.

\bibitem{tenney2019bert}
Ian Tenney, Dipanjan Das, and Ellie Pavlick.
\newblock Bert rediscovers the classical nlp pipeline.
\newblock {\em arXiv preprint arXiv:1905.05950}, 2019.

\bibitem{clark2019does}
Kevin Clark, Urvashi Khandelwal, Omer Levy, and Christopher~D Manning.
\newblock What does bert look at? an analysis of bert's attention.
\newblock {\em arXiv preprint arXiv:1906.04341}, 2019.

\bibitem{attention_is_not_explanation}
Sarthak Jain and Byron~C Wallace.
\newblock Attention is not explanation.
\newblock {\em arXiv preprint arXiv:1902.10186}, 2019.

\bibitem{vashishth2019attention}
Shikhar Vashishth, Shyam Upadhyay, Gaurav~Singh Tomar, and Manaal Faruqui.
\newblock Attention interpretability across nlp tasks.
\newblock {\em arXiv preprint arXiv:1909.11218}, 2019.

\bibitem{wiegreffe2019attention}
Sarah Wiegreffe and Yuval Pinter.
\newblock Attention is not not explanation.
\newblock {\em arXiv preprint arXiv:1908.04626}, 2019.

\bibitem{sabour2017dynamic}
Sara Sabour, Nicholas Frosst, and Geoffrey~E Hinton.
\newblock Dynamic routing between capsules.
\newblock {\em arXiv preprint arXiv:1710.09829}, 2017.

\bibitem{choi2019attention}
Jaewoong Choi, Hyun Seo, Suii Im, and Myungjoo Kang.
\newblock Attention routing between capsules.
\newblock In {\em Proceedings of the IEEE/CVF ICCV Workshops}, pages 0--0,
  2019.

\bibitem{huang2020capsnet}
Wenkai Huang and Fobao Zhou.
\newblock Da-capsnet: dual attention mechanism capsule network.
\newblock {\em Scientific Reports}, 10(1):1--13, 2020.

\bibitem{hoogi2019self}
Assaf Hoogi, Brian Wilcox, Yachee Gupta, and Daniel~L Rubin.
\newblock Self-attention capsule networks for image classification.
\newblock {\em arXiv preprint arXiv:1904.12483}, 2019.

\bibitem{tsai2020capsules}
Yao-Hung~Hubert Tsai, Nitish Srivastava, Hanlin Goh, and Ruslan Salakhutdinov.
\newblock Capsules with inverted dot-product attention routing.
\newblock {\em arXiv preprint arXiv:2002.04764}, 2020.

\bibitem{lecun2015deep}
Yann LeCun, Yoshua Bengio, and Geoffrey Hinton.
\newblock Deep learning.
\newblock {\em nature}, 521(7553):436--444, 2015.

\bibitem{khosravi2010survey}
Hassan Khosravi and Bahareh Bina.
\newblock A survey on statistical relational learning.
\newblock In {\em Canadian conference on artificial intelligence}, pages
  256--268. Springer, 2010.

\bibitem{besold2017neural}
Tarek~R Besold, Artur~d'Avila Garcez, Sebastian Bader, Howard Bowman, Pedro
  Domingos, Pascal Hitzler, Kai-Uwe K{\"u}hnberger, Luis~C Lamb, Daniel Lowd,
  Priscila Machado~Vieira Lima, et~al.
\newblock Neural-symbolic learning and reasoning: A survey and interpretation.
\newblock {\em arXiv preprint arXiv:1711.03902}, 2017.

\bibitem{yang2017differentiable}
Fan Yang, Zhilin Yang, and William~W Cohen.
\newblock Differentiable learning of logical rules for knowledge base
  reasoning.
\newblock In {\em Advances in Neural Information Processing Systems}, pages
  2319--2328, 2017.

\bibitem{wang2019logic}
Peifeng Wang, Jialong Han, Chenliang Li, and Rong Pan.
\newblock Logic attention based neighborhood aggregation for inductive
  knowledge graph embedding.
\newblock In {\em Proceedings of the AAAI Conference on Artificial
  Intelligence}, volume~33, pages 7152--7159, 2019.

\bibitem{harsha2020probabilistic}
L~Vivek Harsha~Vardhan, Guo Jia, and Stanley Kok.
\newblock Probabilistic logic graph attention networks for reasoning.
\newblock In {\em Companion Proceedings of the Web Conference 2020}, pages
  669--673, 2020.

\bibitem{ferret2019self}
Johan Ferret, Rapha{\"e}l Marinier, Matthieu Geist, and Olivier Pietquin.
\newblock Self-attentional credit assignment for transfer in reinforcement
  learning.
\newblock {\em arXiv preprint arXiv:1907.08027}, 2019.

\bibitem{rudin2018please}
Cynthia Rudin.
\newblock Please stop explaining black box models for high stakes decisions.
\newblock {\em arXiv preprint arXiv:1811.10154}, 1, 2018.

\bibitem{riedl2019human}
Mark~O Riedl.
\newblock Human-centered artificial intelligence and machine learning.
\newblock {\em Human Behavior and Emerging Technologies}, 1(1):33--36, 2019.

\bibitem{he2019attgan}
Zhenliang He, Wangmeng Zuo, Meina Kan, Shiguang Shan, and Xilin Chen.
\newblock Attgan: Facial attribute editing by only changing what you want.
\newblock {\em IEEE Transactions on Image Processing}, 28(11):5464--5478, 2019.

\bibitem{zhang2017battrae}
Biao Zhang, Deyi Xiong, and Jinsong Su.
\newblock Battrae: Bidimensional attention-based recursive autoencoders for
  learning bilingual phrase embeddings.
\newblock In {\em Proceedings of the AAAI Conference on Artificial
  Intelligence}, 2017.

\bibitem{tian2019attention}
Tian Tian and Zheng~Felix Fang.
\newblock Attention-based autoencoder topic model for short texts.
\newblock {\em Procedia Computer Science}, 151:1134--1139, 2019.

\end{thebibliography}

\end{document}